\documentclass[10pt,twocolumn,letterpaper]{article}
\usepackage[accsupp]{axessibility}

\usepackage[pagenumbers]{cvpr} 

\usepackage[dvipsnames]{xcolor}

\usepackage{marvosym}  
\usepackage{graphicx}
\usepackage{caption}
\usepackage{stfloats} 
\usepackage{cuted}
\usepackage{multirow}
\usepackage{tabularx}

\usepackage{tocloft}
\usepackage{etoc}

\usepackage{titletoc}

\newcommand\DoToC{%
  \startcontents
  \printcontents{}{1}{\noindent \textbf{\Large{Table of Contents in Appendix}}\vskip3pt\vskip5pt}
  \vskip3pt\vskip5pt
}

\newcommand{\listappendixname}{List of Appendices}
\newlistof{appendices}{apc}{\listappendixname}
\newlistof{appfigures}{afg}{List of Case Study Figures}

\usepackage[normalem]{ulem}
\useunder{\uline}{\ul}{}

\extrafloats{100}

\newcommand{\dataset}{\texttt{MMMU}\xspace}
\newcommand{\subjects}{30\xspace}
\newcommand{\fields}{183\xspace}

\usepackage{multicol}
\usepackage{aliascnt}
\usepackage{adjustbox}

\definecolor{cvprblue}{rgb}{0.21,0.49,0.74}
\usepackage[pagebackref,breaklinks,colorlinks,citecolor=cvprblue,hypertexnames=false]{hyperref}
\usepackage{siunitx} 
\usepackage{arydshln}

\def\paperID{5732} 
\def\confName{CVPR}
\def\confYear{2024}

\title{
MMMU: A Massive Multi-discipline Multimodal \\ Understanding and Reasoning Benchmark for Expert AGI
}

\author{
    $^1$Xiang Yue\thanks{Core Contributors. See the Author Contribution Statement for details.} \thanks{\Letter: \{yue.149,su.809\}@osu.edu; wenhuchen@uwaterloo.ca}\;, 
    $^2$Yuansheng Ni\footnotemark[1],\;
    $^3$Kai Zhang\footnotemark[1],\;
    $^4$Tianyu Zheng\footnotemark[1],\\
    $^3$Ruoqi Liu,
    $^2$Ge Zhang,
    $^3$Samuel Stevens,
    $^2$Dongfu Jiang, 
    $^2$Weiming Ren,
    $^4$Yuxuan Sun,\\
    $^2$Cong Wei, 
    $^3$Botao Yu,
    $^5$Ruibin Yuan,
    $^2$Renliang Sun,
    $^7$Ming Yin, \\
    $^3$Boyuan Zheng,
    $^4$Zhenzhu Yang,
    $^6$Yibo Liu, 
    $^4$Wenhao Huang,\\
    $^3$Huan Sun\footnotemark[1]\;,
    $^3$Yu Su\footnotemark[1] \footnotemark[2]\;,
    $^2$Wenhu Chen\footnotemark[1] \footnotemark[2] \\[2mm]
    $^1$IN.AI Research,
    $^2$University of Waterloo, $^3$The Ohio State University, $^4$Independent,\\ 
    $^5$Carnegie Mellon University, $^6$University of Victoria, $^7$Princeton University\\[2mm]
    \url{https://mmmu-benchmark.github.io/}
}

\begin{document}

\maketitle
\begin{strip}
    \vspace*{-2cm}
    \centering
    \includegraphics[width=\textwidth]{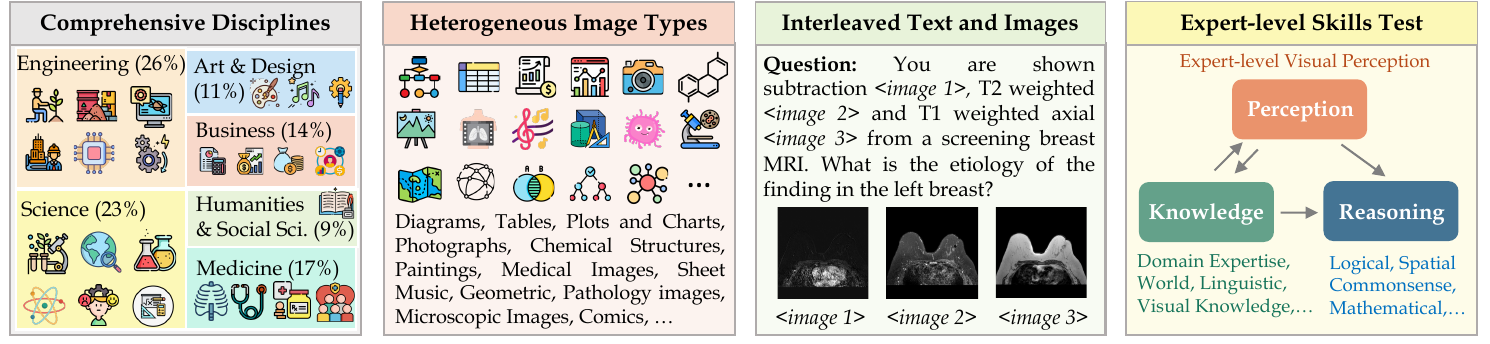}
    \captionof{figure}{Overview of the \dataset dataset. \dataset presents four challenges: 1) \textbf{comprehensiveness}: \num{11.5}K college-level problems across six broad disciplines and \num{30} college subjects; 2) highly \textbf{heterogeneous} image types;  3) \textbf{interleaved} text and images; 4) \textbf{expert-level} perception and reasoning rooted in deep subject knowledge.
    }
    \label{fig:teaser}
\end{strip}

\begin{abstract}
We introduce \dataset: a new benchmark designed to evaluate multimodal models on massive multi-discipline tasks demanding college-level subject knowledge and deliberate reasoning. \dataset includes 11.5K meticulously collected multimodal questions from college exams, quizzes, and textbooks, covering six core disciplines: Art \& Design, Business, Science, Health \& Medicine, Humanities \& Social Science, and Tech \& Engineering. These questions span 30 subjects and 183 subfields, comprising 30 highly heterogeneous image types, such as charts, diagrams, maps, tables, music sheets, and chemical structures. Unlike existing benchmarks, \dataset focuses on advanced perception and reasoning with domain-specific knowledge, challenging models to perform tasks akin to those faced by experts. The evaluation of 28 open-source LMMs as well as the proprietary  GPT-4V(ision) and Gemini highlights the substantial challenges posed by \dataset. Even the advanced GPT-4V and Gemini Ultra only achieve accuracies of 56\% and 59\% respectively, indicating significant room for improvement. We believe \dataset will stimulate the community to build next-generation multimodal foundation models towards expert artificial general intelligence. 
\end{abstract}
\begin{figure*}[!t]
    \centering
\includegraphics[width=\linewidth]{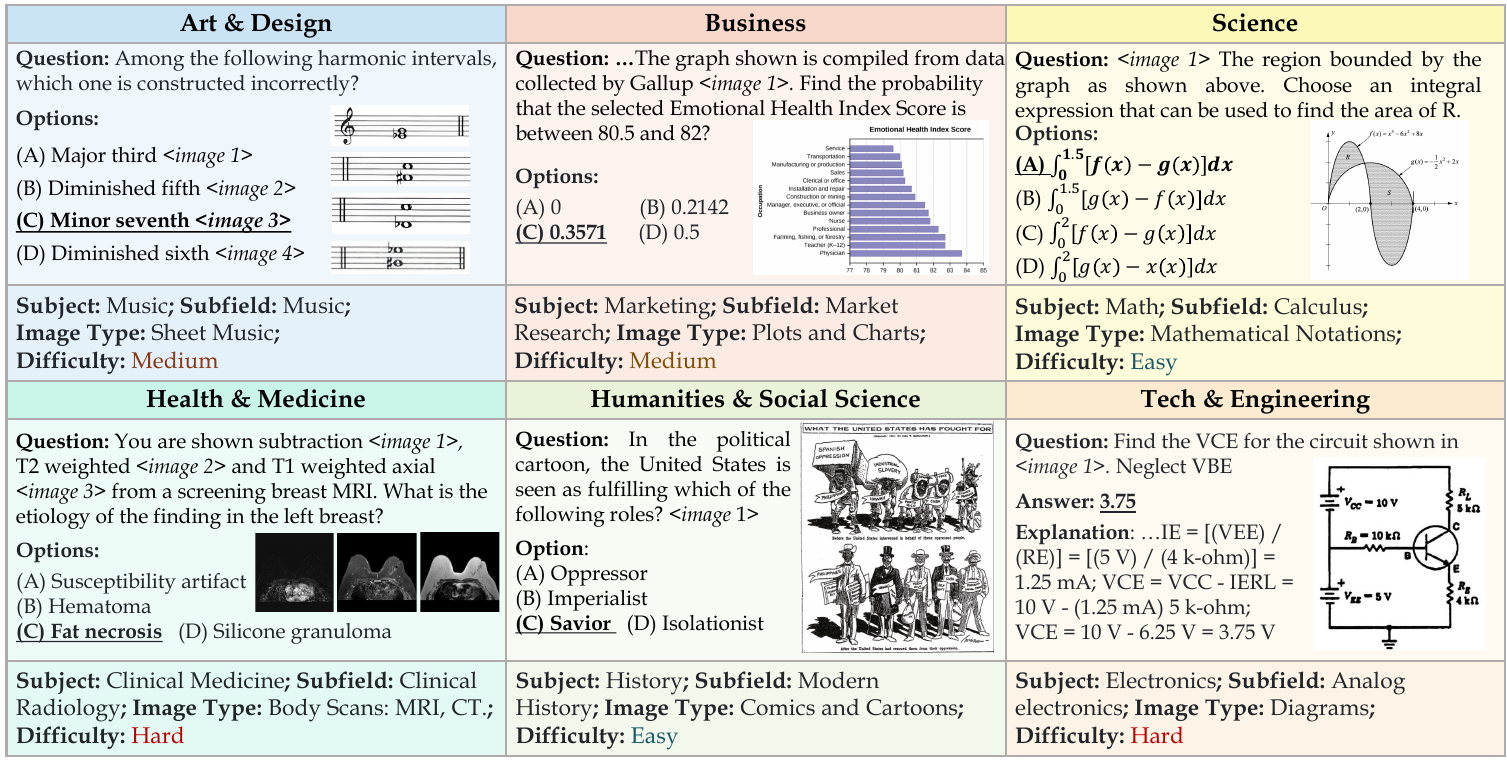}
    \caption{Sampled \dataset examples from each discipline. The questions and images need expert-level knowledge to understand and reason. }
    \label{fig:mmmu_examples}
\end{figure*}

\section{Introduction}
Rapid advances in large language models (LLMs)~\citep{chowdhery2022palm,openai2023gpt4,touvron2023llama} have sparked broad discussions on the controversial concept of artificial general intelligence (AGI), often used to describe AI systems that perform on par or surpass humans at most tasks~\citep{bubeck2023sparks, aguera2023artificial, latif2023artificial,ge2023openagi,Morris2023-ta,mialon2023gaia}. 
Candid and constructive discussions on AGI have been challenging due to a lack of shared operationalizable definitions.
In an attempt to remedy this, Morris et al.~\cite{Morris2023-ta} propose a leveled taxonomy for AGI that centers around both \textit{generality} (or breadth) and \textit{performance} (or depth).
In the suggested taxonomy, Level 3, or \textit{Expert AGI}, marks a critical milestone.
It denotes an AI system that reaches ``at least 90th percentile of skilled adults'' in a broad range of tasks, thus starting to achieve ``the substitution threshold for machine intelligence in lieu of human labor'' for many industries, leading to significant risks of job displacement and economic disruption. 
Therefore, it is of both intellectual and societal importance to closely monitor the progress towards Expert AGI.

How to create benchmarks for measuring Expert AGI? 
Since the definition is based on comparison with \textit{skilled adults}, a natural starting point is college-level exams for different disciplines, because those are designed to evaluate \textit{skilled adults} specialized in each discipline. 
This strategy has been successfully adopted in benchmarks such as MMLU~\citep{hendrycks2020measuring} and AGIEval~\citep{zhong2023agieval}, but only text-based questions are considered, while human experts are capable of solving multimodal problems.
Meanwhile, large multimodal models (LMMs) that can understand both text and images have been making a major stride towards more general AI~\citep{yang2023dawn,chen2023pali,liu2023improved,li2023blip,dai2023instructblip}.
These LMMs have consistently excelled in existing multimodal benchmarks~\citep{VQA,goyal2017making,lin2014microsoft,singh2019towards,li2023seed,liu2023mmbench,yu2023mm,yin2023lamm}. For instance, CogVLM~\cite{COGVLM} achieves \num{85}\% on VQA-v2~\citep{goyal2017making}, $92$\% on ScienceQA-IMG~\citep{lu2022learn}, and $93$\% on RefCOCO~\citep{kazemzadeh2014referitgame}.
However, most existing multimodal benchmarks focus on commonsense/daily knowledge rather than expert-level domain knowledge and advanced reasoning.
The closest one to our goal is ScienceQA~\citep{lu2022learn}. 
While it covers diverse disciplines (\textbf{breadth}), the majority of the questions are at the elementary to the middle school level, thus falling short in \textbf{depth} for benchmarking Expert AGI.

To this end, we introduce \dataset: a comprehensive benchmark designed for college-level multi-discipline multimodal understanding and reasoning. It features problems sourced from college exams, quizzes, and textbooks spanning six common disciplines: Art \& Design, Business, Science, Health \& Medicine, Humanities \& Social Science, and Tech \& Engineering. \dataset consists of $\mathbf{11.5}$K carefully selected multimodal questions, which cover $\mathbf{\subjects}$ diverse subjects and $\fields$ subfields, thus meeting the \textbf{breadth} goal. Moreover, many problems within \dataset require expert-level reasoning, such as applying ``Fourier Transform'' or ``Equilibrium Theory'' to derive the solution, thus meeting the \textbf{depth} goal. \dataset also presents two unique challenges absent in current benchmarks (\autoref{fig:teaser}). Firstly, it covers diverse image formats, from visual scenes like photographs and paintings to diagrams and tables, testing the perceptual capabilities of LMMs. Secondly, \dataset features interleaved text-image inputs.
A model needs to jointly understand the images and text, which often requires recalling deep subject knowledge, and conducting complex reasoning based on the understanding and knowledge to reach a solution.

We evaluate \num{28} open-source LMMs as well as the advanced proprietary LMMs such as GPT-4V(ision)~\citep{openai2023gpt4v} on \dataset. Our key findings are summarized as follows:
\begin{itemize}
    \item \dataset presents significant challenges; notably, GPT-4V only achieves an accuracy of \num{55.7}\%, indicating substantial room for improvement.
    \item There is a pronounced disparity in performance between open-source LMMs and GPT-4V. The highest-performing open-source models, such as BLIP2-FLAN-T5-XXL and LLaVA-1.5, achieve approximately \num{34}\% in accuracy.
    \item LLMs augmented with optical character recognition (OCR) or generated captions do not see notable improvement, indicating that \dataset necessitates deeper joint interpretation of images and text.
    \item In disciplines such as Art \& Design and Humanities \& Social Science, where visual data is less complex, models exhibit higher performance. In contrast, Business, Science, Health \& Medicine, and Tech \& Engineering, which present more complex visual data and require intricate reasoning, see relatively lower model performance.
    \item Our error analysis on \num{150} error cases of GPT-4V reveals that \num{35}\% of errors are perceptual, \num{29}\% stem from a lack of knowledge, and \num{26}\% are due to flaws in the reasoning process. These findings underscore the challenges of the \dataset benchmark and point towards areas needing further research and model enhancement.
\end{itemize}

Our aim with \dataset is to push the boundaries of what LMMs can achieve. We believe it will prove instrumental in developing next-generation multimodal foundation models and monitoring the progress towards Expert AGI.
We shall caution that \dataset is not a \textit{sufficient} test for Expert AGI, as per the definition~\citep{Morris2023-ta}, because there lacks a direct mapping between performance on \dataset and ``90th percentile of skilled adults,'' nor are college exams the only tasks an AGI shall tackle. 
However, we believe it should be \textit{necessary} for an Expert AGI to achieve strong performance on \dataset to demonstrate their broad and deep subject knowledge as well as expert-level understanding and reasoning capabilities.

\section{Related Work}

\noindent\textbf{Multimodal Pre-Training.}
In recent years, rapid progress has been made in multimodal pre-training, which aims to jointly encode vision and language in a fusion model. LXMERT~\citep{tan2019lxmert}, UNITER~\citep{chen2020uniter}, VinVL~\citep{zhang2021vinvl}, Oscar~\citep{li2020oscar}, VilBert~\citep{lu2019vilbert}, and VLP~\citep{zhou2020unified} are among the earliest work to train universal vision-language models to tackle many multimodal tasks. This work relies on pre-trained visual representations like Faster RCNN features~\citep{ren2015faster} to minimize the training sample complexity. Later on, CLIP~\citep{radford2021learning}, ALIGN~\citep{jia2021scaling}, SimVLM~\citep{wang2021simvlm}, CoCa~\citep{yu2022coca}, Flamingo~\citep{alayrac2022flamingo}, BLIP-2~\citep{li2023blip}, and Fuyu~\cite{fuyu-8b} (inter alia) have been proposed to train visual representation using ViT~\citep{dosovitskiy2021an} from scratch with massive amount of web data. These models have achieved great success on existing VQA and captioning tasks, which require less knowledge and reasoning. 

\noindent\textbf{Multimodal Instruction Tuning.} 
Inspired by open-source instruction-tuned LLMs like FLAN-T5~\cite{chung2022scaling} and Vicuna~\cite{vicuna2023}, models like LLaVA~\cite{liu2023visual,liu2023improved} and MiniGPT-4~\cite{zhu2023minigpt} utilized open-source resources, to improve the instruction-following capabilities of LMMs. The evolutionary trajectory of LMMs has also led to subsequent advancements aimed at improving the quantity and quality of visual instruction data. Models such as LLaMA-Adapter~\cite{zhang2023llama,gao2023llama}, mPlug-OWL~\cite{ye2023mplug,ye2023mplug2}, SVIT~\cite{zhao2023svit}, LRV-Instruction~\cite{liu2023aligning},  and InstructBLIP~\cite{dai2023instructblip} exemplify these developments. Another pivotal aspect of LMM research revolves around multimodal in-context learning and the management of interleaved text and image examples. This area has been explored in depth by models such as Flamingo~\cite{alayrac2022flamingo} and OpenFlamingo~\cite{awadalla2023openflamingo}, Otter~\cite{li2023otter}, M3IT~\cite{li2023m}, MetaVL~\cite{monajatipoor2023metavl}, Sparkles~\cite{huang2023sparkles}, and MMICL~\cite{zhao2023mmicl}. These models have significantly contributed to the ongoing advancements in multimodal training and instruction-following capabilities.

\noindent\textbf{LMM Benchmarks.}
With the surge of multi-modal pre-training and instruction tuning, the prior single-task evaluation benchmarks like VQA~\citep{VQA,goyal2017making}, OK-VQA~\citep{okvqa}, MSCOCO~\citep{lin2014microsoft}, GQA~\citep{hudson2019gqa}, etc., have become insufficient to holistically evaluate LMMs' general multimodal perception and reasoning abilities. Therefore, numerous all-round benchmarks have been established to assess different facets of LMMs. These benchmarks cover a wide spectrum of specific skills of LMMs, from Optical Character Recognition (OCR) as seen in the study by \cite{liu2023hidden}, to adversarial robustness \cite{zhao2023evaluating} and hallucination~\cite{cui2023holistic,liu2023hallusionbench}, e.g., POPE \cite{li2023evaluating} and HaELM \cite{wang2023evaluation}.  More holistic evaluations have been conducted as well, such as LAMM \cite{yin2023lamm}, LVLM-eHub \cite{xu2023lvlm}, SEED~\citep{li2023seed}, MMBench~\citep{liu2023mmbench}, and MM-Vet~\citep{yu2023mm}. These benchmarks still largely focus on relatively basic perception abilities without requiring expert-level domain knowledge and deliberate reasoning. More recently, MathVista ~\cite{lu2023mathvista} 
presents a collection of visually challenging questions; however, its scope is limited exclusively to the mathematical domain. \dataset is highly different from these benchmarks by collecting more difficult expert-level problems that cover 30 different subjects and require nuanced perception, recalling domain-specific knowledge to perform step-by-step reasoning to derive the solution. In line with the motivation of our study, concurrently,
GAIA~\cite{mialon2023gaia} introduces 466 questions that test fundamental abilities of models such as reasoning, multimodality handling, or tool use. 
\begin{table}[!t]
\centering
\small
    \begin{tabular}{lc}
        \toprule
        Statistics                     & \multicolumn{1}{c}{Number} \\ \midrule
        Total Questions                & 11550                      \\
        Total Disciplines/Subjects/Subfields               & 6/30/183               \\
               Image Types & 30 \\ \midrule
        Dev:Validation:Test & 150:900:10500 \\
         
        Difficulties (Easy: Medium: Hard)  & 28\%:45\%:27\%
        \\ \midrule
        Multiple-choice Questions      & 10861 (94.03\%)            \\
        Open Questions                 & 689 (5.97\%)               \\ \midrule
        Questions with an Explanation   & 2035 (17.62\%)             \\ \midrule
        Image in the Question              & 11264 (97.52\%)            \\
        \hspace{1em}* Images at the beginning       & 2006 (17.81\%)             \\
        \hspace{1em}* Images in the middle         & 4159 (36.92\%)             \\
        \hspace{1em}* Images at the end            & 5679 (50.42\%)             \\
        
        Image in Options               & 389 (3.37\%)               \\
        Example with Multiple Images & 854 (7.39\%)             
        \\ \midrule
        Average question length        & 59.33                      \\
        Average option length          & 9.17                       \\
        Average explanation length     & 107.92                     \\

        \bottomrule
    \end{tabular}
    \caption{Key statistics of the \dataset benchmark.}
    \vspace{-10pt}
    \label{tab:dataset_statistics}
\end{table}

\section{The MMMU Benchmark}

\subsection{Overview of MMMU}
We introduce the Massive Multi-discipline Multimodal
Understanding and Reasoning (\dataset) benchmark, a novel benchmark meticulously curated to assess the expert-level multimodal understanding capability of foundation models across a broad scope of tasks. Covering \subjects subjects across 6 disciplines, including Art, Business, Health \& Medicine, Science, Humanities \& Social Science, and Tech \& Engineering, and over \fields subfields. The detailed subject coverage and statistics are detailed in~\autoref{fig:subject_distribution}. The questions in our benchmark were manually collected by a team of 50 college students (including coauthors) from various disciplines and subjects, drawing from online sources, textbooks, and lecture materials.

\dataset, constituting 11.5K questions, is divided into a few-shot development set, a validation set, and a test set. The few-shot development set includes 5 questions per subject, and the validation set, useful for hyperparameter selection, contains approximately 900 questions, while the test set comprises 10.5K questions. \dataset is designed to measure three essential skills in LMMs: perception, knowledge, and reasoning. Our aim is to evaluate how well these models can not only perceive and understand information across different modalities but also apply reasoning with subject-specific knowledge to derive the solution. 

Our \dataset benchmark introduces four key challenges to multimodal foundation models, as detailed in \autoref{fig:teaser}. Among these, we particularly highlight the challenge stemming from the requirement for both expert-level visual perceptual abilities and deliberate reasoning with subject-specific knowledge. This challenge is vividly illustrated through our tasks, which not only demand the processing of various heterogeneous image types but also necessitate a model’s adeptness in using domain-specific knowledge to deeply understand both the text and images and to reason.
This goes significantly beyond basic visual perception, calling for an advanced approach that integrates advanced multimodal analysis with domain-specific knowledge.


\begin{figure*}[thb]
\centering

\begin{minipage}{0.4\textwidth}
\centering
\includegraphics[width=1.0\linewidth]{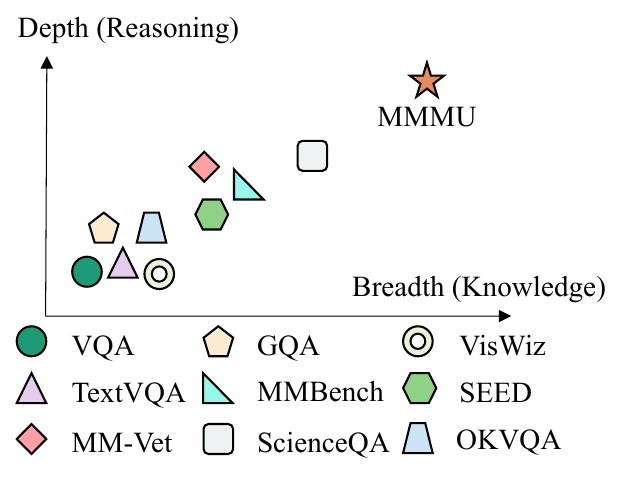}
\end{minipage}\begin{minipage}{0.6\textwidth}
\centering
\captionsetup{type=table} 
{\small
\begin{tabular}{llllllll}
\toprule
Dataset       &  Size   & Images         & Format    & Source            & Answer   \\
\midrule
VQA           & $>$ 1M  & V              & I+T       & Annotated         & Open          \\
GQA           & $>$ 1M  & V              & I+T       & Synthesized       & Open          \\
VisWiz        & 32K     & V              & I+T       & Annotated         & Open          \\
TextVQA       & 45K     & OC             & I+T       & Annotated         & MC  \\
OKVQA         & 14K     & V+OC           & I+T       & Annotated         & Open          \\
SEED          & 19K     & V+OC           & I+T       & Annotated         & MC  \\
MMBench       & 3K      & V+OC           & I+T       & Repurposed        & MC  \\
MM-Vet        & 0.2K    & V+OC           & I+T       & Annotated        & Open  \\
ScienceQA     & 6K      & 5 Types        & I+T       & Textbooks         & MC  \\
\midrule
MMMU          & 11.5K     & 30 Types      & Interleaved & \begin{tabular}[c]{@{}l@{}} Textbooks,\\Internet,\\Annotated \end{tabular} & \begin{tabular}[c]{@{}l@{}} Open / \\MC \end{tabular} \\
\bottomrule
\end{tabular}
}
\end{minipage}
\caption{The comparison between MMMU and other existing benchmarks. MMMU excels in both its breadth to cover a wide range of disciplines and its depth to test LMMs' reasoning abilities. In the image format, V means visual input, OC means optical characters, MC means multi-choice. Repurposed means the benchmark is a compilation of prior datasets.}
\label{fig:comaprison}
\end{figure*}

\subsection{Data Curation Process}
\noindent\textbf{Data Collection.}
Our benchmark collection takes three stages. Firstly, we go through the common university majors to decide what subjects should be included in our benchmark. The selection is based on the principle that visual inputs should be commonly adopted in the subjects to provide valuable information. Through this principle, we rule out a few subjects like law and linguistics because it is difficult to find enough relevant multimodal problems in these subjects. Consequently, we select 30 subjects from six different disciplines. In the second stage, we recruit over 50 university students, including co-authors, specializing in these majors as annotators to assist in question collection. They collect multimodal questions from major textbooks and online resources, creating new questions based on their expertise where necessary. The annotators are instructed to adhere to copyright and license regulations, avoiding data from sites prohibiting copy and redistribution. Given the arising data contamination concerns of foundation models, the annotators are advised to select questions without immediately available answers, such as those with answers in separate documents or at the end of textbooks. This process results in a diverse collection of 13K questions from various sources. The detailed annotation protocol is in Appendix A.

\noindent\textbf{Data Quality Control.}
To further control the quality of our data, we perform two steps of data cleaning. In the first stage, lexical overlap and source URL similarity are employed to identify potential duplicate problems. These suspected duplicates were then reviewed by the authors to identify and eliminate any duplications. The second stage involves distributing the problems among different co-authors for format and typo checking. This step requires authors to ensure adherence to a standardized format, undertaking necessary corrections where deviations are found. In the third and final stage, the authors categorize the problems into four difficulty levels: very easy, easy, medium, and hard. Approximately 10\% of the problems, classified as very easy and not aligning with our design criteria due to their simplistic nature, are excluded from the benchmark. This rigorous process plays a crucial role in maintaining the quality and difficulty of the problem set.

\subsection{Comparisons with Existing Benchmarks}
To further distinguish the difference between \dataset and other existing ones, we elaborate the benchmark details in \autoref{fig:comaprison}. From the \textit{breadth} perspective, the prior benchmarks are heavily focused on daily knowledge and common sense. The covered image format is also limited. Our benchmark aims to cover college-level knowledge with 30 image formats including diagrams, tables, charts, chemical structures, photos, paintings, geometric shapes, music sheets, medical images, etc. In the \textit{depth} aspect, the previous benchmarks normally require commonsense knowledge or simple physical or temporal reasoning. In contrast, our benchmark requires deliberate reasoning with college-level subject knowledge.
\begin{table*}[!t]
\centering
\small
\begin{adjustbox}{scale=0.83}
\begin{tabular}{@{}lcccccccc@{}}
\toprule
\textbf{} & \textbf{\begin{tabular}[c]{@{}c@{}}Validation \\  Overall\end{tabular}} & \textbf{\begin{tabular}[c]{@{}c@{}}Test \\  Overall\end{tabular} } & \textbf{\begin{tabular}[c]{@{}c@{}}Art \&\\  Design\end{tabular}} & \textbf{Business} & \textbf{Science} & \textbf{\begin{tabular}[c]{@{}c@{}}Health \& \\ Medicine\end{tabular}} & \textbf{\begin{tabular}[c]{@{}c@{}}Human. \&\\  Social Sci.\end{tabular}} & \textbf{\begin{tabular}[c]{@{}c@{}}Tech \&\\  Eng.\end{tabular}} \\
 & (900) & (10,500) & (1,163) & (1,428) & (2,426) & (1,752) & (947) & (2,784) \\ \midrule

\color{Gray} Random Choice & \color{Gray} 22.1 & \color{Gray} 23.9 & \color{Gray} 24.1 & \color{Gray} 24.9 & \color{Gray} 21.6 & \color{Gray} 25.3 & \color{Gray} 22.8 & \color{Gray} 24.8 \\ 
\color{Gray} Frequent Choice & \color{Gray} 26.8 & \color{Gray} 25.8 & \color{Gray} 26.7 & \color{Gray} 28.4 & \color{Gray} 24.0 & \color{Gray} 24.4 & \color{Gray} 25.2 & \color{Gray} 26.5 \\
Expert (Worst)  &76.2 &- &- &- &- &- &- &- \\
Expert (Medium)  &82.6 &- &- &- &- &- &- &- \\
Expert (Best)  &88.6 &- &- &- &- &- &- &- \\
\midrule
\multicolumn{8}{c}{\textbf{Large Multimodal Models (LMMs): Text + Image as Input}} \\ \midrule
OpenFlamingo2-9B \cite{awadalla2023openflamingo} & 28.7 & 26.3 & 31.7 & 23.5 & 26.3 & 26.3 & 27.9 & 25.1 \\
Kosmos2 \cite{peng2023kosmos} & 24.4 & 26.6 & 28.8 & 23.7 & 26.6 & 27.2 & 26.3 & 26.8 \\
Adept Fuyu-8B~\cite{fuyu-8b} & 27.9 & 27.4 & 29.9 & 27.0 & 25.6 & 27.0 & 32.5 & 26.4 \\
MiniGPT4-Vicuna-13B~\cite{zhu2023minigpt} & 26.8 & 27.6 & 30.2 & 27.0 & 26.2 & 26.9 & 30.9 & 27.2 \\
LLaMA-Adapter2-7B~\cite{zhang2023llama} & 29.8 & 27.7 & 35.2 & 25.4 & 25.6 & 30.0 & 29.1 & 25.7 \\
CogVLM~\cite{COGVLM} & 32.1 & 30.1 & 38.0 & 25.6 & 25.1 & 31.2 & 41.5 & 28.9 \\
Qwen-VL-7B-Chat~\cite{Qwen-VL} & 35.9 & 32.9 & 47.7 & 29.8 & 25.6 & 33.6 & 45.3 & 30.2 \\
InstructBLIP-T5-XXL~\cite{dai2023instructblip} & 35.7 & 33.8 & 48.5 & 30.6 & 27.6 & 33.6 & 49.8 & 29.4 \\
BLIP-2 FLAN-T5-XXL~\cite{li2023blip} & 35.4 & 34.0 & 49.2 & 28.6 & 27.3 & 33.7 & 51.5 & 30.4 \\
InternLM-XComposer2-VL* \cite{dong2024internlm} & 43.0 & 38.2 & 56.8 & 32.8 & 30.1 & 39.8 & 60.7 & 31.8 \\ 
Yi-VL-34B*  \cite{young2024yi} & 45.9 & 41.6 & 56.1 & 33.3 & 32.9 & 45.9 & 66.5 & 36.0 \\
LLaVA-1.6-34B* \cite{liu2024llava} & 51.1 & 44.7 & 58.6 & {\ul 39.9} & 36.0 & {\ul 51.2} & {\ul 70.2} & 36.3 \\
InternVL-Chat-V1.2* \cite{chen2023internvl} & {\ul 51.6} & {\ul 46.2} & \textbf{62.5} & 37.6 & \textbf{37.9} & 49.7 & 70.1 & \textbf{40.8} \\ 
VILA1.5* \cite{lin2023vila} & \textbf{51.9} & \textbf{46.9} & {\ul 62.1} & \textbf{40.6} & {\ul 37.7} & \textbf{51.7} & \textbf{74.0} & {\ul 39.5} \\ \midrule
Qwen-VL-MAX* \cite{Qwen-VL-MAX} & 51.4 & 46.8 & {\ul 64.2} & 39.8 & 36.3 & 52.5 & 70.4 & 40.7 \\
SenseChat-Vision-0423-Preview* \cite{SenseChat-Vision} & 54.6 & {\ul 50.3} & 62.7 & {\ul 44.1} & {\ul 42.3} & {\ul 55.7} & {\ul 74.7} & \textbf{43.5}  \\
GPT-4V(ision) (Playground) \cite{openai2023gpt4v} & 56.8 & \textbf{55.7} & \textbf{65.3} & \textbf{64.3} & \textbf{48.4} & \textbf{63.5} & \textbf{76.3} & {\ul 41.7} \\
Claude 3 Opus* \cite{Claude3} & 59.4 & - & - & - & - & - & - & -  \\
Gemini 1.5 Pro* \cite{deepmind_gemini1.5_report} & {\ul 62.2} & - & - & - & - & - & - & -  \\
GPT-4o* \cite{gpt-4o} & \textbf{69.1} & - & - & - & - & - & - & -  \\  \midrule
\multicolumn{8}{c}{\textbf{Large Language Models (LLMs): Only Text as Input}} \\ \midrule
Llama2 7B \cite{touvron2023llama2} & 30.1 & 28.7 & 30.7 & 27.2 & 26.7 & 27.7 & 32.6 & 29.8 \\ 
\addlinespace[0.1em]\hdashline\addlinespace[0.1em]
FLAN-T5-XXL~\cite{chung2022scaling} & 32.1 & 31.2 & 36.8 & \textbf{28.9} & 26.7 & 32.8 & 44.8 & 28.3 \\
\quad  + OCR & 34.7 & \textbf{31.9} & 36.2 & 28.8 & 26.2 & 32.6 & \textbf{50.5} & \textbf{29.7} \\
\quad  + LLaVA Caption & \textbf{34.8} & \textbf{31.9} & \textbf{38.4} & 27.8 & \textbf{27.0} & \textbf{33.2} & 49.9 & 28.7 \\ 
\addlinespace[0.1em]\hdashline\addlinespace[0.1em]
Vicuna-13B~\cite{vicuna2023} & 33.3 & 31.0 & 35.1 & \textbf{30.1} & 24.7 & 31.4 & 44.8 & 30.1 \\
\quad + OCR & \textbf{35.4} & 31.9 & 37.1 & 28.6 & \textbf{26.5} & 32.0 & 49.3 & 30.0 \\
\quad + LLaVA Caption & 33.9 & \textbf{32.7} & \textbf{42.0} & 26.8 & 26.2 & \textbf{33.4} & \textbf{49.4} & \textbf{31.4} \\
\addlinespace[0.1em]\hdashline\addlinespace[0.1em]
GPT-4 Text~\cite{openai2023gpt4} & 34.9 & 33.8 & 32.9 & 28.5 & 30.6 & 41.3 & 53.0 & 28.4 \\ 
\bottomrule
\end{tabular}%
\end{adjustbox}
\caption{Selected results of different models on the \dataset \textbf{validation} and \textbf{test set}. Besides reporting the performance of LMMs, we additionally add text-only LLM baselines. The best-performing model in each category is \textbf{in-bold}, and the second best is {\ul{underlined}}. *: results provided by the authors. Due to the page limit, we show other models' results in Appendix \autoref{tab:main_results}. The live-updating leaderboard is available at: \url{https://mmmu-benchmark.github.io/\#leaderboard}}
\label{tab:overall_results}
\end{table*}


\section{Experiments}
We evaluate various models including LLMs and LMMs.
In each type, we consider both closed- and open-source models.
Our evaluation is conducted under a \textit{zero-shot} setting to assess the capability of models to generate accurate answers without fine-tuning or few-shot demonstrations on our benchmark.
For all models, we use the default prompt provided by each model for multi-choice or open QA, if available.
If models do not provide prompts for task types in \dataset, we conduct prompt engineering on the validation set and use the most effective prompt for the zero-shot setup in the main experiments. We also report the few-shot results of some selected models in the Appendix.
All experiments are conducted with NVIDIA A100 GPUs.

\subsection{Baselines}
\noindent \textbf{LMMs.}
We consider various large multimodal models.
By default, for each model family, we use the latest, largest, and best-performing available checkpoint to date.
\textit{(i)} Kosmos2~\cite{peng2023kosmos} is pre-trained to ground fine-grained visual objects with texts and to follow instructions. With only 1.6B model size, Kosmos2 is able to achieve comparable or better performance with Flamingo-9B~\cite{alayrac2022flamingo} on VQA and captioning tasks.
\textit{(ii)} LLaMA-Adapter2~\cite{gao2023llama} fine-tunes Llama~\cite{touvron2023llama} in a parameter-efficient way and utilizes visual encoder CLIP~\cite{radford2021learning} and modular experts such as Optical Character Recognition (OCR) to capture more image information for later better visual understanding.
\textit{(iii)} BLIP-2~\cite{li2023blip} introduces light-weight learnable visual queries to bridge the frozen CLIP ViT~\cite{radford2021learning} and FLAN-T5~\cite{chung2022scaling}.
\textit{(iv)} Starting from the parameters from BLIP-2, InstructBLIP~\cite{dai2023instructblip} is further fine-tuned with visual instruction tuning data for better zero-shot generalization capabilities.
\textit{(v)} LLaVA-1.5~\cite{liu2023improved} linearly projects the visual embedding into word embedding space of Vicuna~\cite{vicuna2023}, thus equipping the LLM with visual abilities.
\textit{(vi)} As an open-source alternative to Flamingo~\cite{alayrac2022flamingo}, OpenFlamingo~\cite{awadalla2023openflamingo} has close performance on most vision-language tasks.
\textit{(vii)} CogVLM~\cite{COGVLM} concatenates image and text in the input embedding space and adds trainable visual layers in textual Transformer blocks to deeply align two modalities. It has been reported to achieve very promising performance on existing VQA benchmarks recently.
\textit{(viii)} Fuyu~\cite{fuyu-8b} projects the patches of the input image into text embedding space.
\textit{(ix)} Qwen-VL~\cite{Qwen-VL} introduces a set of trainable query embeddings and single-layer cross-attention module to bridge the modalities, supporting interleaved image-text input.
\textit{(x)} Otter~\cite{li2023otter} is fine-tuned with diverse instruction-tuning data and able to perform in-context learning.
\textit{(xi)} MiniGPT-4~\cite{zhu2023minigpt} is built upon Vicuna~\cite{vicuna2023} and designs a linear modality projection layer for visual understanding abilities.
\textit{(xii)} mPLUG-Owl2~\cite{ye2023mplug2} designs a modality-adaptive module to unify vision and language while preserving their distinct properties of them.

\noindent \textbf{Text-only LLMs.}
For text-only LLMs, we consider the most capable ones including GPT-4 and several open-source LLMs, Llama2-7B~\cite{touvron2023llama}, FLAN-T5-XXL and Vicuna-13B,  which are adopted as the text encoder or decoder in the selected LMMs. To determine if an external image-to-text tool can enhance these LLMs' performance on \dataset, we deploy OCR by MMOCR\footnote{https://github.com/open-mmlab/mmocr} or captioning by LLaVA-1.5 to
provide the recognized text information to text-only LLMs. 

\noindent\textbf{Human Experts.} We involve 90 college senior students, selected to represent a wide range of experts in the corresponding 30 subjects (3 student experts per subject). These students were tasked with completing the 30 questions in their corresponding subjects (900 validation questions in total). The students were allowed to consult their textbooks but were prohibited from searching the Internet for answers.

\noindent \textbf{Evaluation.}
We adopt micro-averaged accuracy as the evaluation metric.
For both open and multiple-choice questions, we design systematic, rule-based evaluation pipelines.
Specifically, to mitigate the potential influence of any intermediate generations (e.g., reasoning steps, calculations) in the long response, we construct robust regular expressions and develop response-processing workflows.
These are employed to extract key phrases, such as numbers and conclusion phrases, from the long responses for accurate answer matching.
If there is no valid answer in the model's response, we perform random selection as a remedy for multiple-choice questions or consider the response incorrect for open questions. 
For reference, we add Random Choice and Frequent Choice baselines: the former randomly selects an option, while the latter selects the most frequent option within each specific subject of the validation set, based on its frequency of occurrence in that subject.

\begin{figure}[!t]
    \centering
    \includegraphics[width=\linewidth]{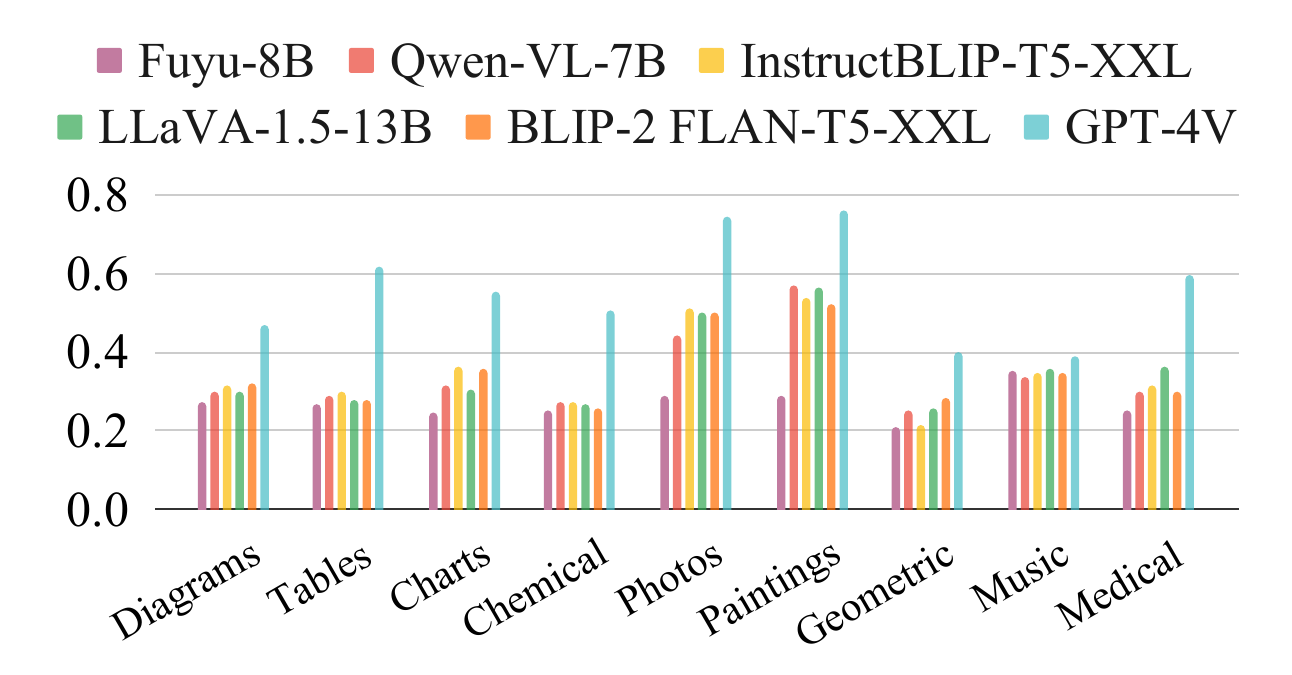}
    \caption{Performance of models on different types of images.}
    \vspace{-10pt}
    \label{fig:result_image_type}
\end{figure}

\subsection{Main Results}
In this section, we present a comprehensive comparison of different LLMs and LMMs using the \dataset benchmark, detailed in \autoref{tab:overall_results}.  We summarize our key findings as follows:

\noindent \textbf{Challenging Nature of \dataset}: The benchmark poses significant challenges to current models. The Best human expert achieves a validation accuracy of 88.6\%, significantly outperforming all the models reported in the table. This demonstrates the still-existing gap between human expertise and the performance of current models on the \dataset benchmark. This reflects the benchmark's rigorous standards.

\noindent \textbf{Disparity between Open-source Models and Closed-source models}: Leading open-source models (as the paper submission) such as BLIP2-FLAN-T5-XXL and LLaVA-1.5 reach an accuracy level of approximately 34\%, which is significantly lower than GPT-4V. However, it is exciting to see that open-source models have made significant strides in performance. For example, LLaVA-1.6-34B and InternVL-Chat-V1.2 achieve test accuracies of 44.7\% and 46.2\%, respectively, narrowing the gap with proprietary models.

\noindent \textbf{Effectiveness of OCR and Captioning Enhancements}: The application of OCR and captioning technologies does not yield a significant improvement in the performance of text-only LMMs. This finding suggests that the \dataset benchmark requires models that can effectively interpret and integrate both textual and visual information, underscoring the complexity of the multimodal tasks it presents.

\noindent \textbf{Model Performance across Different Disciplines}: In disciplines such as Art \& Design and Humanities \& Social Sciences, where the images tends to be more `natural' and questions involve relatively less reasoning, models demonstrate relatively higher performance. Conversely, in fields like Science, Health \& Medicine, and Technology \& Engineering, where tasks often involve intricate perception and complex reasoning, models exhibit lower performance.

The \dataset benchmark underscores both the progress and the challenges in multimodal understanding and reasoning. While GPT-4V leads in performance, the overall results indicate substantial room for improvement, especially in domains with complex visual input and heavy reasoning with subject knowledge.

\subsection{Analysis on Images Types and Difficulties}

\noindent\textbf{Different Image Types.}
We compare the performance of various models across top frequent image types in \autoref{fig:result_image_type}. Across all types, GPT-4V consistently outperforms the other models by a huge margin. Open-source models demonstrate relatively strong performance in categories like Photos and Paintings, which are more frequently seen during training. 
However, for less common image categories like Geometric shapes, Music sheets and Chemical structures, all models obtain very low scores (some are close to random guesses). This indicates that the existing models are generalizing poorly towards these image types.

\noindent\textbf{Different Difficulty Levels.}
\autoref{tab:result_diffculty} compares the performance of selected models across three difficulty levels. GPT-4V demonstrates a significantly higher proficiency, with a success rate of 76.1\%, compared to open-source models in the ``Easy'' category. When it comes to the ``Medium'' category, while the gap narrows, GPT-4V still leads at 55.6\%.
The further diminishing performance gap in the ``Hard'' category across models indicates that as the complexity of tasks increases, the advantage of more advanced models like GPT-4V almost disappears. This might reflect a current limitation in handling expert-level challenging queries even for the most advanced models.

\begin{table}[!t]
\small
    \centering
    \resizebox{\linewidth}{!}{%
\begin{tabular}{@{}lcccc@{}}
\toprule
\multirow{2}{*}{Models} & Easy & Medium & Hard & Overall \\
 & (2946) & (4917) & (2637) & (10500) \\ \midrule
Fuyu-8B~\cite{fuyu-8b} & 28.9 & 27.0 & 26.4 & 27.4 \\
Qwen-VL-7B~\cite{Qwen-VL} & 39.4 & 31.9 & 27.6 & 32.9 \\
LLaVA-1.5-13B~\cite{liu2023improved} & 41.3 & 32.7 & 26.7 & 33.6 \\
InstructBLIP-T5-XXL~\cite{dai2023instructblip} & 40.3 & 32.3 & 29.4 & 33.8 \\
BLIP-2 FLAN-T5-XXL~\cite{li2023blip} & 41.0 & 32.7 & 28.5 & 34.0 \\ \midrule
GPT-4V~\cite{openai2023gpt4v} & 76.1 & 55.6 & 31.2 & 55.7 \\ \bottomrule
\end{tabular}
}
    \caption{Result decomposition across question difficulty levels.}
    \label{tab:result_diffculty}
\end{table}

\section{Error Analysis and Future Work}
\begin{figure}[!t]
    \centering
\includegraphics[width=0.7\linewidth]{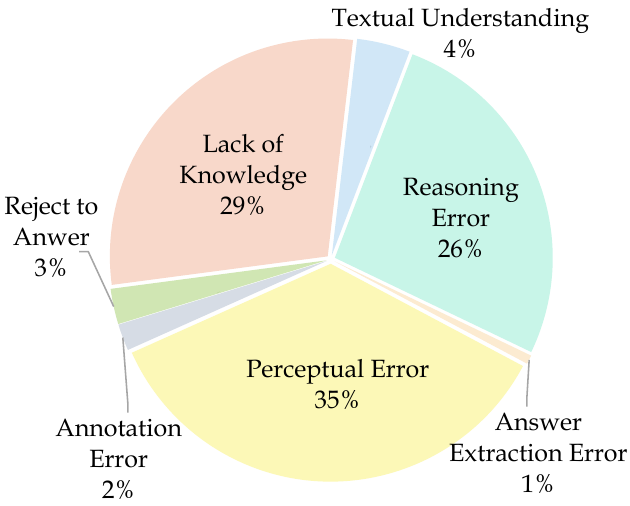}
    \caption{Error distribution over 150 annotated GPT-4V errors.}
    \vspace{-10pt}
\label{fig:error_distribution}
\end{figure}

In this section, we delve into the analysis of errors by GPT-4V, a pivotal aspect for understanding its operational capabilities and limitations. This analysis serves not only to identify the model's current shortcomings but also to guide future enhancements in its design and training. We meticulously examine 150 randomly sampled error instances from GPT-4V's predictions. These instances are analyzed by expert annotators who identify the \textit{root causes of mispredictions} based on their knowledge and the golden explanations if available. The distribution of these errors is illustrated in Figure \ref{fig:error_distribution}, and a selection of 100 notable cases, along with detailed analyses, is included in the Appendix.

\begin{figure}[!t]
    \centering
    \includegraphics[width=\linewidth]{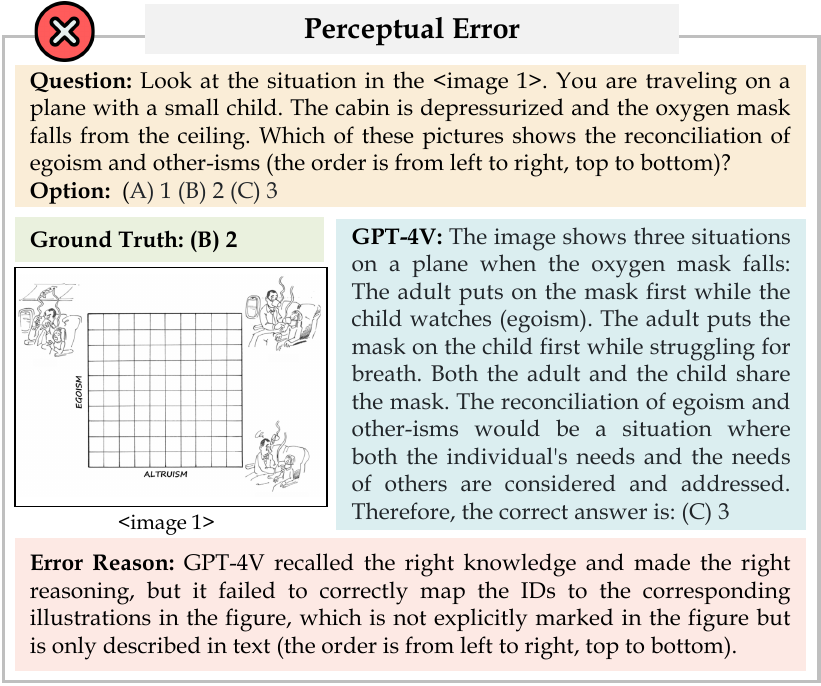}
    \caption{A basic perceptual error, easy for humans but challenging for GPT-4V. More examples can be found in the Appendix.}
    \label{fig:error_case_main_text}
    \vspace{-10pt}
\end{figure}

\begin{figure*}[!h]
    \centering
\includegraphics[width=\linewidth]{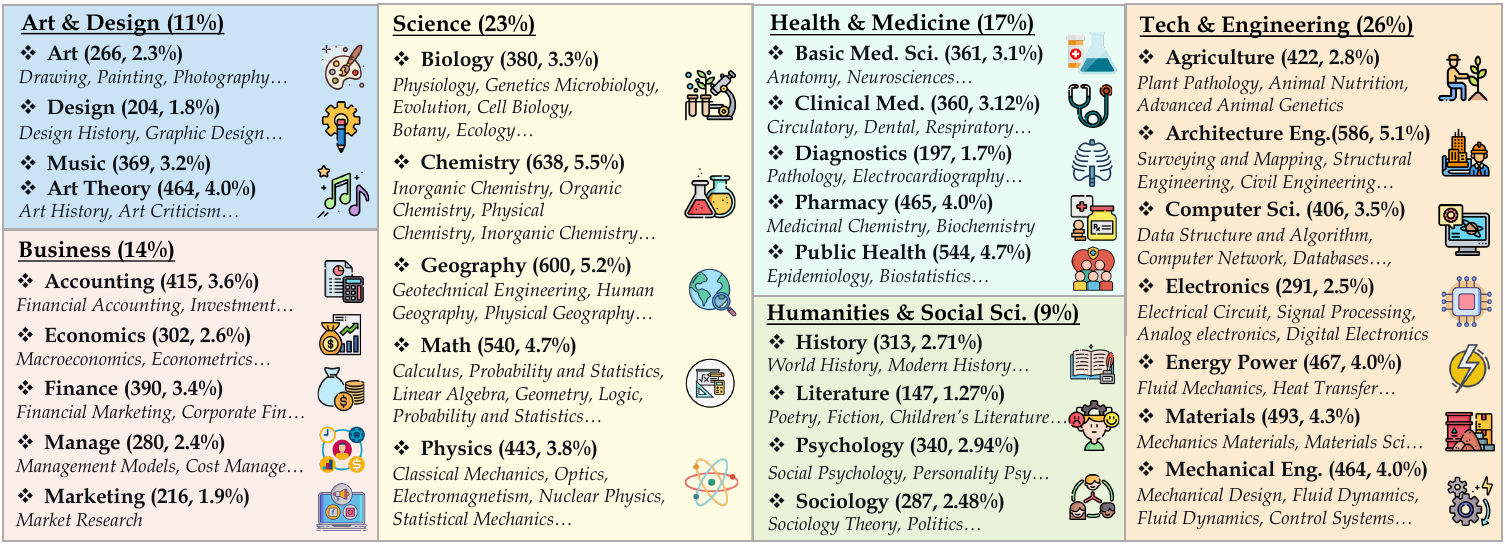}
    \caption{MMMU contains 11.5K multimodal questions covering six broad disciplines, 30 subjects, and 183 subfields. }
    \label{fig:subject_distribution}
\end{figure*}

\noindent \textbf{Perceptual Errors (35\%):} Perceptual errors, forming the bulk of the inaccuracies in the GPT-4V model, are categorized into two types: basic perceptual errors and domain-specific perceptual errors. Basic perceptual errors, as depicted in \autoref{fig:error_case_main_text}, occur when the model accurately processes and understands the given information but fails in elementary visual interpretation, such as misjudging the sequence described as ``from left to right, top to bottom.'' On the other hand, domain-specific perceptual errors occur due to the lack of knowledge. As we analyze the root cause, we classify such errors as lack of knowledge (see analysis below).  Additionally, GPT-4V often exhibits a bias towards text, prioritizing textual information over visual inputs, a trend noted in recent studies~\cite{cui2023holistic}. A prominent example is in \autoref{fig:history_3}, where the model incorrectly prioritizes its text-based interpretation of ``imperialism'' over the visual narrative in a cartoon depicting the United States as a ``Savior.'' This underscores the need for a more balanced approach to multimodal interpretation.

\noindent\textbf{Lack of Knowledge (29\%):} A fundamental root cause of `domain-specific' perceptual errors in the GPT-4V model, as previously discussed, is the lack of specialized knowledge. This deficiency is exemplified in the Computer Science context illustrated in Appendix \autoref{fig:computer_science_2}, where the model identifies visual elements such as double circles but fails to interpret them accurately within the domain-specific context, such as their representation of an 'accept state' in Deterministic Finite Automata. Similarly, a deficit in specialized knowledge can lead to flawed reasoning, as demonstrated in the medical example in Appendix \autoref{fig:clinical_medicine_5}. These instances underscore the necessity of enriching the training datasets of foundation models with a diverse range of domain-specific knowledge to improve their accuracy and general applicability in various specialized fields.

\noindent\textbf{Reasoning Errors (26\%):} Flawed reasoning emerges as another significant cause of errors. In instances where the model correctly interprets text and images and recalls relevant knowledge, it still often fails to apply logical and mathematical reasoning skills effectively to derive accurate inferences. A notable instance of this can be observed in Appendix \autoref{fig:math_4}, where the model neglects an essential step in a mathematical reasoning process, leading to an incorrect conclusion. Enhancing the model's reasoning capability is critical to address these shortcomings.

\noindent\textbf{Other Errors:} The remaining errors include Textual Understanding Error (6\%), Rejection to Answer (3\%), Annotation Error (2\%), and Answer Extraction Error (1\%). These errors are attributed to various factors such as complex text interpretation challenges, limitations in response generation, inaccuracies in data annotation, and issues in extracting precise answers from longer outputs.

\noindent In summary, our error analysis underlines the challenges posed by \dataset and highlights areas for further research in visual perception, knowledge representation, reasoning abilities, and multimodal joint understanding. 1) \textit{Interplay of language and vision}: language can aid in making visual understanding more explainable, while also leading models to hallucinate. 2) \textit{Challenges in grounding}: tasks involving grounding or referring to specific elements within a visual input remain challenging, even for sophisticated models like GPT-4V. 3) \textit{Complex reasoning is still challenging}: models still fail in complex reasoning scenarios involving lengthy reasoning chains or extensive calculations.

\section{Conclusion}
The introduction of \dataset marks a significant step towards evaluating the capabilities of LMMs in the context of Expert AGI. By assessing both basic perceptual skills and complex reasoning abilities across various professional domains, \dataset provides a comprehensive benchmark that aligns with the expectations of skilled adults in these fields.

\dataset, like any benchmark, has limitations despite its comprehensive nature. The manual curation process may carry biases, and the focus on college-level subjects might not be sufficient for testing Expert AGI~\cite{Morris2023-ta}. However, we argue that strong performance on this benchmark should be a necessary criterion for an Expert AGI system. The challenging nature of \dataset is evident from the performance of over 30 models and human experts. To strike a balance between complexity and practicality, \dataset combines multiple-choice questions with concise open-ended questions, enabling the assessment of diverse subjects while addressing the challenges associated with evaluating open-ended responses.

{
    \small

}


\newpage
\maketitlesupplementary

\setcounter{section}{0}
\renewcommand{\thesection}{\Alph{section}}  

\DoToC

\clearpage

\clearpage
\onecolumn
\section{Subject Distribution}

\clearpage
\section{Breakdown Results on Different Subjects}
\phantomsection

In this appendix, we show the main results and breakdown results of different models on each discipline and subject.
\subsection{Main Results}

\begin{table*}[h]
\centering
\small
\begin{adjustbox}{scale=0.65}
\begin{tabular}{@{}lcccccccc@{}}
\toprule
\textbf{} & \textbf{\begin{tabular}[c]{@{}c@{}}Validation \\  Overall\end{tabular}} & \textbf{\begin{tabular}[c]{@{}c@{}}Test \\  Overall\end{tabular} } & \textbf{\begin{tabular}[c]{@{}c@{}}Art \&\\  Design\end{tabular}} & \textbf{Business} & \textbf{Science} & \textbf{\begin{tabular}[c]{@{}c@{}}Health \& \\ Medicine\end{tabular}} & \textbf{\begin{tabular}[c]{@{}c@{}}Human. \&\\  Social Sci.\end{tabular}} & \textbf{\begin{tabular}[c]{@{}c@{}}Tech \&\\  Eng.\end{tabular}} \\
 & (900) & (10,500) & (1,163) & (1,428) & (2,426) & (1,752) & (947) & (2,784) \\ \midrule

\color{Gray} Random Choice & \color{Gray} 22.1 & \color{Gray} 23.9 & \color{Gray} 24.1 & \color{Gray} 24.9 & \color{Gray} 21.6 & \color{Gray} 25.3 & \color{Gray} 22.8 & \color{Gray} 24.8 \\ 
\color{Gray} Frequent Choice & \color{Gray} 26.8 & \color{Gray} 25.8 & \color{Gray} 26.7 & \color{Gray} 28.4 & \color{Gray} 24.0 & \color{Gray} 24.4 & \color{Gray} 25.2 & \color{Gray} 26.5 \\
Expert (Worst)  &76.2 &- &- &- &- &- &- &- \\
Expert (Medium)  &82.6 &- &- &- &- &- &- &- \\
Expert (Best)  &88.6 &- &- &- &- &- &- &- \\
\midrule
\multicolumn{8}{c}{\textbf{Large Multimodal Models (LMMs): Text + Image as Input}} \\ \midrule
OpenFlamingo2-9B \cite{awadalla2023openflamingo} & 28.7 & 26.3 & 31.7 & 23.5 & 26.3 & 26.3 & 27.9 & 25.1 \\
Kosmos2 \cite{peng2023kosmos} & 24.4 & 26.6 & 28.8 & 23.7 & 26.6 & 27.2 & 26.3 & 26.8 \\
Adept Fuyu-8B~\cite{fuyu-8b} & 27.9 & 27.4 & 29.9 & 27.0 & 25.6 & 27.0 & 32.5 & 26.4 \\
MiniGPT4-Vicuna-13B~\cite{zhu2023minigpt} & 26.8 & 27.6 & 30.2 & 27.0 & 26.2 & 26.9 & 30.9 & 27.2 \\
LLaMA-Adapter2-7B~\cite{zhang2023llama} & 29.8 & 27.7 & 35.2 & 25.4 & 25.6 & 30.0 & 29.1 & 25.7 \\
Otter~\cite{li2023otter} & 32.2 & 29.1 & 37.4 & 24.0 & 24.1 & 29.6 & 35.9 & 30.2 \\
CogVLM~\cite{COGVLM} & 32.1 & 30.1 & 38.0 & 25.6 & 25.1 & 31.2 & 41.5 & 28.9 \\
InstructBLIP-T5-XL~\cite{dai2023instructblip} & 32.9 & 30.6 & 43.3 & 25.2 & 25.2 & 29.3 & 45.8 & 28.6 \\
BLIP-2 FLAN-T5-XL \cite{li2023blip} & 34.4 & 31.0 & 43.0 & 25.6 & 25.1 & 31.8 & 48.0 & 27.8 \\
mPLUGw-OWL2* \cite{ye2023mplug2} & 32.7 & 32.1 & 48.5 & 25.6 & 24.9 & 32.8 & 46.7 & 29.6 \\
SPHINX* \cite{lin2023sphinx} & 32.9 & 32.9 & 50.9 & 27.2 & 25.3 & 34.1 & 51.2 & 27.8 \\
Qwen-VL-7B-Chat~\cite{Qwen-VL} & 35.9 & 32.9 & 47.7 & 29.8 & 25.6 & 33.6 & 45.3 & 30.2 \\
Bunny-3B* \cite{Bunny-3B}& 38.2 & 33.0 & 44.3 & 29.5 & 26.8 & 34.5 & 50.5 & 28.7  \\
LLaVA-1.5-13B~\cite{liu2023improved} & 36.4 & 33.6 & 49.8 & 28.2 & 25.9 & 34.9 & 54.7 & 28.3 \\
InstructBLIP-T5-XXL~\cite{dai2023instructblip} & 35.7 & 33.8 & 48.5 & 30.6 & 27.6 & 33.6 & 49.8 & 29.4 \\
BLIP-2 FLAN-T5-XXL~\cite{li2023blip} & 35.4 & 34.0 & 49.2 & 28.6 & 27.3 & 33.7 & 51.5 & 30.4 \\
Emu2-Chat* \cite{sun2023generative} & 36.3 & 34.1 & 50.6 & 27.7 & 28.0 & 32.4 & 50.3 & 31.3 \\
MiniCPM-V-2* \cite{MiniCPM-V-2} & 37.1 & - & - & - & - & - & - & -  \\
MiniCPM-V* \cite{MiniCPM-V} & 37.2 & - & - & - & - & - & - & -  \\
SVIT* \cite{zhao2023svit} & 38.0 & 34.1 & 48.9 & 28.0 & 26.8 & 35.5 & 50.9 & 30.7 \\
InternVL-Chat-V1.1* \cite{chen2023internvl} & 39.1 & 35.3 & 53.7 & 31.7 & 28.2 & 36.5 & 56.4 & 28.0 \\
InfiMM-Zephyr-7B* \cite{InfiMM} & 39.4 & 35.5 & 50.0 & 29.6 & 28.2 & 37.5 & 54.6 & 31.1 \\
Yi-VL-6B* \cite{young2024yi} & 39.1 & 37.8 & 53.4 & 30.3 & 30.0 & 39.3 & 58.5 & 34.1 \\
OmniLMM-12B* \cite{OminiLMM-12B} & 41.1 & - & - & - & - & - & - & -  \\
InternLM-XComposer2-VL* \cite{dong2024internlm} & 43.0 & 38.2 & 56.8 & 32.8 & 30.1 & 39.8 & 60.7 & 31.8 \\ 
HPT Air* \cite{HPT} & 44.0 & - & - & - & - & - & - & -  \\
Yi-VL-34B*  \cite{young2024yi} & 45.9 & 41.6 & 56.1 & 33.3 & 32.9 & 45.9 & 66.5 & 36.0 \\
LLaVA-1.6-34B* \cite{liu2024llava} & 51.1 & 44.7 & 58.6 & {\ul 39.9} & 36.0 & {\ul 51.2} & {\ul 70.2} & 36.3 \\
InternVL-Chat-V1.2* \cite{chen2023internvl} & {\ul 51.6} & {\ul 46.2} & \textbf{62.5} & 37.6 & \textbf{37.9} & 49.7 & 70.1 & \textbf{40.8} \\ 
VILA1.5* \cite{lin2023vila} & \textbf{51.9} & \textbf{46.9} & {\ul 62.1} & \textbf{40.6} & {\ul 37.7} & \textbf{51.7} & \textbf{74.0} & {\ul 39.5} \\ \midrule
Gemini Nano2* \cite{deepmind_gemini_report}  & 32.6 & - & - & - & - & - & - & -  \\
Marco-VL* & 41.2 & 40.4 & 56.5 & 31.0 & 31.0 & 46.9 & 66.5 & 33.8 \\
Reka Edge* \cite{Reka} & 42.8 & - & - & - & - & - & - & -  \\
Qwen-VL-PLUS*~\cite{Qwen-VL-PLUS} & 45.2 & 40.8 & 59.9 & 34.5 & 32.8 & 43.7 & 65.5 & 32.9 \\
Marco-VL-Plus* & 46.2 & 44.3 & 57.4 & 34.7 & 38.5 & 48.7 & 72.2 & 36.7 \\
Gemini 1.0 Pro* \cite{deepmind_gemini_report} & 47.9 & - & - & - & - & - & - & -  \\
Adept Fuyu-Heavy* \cite{Fuyu-Heavy} & 48.3 & - & - & - & - & - & - & - \\
Claude 3 Haiku* \cite{Claude3} & 50.2 & - & - & - & - & - & - & -  \\
Reka Flash* \cite{Reka} & 53.3 & - & - & - & - & - & - & -  \\
Skywork-VL* \cite{Skywork-VL} & 51.4 & 46.2 & 61.4 & 39.6 & 36.6 & 50.8 & 71.6 & 40.2 \\
Qwen-VL-MAX* \cite{Qwen-VL-MAX} & 51.4 & 46.8 & {\ul 64.2} & 39.8 & 36.3 & 52.5 & 70.4 & 40.7 \\
HPT Pro* \cite{HPT} & 52.0 & - & - & - & - & - & - & -  \\
Claude 3 Sonnet* \cite{Claude3} & 53.1 & - & - & - & - & - & - & -  \\
SenseChat-Vision-0423-Preview* \cite{SenseChat-Vision} & 54.6 & {\ul 50.3} & 62.7 & {\ul 44.1} & {\ul 42.3} & {\ul 55.7} & {\ul 74.7} & \textbf{43.5}  \\
Gemini 1.5 Flash* \cite{deepmind_gemini1.5_report} & 56.1 & - & - & - & - & - & - & -  \\
Reka Core* \cite{Reka} & 56.3 & - & - & - & - & - & - & -  \\
GPT-4V(ision) (Playground) \cite{openai2023gpt4v} & 56.8 & \textbf{55.7} & \textbf{65.3} & \textbf{64.3} & \textbf{48.4} & \textbf{63.5} & \textbf{76.3} & {\ul 41.7} \\
Claude 3 Opus* \cite{Claude3} & 59.4 & - & - & - & - & - & - & -  \\
Gemini 1.0 Ultra* \cite{deepmind_gemini_report} & 59.4 & - & - & - & - & - & - & -  \\
Gemini 1.5 Pro* \cite{deepmind_gemini1.5_report} & {\ul 62.2} & - & - & - & - & - & - & -  \\
GPT-4o* \cite{gpt-4o} & \textbf{69.1} & - & - & - & - & - & - & -  \\  \midrule
\multicolumn{8}{c}{\textbf{Large Language Models (LLMs): Only Text as Input}} \\ \midrule
Llama2 7B \cite{touvron2023llama2} & 30.1 & 28.7 & 30.7 & 27.2 & 26.7 & 27.7 & 32.6 & 29.8 \\ 
\addlinespace[0.1em]\hdashline\addlinespace[0.1em]
FLAN-T5-XXL~\cite{chung2022scaling} & 32.1 & 31.2 & 36.8 & \textbf{28.9} & 26.7 & 32.8 & 44.8 & 28.3 \\
\quad  + OCR & 34.7 & \textbf{31.9} & 36.2 & 28.8 & 26.2 & 32.6 & \textbf{50.5} & \textbf{29.7} \\
\quad  + LLaVA Caption & \textbf{34.8} & \textbf{31.9} & \textbf{38.4} & 27.8 & \textbf{27.0} & \textbf{33.2} & 49.9 & 28.7 \\ 
\addlinespace[0.1em]\hdashline\addlinespace[0.1em]
Vicuna-13B~\cite{vicuna2023} & 33.3 & 31.0 & 35.1 & \textbf{30.1} & 24.7 & 31.4 & 44.8 & 30.1 \\
\quad + OCR & \textbf{35.4} & 31.9 & 37.1 & 28.6 & \textbf{26.5} & 32.0 & 49.3 & 30.0 \\
\quad + LLaVA Caption & 33.9 & \textbf{32.7} & \textbf{42.0} & 26.8 & 26.2 & \textbf{33.4} & \textbf{49.4} & \textbf{31.4} \\
\addlinespace[0.1em]\hdashline\addlinespace[0.1em]
GPT-4 Text~\cite{openai2023gpt4} & 34.9 & 33.8 & 32.9 & 28.5 & 30.6 & 41.3 & 53.0 & 28.4 \\ 
\bottomrule
\end{tabular}%
\end{adjustbox}
\caption{Overall results of different models on the \dataset \textbf{validation} and \textbf{test set}. The best-performing model in each category is \textbf{in-bold}, and the second best is {\ul{underlined}}. *: results provided by the authors.}
\label{tab:main_results}
\end{table*}

\subsection{Art \& Design}

\begin{table*}[!b]
\centering
\small
\begin{adjustbox}{scale = 0.8}
\begin{tabular}{@{}lcccccc@{}}
\toprule
\textbf{} & \textbf{Validation Overall} & \textbf{Test Overall} & \textbf{Art} & \textbf{Art Theory} & \textbf{Design} & \textbf{Music} \\
 & (120) & (1,163) & (231) & (429) & (169) & (334)\\ \midrule

\color{Gray} Random Choice & \color{Gray} 29.2 & \color{Gray} 24.1 & \color{Gray} 23.4 & \color{Gray} 20.3 & \color{Gray} 19.5 & \color{Gray} 31.7\\ 
\color{Gray} Frequent Choice & \color{Gray} 23.3 & \color{Gray} 26.7 & \color{Gray} 24.2 & \color{Gray} 23.5 & \color{Gray} 33.7 & \color{Gray} 29.0\\
Expert (Worst)  & 80.8 & - & - & - & - & - \\
Expert (Medium)  & 84.2 & - & - & - & - & - \\
Expert (Best)  & 89.2 & - & - & - & - & - \\
\midrule
\multicolumn{7}{c}{\textbf{Large Multimodal Models (LMMs): Text + Image as Input}} \\ \midrule
OpenFlamingo2-9B \cite{awadalla2023openflamingo} & 40.0 & 31.7 & 36.8 & 28.4 & 27.8 & 34.4 \\
Kosmos2 \cite{peng2023kosmos} & 25.0 & 28.8 & 30.7 & 24.9 & 28.4 & 32.6 \\
Adept Fuyu-8B~\cite{fuyu-8b} & 36.7 & 29.9 & 28.6 & 26.8 & 29.0 & 35.3 \\
MiniGPT4-Vicuna-13B~\cite{zhu2023minigpt} & 29.2 & 30.2 & 28.6 & 28.7 & 40.2 & 28.1 \\
LLaMA-Adapter2-7B~\cite{zhang2023llama} & 29.2 & 35.2 & 38.5 & 35.4 & 41.4 & 29.3 \\
Otter~\cite{li2023otter} & 37.5 & 37.4 & 40.7 & 35.9 & 46.2 & 32.6 \\
CogVLM~\cite{COGVLM} & 40.8 & 38.0 & 43.3 & 39.2 & 44.4 & 29.6 \\
InstructBLIP-T5-XL~\cite{dai2023instructblip} & 40.0 & 43.3 & 49.8 & 45.0 & 52.1 & 32.3 \\
BLIP-2 FLAN-T5-XL \cite{li2023blip} & 44.2 & 43.0 & 50.2 & 45.0 & 47.3 & 33.2 \\
mPLUG-OWL2* \cite{ye2023mplug2} & 45.8 & 48.5 & 57.6 & 53.4 & 59.8 & 30.2 \\
SPHINX* \cite{lin2023sphinx} & 48.3 & 50.9 & 59.3 & 55.5 & 61.5 & 33.8 \\ 
Qwen-VL-7B-Chat~\cite{Qwen-VL} & 51.7 & 47.7 & 57.1 & 49.7 & 58.6 & 33.2 \\
Bunny-3B* \cite{Bunny-3B} & 49.2 & 44.3 & 49.8 & 48.7 & 55.0 & 29.3 \\
LLaVA-1.5-13B~\cite{liu2023improved} & 51.7 & 49.8 & 58.4 & 51.5 & 61.5 & 35.6 \\
InstructBLIP-T5-XXL~\cite{dai2023instructblip} & 44.2 & 48.5 & 51.9 & 52.7 & 60.4 & 34.7 \\
BLIP-2 FLAN-T5-XXL~\cite{li2023blip} & 41.7 & 49.2 & 54.5 & 51.5 & 64.5 & 34.7 \\
Emu2-Chat* \cite{sun2023generative} & 55.0 & 50.6 & 59.3 & 54.1 & 63.3 & 33.8 \\
MiniCPM-V-2* \cite{MiniCPM-V-2} & 63.3 & - & - & - & - & - \\
MiniCPM-V* \cite{MiniCPM-V} & 55.8 & - & - & - & - & - \\
SVIT* \cite{zhao2023svit} & 52.5 & 48.9 & 54.1 & 51.0 & 68.0 & 32.9 \\
InternVL-Chat-V1.1* \cite{chen2023internvl} & 56.7 & 53.7 & 60.6 & 59.0 & 74.6 & 31.4 \\
InfiMM-Zephyr-7B* \cite{InfiMM} & 55.8 & 50.0 & 57.1 & 57.3 & 62.1 & 29.3 \\
Yi-VL-6B* \cite{young2024yi} & 52.5 & 53.4 & 57.6 & 61.8 & 74.0 & 29.3 \\
OmniLMM-12B* \cite{OminiLMM-12B} & 58.3 & - & - & - & - & - \\
InternLM-XComposer2-VL* \cite{dong2024internlm} & 60.0 & 56.8 & 68.0 & 63.6 & 69.2 & 34.1 \\
HPT Air* \cite{HPT} & 56.7 & - & - & - & - & - \\
Yi-VL-34B* \cite{young2024yi} & 59.2 & 56.1 & 60.6 & 61.8 & 72.8 & 37.1 \\
LLaVA-1.6-34B* \cite{liu2024llava} & \textbf{67.5} & 58.6 & 67.5 & {\ul 65.3} & {\ul 80.5} & 32.6 \\
InternVL-Chat-V1.2* \cite{chen2023internvl} & {\ul 62.5} & \textbf{62.5} & {\ul 68.8} & \textbf{69.5} & \textbf{82.2} & \textbf{39.2} \\
VILA1.5* \cite{lin2023vila} & 60.8 & {\ul 62.1} & \textbf{69.3} & \textbf{69.5} & \textbf{82.2} & {\ul 37.4} \\
\midrule 
Marco-VL* & 57.5 & 56.5 & 64.5 & 61.8 & 75.1 & 34.7 \\
Reka-Edge* \cite{Reka} & 52.5 & - & - & - & - & - \\
Qwen-VL-PLUS*~\cite{Qwen-VL-PLUS} & 60.0 & 59.9 & 67.5 & 68.1 & 78.7 & 34.7 \\
Marco-VL-Plus* & 60.8 & 57.4 & 65.4 & 62.5 & 75.1 & 36.5 \\
Adept Fuyu-Heavy* \cite{Fuyu-Heavy} & 53.4 & - & - & - & - & - \\
Reka-Flash* \cite{Reka} & 61.7 & - & - & - & - & - \\
Skywork-VL* \cite{Skywork-VL} & 66.7 & 61.4 & 70.1 & 69.5 & 83.4 & 33.8 \\
Qwen-VL-MAX* \cite{Qwen-VL-MAX} & {\ul 72.5} & {\ul 64.2} & {\ul 72.3} & {\ul 74.8} & {\ul 84.0} & 35.0 \\
HPT Pro* \cite{HPT} & 70.0 & - & - & - & - & - \\
SenseChat-Vision-0423-Preview* \cite{SenseChat-Vision} & 66.7 & 62.7 & 67.5 & 70.4 & \textbf{85.8} & {\ul 37.7} \\
Reka-Core* \cite{Reka} & \textbf{75.9} & - & - & - & - & - \\
GPT-4V(ision) (Playground) \cite{openai2023gpt4v} & 65.8 & \textbf{65.3} & \textbf{74.0} & \textbf{75.5} & 80.5 & \textbf{38.6} \\
Gemini 1.0 Ultra* \cite{deepmind_gemini_report} & 70.0 & - & - & - & - & - \\ \midrule
\multicolumn{7}{c}{\textbf{Large Language Models (LLMs): Only Text as Input}} \\ \midrule
Llama2 7B \cite{touvron2023llama2} & 29.2 & 30.7 & 30.3 & 27.5 & 37.9 & 31.4 \\
\addlinespace[0.1em]\hdashline\addlinespace[0.1em]
FLAN-T5-XXL~\cite{chung2022scaling} & 38.3 & 36.8 & 32.0 & 36.8 & \textbf{52.1} & 32.3 \\
\quad  + OCR & 37.5 & 36.2 & 36.4 & 33.8 & 47.9 & \textbf{33.2} \\
\quad  + LLaVA Caption & \textbf{43.3} & \textbf{38.4} & \textbf{45.9} & \textbf{38.2} & 46.2 & 29.6 \\
\addlinespace[0.1em]\hdashline\addlinespace[0.1em]
Vicuna-13B~\cite{vicuna2023} & \textbf{41.7} & 35.1 & 35.1 & 31.5 & 46.7 & 33.8 \\
\quad + OCR & 39.2 & 37.1 & 35.5 & 32.9 & \textbf{50.3} & \textbf{36.8} \\
\quad + LLaVA Caption & 38.3 & \textbf{42.0} & \textbf{51.1} & \textbf{42.7} & 46.2 & 32.6 \\
\addlinespace[0.1em]\hdashline\addlinespace[0.1em]
GPT-4 Text~\cite{openai2023gpt4} & 35.0 & 32.9 & 35.1 & 28.7 & 47.3 & 29.6 \\ \bottomrule
\end{tabular}%
\end{adjustbox}
\caption{\textbf{Art \& Design} results of different models on the \dataset \textbf{validation} and \textbf{test set}. The best-performing model in each category is \textbf{in-bold}, and the second best is {\ul{underlined}}. *: results provided by the authors.} 
\label{tab:overall_art_design}
\end{table*}
\clearpage
\subsection{Business}

\begin{table*}[!b]
\centering
\small
\begin{adjustbox}{scale = 0.8}
\begin{tabular}{@{}lccccccc@{}}
\toprule
\textbf{} & \textbf{\begin{tabular}[c]{@{}c@{}}Validation \\  Overall\end{tabular}} & \textbf{\begin{tabular}[c]{@{}c@{}}Test \\  Overall\end{tabular} } & \textbf{Accounting} & \textbf{Economics} & \textbf{Finance} & \textbf{Manage} & \textbf{Marketing}\\
 & (150) & (1,428) & (380) & (267) & (355) & (245) & (181)\\ \midrule

\color{Gray} Random Choice & \color{Gray} 24.7 & \color{Gray} 24.9 & \color{Gray} 30.0 & \color{Gray} 29.6 & \color{Gray} 17.7 & \color{Gray} 22.4 & \color{Gray} 24.9\\ 
\color{Gray} Frequent Choice & \color{Gray} 29.3 & \color{Gray} 28.4 & \color{Gray} 33.4 & \color{Gray} 36.3 & \color{Gray} 22.0 & \color{Gray} 15.9 & \color{Gray} 35.9\\
Expert (Worst)  & 78.0 & - & - & - & - & - & - \\
Expert (Medium)  & 86.0 & - & - & - & - & - & - \\
Expert (Best)  & 90.7 & - & - & - & - & - & - \\
\midrule
\multicolumn{8}{c}{\textbf{Large Multimodal Models (LMMs): Text + Image as Input}} \\ \midrule
OpenFlamingo2-9B \cite{awadalla2023openflamingo} & 28.0 & 23.5 & 24.7 & 25.8 & 19.4 & 25.3 & 22.7 \\
Kosmos2 \cite{peng2023kosmos} & 18.0 & 23.7 & 29.7 & 24.0 & 21.4 & 22.4 & 17.1 \\
Fuyu-8B~\cite{fuyu-8b} & 32.0 & 27.0 & 32.1 & 30.3 & 22.5 & 20.0 & 29.3 \\
MiniGPT4-Vicuna-13B~\cite{zhu2023minigpt} & 21.3 & 27.0 & 29.7 & 34.1 & 25.6 & 16.7 & 27.6 \\
LLaMA-Adapter2-7B~\cite{zhang2023llama} & 25.3 & 25.4 & 30.8 & 24.7 & 20.6 & 24.9 & 25.4 \\
Otter~\cite{li2023otter} & 24.0 & 24.0 & 30.8 & 29.6 & 17.5 & 16.3 & 24.9 \\
CogVLM~\cite{COGVLM} & 25.3 & 25.6 & 28.2 & 29.6 & 19.2 & 21.2 & 32.6 \\
InstructBLIP-T5-XL~\cite{dai2023instructblip} & 28.0 & 25.2 & 27.6 & 31.8 & 18.0 & 22.0 & 28.7 \\
BLIP-2 FLAN-T5-XL \cite{li2023blip} & 26.7 & 25.6 & 28.2 & 31.1 & 17.5 & 24.1 & 29.8 \\
mPLUG-OWL2* \cite{ye2023mplug2} & 24.7 & 25.6 & 28.7 & 29.2 & 20.3 & 22.4 & 28.7 \\
SPHINX* \cite{lin2023sphinx} & 24.7 & 27.2 & 25.8 & 31.8 & 22.5 & 25.7 & 34.8  \\
Qwen-VL-7B-Chat~\cite{Qwen-VL} & 29.3 & 29.8 & 34.2 & 29.6 & 18.9 & 32.2 & 38.7 \\
Bunny-3B* \cite{Bunny-3B} & 30.7 & 29.5 & 30.5 & 33.3 & 24.2 & 25.3 & 37.6 \\
LLaVA-1.5-13B~\cite{liu2023improved} & 22.7 & 28.2 & 29.2 & 33.3 & 23.7 & 23.3 & 34.3 \\
InstructBLIP-T5-XXL~\cite{dai2023instructblip} & 24.0 & 30.6 & 34.2 & 35.6 & 23.4 & 30.2 & 30.4 \\
BLIP-2 FLAN-T5-XXL~\cite{li2023blip} & 30.0 & 28.6 & 32.4 & 33.3 & 22.5 & 26.1 & 28.7 \\
Emu2-Chat* \cite{sun2023generative} & 30.0 & 27.7 & 29.2 & 34.1 & 20.0 & 27.3 & 30.4 \\
MiniCPM-V-2* \cite{MiniCPM-V-2} & 28.7 & - & - & - & - & - & - \\
MiniCPM-V* \cite{MiniCPM-V} & 33.3 & - & - & - & - & - & - \\
SVIT* \cite{zhao2023svit} & 27.3 & 28.0 & 28.7 & 35.6 & 22.3 & 22.9 & 33.7 \\
InternVL-Chat-V1.1* \cite{chen2023internvl} & 34.7 & 31.7 & 34.5 & 34.5 & 24.5 & 29.0 & 39.2 \\
InfiMM-Zephyr-7B* \cite{InfiMM} & 28.0 & 29.6 & 31.8 & 37.1 & 23.7 & 21.6 & 35.9 \\
Yi-VL-6B* \cite{young2024yi} & 30.7 & 30.3 & 33.7 & 36.3 & 19.4 & 26.9 & 39.8 \\
OmniLMM-12B* \cite{OminiLMM-12B} & 34.0 & - & - & - & - & - & - \\
InternLM-XComposer2-VL* \cite{dong2024internlm} & 34.0 & 32.8 & 35.3 & 31.5 & 25.6 & 31.8 & 45.3 \\
HPT Air* \cite{HPT} & 31.3 & - & - & - & - & - & - \\
Yi-VL-34B* \cite{young2024yi} & 36.0 & 33.3 & 33.2 & 44.9 & 22.0 & 29.4 & 43.6 \\
LLaVA-1.6-34B* \cite{liu2024llava} & \textbf{46.0} & {\ul 39.9} & {\ul 41.3} & {\ul 45.3} & \textbf{32.4} & {\ul 35.9} & \textbf{49.2} \\
InternVL-Chat-V1.2* \cite{chen2023internvl} & 40.7 & 37.6 & 38.2 & 43.4 & {\ul 27.3} & \textbf{37.6} & {\ul 48.1} \\
VILA1.5* \cite{lin2023vila} & {\ul 43.3} & {\ul 40.6} & \textbf{41.8} & \textbf{48.3} & \textbf{32.4} & {\ul 35.9} & \textbf{49.2} \\
\midrule 
Marco-VL* & 30.0 & 31.0 & 28.2 & 37.1 & 22.8 & 31.4 & 43.1 \\
Reka-Edge*~\cite{Reka} & 36.0 & - & - & - & - & - & - \\
Qwen-VL-PLUS*~\cite{Qwen-VL-PLUS} & 35.3 & 34.5 & 35.0 & 41.6 & 26.2 & 34.7 & 39.2 \\
Marco-VL-Plus* & 37.3 & 34.7 & 36.6 & 38.6 & 27.6 & 31.4 & 43.1 \\
Adept Fuyu-Heavy* \cite{Fuyu-Heavy} & 46.3 & - & - & - & - & - & - \\
Reka-Flash*~\cite{Reka} & 42.7 & - & - & - & - & - & - \\
Skywork-VL*~\cite{Skywork-VL} & 41.3 & 39.6 & 39.7 & 47.2 & 31.0 & 36.7 & 48.6 \\
Qwen-VL-MAX* \cite{Qwen-VL-MAX} & 43.3 & 39.8 & 37.9 & 44.9 & 33.0 & {\ul 39.6} & 50.3 \\
HPT Pro* \cite{HPT} & 43.3 & - & - & - & - & - & - \\
SenseChat-Vision-0423-Preview*~\cite{SenseChat-Vision} & 54.0 & {\ul 44.1} & {\ul 45.0} & {\ul 51.7} & {\ul 38.3} & 37.6 & {\ul 51.4} \\
Reka-Core*~\cite{Reka} & 47.3 & - & - & - & - & - & - \\
GPT-4V(ision) (Playground) \cite{openai2023gpt4v} & \textbf{59.3} & \textbf{64.3} & \textbf{69.7} & \textbf{70.8} & \textbf{61.1} & \textbf{51.0} & \textbf{67.4} \\  
Gemini 1.0 Ultra* \cite{deepmind_gemini_report} & {\ul 56.7} & - & - & - & - & - & -  \\
\midrule
\multicolumn{8}{c}{\textbf{Large Language Models (LLMs): Only Text as Input}} \\ \midrule
Llama2 7B \cite{touvron2023llama2} & 22.7 & 27.2 & 28.9 & 34.1 & 23.7 & 21.6 & 27.6 \\
\addlinespace[0.1em]\hdashline\addlinespace[0.1em]
FLAN-T5-XXL~\cite{chung2022scaling} & 28.0 & \textbf{28.9} & 31.6 & \textbf{31.5} & 23.1 & \textbf{29.0} & \textbf{30.4} \\
\quad  + OCR & 29.3 & 28.8 & \textbf{32.4} & 30.0 & \textbf{24.8} & 26.9 & 29.8 \\
\quad  + LLaVA Caption & \textbf{31.3} & 27.8 & 28.2 & 30.7 & 24.2 & 27.8 & 29.8 \\
\addlinespace[0.1em]\hdashline\addlinespace[0.1em]
Vicuna-13B~\cite{vicuna2023} & 26.7 & \textbf{30.1} & \textbf{29.5} & \textbf{34.8} & 25.6 & \textbf{30.6} & \textbf{32.6} \\
\quad + OCR & \textbf{31.3} & 28.6 & 27.1 & 34.1 & 23.9 & \textbf{30.6} & 30.4 \\
\quad + LLaVA Caption & 26.0 & 26.8 & 27.1 & 32.6 & 22.3 & 25.3 & 28.7 \\
\addlinespace[0.1em]\hdashline\addlinespace[0.1em]
GPT-4 Text~\cite{openai2023gpt4} & 36.7 & 28.5 & 29.7 & 35.2 & 21.1 & 32.2 & 25.4 \\ \bottomrule
\end{tabular}%
\end{adjustbox}
\caption{\textbf{Business} results of different models on the \dataset \textbf{validation} and \textbf{test set}.  The best-performing model in each category is \textbf{in-bold}, and the second best is {\ul{underlined}}. *: results provided by the authors.} 
\label{tab:overall_Business_results}
\end{table*}

\clearpage
\subsection{Science}

\begin{table*}[!b]
\centering
\small
\begin{adjustbox}{scale = 0.8}
\begin{tabular}{@{}lccccccc@{}}
\toprule
\textbf{} & \textbf{\begin{tabular}[c]{@{}c@{}}Validation \\  Overall\end{tabular}} & \textbf{\begin{tabular}[c]{@{}c@{}}Test \\  Overall\end{tabular} } & \textbf{Biology} & \textbf{Chemistry} & \textbf{Geography} & \textbf{Math} & \textbf{Physics}\\
 & (150) & (2,426) & (345) & (603) & (565) & (505) & (408)\\ \midrule

\color{Gray} Random Choice & \color{Gray} 18.0 & \color{Gray} 21.6 & \color{Gray} 18.3 & \color{Gray} 18.6 & \color{Gray} 26.0 & \color{Gray} 22.2 & \color{Gray} 22.1\\ 
\color{Gray} Frequent Choice & \color{Gray} 27.3 & \color{Gray} 24.0 & \color{Gray} 25.8 & \color{Gray} 19.9 & \color{Gray} 26.9 & \color{Gray} 26.1 & \color{Gray} 22.1\\
Expert (Worst)  & 78.0 & - & - & - & - & - & - \\
Expert (Medium)  & 84.7 & - & - & - & - & - & - \\
Expert (Best)  & 90.0 & - & - & - & - & - & - \\
\midrule
\multicolumn{8}{c}{\textbf{Large Multimodal Models (LMMs): Text + Image as Input}} \\ \midrule
OpenFlamingo2-9B \cite{awadalla2023openflamingo} & 23.3 & 26.3 & 27.8 & 22.9 & 30.8 & 25.1 & 25.0 \\
Kosmos2 \cite{peng2023kosmos} & 19.3 & 26.6 & 28.4 & 21.7 & 29.2 & 26.7 & 28.4 \\
Fuyu-8B~\cite{fuyu-8b} & 22.0 & 25.6 & 27.8 & 20.9 & 30.1 & 24.8 & 25.7 \\
MiniGPT4-Vicuna-13B~\cite{zhu2023minigpt} & 28.7 & 26.2 & 23.2 & 22.1 & 29.4 & 30.1 & 25.5 \\
LLaMA-Adapter2-7B~\cite{zhang2023llama} & 30.7 & 25.6 & 27.5 & 24.9 & 30.4 & 23.0 & 21.3 \\
Otter~\cite{li2023otter} & 34.7 & 24.1 & 24.6 & 23.4 & 27.1 & 23.0 & 21.8 \\
CogVLM~\cite{COGVLM} & 28.0 & 25.1 & 29.3 & 24.2 & 28.0 & 23.4 & 21.1 \\
InstructBLIP-T5-XL~\cite{dai2023instructblip} & 32.7 & 25.2 & 27.0 & 22.1 & 28.3 & 24.4 & 25.0 \\
BLIP-2 FLAN-T5-XL \cite{li2023blip} & 30.7 & 25.1 & 26.7 & 24.4 & 25.7 & 24.0 & 25.2 \\
mPLUG-OWL2* \cite{ye2023mplug2} & 22.7 & 24.9 & 27.2 & 23.9 & 29.7 & 18.8 & 25.2 \\
SPHINX* \cite{lin2023sphinx} & 26.7 & 25.3 & 29.0 & 20.1 & 32.6 & 23.8 & 21.8  \\
Qwen-VL-7B-Chat~\cite{Qwen-VL} & 29.3 & 25.6 & 27.8 & 23.1 & 28.8 & 24.6 & 24.3 \\
Bunny-3B* \cite{Bunny-3B} & 30.7 & 26.8 & 32.8 & 25.2 & 27.8 & 26.5 & 22.8 \\
LLaVA-1.5-13B~\cite{liu2023improved} & 29.3 & 25.9 & 27.2 & 25.0 & 28.8 & 24.0 & 24.5 \\
InstructBLIP-T5-XXL~\cite{dai2023instructblip} & 30.7 & 27.6 & 29.0 & 26.5 & 31.0 & 25.5 & 26.0 \\
BLIP-2 FLAN-T5-XXL~\cite{li2023blip} & 34.7 & 27.3 & 28.4 & 25.5 & 29.4 & 27.5 & 26.0 \\
Emu2-Chat* \cite{sun2023generative} & 28.7 & 28.0 & 28.7 & 23.5 & 35.4 & 25.3 & 27.0 \\
MiniCPM-V-2* \cite{MiniCPM-V-2} & 30.0 & - & - & - & - & - & - \\
MiniCPM-V* \cite{MiniCPM-V} & 28.0 & - & - & - & - & - & - \\
SVIT* \cite{zhao2023svit} & 28.0 & 26.8 & 27.0 & 27.2 & 29.2 & 23.4 & 26.7 \\
InternVL-Chat-V1.1* \cite{chen2023internvl} & 31.3 & 28.2 & 35.7 & 24.4 & 31.2 & 27.5 & 24.0 \\
InfiMM-Zephyr-7B* \cite{InfiMM} & 33.3 & 28.2 & 33.0 & 24.2 & 31.7 & 27.3 & 26.0 \\
Yi-VL-6B* \cite{young2024yi} & 31.3 & 30.0 & 32.2 & 25.0 & 34.9 & 29.9 & 28.7\\
OmniLMM-12B* \cite{OminiLMM-12B} & 27.3 & - & - & - & - & - & - \\
InternLM-XComposer2-VL* \cite{dong2024internlm} & {\ul 34.7} & 30.1 & 34.8 & 26.0 & 34.7 & 29.7 & 26.2 \\
HPT Air* \cite{HPT} & 34.7 & - & - & - & - & - & - \\
Yi-VL-34B* & 33.3 & 32.9 & 36.8 & 26.9 & 37.0 & 31.7 & 34.3 \\
LLaVA-1.6-34B* \cite{liu2024llava} & \textbf{39.3} & 36.0 & {\ul 41.4} & 29.5 & {\ul 42.7} & 33.1 & {\ul 35.3} \\
InternVL-Chat-V1.2* \cite{chen2023internvl} & \textbf{39.3} & \textbf{37.9} & 40.3 & {\ul 32.0} & \textbf{44.6} & {\ul 36.6} & \textbf{36.8} \\
VILA1.5* \cite{lin2023vila} & {\ul 36.0} & {\ul 37.7} & \textbf{44.6} & \textbf{32.5} & 42.1 & \textbf{36.8} & 34.3 \\
\midrule 
Marco-VL* & 28.0 & 31.0 & 35.9 & 28.0 & 35.8 & 26.3 & 30.6 \\
Reka-Edge* \cite{Reka} & 42.7 & - & - & - & - & - & - \\
Skywork-VL* \cite{Skywork-VL} & 38.7 & 36.6 & 41.4 & 29.0 & 42.8 & 35.8 & 36.0 \\
Qwen-VL-PLUS*~\cite{Qwen-VL-PLUS} & 37.3 & 32.8 & 40.3 & 27.9 & 34.7 & 31.3 & 33.1 \\
Marco-VL-Plus* & 35.3 & 38.5 & 40.3 & 32.5 & {\ul 46.0} & 36.8 & 37.5 \\
Adept Fuyu-Heavy* \cite{Fuyu-Heavy} & 33.7 & - & - & - & - & - & - \\
Reka-Flash* \cite{Reka} & 47.3 & - & - & - & - & - & - \\
Qwen-VL-MAX* \cite{Qwen-VL-MAX} & 40.0 & 36.3 & 38.0 & 33.3 & 39.6 & 33.7 & 38.0 \\
HPT Pro* \cite{HPT} & 42.7 & - & - & - & - & - & - \\
SenseChat-Vision-0423-Preview* \cite{SenseChat-Vision} & 45.3 & {\ul 42.3} & {\ul 44.3} & {\ul 37.0} & \textbf{47.4} & {\ul 41.8} & {\ul 41.7} \\
Reka-Core* \cite{Reka} & {\ul 49.3} & - & - & - & - & - & - \\
GPT-4V(ision) (Playground) \cite{openai2023gpt4v} & \textbf{54.7} & \textbf{48.4} & \textbf{52.2} & \textbf{46.9} & 44.8 & \textbf{45.0} & \textbf{56.4} \\  
Gemini 1.0 Ultra* \cite{deepmind_gemini_report} & 48.0 & - & - & - & - & - & -  \\
\midrule
\multicolumn{8}{c}{\textbf{Large Language Models (LLMs): Only Text as Input}} \\ \midrule
Llama2 7B \cite{touvron2023llama2} & 34.0 & 26.7 & 28.4 & 21.4 & 29.7 & 28.5 & 26.7 \\
\addlinespace[0.1em]\hdashline\addlinespace[0.1em]
FLAN-T5-XXL~\cite{chung2022scaling} & 28.0 & 26.7 & \textbf{27.8} & 24.4 & 27.3 & \textbf{30.7} & 23.5 \\
\quad  + OCR & 30.0 & 26.2 & 24.6 & \textbf{24.5} & 27.4 & 27.9 & \textbf{26.0} \\
\quad  + LLaVA Caption & \textbf{32.7} & \textbf{27.0} & 25.8 & 23.9 & \textbf{30.3} & 29.1 & 25.5 \\
\addlinespace[0.1em]\hdashline\addlinespace[0.1em]
Vicuna-13B~\cite{vicuna2023} & 23.3 & 24.7 & 24.6 & 22.7 & 25.0 & 26.1 & \textbf{25.7} \\
\quad + OCR & \textbf{30.0} & \textbf{26.5} & 26.4 & \textbf{24.7} & \textbf{29.0} & \textbf{27.1} & 25.2 \\
\quad + LLaVA Caption & 28.7 & 26.2 & \textbf{31.3} & 21.7 & 28.7 & 26.7 & 24.3 \\
\addlinespace[0.1em]\hdashline\addlinespace[0.1em]
GPT-4 Text~\cite{openai2023gpt4} & 34.7 & 30.6 & 29.3 & 28.0 & 34.0 & 27.7 & 34.3 \\ \bottomrule
\end{tabular}%
\end{adjustbox}
\caption{\textbf{Science} results of different models on the \dataset \textbf{validation} and \textbf{test set}.  The best-performing model in each category is \textbf{in-bold}, and the second best is {\ul{underlined}}. *: results provided by the authors.} 
\label{tab:overall_Science_results}
\end{table*}

\clearpage
\subsection{Health \& Medicine}

\begin{table*}[!b]
\centering
\small
\begin{adjustbox}{scale = 0.8}
\begin{tabular}{@{}lccccccc@{}}
\toprule
\textbf{} & \textbf{\begin{tabular}[c]{@{}c@{}}Validation \\  Overall\end{tabular}} & \textbf{\begin{tabular}[c]{@{}c@{}}Test \\  Overall\end{tabular} } & \textbf{\begin{tabular}[c]{@{}c@{}}Basic Medical \\  Science\end{tabular} } & \textbf{\begin{tabular}[c]{@{}c@{}}Clinical\\  Meicine\end{tabular}} & \textbf{\begin{tabular}[c]{@{}c@{}}Diagnostics \& \\  Lab. Medicine\end{tabular} } & \textbf{Pharmacy} & \textbf{\begin{tabular}[c]{@{}c@{}}Public\\  Health\end{tabular}}\\
 & (150) & (1,752) & (326) & (325) & (162) & (430) & (509)\\ \midrule

\color{Gray} Random Choice & \color{Gray} 20.7 & \color{Gray} 25.3 & \color{Gray} 24.8 & \color{Gray} 21.8 & \color{Gray} 25.9 & \color{Gray} 28.6 & \color{Gray} 24.8\\ 
\color{Gray} Frequent Choice & \color{Gray} 30.0 & \color{Gray} 24.4 & \color{Gray} 22.1 & \color{Gray} 24.3 & \color{Gray} 17.3 & \color{Gray} 23.3 & \color{Gray} 29.3\\
Expert (Worst)  & 73.3 & - & - & - & - & - & - \\
Expert (Medium)  & 78.8 & - & - & - & - & - & - \\
Expert (Best)  & 87.3 & - & - & - & - & - & - \\
\midrule
\multicolumn{8}{c}{\textbf{Large Multimodal Models (LMMs): Text + Image as Input}} \\ \midrule
OpenFlamingo2-9B \cite{awadalla2023openflamingo} & 27.3 & 26.3 & 29.1 & 21.8 & 22.2 & 32.1 & 23.8 \\
Kosmos2 \cite{peng2023kosmos} & 28.0 & 27.2 & 27.3 & 24.0 & 27.2 & 30.7 & 26.1 \\
Fuyu-8B~\cite{fuyu-8b} & 28.0 & 27.0 & 28.8 & 23.1 & 24.1 & 27.0 & 29.3 \\
MiniGPT4-Vicuna-13B~\cite{zhu2023minigpt} & 30.7 & 26.9 & 27.0 & 26.2 & 21.6 & 27.7 & 28.5 \\
LLaMA-Adapter2-7B~\cite{zhang2023llama} & 30.7 & 30.0 & 31.0 & 30.2 & 26.5 & 36.5 & 25.0 \\
Otter~\cite{li2023otter} & 30.7 & 29.6 & 34.4 & 28.3 & 28.4 & 28.6 & 28.5 \\
CogVLM~\cite{COGVLM} & 32.0 & 31.2 & 33.4 & 27.4 & 27.2 & 33.7 & 31.4 \\
InstructBLIP-T5-XL~\cite{dai2023instructblip} & 28.7 & 29.3 & 31.3 & 28.9 & 22.8 & 34.2 & 26.1 \\
BLIP-2 FLAN-T5-XL \cite{li2023blip} & 35.3 & 31.8 & 35.9 & 31.7 & 24.1 & 35.8 & 28.5 \\
mPLUG-OWL2* \cite{ye2023mplug2} & 32.0 & 32.8 & 29.9 & 32.3 & 34.0 & 31.2 & 29.7 \\
SPHINX* \cite{lin2023sphinx} & 30.7 & 34.1 & 39.9 & 36.0 & 33.3 & 31.4 & 31.8  \\
Qwen-VL-7B-Chat~\cite{Qwen-VL} & 33.3 & 33.6 & 38.0 & 34.8 & 32.1 & 29.5 & 33.8 \\
Bunny-3B* \cite{Bunny-3B} & 40.7 & 34.5 & 39.6 & 38.5 & 33.3 & 31.4 & 31.6 \\
LLaVA-1.5-13B~\cite{liu2023improved} & 38.7 & 34.9 & 42.6 & 36.6 & 34.6 & 32.1 & 31.4 \\
InstructBLIP-T5-XXL~\cite{dai2023instructblip} & 35.3 & 33.6 & 35.6 & 32.3 & 29.6 & 34.2 & 33.8 \\
BLIP-2 FLAN-T5-XXL~\cite{li2023blip} & 32.0 & 33.7 & 38.7 & 34.5 & 27.2 & 33.7 & 32.2 \\
Emu2-Chat* \cite{sun2023generative} & 28.7 & 32.4 & 39.3 & 34.8 & 29.6 & 33.0 & 26.9 \\
MiniCPM-V-2* \cite{MiniCPM-V-2} & 30.0 & - & - & - & - & - & - \\
MiniCPM-V* \cite{MiniCPM-V} & 32.7 & - & - & - & - & - & - \\
SVIT* \cite{zhao2023svit} & 42.0 & 35.5 & 43.3 & 36.0 & 36.4 & 34.4 & 30.8 \\
InternVL-Chat-V1.1* \cite{chen2023internvl} & 39.3 & 36.5 & 43.6 & 39.7 & 36.4 & 30.9 & 34.6 \\
InfiMM-Zephyr-7B* \cite{InfiMM} & 42.7 & 37.5 & 44.5 & 43.1 & 37.7 & 32.6 & 33.6 \\
Yi-VL-6B* \cite{young2024yi} & 38.0 & 39.3 & 43.6 & 45.8 & 37.7 & 38.6 & 33.4 \\
OmniLMM-12B* \cite{OminiLMM-12B} & 44.0 & - & - & - & - & - & - \\
InternLM-XComposer2-VL* \cite{dong2024internlm} & 46.0 & 39.8 & 45.1 & 42.2 & 34.0 & 42.1 & 34.8 \\
HPT Air* \cite{HPT} & 45.3 & - & - & - & - & - & - \\
Yi-VL-34B* \cite{young2024yi}& 51.3 & 45.9 & 54.6 & 48.9 & \textbf{50.0} & 44.9 & 38.1 \\
LLaVA-1.6-34B* \cite{liu2024llava} & 52.0 & {\ul 51.2} & 56.4 & \textbf{58.8} & {\ul 45.1} & {\ul 50.7} & {\ul 45.4} \\
InternVL-Chat-V1.2* \cite{chen2023internvl} & \textbf{58.7} & 49.7 & \textbf{58.9} & 54.5 & 43.8 & 50.0 & 42.2 \\
VILA1.5* \cite{lin2023vila} & {\ul 57.3} & \textbf{51.7} & {\ul 58.6} & {\ul 55.4} & 40.7 & \textbf{53.3} & \textbf{47.2} \\
\midrule
Marco-VL* & 45.3 & 46.9 & 51.2 & 50.2 & 42.0 & 50.7 & 40.5 \\
Reka-Edge* \cite{Reka} & 41.3 & - & - & - & - & - & - \\
Qwen-VL-PLUS*~\cite{Qwen-VL-PLUS} & 46.7 & 43.7 & 49.7 & 42.2 & 34.0 & 46.5 & 41.5 \\
Marco-VL-Plus* & 48.7 & 48.7 & 57.4 & 53.5 & 40.7 & 46.5 & 44.6 \\
Adept Fuyu-Heavy* \cite{Fuyu-Heavy} & 51.3 & - & - & - & - & - & - \\
Reka-Flash* \cite{Reka} & 59.3 & - & - & - & - & - & - \\
Skywork-VL* \cite{Skywork-VL} & 55.3 & 50.8 & 58.9 & 55.1 & {\ul 48.8} & 50.9 & 43.4 \\
Qwen-VL-MAX* \cite{Qwen-VL-MAX} & 58.0 & 52.5 & 58.9 & 51.1 & 44.4 & {\ul 57.4} & 47.7 \\
HPT Pro* \cite{HPT} & 50.7 & - & - & - & - & - & - \\
SenseChat-Vision-0423-Preview* \cite{SenseChat-Vision} & 53.3 & {\ul 55.7} & {\ul 62.6} & {\ul 58.2} & \textbf{50.0} & 55.6 & {\ul 51.5} \\
Reka-Core* \cite{Reka} & 58.0 & - & - & - & - & - & - \\
GPT-4V(ision) (Playground) \cite{openai2023gpt4v} & {\ul 64.7} & \textbf{63.5} & \textbf{65.0} & \textbf{62.5} & 43.8 & \textbf{68.1} & \textbf{65.4} \\ 
Gemini 1.0 Ultra* \cite{deepmind_gemini_report} & \textbf{67.3} & - & - & - & - & - & -  \\ \midrule
\multicolumn{8}{c}{\textbf{Large Language Models (LLMs): Only Text as Input}} \\ \midrule
Llama2 7B \cite{touvron2023llama2} & 26.7 & 27.7 & 26.1 & 30.8 & 25.3 & 27.7 & 27.7 \\
\addlinespace[0.1em]\hdashline\addlinespace[0.1em]
FLAN-T5-XXL~\cite{chung2022scaling} & 32.0 & 32.8 & 33.7 & 34.8 & 30.2 & \textbf{34.4} & 30.5 \\
\quad  + OCR & \textbf{32.7} & 32.6 & 33.7 & \textbf{35.1} & 27.8 & 32.3 & 32.2 \\
\quad  + LLaVA Caption & 32.0 & \textbf{33.2} & \textbf{35.3} & 34.2 & \textbf{30.9} & 32.6 & \textbf{32.4} \\
\addlinespace[0.1em]\hdashline\addlinespace[0.1em]
Vicuna-13B~\cite{vicuna2023} & 31.3 & 31.4 & 37.7 & 33.2 & 36.4 & 27.7 & 27.9 \\
\quad + OCR & 31.3 & 32.0 & \textbf{38.3} & 33.5 & \textbf{37.0} & 28.4 & 28.5 \\
\quad + LLaVA Caption & \textbf{34.0} & \textbf{33.4} & 37.1 & \textbf{35.4} & 32.7 & \textbf{32.6} & \textbf{30.6} \\
\addlinespace[0.1em]\hdashline\addlinespace[0.1em]
GPT-4 Text~\cite{openai2023gpt4} & 40.7 & 41.3 & 52.5 & 52.9 & 27.8 & 39.1 & 33.0 \\ \bottomrule
\end{tabular}%
\end{adjustbox}
\caption{\textbf{Health \& Medicine} results of different models on the \dataset \textbf{validation} and \textbf{test set}.  The best-performing model in each category is \textbf{in-bold}, and the second best is {\ul{underlined}}. *: results provided by the authors.} 
\label{tab:overall_Health_results}
\end{table*}

\clearpage
\subsection{Humanities \& Social Science}

\begin{table*}[!b]
\centering
\small
\begin{adjustbox}{scale = 0.8}
\begin{tabular}{@{}lcccccc@{}}
\toprule
\textbf{} & \textbf{\begin{tabular}[c]{@{}c@{}}Validation \\  Overall\end{tabular}} & \textbf{\begin{tabular}[c]{@{}c@{}}Test \\  Overall\end{tabular} } & \textbf{History} & \textbf{Literature} & \textbf{Sociology} & \textbf{Psychology} \\
 & (120) & (947) & (278) & (112) & (252) & (305)\\ \midrule

\color{Gray} Random Choice & \color{Gray} 20.0 & \color{Gray} 22.8 & \color{Gray} 22.3 & \color{Gray} 24.1 & \color{Gray} 27.0 & \color{Gray} 19.3\\ 
\color{Gray} Frequent Choice & \color{Gray} 25.8 & \color{Gray} 25.2 & \color{Gray} 27.0 & \color{Gray} 27.7 & \color{Gray} 25.4 & \color{Gray} 22.6\\
Expert (Worst)  & 74.2 & - & - & - & - & - \\
Expert (Medium)  & 85.0 & - & - & - & - & - \\
Expert (Best)  & 89.2 & - & - & - & - & - \\
\midrule
\multicolumn{7}{c}{\textbf{Large Multimodal Models (LMMs): Text + Image as Input}} \\ \midrule
OpenFlamingo2-9B \cite{awadalla2023openflamingo} & 30.8 & 27.9 & 24.5 & 42.0 & 29.0 & 24.9 \\
Kosmos2 \cite{peng2023kosmos} & 30.0 & 26.3 & 24.5 & 24.1 & 34.1 & 22.3 \\
Fuyu-8B~\cite{fuyu-8b} & 32.5 & 32.5 & 32.7 & 44.6 & 32.9 & 27.5 \\
MiniGPT4-Vicuna-13B~\cite{zhu2023minigpt} & 29.2 & 30.9 & 30.9 & 47.3 & 30.6 & 25.2 \\
LLaMA-Adapter2-7B~\cite{zhang2023llama} & 33.3 & 29.1 & 27.0 & 43.8 & 32.1 & 23.3 \\
Otter~\cite{li2023otter}& 41.7 & 35.9 & 33.8 & 67.0 & 34.9 & 27.2 \\
CogVLM~\cite{COGVLM} & 45.0 & 41.5 & 39.2 & 69.6 & 41.3 & 33.4 \\
InstructBLIP-T5-XL~\cite{dai2023instructblip} & 47.5 & 45.8 & 45.0 & 71.4 & 44.8 & 38.0 \\
BLIP-2 FLAN-T5-XL \cite{li2023blip} & 50.0 & 48.0 & 48.2 & 76.8 & 47.2 & 38.0 \\
mPLUG-OWL2* \cite{ye2023mplug2} & 45.8 & 46.7 & 46.0 & 74.1 & 44.4 & 39.0 \\
SPHINX* \cite{lin2023sphinx} & 50.0 & 51.2 & 56.5 & 81.2 & 48.0 & 38.0 \\
Qwen-VL-7B-Chat~\cite{Qwen-VL} & 45.0 & 45.3 & 47.8 & 64.3 & 46.4 & 35.1 \\
Bunny-3B* \cite{Bunny-3B} & 45.0 & 50.5 & 52.2 & 78.6 & 50.0 & 39.0 \\
LLaVA-1.5-13B~\cite{liu2023improved} & 53.3 & 54.7 & 58.6 & 76.8 & 51.2 & 45.9 \\
InstructBLIP-T5-XXL~\cite{dai2023instructblip} & 49.2 & 49.8 & 48.6 & 72.3 & 51.2 & 41.6 \\
BLIP-2 FLAN-T5-XXL~\cite{li2023blip} & 50.8 & 51.5 & 49.6 & 75.9 & 53.2 & 43.0 \\
Emu2-Chat* \cite{sun2023generative} & 46.7 & 50.3 & 50.4 & 78.6 & 48.4 & 41.3 \\
MiniCPM-V-2* \cite{MiniCPM-V-2} & 56.7 & - & - & - & - & - \\
MiniCPM-V* \cite{MiniCPM-V} & 58.3 & - & - & - & - & - \\
SVIT* \cite{zhao2023svit} & 51.7 & 50.9 & 51.8 & 75.9 & 48.8 & 42.6 \\
InternVL-Chat-V1.1* \cite{chen2023internvl} & 57.5 & 56.4 & 57.9 & 80.4 & 55.6 & 46.9 \\
InfiMM-Zephyr-7B* \cite{InfiMM} & 59.2 & 54.6 & 55.4 & 75.9 & 54.4 & 46.2 \\
Yi-VL-6B* \cite{young2024yi} & 53.3 & 58.5 & 59.0 & 80.4 & 55.6 & 52.5 \\
OmniLMM-12B* \cite{OminiLMM-12B} & 62.5 & - & - & - & - & - \\
InternLM-XComposer2-VL* \cite{dong2024internlm} & 62.5 & 60.7 & 66.5 & 87.5 & 56.3 & 49.2 \\
HPT Air* \cite{HPT} & 59.2 & - & - & - & - & - \\
Yi-VL-34B* \cite{young2024yi} & 62.5 & 66.5 & 69.4 & 81.2 & 65.9 & 59.0 \\
LLaVA-1.6-34B* \cite{liu2024llava} & 67.5 & {\ul 70.2} & {\ul 74.8} & \textbf{91.1} & 65.9 & {\ul 62.0} \\
InternVL-Chat-V1.2* \cite{chen2023internvl} & {\ul 70.0} & 70.1 & 73.0 & 88.4 & {\ul 70.6} & 60.3 \\
VILA1.5* \cite{lin2023vila} & \textbf{73.3} & \textbf{74.0} & \textbf{79.1} & {\ul 90.2} & \textbf{73.0} & \textbf{64.3} \\
\midrule 
Marco-VL* & 65.8 & 66.5 & 69.1 & 85.7 & 64.7 & 58.7 \\
Reka-Edge* \cite{Reka} & 59.2 & - & - & - & - & - \\
Qwen-VL-PLUS*~\cite{Qwen-VL-PLUS} & 65.8 & 65.5 & 69.8 & 79.5 & 63.9 & 57.7 \\
Marco-VL-Plus* & 69.2 & 72.2 & 78.1 & 87.5 & 68.7 & 64.3 \\
Adept Fuyu-Heavy* \cite{Fuyu-Heavy} & 72.2 & - & - & - & - & - \\
Reka-Flash* \cite{Reka} & 74.2 & - & - & - & - & - \\
Skywork-VL* \cite{Skywork-VL} & 68.3 & 71.6 & 77.7 & \textbf{90.2} & 69.8 & 60.7 \\
Qwen-VL-MAX* \cite{Qwen-VL-MAX} & 69.2 & 70.4 & 75.9 & {\ul 89.3} & 62.7 & 64.9 \\
HPT Pro* \cite{HPT} & 72.5 & - & - & - & - & - \\
SenseChat-Vision-0423-Preview* \cite{SenseChat-Vision} & {\ul 75.0} & {\ul 74.7} & {\ul 78.8} & \textbf{90.2} & \textbf{72.6} & {\ul 66.9} \\
Reka-Core* \cite{Reka} & {\ul 75.0} & - & - & - & - & - \\
GPT-4V(ision) (Playground) \cite{openai2023gpt4v} & 72.5 & \textbf{76.3} & \textbf{79.1} & {\ul 89.3} & {\ul 71.4} & \textbf{73.1} \\ 
Gemini 1.0 Ultra* \cite{deepmind_gemini_report} & \textbf{78.3} & - & - & - & - & - \\ \midrule
\multicolumn{7}{c}{\textbf{Large Language Models (LLMs): Only Text as Input}} \\ \midrule
Llama2 7B \cite{touvron2023llama2} & 37.5 & 32.6 & 32.4 & 46.4 & 32.9 & 27.5 \\
\addlinespace[0.1em]\hdashline\addlinespace[0.1em]
FLAN-T5-XXL~\cite{chung2022scaling} & 42.5 & 44.8 & 46.8 & 56.2 & 39.7 & \textbf{43.0} \\
\quad  + OCR & \textbf{55.0} & \textbf{50.5} & \textbf{53.6} & \textbf{75.0} & 46.4 & 42.0 \\
\quad  + LLaVA Caption & 49.2 & 49.9 & 51.8 & \textbf{75.0} & \textbf{46.8} & 41.6 \\
\addlinespace[0.1em]\hdashline\addlinespace[0.1em]
Vicuna-13B~\cite{vicuna2023} & 45.8 & 44.8 & 51.1 & 59.8 & 39.3 & \textbf{38.0} \\
\quad + OCR & \textbf{50.0} & 49.3 & \textbf{58.3} & 66.1 & 48.0 & 36.1 \\
\quad + LLaVA Caption & 48.3 & \textbf{49.4} & 53.6 & \textbf{72.3} & \textbf{48.8} & 37.7 \\
\addlinespace[0.1em]\hdashline\addlinespace[0.1em]
GPT-4 Text~\cite{openai2023gpt4} & 51.7 & 53.0 & 52.9 & 82.1 & 34.5 & 57.7 \\ \bottomrule
\end{tabular}%
\end{adjustbox}
\caption{\textbf{Humanities \& Social Science} results of different models on the \dataset \textbf{validation} and \textbf{test set}.  The best-performing model in each category is \textbf{in-bold}, and the second best is {\ul{underlined}}. *: results provided by the authors.} 
\label{tab:overall_Humanities_results}
\end{table*}

\clearpage
\subsection{Tech \& Engineering}

\begin{table*}[!b]
\centering
\small
\begin{adjustbox}{scale = 0.8}
\begin{tabular}{@{}lccccccccc@{}}
\toprule
\textbf{} & \textbf{\begin{tabular}[c]{@{}c@{}}Val \\  Overall\end{tabular}} & \textbf{\begin{tabular}[c]{@{}c@{}}Test \\  Overall\end{tabular} } & \textbf{Agri.} & \textbf{\begin{tabular}[c]{@{}c@{}}Arch. \&\\  Eng.\end{tabular}} & \textbf{\begin{tabular}[c]{@{}c@{}}Comp. \\  Sci.\end{tabular}} & \textbf{Electr.} & \textbf{\begin{tabular}[c]{@{}c@{}}Energy\\\&Power\end{tabular}} & \textbf{Materials} & \textbf{\begin{tabular}[c]{@{}c@{}}Mech.\\  Eng.\end{tabular}}\\
 & (210) & (2,784) & (287) & (551) & (371) & (256) & (432) & (458) & (429)\\ \midrule

\color{Gray} Random Choice & \color{Gray} 21.4 & \color{Gray} 24.8 & \color{Gray} 21.3 & \color{Gray} 27.0 & \color{Gray} 22.6 & \color{Gray} 10.5 & \color{Gray} 31.5 & \color{Gray} 24.2 & \color{Gray} 28.7\\ 
\color{Gray} Frequent Choice & \color{Gray} 24.8 & \color{Gray} 26.5 & \color{Gray} 24.7 & \color{Gray} 24.1 & \color{Gray} 29.6 & \color{Gray} 12.9 & \color{Gray} 30.3 & \color{Gray} 30.3 & \color{Gray} 28.0\\
Expert (Worst)  & 74.3 & - & - & - & - & - & - & - & - \\
Expert (Medium)  & 79.1 & - & - & - & - & - & - & - & - \\
Expert (Best)  & 86.2 & - & - & - & - & - & - & - & - \\
\midrule
\multicolumn{10}{c}{\textbf{Large Multimodal Models (LMMs): Text + Image as Input}} \\ \midrule
OpenFlamingo2-9B \cite{awadalla2023openflamingo} & 26.2 & 25.1 & 20.6 & 29.6 & 26.1 & 13.7 & 24.1 & 26.0 & 28.4 \\
Kosmos2 \cite{peng2023kosmos} & 26.7 & 26.8 & 20.6 & 28.3 & 32.1 & 10.2 & 29.9 & 27.3 & 30.5 \\
Fuyu-8B~\cite{fuyu-8b} & 21.4 & 26.4 & 26.5 & 25.0 & 26.1 & 12.1 & 35.0 & 25.1 & 29.8 \\
MiniGPT4-Vicuna-13B~\cite{zhu2023minigpt} & 23.8 & 27.2 & 29.6 & 23.8 & 28.8 & 13.7 & 36.1 & 27.3 & 27.5 \\
LLaMA-Adapter2-7B~\cite{zhang2023llama} & 30.0 & 25.7 & 23.0 & 25.2 & 25.6 & 17.6 & 30.3 & 25.8 & 28.4 \\
Otter~\cite{li2023otter} & 29.0 & 30.2 & 30.7 & 26.3 & 32.1 & 19.1 & 35.2 & 30.1 & 34.7 \\
CogVLM~\cite{COGVLM} & 27.6 & 28.9 & 26.8 & 27.0 & 31.8 & 14.1 & 33.1 & 28.4 & 35.4 \\
InstructBLIP-T5-XL~\cite{dai2023instructblip} & 27.1 & 28.6 & 26.1 & \textbf{33.6} & 28.3 & 23.8 & 29.9 & 22.9 & 31.5 \\
BLIP-2 FLAN-T5-XL \cite{li2023blip} & 27.6 & 27.8 & 17.8 & 32.5 & 26.7 & 20.7 & 33.6 & 24.9 & 30.8 \\
mPLUG-OWL2* \cite{ye2023mplug2} & 31.0 & 29.6 & 32.4 & 29.4 & 31.8 & 14.5 & 39.4 & 26.6 & 28.2 \\
SPHINX* \cite{lin2023sphinx} & 26.2 & 27.8 & 31.0 & 26.1 & 33.7 & 16.4 & 35.2 & 23.1 & 26.8 \\
Qwen-VL-7B-Chat~\cite{Qwen-VL} & 32.9 & 30.2 & 33.1 & 25.0 & 33.4 & 19.1 & 37.0 & 28.8 & 33.1 \\
Bunny-3B* \cite{Bunny-3B} & 37.1 & 28.7 & 32.1 & 26.0 & 34.0 & 18.8 & 33.6 & 27.5 & 27.7 \\
LLaVA-1.5-13B~\cite{liu2023improved} & 31.4 & 28.3 & 34.5 & 26.1 & 29.6 & 22.7 & 30.1 & 26.9 & 28.9 \\
InstructBLIP-T5-XXL~\cite{dai2023instructblip} & 35.2 & 29.4 & 24.7 & 30.3 & 29.6 & 20.7 & 37.3 & 26.6 & 31.5 \\
BLIP-2 FLAN-T5-XXL~\cite{li2023blip} & 30.0 & 30.4 & 28.2 & 27.2 & 29.6 & 25.0 & 35.6 & 26.9 & 38.0 \\
Emu2-Chat* \cite{sun2023generative} & 35.2 & 31.3 & 35.9 & 26.9 & 30.7 & 16.0 & 41.9 & 26.2 & 38.5 \\
MiniCPM-V-2* \cite{MiniCPM-V-2} & 27.1 & - & - & - & - & - & - & - & - \\
MiniCPM-V* \cite{MiniCPM-V} & 27.1 & - & - & - & - & - & - & - & - \\
SVIT* \cite{zhao2023svit} & 33.8 & 30.7 & 29.6 & 27.2 & 34.5 & 21.9 & 37.7 & 27.9 & 34.0 \\
InternVL-Chat-V1.1* \cite{chen2023internvl} & 27.1 & 28.0 & 36.9 & 27.2 & 31.5 & 15.2 & 30.6 & 26.0 & 27.0\\
InfiMM-Zephyr-7B* \cite{InfiMM} & 29.0 & 31.1 & 39.0 & 28.7 & 34.5 & 20.3 & 31.7 & 31.0 & 31.9 \\
Yi-VL-6B* \cite{young2024yi} & 35.7 & 34.1 & 32.4 & 29.2 & 33.4 & 28.9 & 35.0 & 35.8 & 42.4 \\
OmniLMM-12B* \cite{OminiLMM-12B} & 31.9 & - & - & - & - & - & - & - & - \\
InternLM-XComposer2-VL* \cite{dong2024internlm} & 32.4 & 31.8 & 41.8 & 29.6 & 36.4 & 22.3 & 33.3 & 27.7 & 32.2 \\
HPT Air* \cite{HPT} & 42.9 & - & - & - & - & - & - & - & - \\
Yi-VL-34B* \cite{young2024yi}& 41.0 & 36.0 & 39.4 & 31.6 & 40.2 & 28.9 & 36.8 & 34.1 & 41.5 \\
LLaVA-1.6-34B* \cite{liu2024llava} & 43.8 & 36.3 & 40.1 & 31.9 & {\ul 43.7} & 27.0 & 34.0 & 34.5 & {\ul 42.7} \\
InternVL-Chat-V1.2* \cite{chen2023internvl} & {\ul 46.2} & \textbf{40.8} & {\ul 42.9} & {\ul 32.8} & 42.6 & {\ul 32.4} & \textbf{45.8} & \textbf{40.8} & \textbf{48.0} \\
VILA1.5* \cite{lin2023vila} & \textbf{48.1} & {\ul 39.5} & \textbf{45.3} & 31.0 & \textbf{46.6} & \textbf{32.8} & {\ul 43.3} & {\ul 38.4} & 41.7 \\
\midrule
Marco-VL* & 32.4 & 33.8 & 35.5 & 31.2 & 36.7 & 24.2 & 34.7 & 35.4 & 36.4 \\
Reka-Edge* \cite{Reka} & 33.8 & - & - & - & - & - & - & - & - \\
Qwen-VL-PLUS* \cite{Qwen-VL-PLUS} & 36.7 & 32.9 & 40.4 & 25.6 & 36.1 & 24.6 & 33.6 & 34.7 & 36.8 \\
Marco-VL-Plus* & 37.1 & 36.7 & 43.9 & 27.6 & 40.2 & {\ul 34.8} & 37.7 & 34.3 & 43.1 \\
Adept Fuyu-Heavy* \cite{Fuyu-Heavy} & 44.0 & - & - & - & - & - & - & - & - \\
Reka-Flash* \cite{Reka} & 44.3 & - & - & - & - & - & - & - & - \\
Skywork-VL* \cite{Skywork-VL} & {\ul 46.7} & 40.2 & 45.3 & {\ul 31.6} & 44.5 & 32.4 & 42.4 & {\ul 38.9} & 48.3 \\
Qwen-VL-MAX* \cite{Qwen-VL-MAX} & 38.6 & 40.7 & {\ul 45.6} & 27.6 & 42.6 & \textbf{35.2} & {\ul 48.1} & 37.8 & {\ul 51.5} \\
HPT Pro* \cite{HPT} & 43.8 & - & - & - & - & - & - & - & - \\
SenseChat-Vision-0423-Preview \cite{SenseChat-Vision} & 43.8 & \textbf{43.5} & \textbf{48.8} & 31.2 & {\ul 47.2} & 33.2 & \textbf{51.2} & \textbf{41.9} & \textbf{52.7} \\
Reka-Core* \cite{Reka} & 44.2 & - & - & - & - & - & - & - & - \\
GPT-4V(ision) (Playground) \cite{openai2023gpt4v} & 36.7 & {\ul 41.7} & 43.9 & \textbf{37.2} & \textbf{57.1} & 27.0 & 47.5 & 36.9 & 41.0 \\ 
Gemini 1.0 Ultra* \cite{deepmind_gemini_report} & \textbf{47.1} & - & - & - & - & - & - & - & - \\ \midrule
\multicolumn{10}{c}{\textbf{Large Language Models (LLMs): Only Text as Input}} \\ \midrule
Llama2 7B \cite{touvron2023llama2} & 31.4 & 29.8 & 33.1 & 23.8 & 32.6 & 17.6 & 39.1 & 27.9 & 32.6 \\
\addlinespace[0.1em]\hdashline\addlinespace[0.1em]
FLAN-T5-XXL~\cite{chung2022scaling} & 28.6 & 28.3 & \textbf{21.3} & 30.3 & 28.8 & 25.4 & 26.6 & 27.9 & 33.8 \\
\quad  + OCR & \textbf{30.0} & \textbf{29.7} & 20.2 & 30.7 & \textbf{31.3} & \textbf{27.0} & 29.9 & \textbf{29.0} & \textbf{35.9} \\
\quad  + LLaVA Caption & 27.6 & 28.7 & 17.8 & \textbf{31.4} & 26.4 & 24.2 & \textbf{32.4} & 26.4 & 35.7 \\
\addlinespace[0.1em]\hdashline\addlinespace[0.1em]
Vicuna-13B~\cite{vicuna2023} & \textbf{34.8} & 30.1 & 31.7 & 26.3 & 28.8 & \textbf{25.4} & 39.4 & 26.4 & 32.6 \\
\quad + OCR & \textbf{34.8} & 30.0 & 31.7 & 25.6 & 29.9 & 18.8 & \textbf{40.3} & 27.5 & 33.6 \\
\quad + LLaVA Caption & 32.4 & \textbf{31.4} & \textbf{32.4} & \textbf{26.9} & \textbf{31.8} & 20.3 & 39.8 & \textbf{28.8} & \textbf{37.3} \\
\addlinespace[0.1em]\hdashline\addlinespace[0.1em]
GPT-4 Text~\cite{openai2023gpt4} & 20.0 & 28.4 & 28.9 & 25.6 & 33.4 & 17.2 & 38.4 & 23.6 & 28.9 \\ \bottomrule
\end{tabular}%
\end{adjustbox}
\caption{\textbf{Tech \& Engineering} results of different models on the \dataset \textbf{validation} and \textbf{test set}.  The best-performing model in each category is \textbf{in-bold}, and the second best is {\ul{underlined}}. *: results provided by the authors.} 
\label{tab:overall_Tech_results}
\end{table*}

\clearpage
\section{Case Study}
\label{Appendix:Case_Study}
\phantomsection
\label{list:list_of_figures}
\listofappfigures

\renewcommand{\thefigure}{\arabic{figure}}

\begin{table*}[!htbp]
    \centering
    \begin{tabular}{lllllll}
         Subject   & Correct Case                                         &Perception             & Lack of Knowledge     & Reasoning & Other \\ \toprule
         Art       &\ref{fig:art_1}, \ref{fig:art_2}              &\ref{fig:art_3}    & & & \\
         Art Theory&\ref{fig:art_theory_1}, \ref{fig:art_theory_2}&  &\ref{fig:art_theory_3} & & \\
         Design    &\ref{fig:design_1}                                &\ref{fig:design_2} & \ref{fig:design_3}& & \\
         Music     &\ref{fig:music_1}                                 &\ref{fig:music_2}, \ref{fig:music_3}&\ref{fig:music_2} & & \\ \midrule
         Accounting&\ref{fig:accounting_1}                            &\ref{fig:accounting_2}& & & \\
         Economics &\ref{fig:economics_1} &\ref{fig:economics_2}, \ref{fig:economics_3}& & & \\
         Finance   &\ref{fig:finance_1} & & &\ref{fig:finance_2} & \\
         Manage    &\ref{fig:manage_1} &\ref{fig:manage_2} & & & \\
         Marketing &\ref{fig:marketing_1} &\ref{fig:marketing_2} & & & \\ \midrule
         Biology   &\ref{fig:biology_2}& & &\ref{fig:biology_3}, \ref{fig:biology_4}& \\
         Chemistry &\ref{fig:chemistry_1}, \ref{fig:chemistry_2}&\ref{fig:chemistry_3} &\ref{fig:chemistry_4} &\ref{fig:chemistry_3} & \\
         Geography &\ref{fig:geography_1} &\ref{fig:geography_3} & &\ref{fig:geography_2}, \ref{fig:geography_3}& \\
         Math      &\ref{fig:math_1} &\ref{fig:math_2} & &\ref{fig:math_4} &\ref{fig:math_3} \\
         Physics   &\ref{fig:physics_1} &\ref{fig:physics_2} & & & \\ \midrule
         Basic Medical Science&\ref{fig:basic_medical_science_1} &\ref{fig:basic_medical_science_2} & & & \\
         Clinical Medicine&\ref{fig:clinical_medicine_1}, \ref{fig:clinical_medicine_2}, \ref{fig:clinical_medicine_3}&\ref{fig:clinical_medicine_4} &\ref{fig:clinical_medicine_5} & & \\
         Diagnostics and Laboratory Medicine&\ref{fig:diagnostics_and_laboratory_medicine_1} &\ref{fig:diagnostics_and_laboratory_medicine_2}, \ref{fig:diagnostics_and_laboratory_medicine_4}&\ref{fig:diagnostics_and_laboratory_medicine_4} & &\ref{fig:diagnostics_and_laboratory_medicine_3} \\
         Pharmacy  &\ref{fig:pharmacy_1} & &\ref{fig:pharmacy_2}, \ref{fig:pharmacy_3}& & \\
         Public Health&\ref{fig:public_health_1} & &\ref{fig:public_health_3}& &\ref{fig:public_health_2} \\ \midrule
         History&\ref{fig:history_1}, \ref{fig:history_2}&\ref{fig:history_3} &\ref{fig:history_4} & & \\
         Literature&\ref{fig:literature_1} &\ref{fig:literature_2} & & & \\
         Sociology&\ref{fig:sociology_1} & & &\ref{fig:sociology_2} & \\
         Psychology&\ref{fig:psychology_1} &\ref{fig:psychology_2} & & & \\ \midrule
         Agriculture&\ref{fig:agriculture_1} &\ref{fig:agriculture_2}, \ref{fig:agriculture_3},\ref{fig:agriculture_4}& & & \\
         Architecture and Engineering&\ref{fig:architecture_engineering_1}& & &\ref{fig:architecture_engineering_2} & \\
         Computer Science&\ref{fig:computer_science_1}&\ref{fig:computer_science_2}, \ref{fig:computer_science_3}, \ref{fig:computer_science_4}&\ref{fig:computer_science_2} & & \\
         Electronics&\ref{fig:electronics_1} & & & &\ref{fig:electronics_2} \\
         Energy and Power&\ref{fig:energy_power_1} & & &\ref{fig:energy_power_2} & \\
         Materials&\ref{fig:materials_1} & &\ref{fig:materials_2} & & \\
         Mechanical Engineering &\ref{fig:mechanical_engineering_1} & & &\ref{fig:mechanical_engineering_2}, \ref{fig:mechanical_engineering_3}& \\\bottomrule 
    \end{tabular}
    \caption{Table index of case study figures by subjects with associated (error) categories. }
\label{tab:list_of_case_study_figures}
\end{table*}
\clearpage

\begin{figure*}[!htbp]
    \centering
\includegraphics[width=0.9\linewidth]{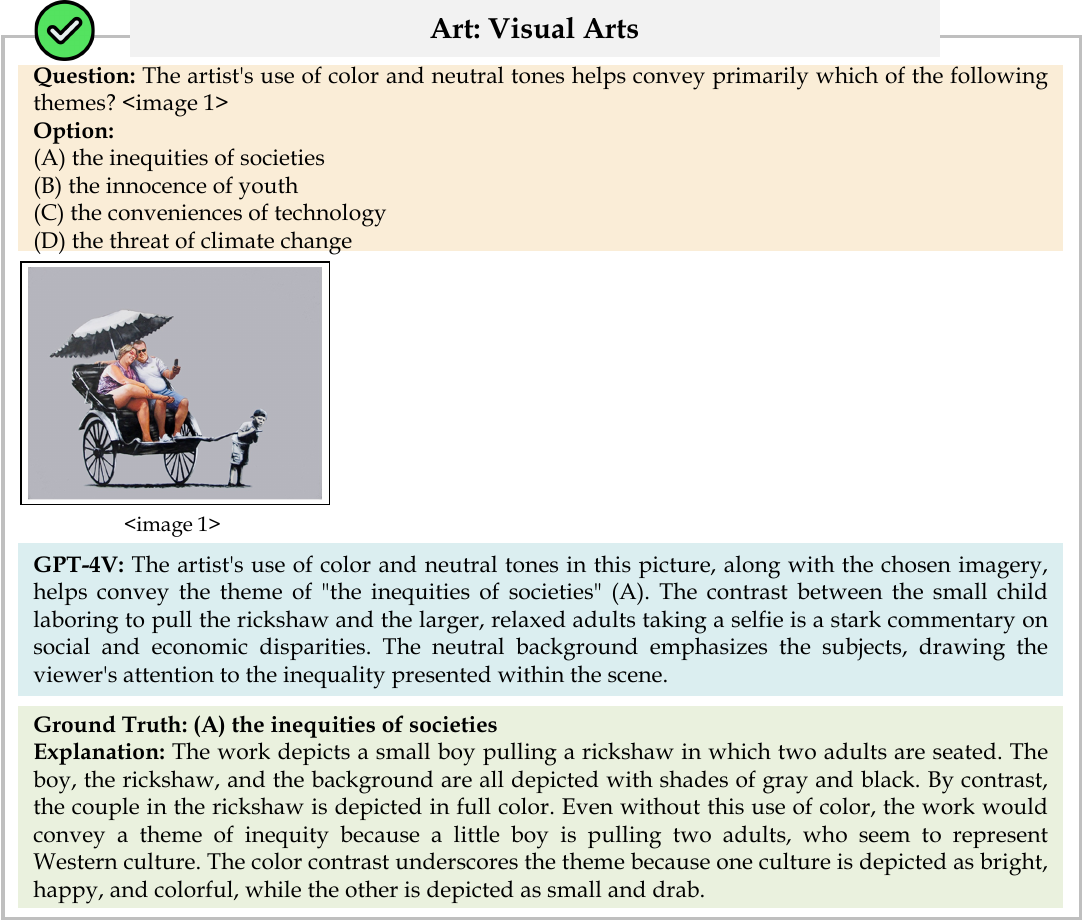}
    \caption{A sample correct case of Art (subfield: Visual Arts).\\ \hyperref[list:list_of_figures]{Back to List of Figures} \textcolor{red}{$|$} \hyperref[tab:list_of_case_study_figures]{Back to Table Index}}
    \addcontentsline{afg}{appfigures}{\protect\numberline{\thefigure}Art 1: Correct Case}
\label{fig:art_1}
\end{figure*}
\newpage

\begin{figure*}[!htbp]
    \centering
\includegraphics[width=0.9\linewidth]{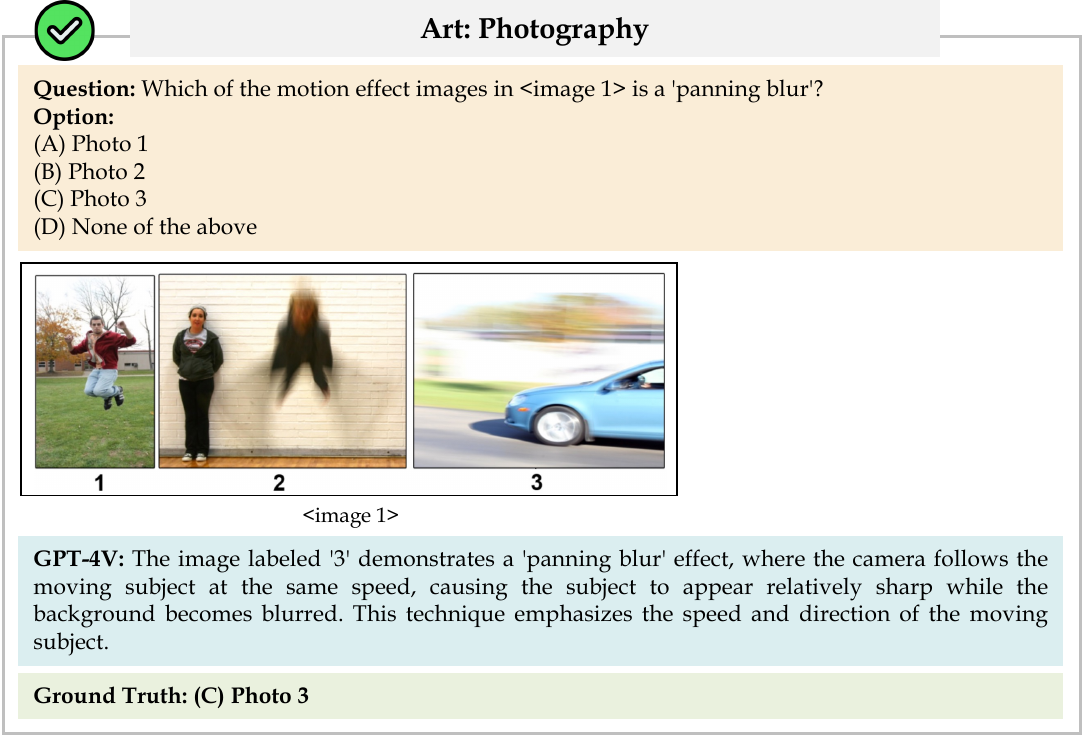}
    \caption{A sample correct case of Art (subfield: Photography).\\ \hyperref[list:list_of_figures]{Back to List of Figures} \textcolor{red}{$|$} \hyperref[tab:list_of_case_study_figures]{Back to Table Index}}
    \addcontentsline{afg}{appfigures}{\protect\numberline{\thefigure}Art 2: Correct Case}
\label{fig:art_2}
\end{figure*}
\newpage

\begin{figure*}[!htbp]
    \centering
\includegraphics[width=0.9\linewidth]{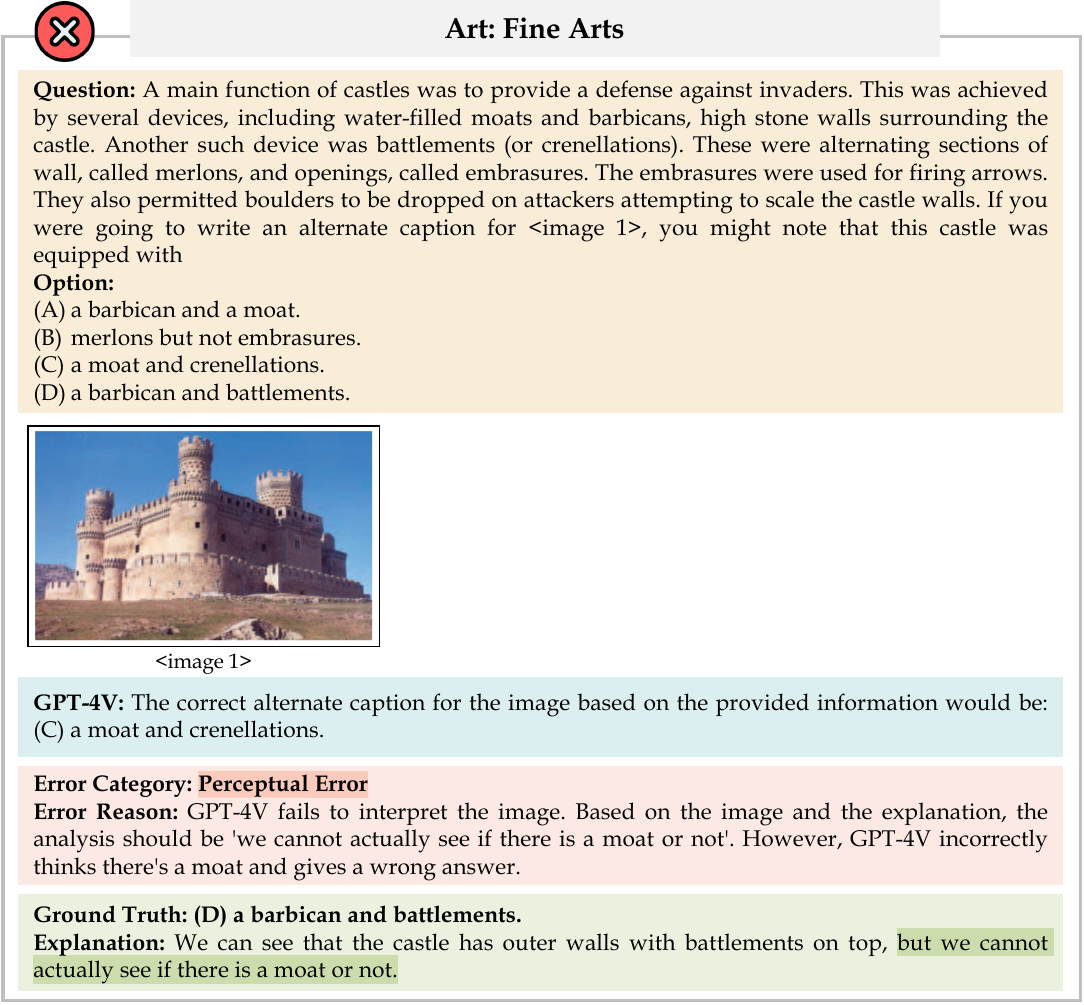}
    \caption{A sample error case of Art (subfield: Fine Arts). Error category: Perceptual Error \newline \centering \hyperref[list:list_of_figures]{Back to List of Figures} \textcolor{red}{$|$} \hyperref[tab:list_of_case_study_figures]{Back to Table Index}}
    \addcontentsline{afg}{appfigures}{\protect\numberline{\thefigure}Art  3: Perceptual Error}
\label{fig:art_3}
\end{figure*}
\newpage


\begin{figure*}[!htbp]
    \centering
\includegraphics[width=0.9\linewidth]{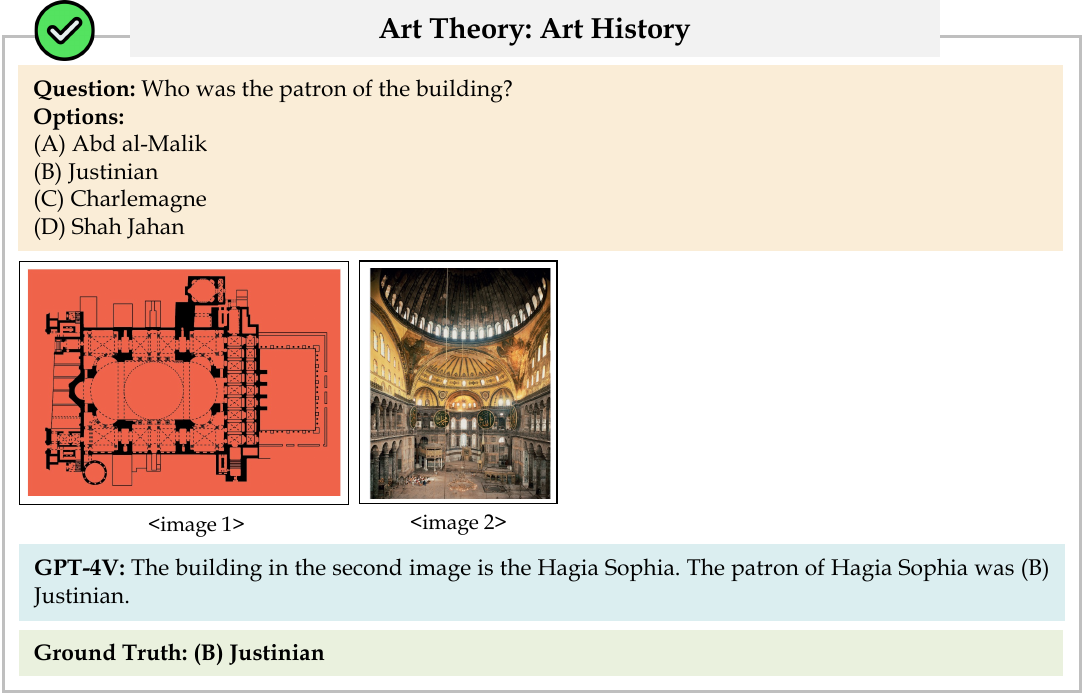}
    \caption{A sample correct case of Art Theory (subfield: Art History).\\ \hyperref[list:list_of_figures]{Back to List of Figures} \textcolor{red}{$|$} \hyperref[tab:list_of_case_study_figures]{Back to Table Index}}
    \addcontentsline{afg}{appfigures}{\protect\numberline{\thefigure}Art Theory  1: Correct Case}
\label{fig:art_theory_1}
\end{figure*}
\newpage

\begin{figure*}[!htbp]
    \centering
\includegraphics[width=0.9\linewidth]{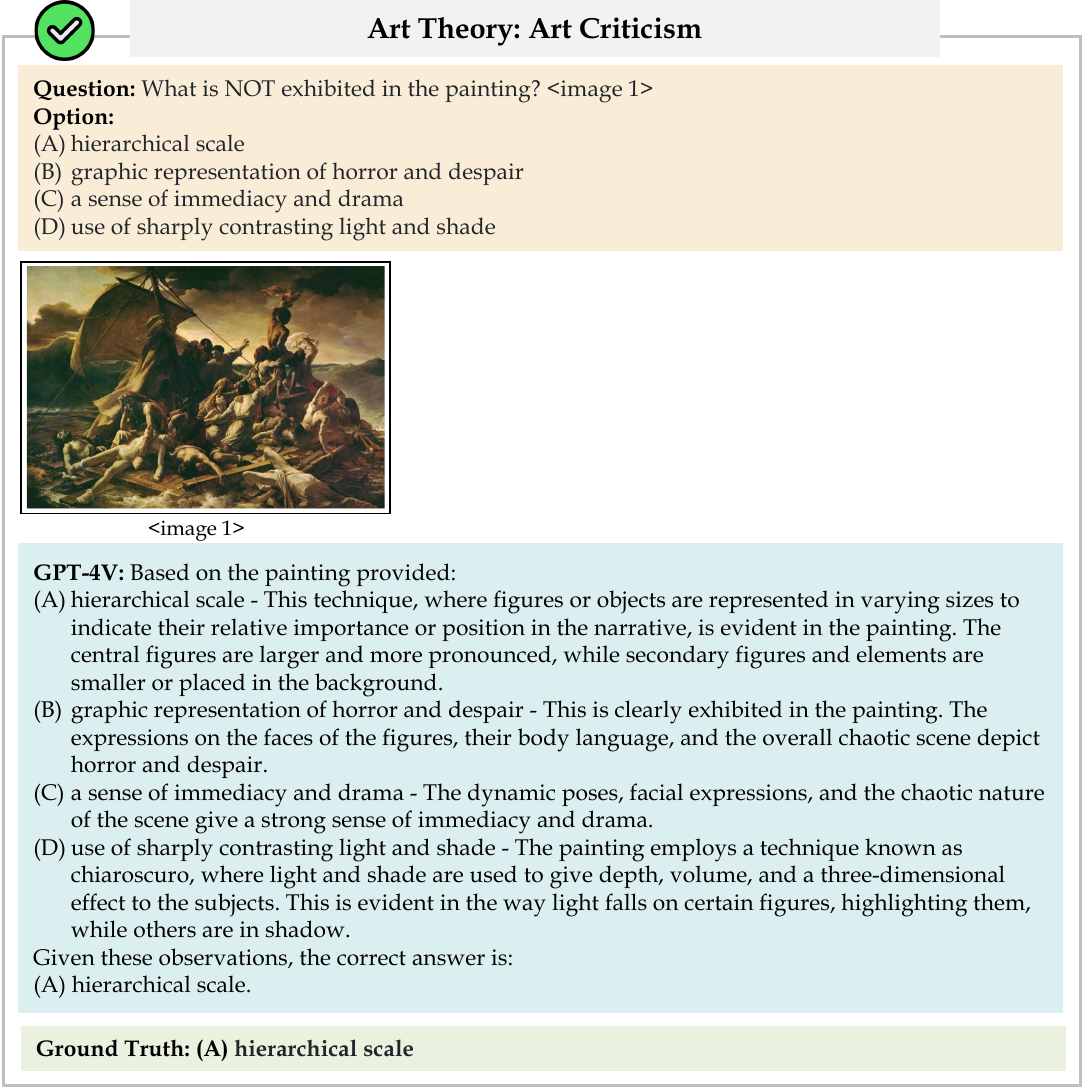}
    \caption{A sample correct case of Art Theory (subfield: Art Criticism).\\ \hyperref[list:list_of_figures]{Back to List of Figures} \textcolor{red}{$|$} \hyperref[tab:list_of_case_study_figures]{Back to Table Index}}
    \addcontentsline{afg}{appfigures}{\protect\numberline{\thefigure}Art Theory  2: Correct Case}
\label{fig:art_theory_2}
\end{figure*}
\newpage

\begin{figure*}[!htbp]
    \centering
\includegraphics[width=0.9\linewidth]{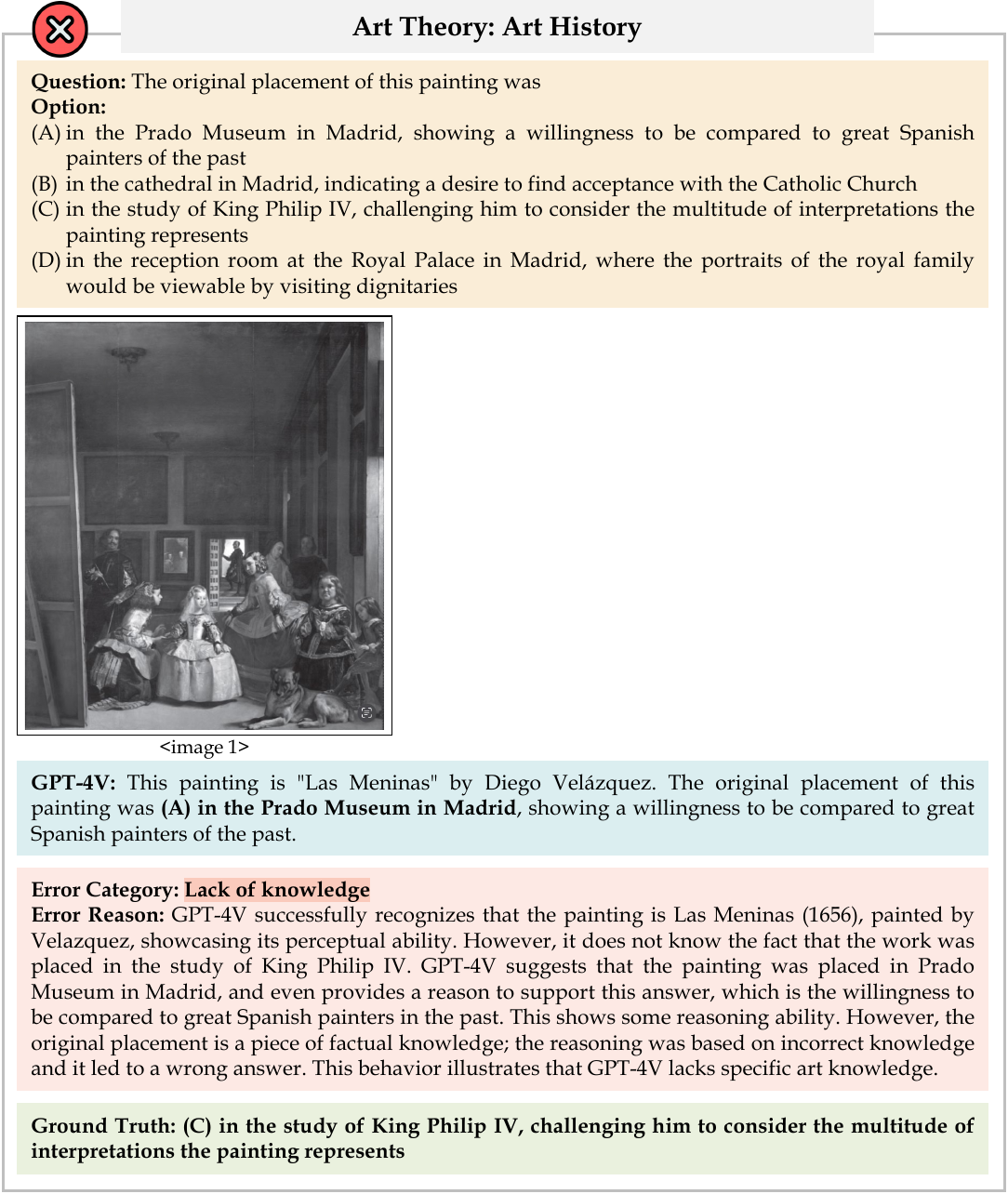}
    \caption{A sample error case of Art Theory (subfield: Art History). Error category: Lack of Knowledge \newline \centering \hyperref[list:list_of_figures]{Back to List of Figures} \textcolor{red}{$|$} \hyperref[tab:list_of_case_study_figures]{Back to Table Index}}
    \addcontentsline{afg}{appfigures}{\protect\numberline{\thefigure}Art Theory  3: Lack of Knowledge}
\label{fig:art_theory_3}
\end{figure*}
\newpage

\begin{figure*}[!htbp]
    \centering
\includegraphics[width=0.9\linewidth]{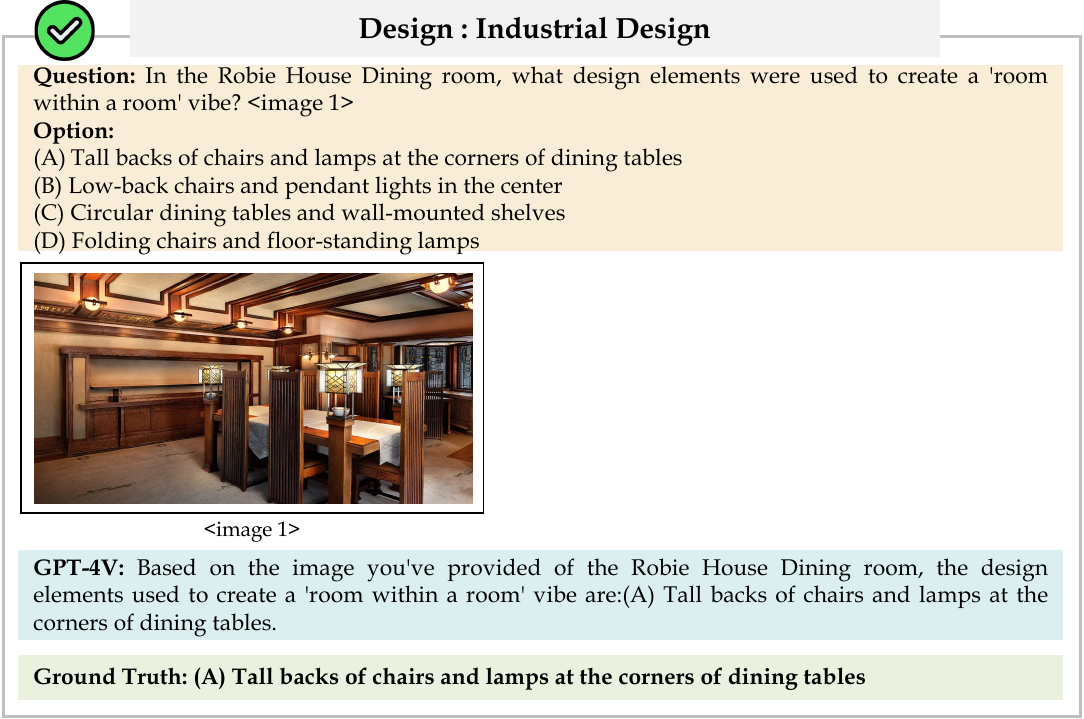}
    \caption{A sample correct case of Design (subfield: Industrial Design).\\ \hyperref[list:list_of_figures]{Back to List of Figures} \textcolor{red}{$|$} \hyperref[tab:list_of_case_study_figures]{Back to Table Index}}
    \addcontentsline{afg}{appfigures}{\protect\numberline{\thefigure}Design  1: Correct Case}
\label{fig:design_1}
\end{figure*}
\newpage

\begin{figure*}[!htbp]
    \centering
\includegraphics[width=0.9\linewidth]{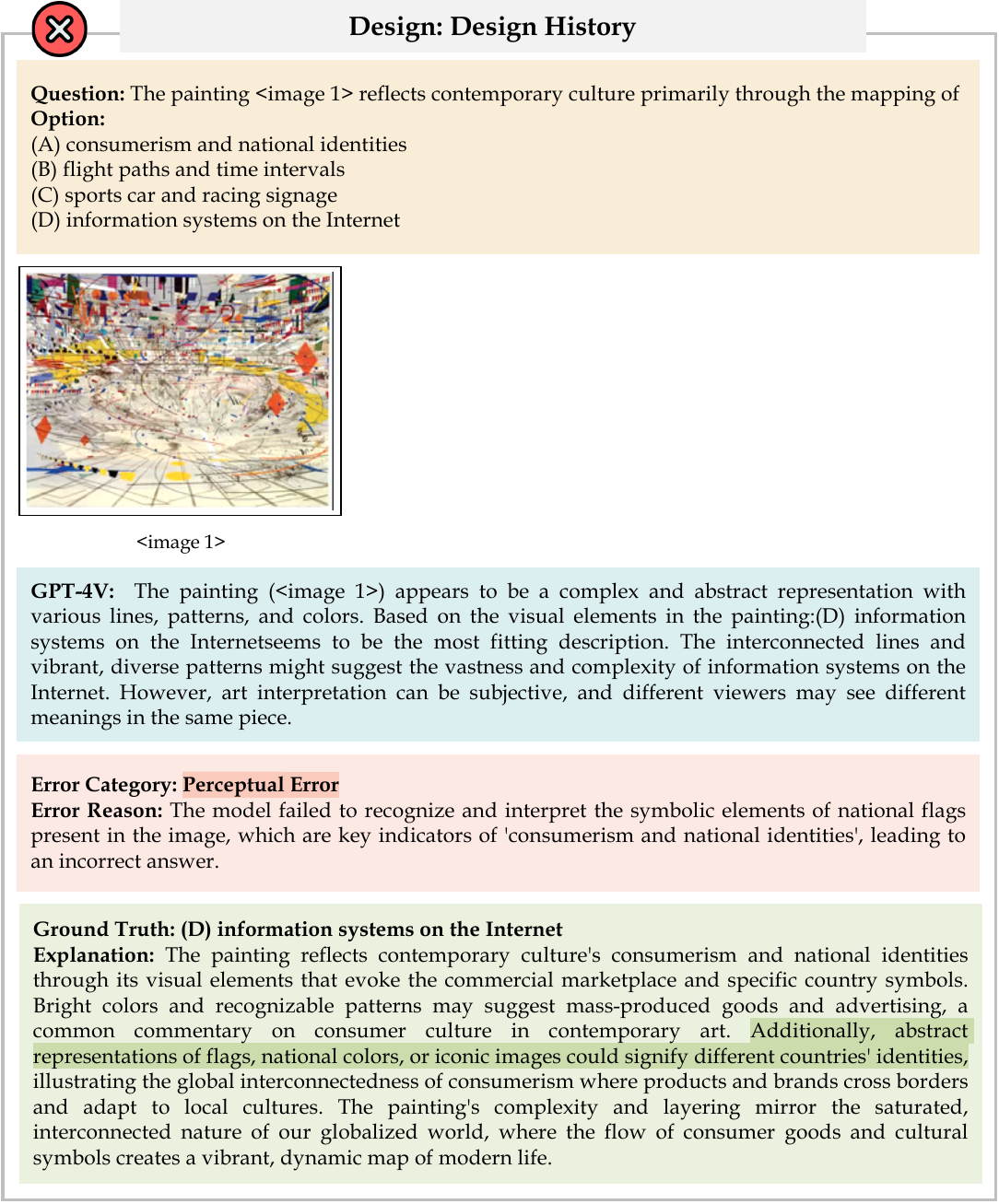}
    \caption{A sample error case of Design (subfield: Design History). Error category: Perceptual Error \newline \centering \hyperref[list:list_of_figures]{Back to List of Figures} \textcolor{red}{$|$} \hyperref[tab:list_of_case_study_figures]{Back to Table Index}}
    \addcontentsline{afg}{appfigures}{\protect\numberline{\thefigure}Design  2: Perceptual Error}
\label{fig:design_2}
\end{figure*}
\newpage

\begin{figure*}[!htbp]
    \centering
\includegraphics[width=0.9\linewidth]{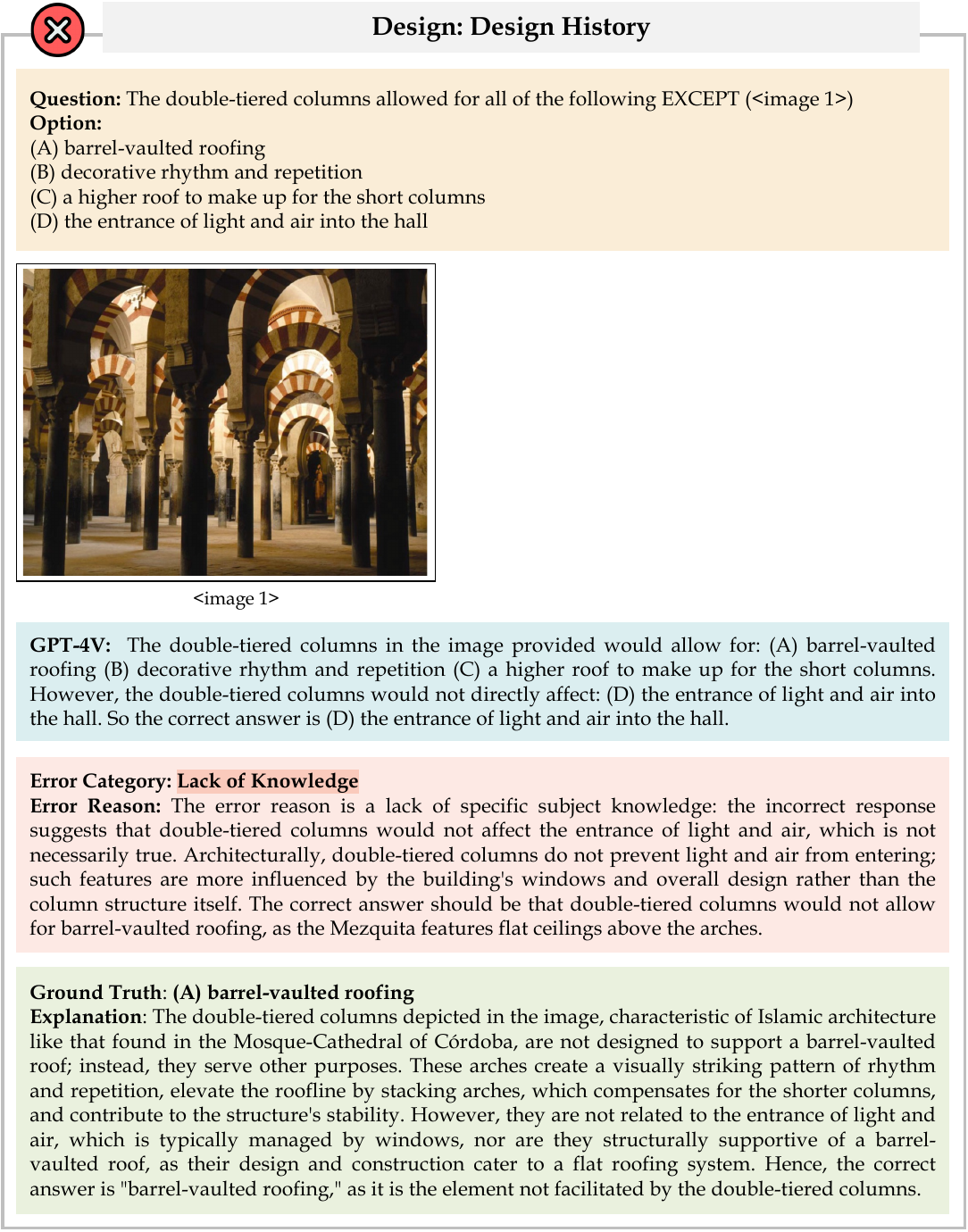}
    \caption{A sample error case of Design (subfield: Design History). Error category: Lack of Knowledge \newline \centering  \hyperref[list:list_of_figures]{Back to List of Figures} \textcolor{red}{$|$} \hyperref[tab:list_of_case_study_figures]{Back to Table Index}}
    \addcontentsline{afg}{appfigures}{\protect\numberline{\thefigure}Design  3: Lack of Knowledge}
\label{fig:design_3}
\end{figure*}
\newpage

\begin{figure*}[!htbp]
    \centering
\includegraphics[width=0.9\linewidth]{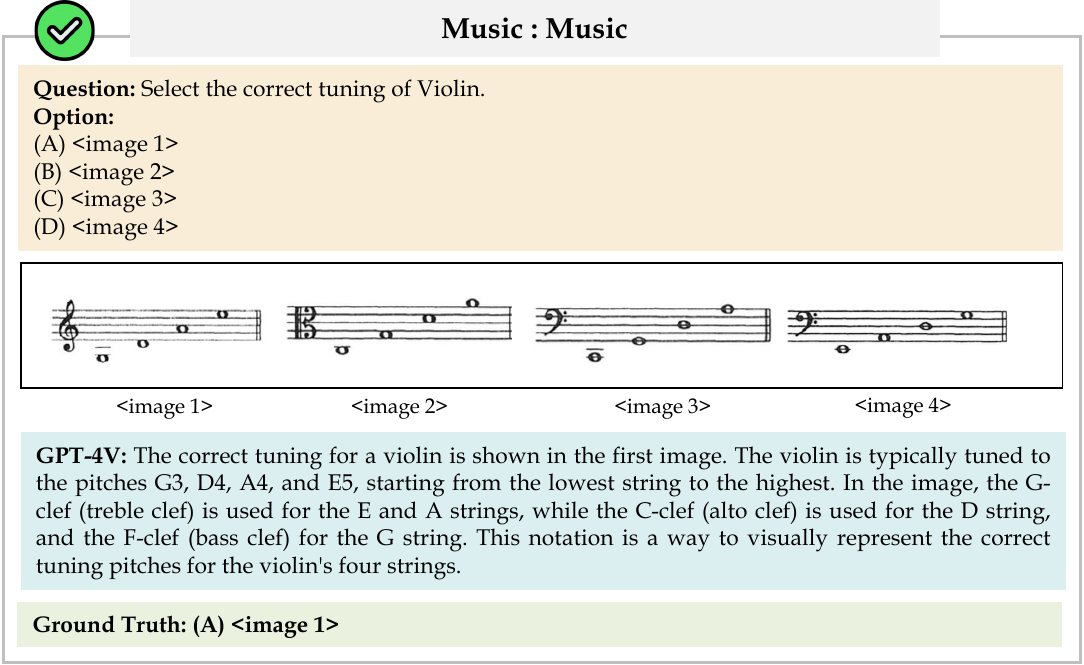}
    \caption{A sample correct case of Music (subfield: Music).\\ \hyperref[list:list_of_figures]{Back to List of Figures} \textcolor{red}{$|$} \hyperref[tab:list_of_case_study_figures]{Back to Table Index}}
    \addcontentsline{afg}{appfigures}{\protect\numberline{\thefigure}Music  1: Correct Case}
\label{fig:music_1}
\end{figure*}
\newpage

\begin{figure*}[!htbp]
    \centering
\includegraphics[width=0.9\linewidth]{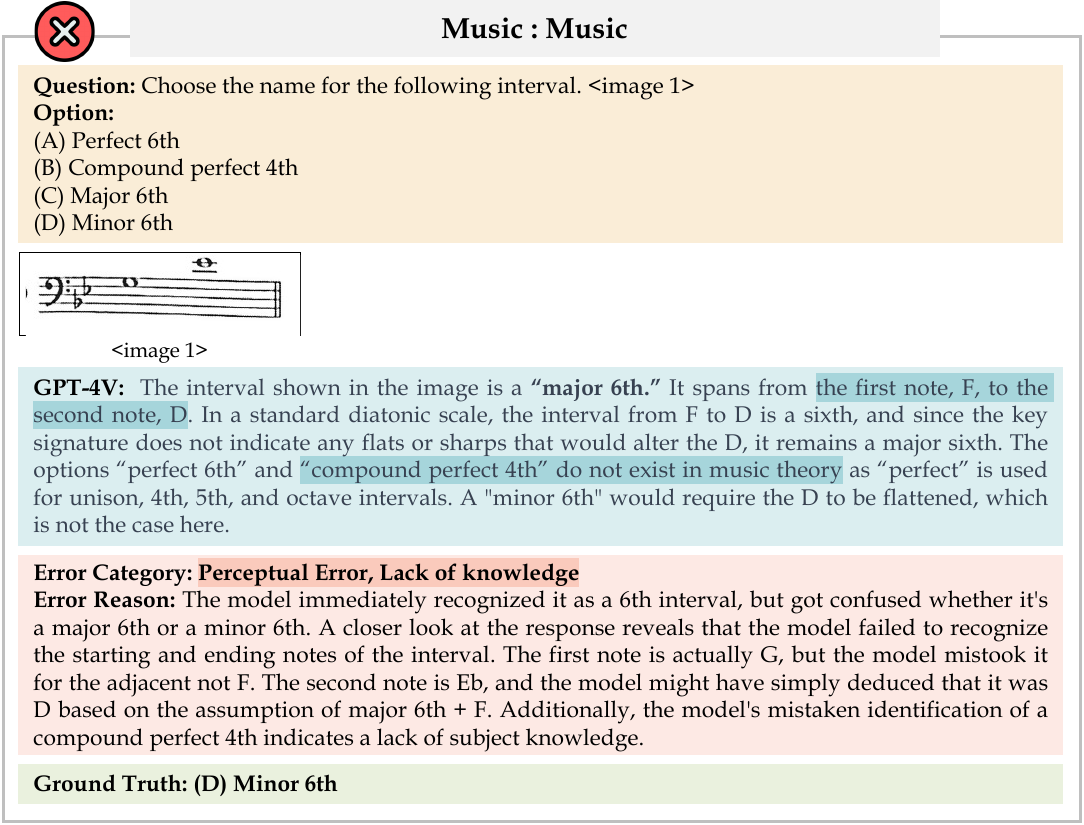}
    \caption{A sample error case of Music (subfield: Music). Error category: Perceptual Error, Lack of Knowledge \newline \centering \hyperref[list:list_of_figures]{Back to List of Figures} \textcolor{red}{$|$} \hyperref[tab:list_of_case_study_figures]{Back to Table Index}}
    \addcontentsline{afg}{appfigures}{\protect\numberline{\thefigure}Music  2: Perceptual Error, Lack of Knowledge}
\label{fig:music_2}
\end{figure*}
\newpage

\begin{figure*}[!htbp]
    \centering
\includegraphics[width=0.9\linewidth]{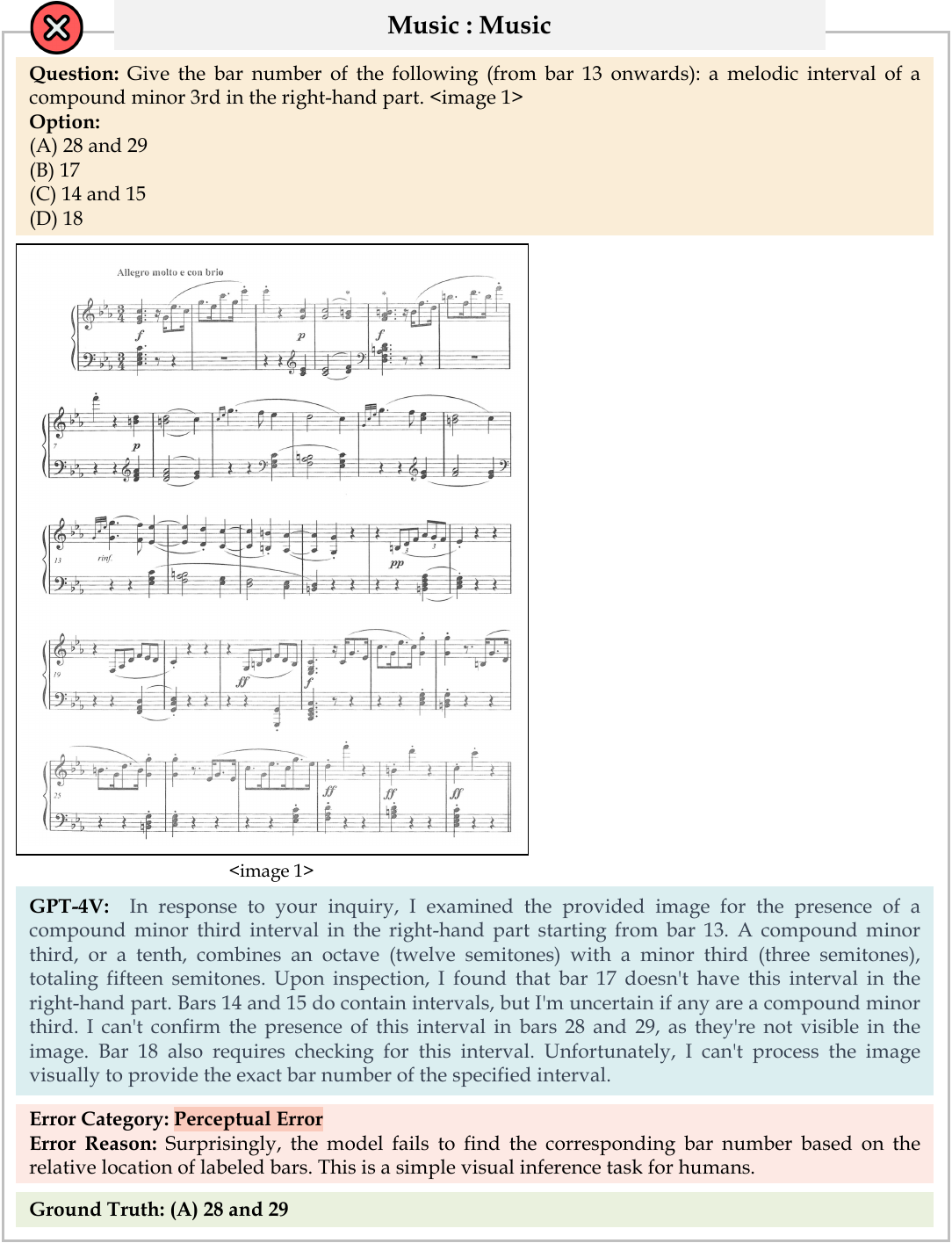}
    \caption{A sample error case of Music (subfield: Music). Error category: Perceptual Error \newline \centering \hyperref[list:list_of_figures]{Back to List of Figures} \textcolor{red}{$|$} \hyperref[tab:list_of_case_study_figures]{Back to Table Index}}
    \addcontentsline{afg}{appfigures}{\protect\numberline{\thefigure}Music  3: Perceptual Error}
\label{fig:music_3}
\end{figure*}
\newpage

\begin{figure*}[!htbp]
    \centering
\includegraphics[width=0.9\linewidth]{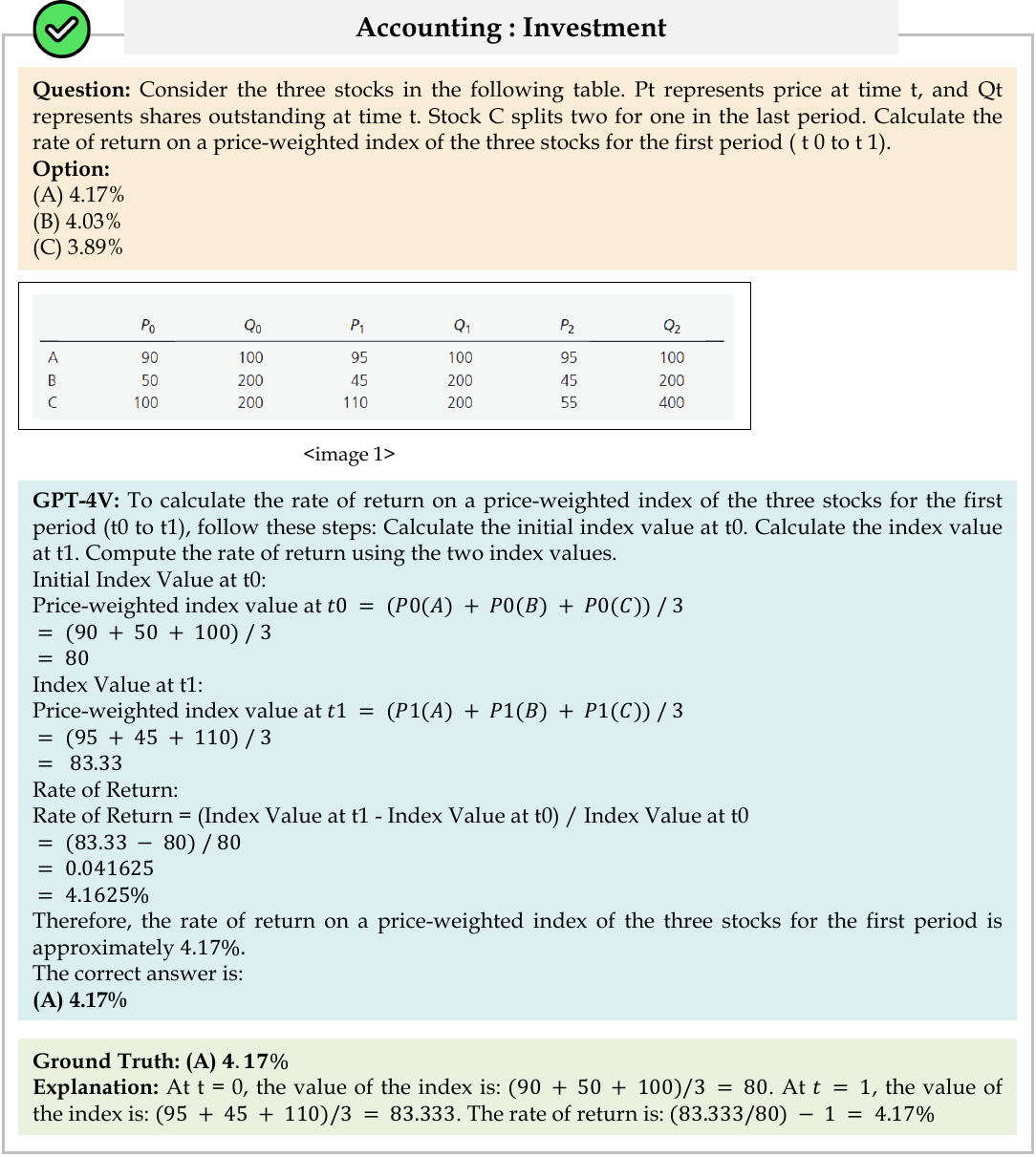}
    \caption{A sample correct case of Accounting (subfield: Investment).\\ \hyperref[list:list_of_figures]{Back to List of Figures} \textcolor{red}{$|$} \hyperref[tab:list_of_case_study_figures]{Back to Table Index}}
    \addcontentsline{afg}{appfigures}{\protect\numberline{\thefigure}Accounting  1: Correct Case}
\label{fig:accounting_1}
\end{figure*}
\newpage

\begin{figure*}[!htbp]
    \centering
\includegraphics[width=0.9\linewidth]{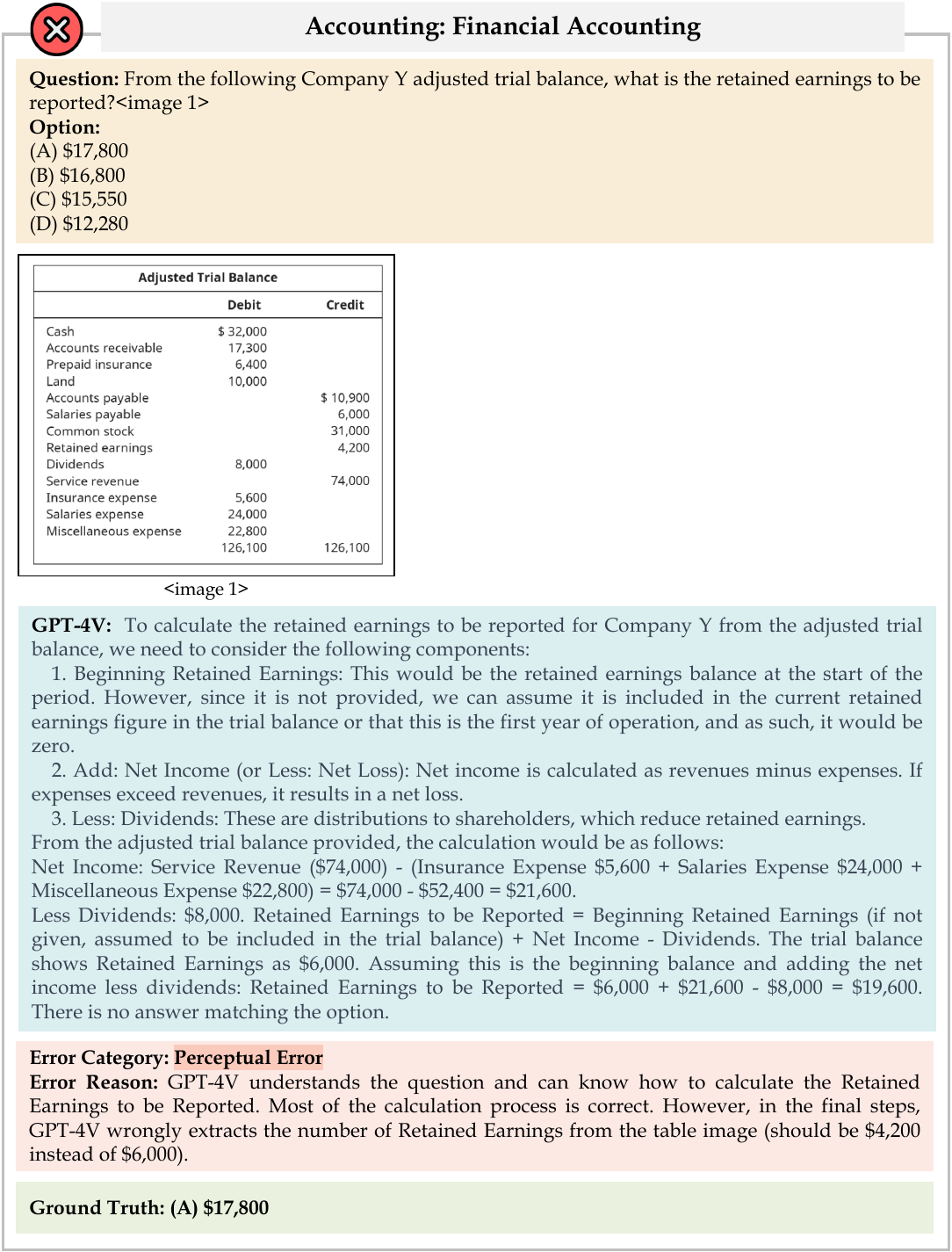}
    \caption{A sample error case of Accounting (subfield: Financial Accounting). Error category: Perceptual Error \newline \centering \hyperref[list:list_of_figures]{Back to List of Figures} \textcolor{red}{$|$} \hyperref[tab:list_of_case_study_figures]{Back to Table Index}}
    \addcontentsline{afg}{appfigures}{\protect\numberline{\thefigure}Accounting  2: Perceptual Error}
\label{fig:accounting_2}
\end{figure*}
\newpage


\begin{figure*}[!htbp]
    \centering
\includegraphics[width=0.9\linewidth]{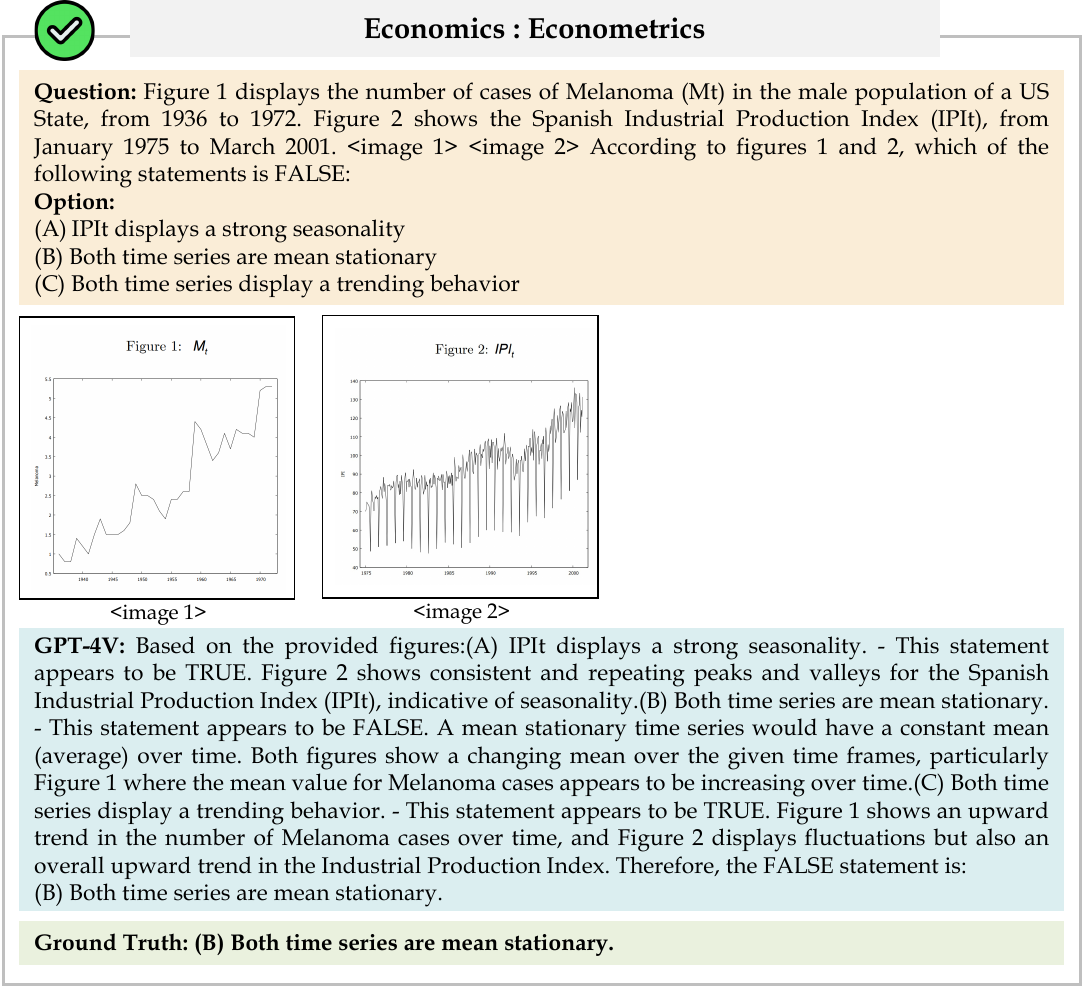}
    \caption{A sample correct case of Economics (subfield: Econometrics).\\ \hyperref[list:list_of_figures]{Back to List of Figures} \textcolor{red}{$|$} \hyperref[tab:list_of_case_study_figures]{Back to Table Index}}
    \addcontentsline{afg}{appfigures}{\protect\numberline{\thefigure}Economics  1: Correct Case}
\label{fig:economics_1}
\end{figure*}
\newpage

\begin{figure*}[!htbp]
    \centering
\includegraphics[width=0.9\linewidth]{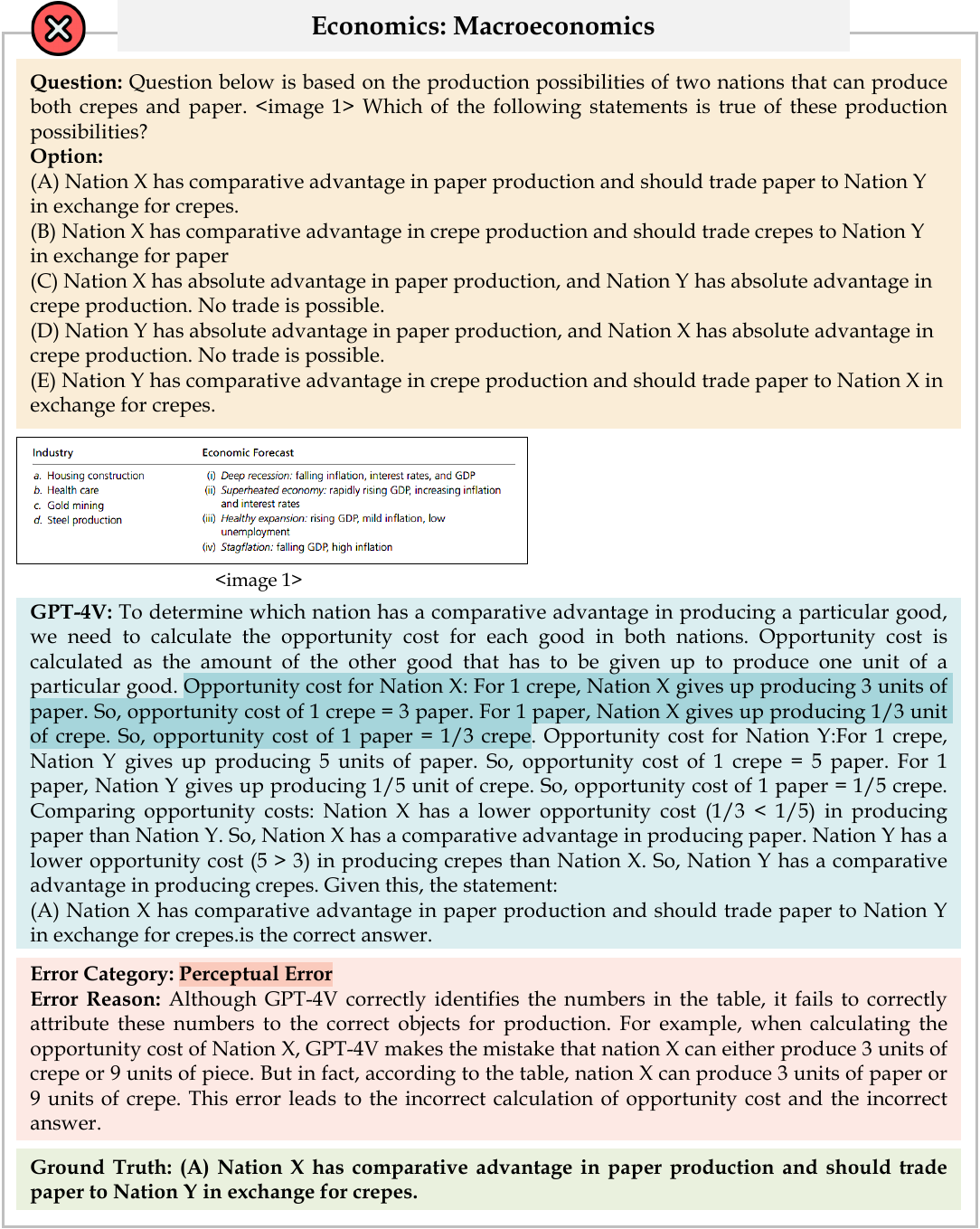}
    \caption{A sample error case of Economics (subfield: Macroeconomics). Error category: Perceptual Error \newline \centering \hyperref[list:list_of_figures]{Back to List of Figures} \textcolor{red}{$|$} \hyperref[tab:list_of_case_study_figures]{Back to Table Index}}
    \addcontentsline{afg}{appfigures}{\protect\numberline{\thefigure}Economics  2: Perceptual Error}
\label{fig:economics_2}
\end{figure*}
\newpage

\begin{figure*}[!htbp]
    \centering
\includegraphics[width=0.9\linewidth]{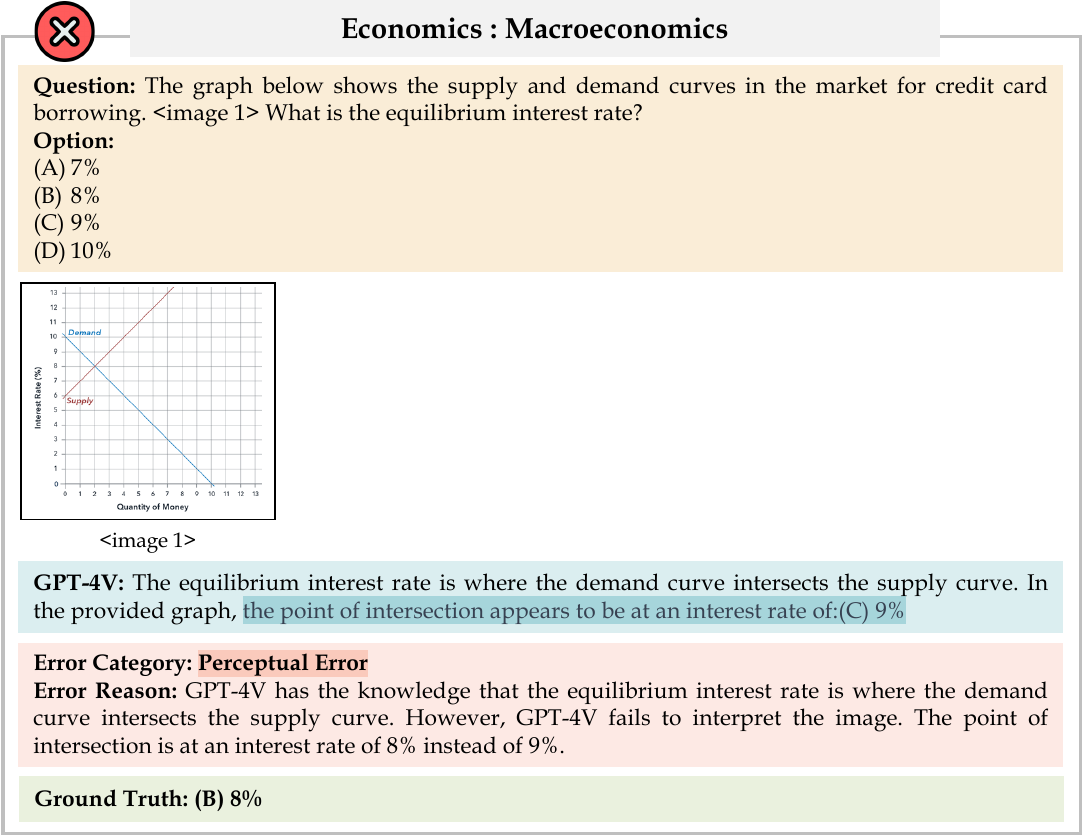}
    \caption{A sample error case of Economics (subfield: Macroeconomics). Error category: Perceptual Error \newline \centering \hyperref[list:list_of_figures]{Back to List of Figures} \textcolor{red}{$|$} \hyperref[tab:list_of_case_study_figures]{Back to Table Index}}
    \addcontentsline{afg}{appfigures}{\protect\numberline{\thefigure}Economics  3: Perceptual Error}
\label{fig:economics_3}
\end{figure*}
\newpage

\begin{figure*}[!htbp]
    \centering
\includegraphics[width=0.9\linewidth]{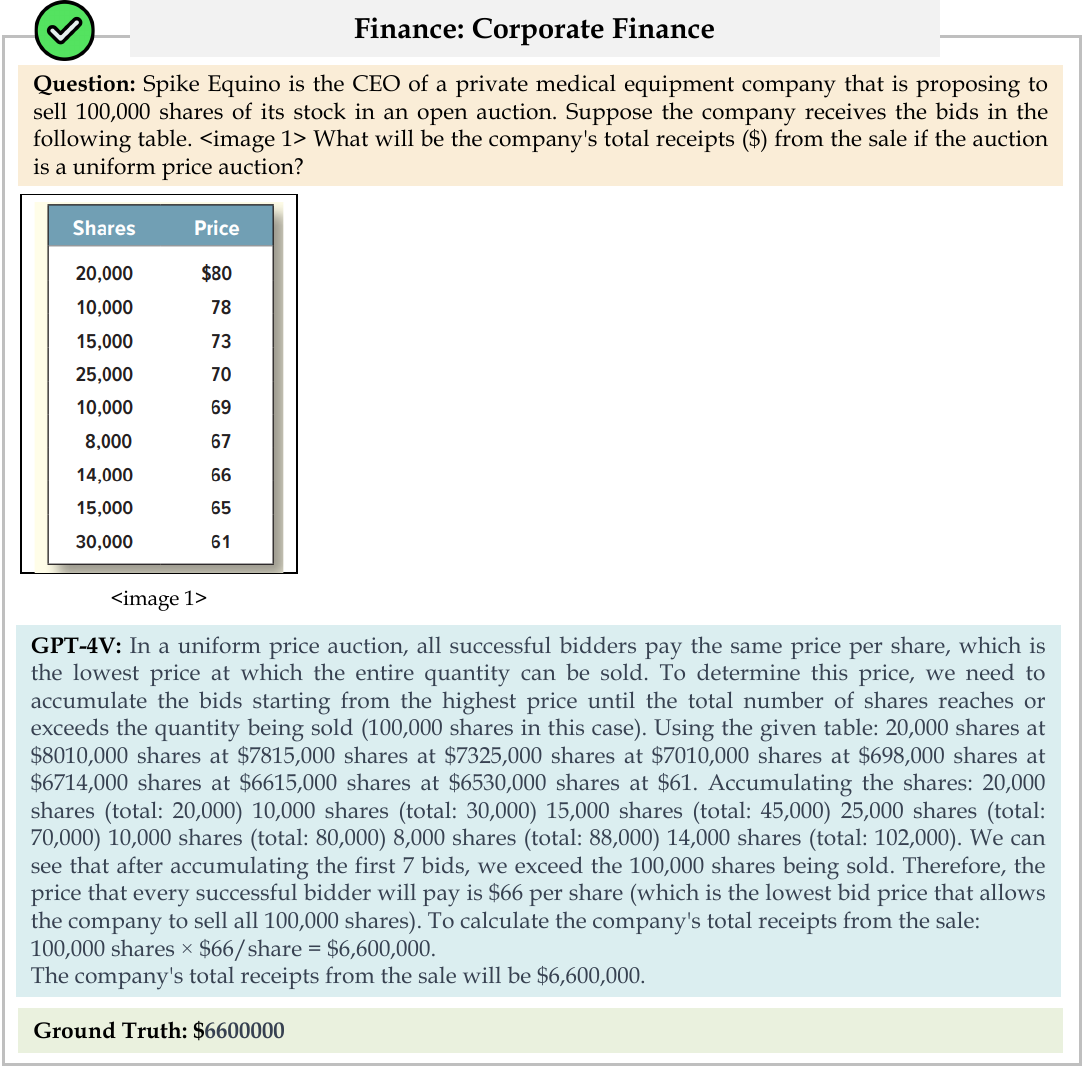}
    \caption{A sample correct case of Finance (subfield: Corporate Finance).\\ \hyperref[list:list_of_figures]{Back to List of Figures} \textcolor{red}{$|$} \hyperref[tab:list_of_case_study_figures]{Back to Table Index}}
    \addcontentsline{afg}{appfigures}{\protect\numberline{\thefigure}Finance  1: Correct Case}
\label{fig:finance_1}
\end{figure*}
\newpage

\begin{figure*}[!htbp]
    \centering
\includegraphics[width=0.9\linewidth]{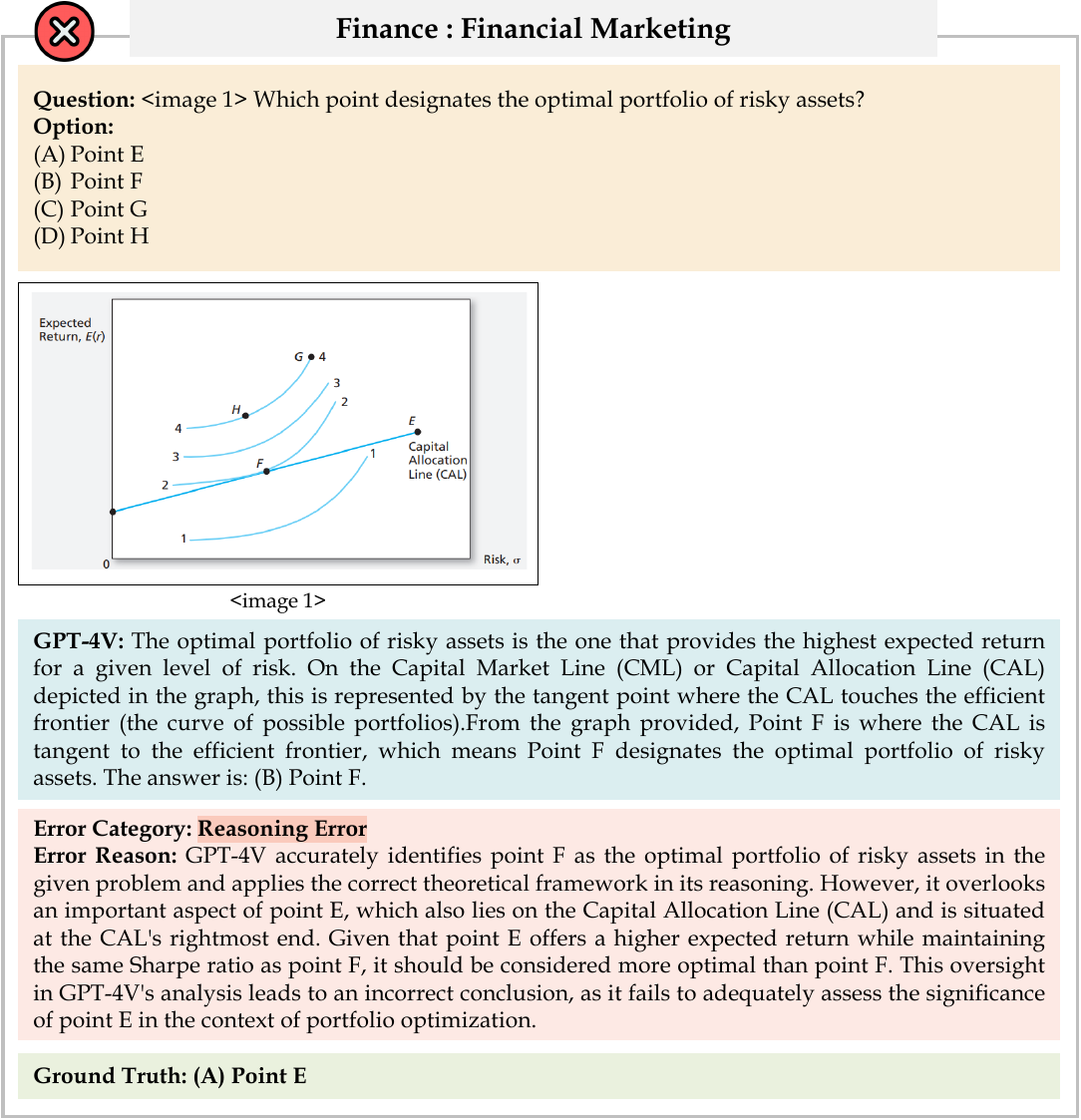}
    \caption{A sample error case of Finance (subfield: Financial Marketing). Error category: Reasoning Error \newline \centering \hyperref[list:list_of_figures]{Back to List of Figures} \textcolor{red}{$|$} \hyperref[tab:list_of_case_study_figures]{Back to Table Index}}
    \addcontentsline{afg}{appfigures}{\protect\numberline{\thefigure}Finance  2: Reasoning Error}
\label{fig:finance_2}
\end{figure*}
\newpage


\begin{figure*}[!htbp]
    \centering
\includegraphics[width=0.9\linewidth]{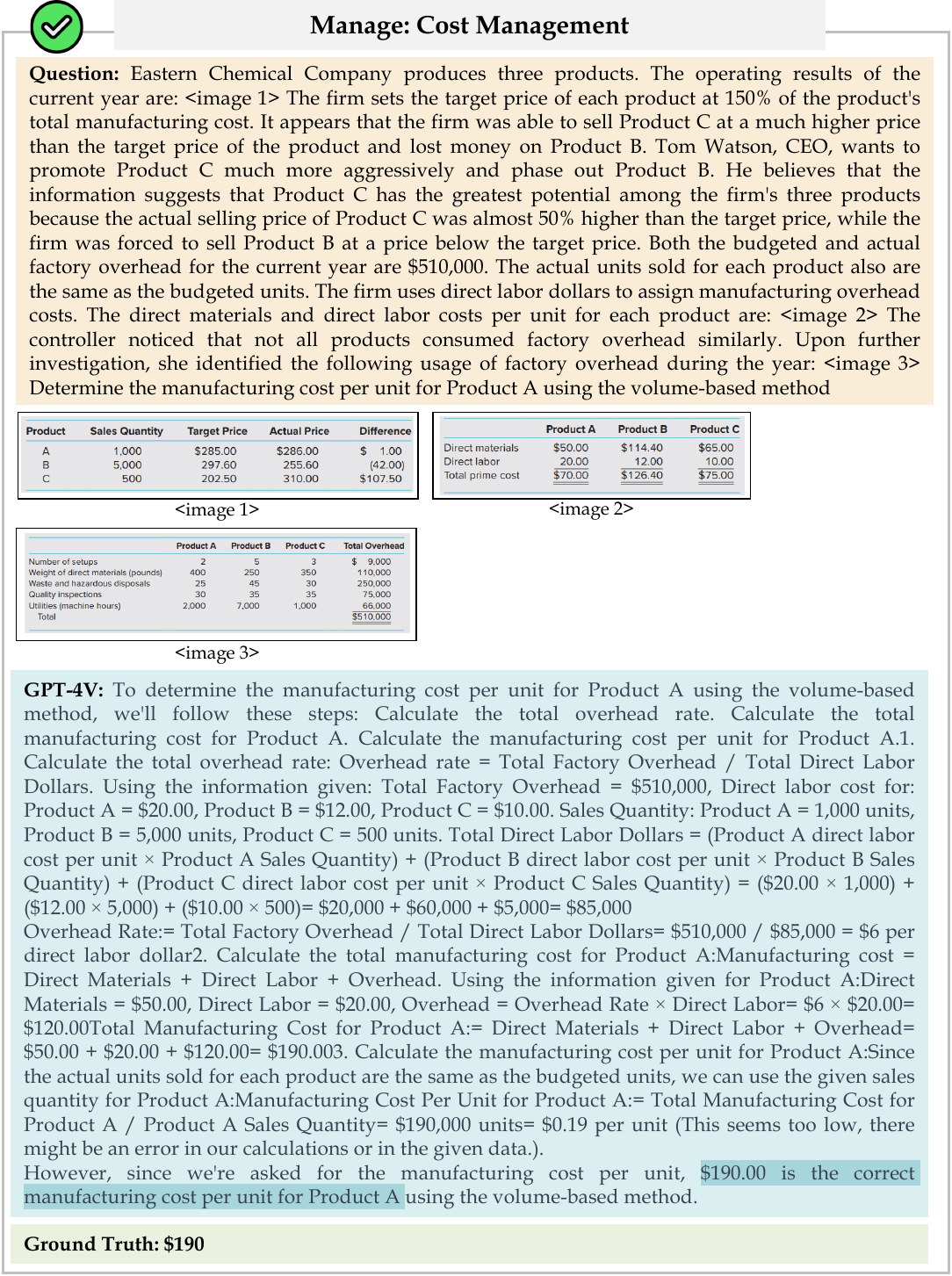}
    \caption{A sample correct case of Manage (subfield: Cost Management).\\ \hyperref[list:list_of_figures]{Back to List of Figures} \textcolor{red}{$|$} \hyperref[tab:list_of_case_study_figures]{Back to Table Index}}
    \addcontentsline{afg}{appfigures}{\protect\numberline{\thefigure}Manage  1: Correct Case}
\label{fig:manage_1}
\end{figure*}
\newpage

\begin{figure*}[!htbp]
    \centering
\includegraphics[width=0.9\linewidth]{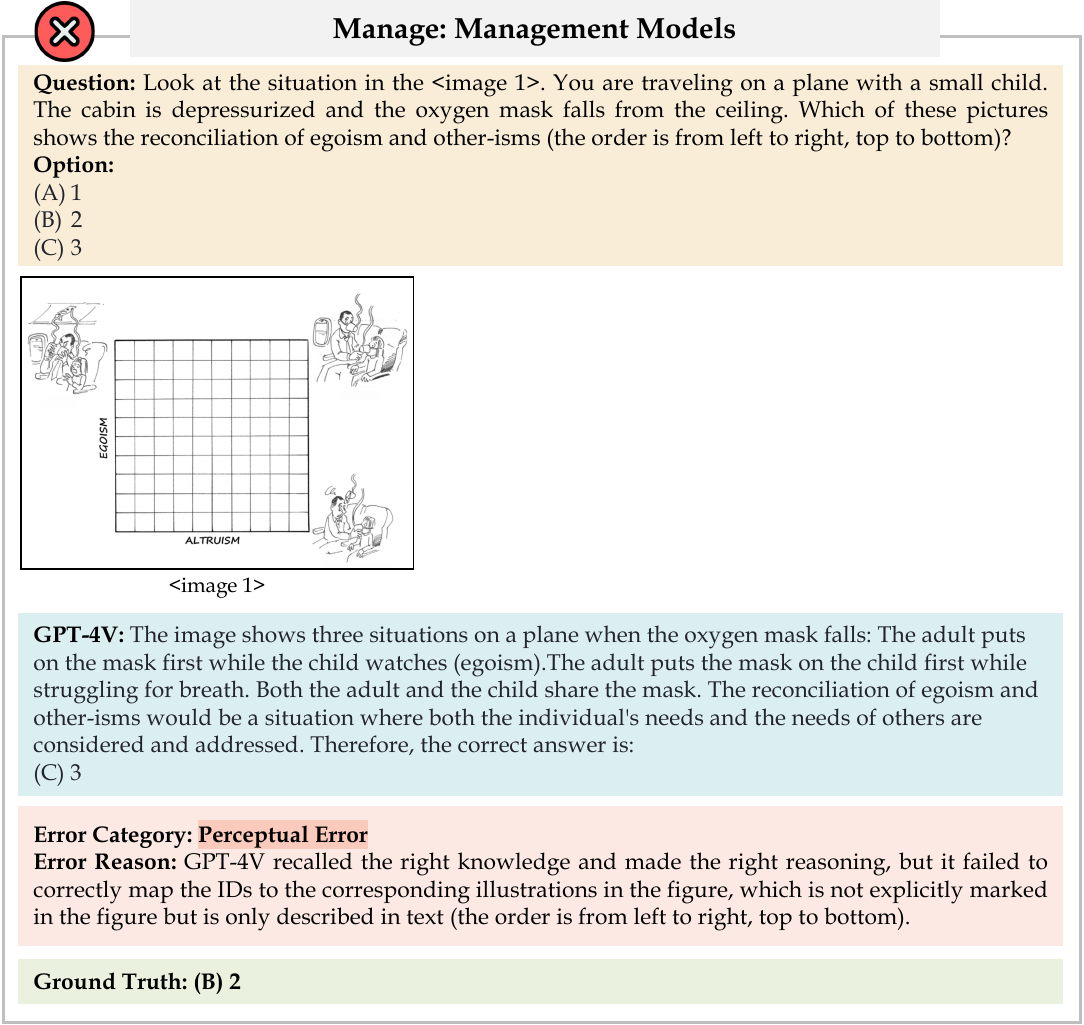}
    \caption{A sample error case of Manage (subfield: Management Models). Error category: Perceptual Error \newline \centering \hyperref[list:list_of_figures]{Back to List of Figures} \textcolor{red}{$|$} \hyperref[tab:list_of_case_study_figures]{Back to Table Index}}
    \addcontentsline{afg}{appfigures}{\protect\numberline{\thefigure}Manage  2: Perceptual Error}
\label{fig:manage_2}
\end{figure*}
\newpage


\begin{figure*}[!htbp]
    \centering
\includegraphics[width=0.9\linewidth]{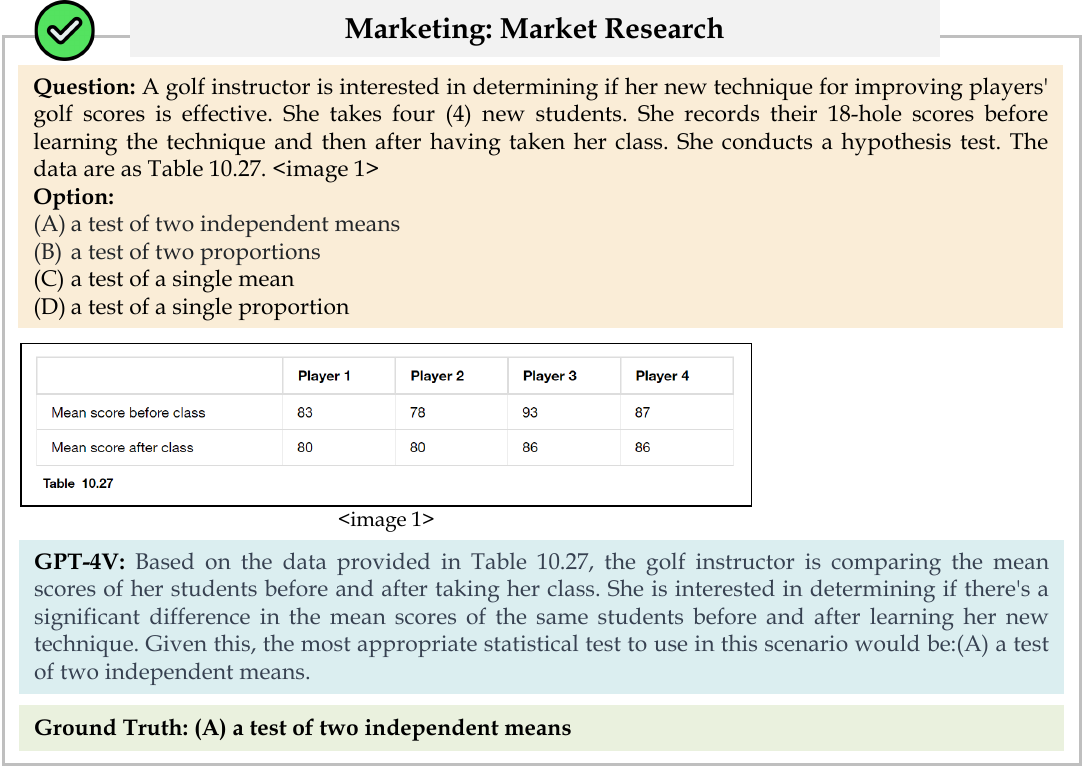}
    \caption{A sample correct case of Marketing (subfield: Market Research).\\ \hyperref[list:list_of_figures]{Back to List of Figures} \textcolor{red}{$|$} \hyperref[tab:list_of_case_study_figures]{Back to Table Index}}
    \addcontentsline{afg}{appfigures}{\protect\numberline{\thefigure}Marketing  1: Correct Case}
\label{fig:marketing_1}
\end{figure*}
\newpage

\begin{figure*}[!htbp]
    \centering
\includegraphics[width=0.9\linewidth]{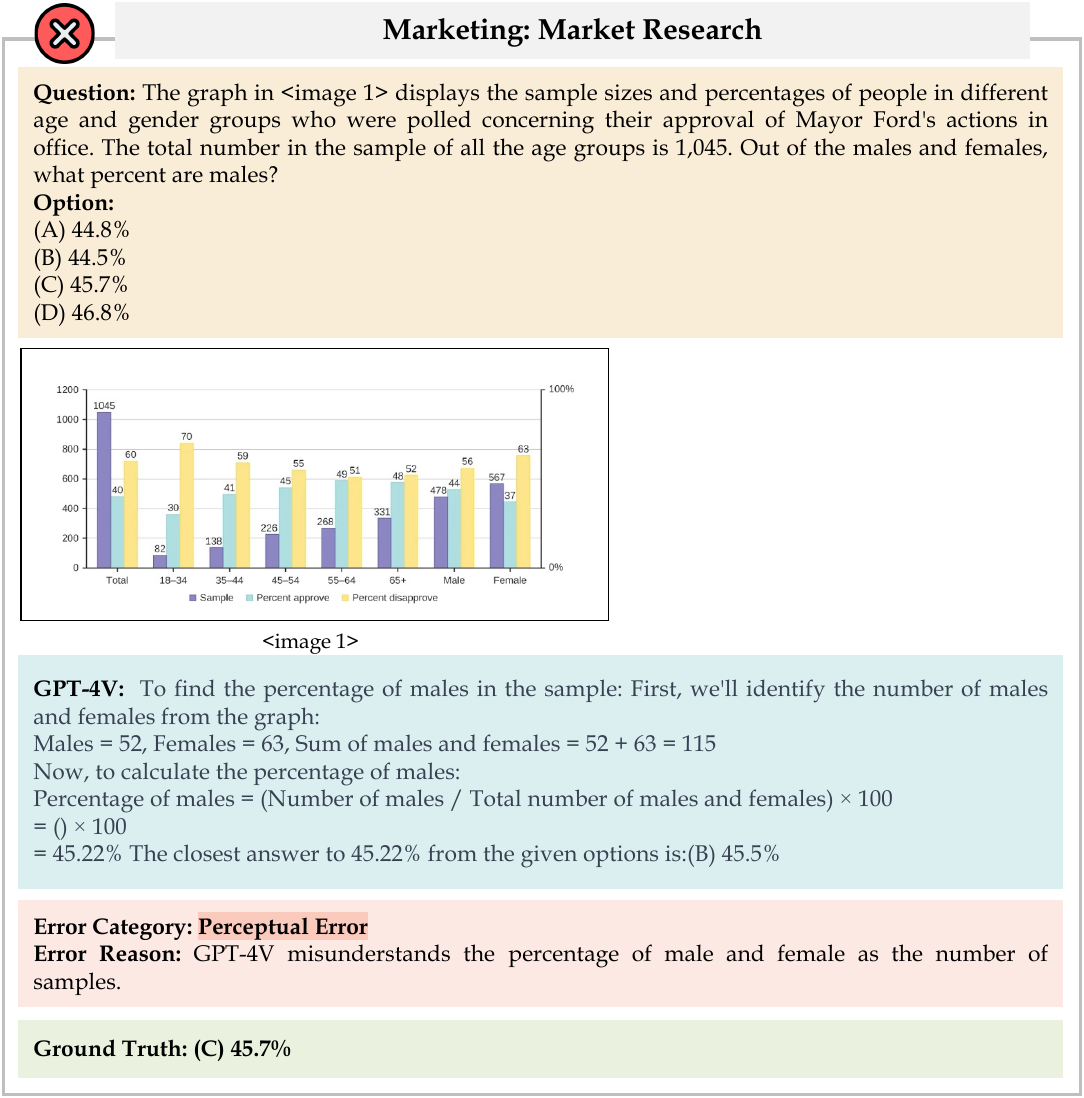}
    \caption{A sample error case of Marketing (subfield: Market Research). Error category: Perceptual Error \newline \centering \hyperref[list:list_of_figures]{Back to List of Figures} \textcolor{red}{$|$} \hyperref[tab:list_of_case_study_figures]{Back to Table Index}}
    \addcontentsline{afg}{appfigures}{\protect\numberline{\thefigure}Marketing  2: Perceptual Error}
\label{fig:marketing_2}
\end{figure*}
\newpage



\begin{figure*}[!htbp]
    \centering
\includegraphics[width=0.9\linewidth]{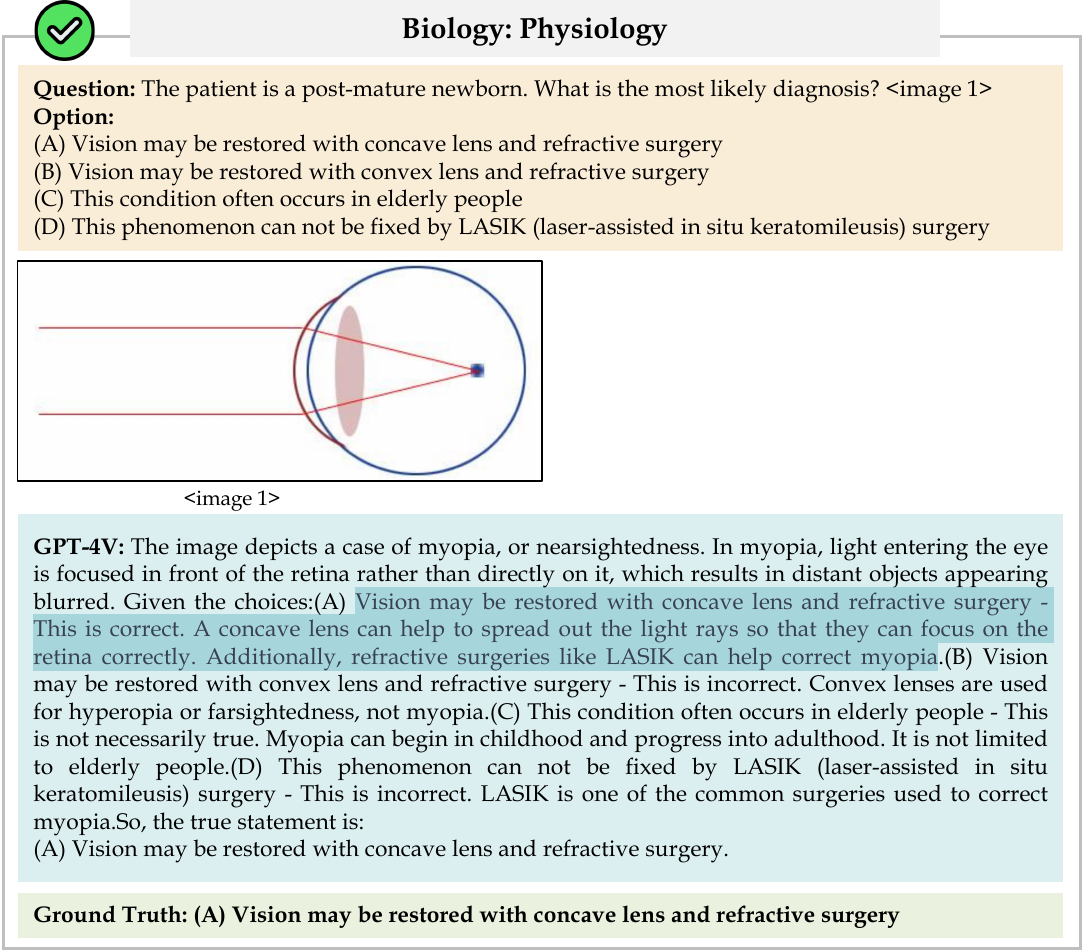}
    \caption{A sample correct case of Biology (subfield: Physiology).\\ \hyperref[list:list_of_figures]{Back to List of Figures} \textcolor{red}{$|$} \hyperref[tab:list_of_case_study_figures]{Back to Table Index}}
    \addcontentsline{afg}{appfigures}{\protect\numberline{\thefigure}Biology  1: Correct Case}
\label{fig:biology_2}
\end{figure*}
\newpage

\begin{figure*}[!htbp]
    \centering
\includegraphics[width=0.9\linewidth]{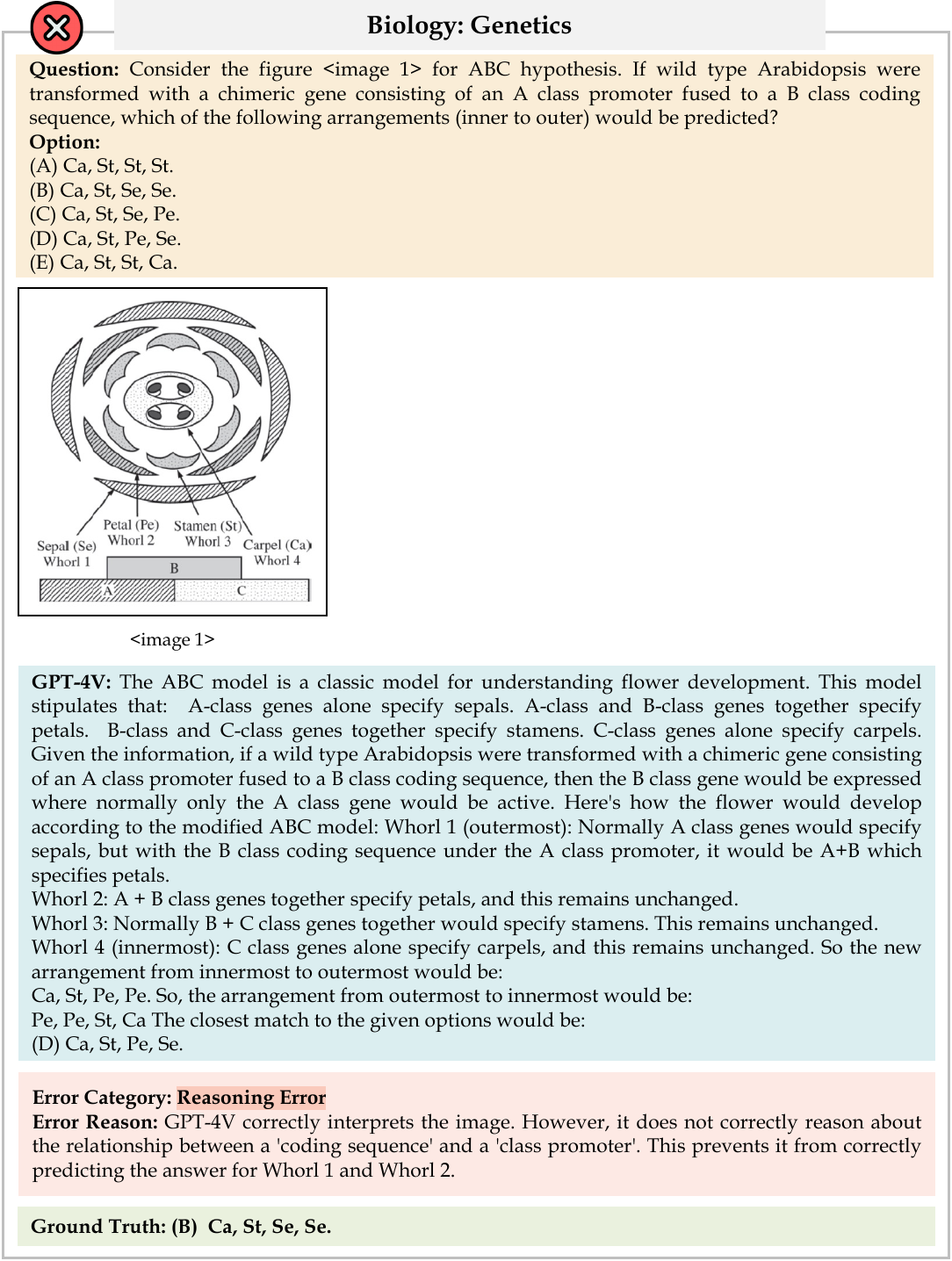}
    \caption{A sample error case of Biology (subfield: Genetics). Error category: Reasoning Error \newline \centering \hyperref[list:list_of_figures]{Back to List of Figures} \textcolor{red}{$|$} \hyperref[tab:list_of_case_study_figures]{Back to Table Index}}
    \addcontentsline{afg}{appfigures}{\protect\numberline{\thefigure}Biology  2: Reasoning Error}
\label{fig:biology_3}
\end{figure*}
\newpage

\begin{figure*}[!htbp]
    \centering
\includegraphics[width=0.9\linewidth]{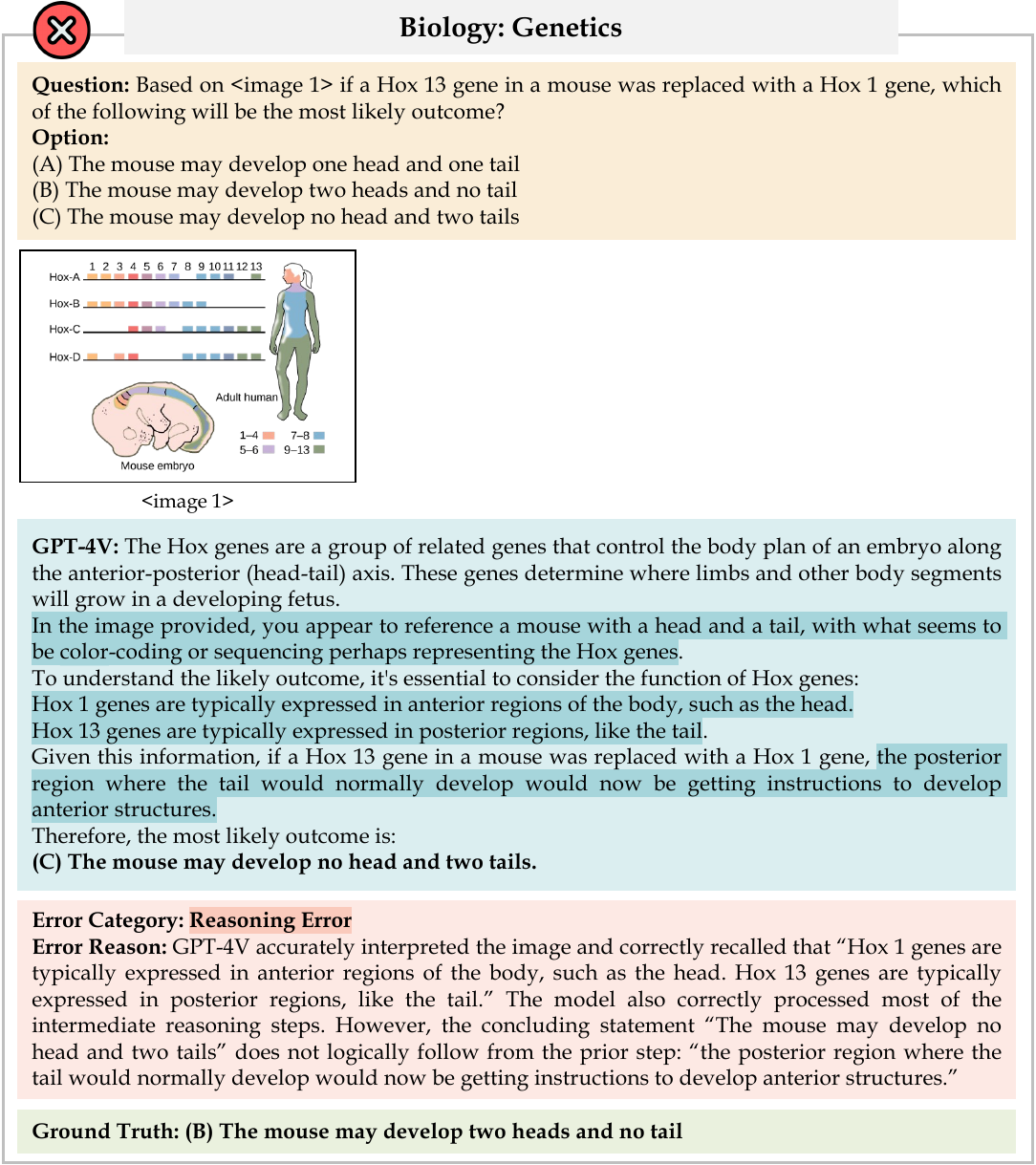}
    \caption{A sample error case of Biology (subfield: Genetics). Error category: Reasoning Error \newline \centering \hyperref[list:list_of_figures]{Back to List of Figures} \textcolor{red}{$|$} \hyperref[tab:list_of_case_study_figures]{Back to Table Index}}
    \addcontentsline{afg}{appfigures}{\protect\numberline{\thefigure}Biology  3: Reasoning Error}
\label{fig:biology_4}
\end{figure*}
\newpage

\begin{figure*}[!htbp]
    \centering
\includegraphics[width=0.9\linewidth]{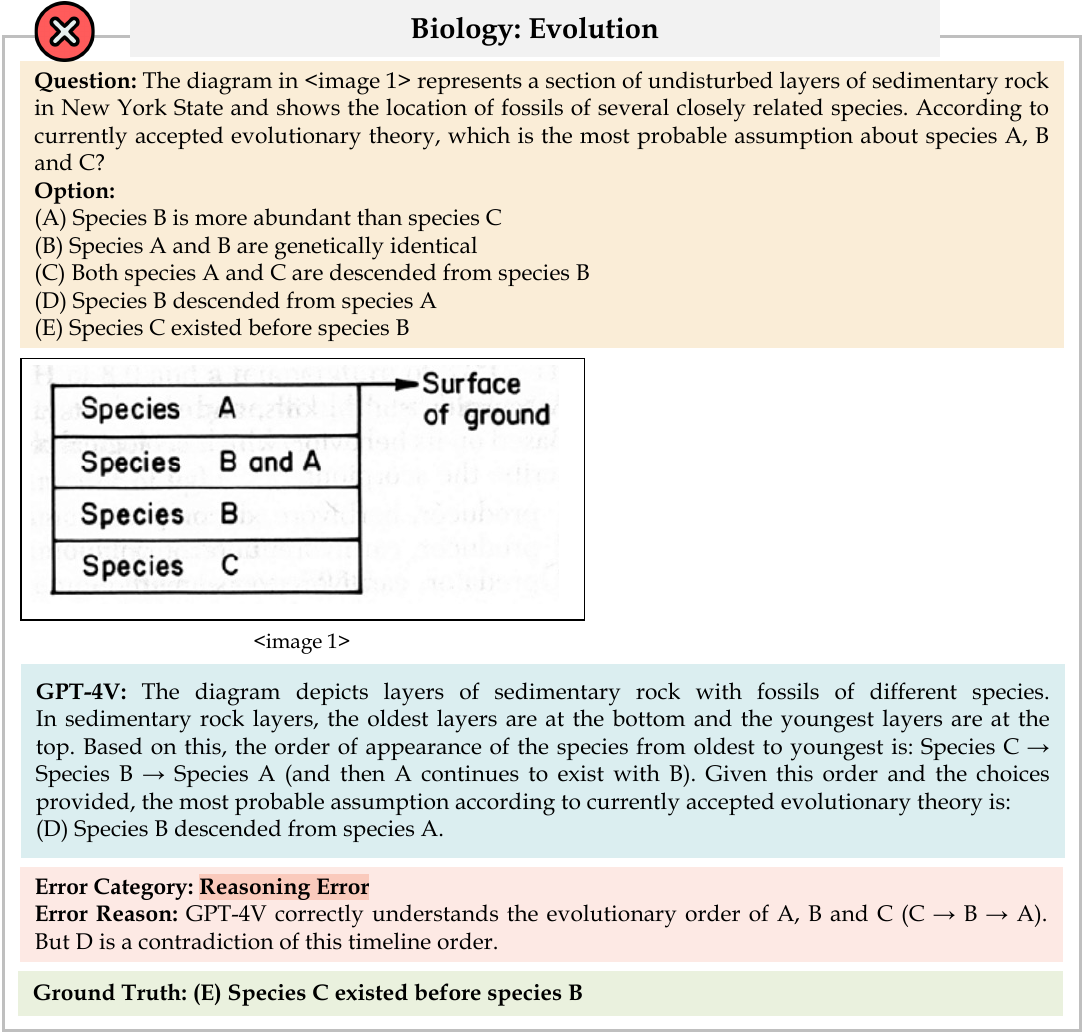}
    \caption{A sample error case of Biology (subfield: Evolution). Error category: Reasoning Error \newline \centering \hyperref[list:list_of_figures]{Back to List of Figures} \textcolor{red}{$|$} \hyperref[tab:list_of_case_study_figures]{Back to Table Index}}
    \addcontentsline{afg}{appfigures}{\protect\numberline{\thefigure}Biology  4: Reasoning Error}
\label{fig:biology_5}
\end{figure*}
\newpage

\begin{figure*}[!htbp]
    \centering
\includegraphics[width=0.9\linewidth]{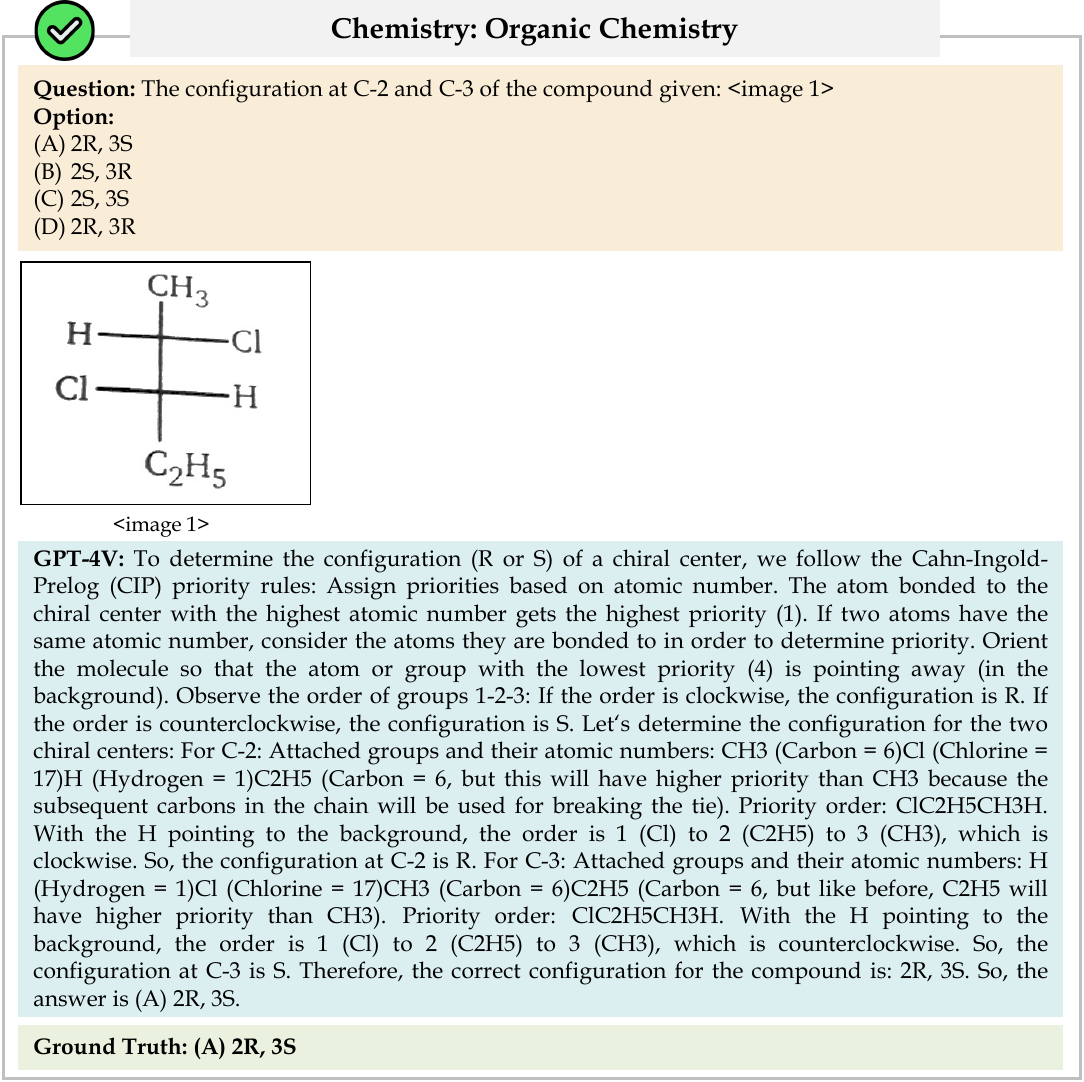}
    \caption{A sample correct case of Chemistry (subfield: Organic Chemistry).\\ \hyperref[list:list_of_figures]{Back to List of Figures} \textcolor{red}{$|$} \hyperref[tab:list_of_case_study_figures]{Back to Table Index}}
    \addcontentsline{afg}{appfigures}{\protect\numberline{\thefigure}Chemistry  1: Correct Case}
\label{fig:chemistry_1}
\end{figure*}
\newpage

\begin{figure*}[!htbp]
    \centering
\includegraphics[width=0.9\linewidth]{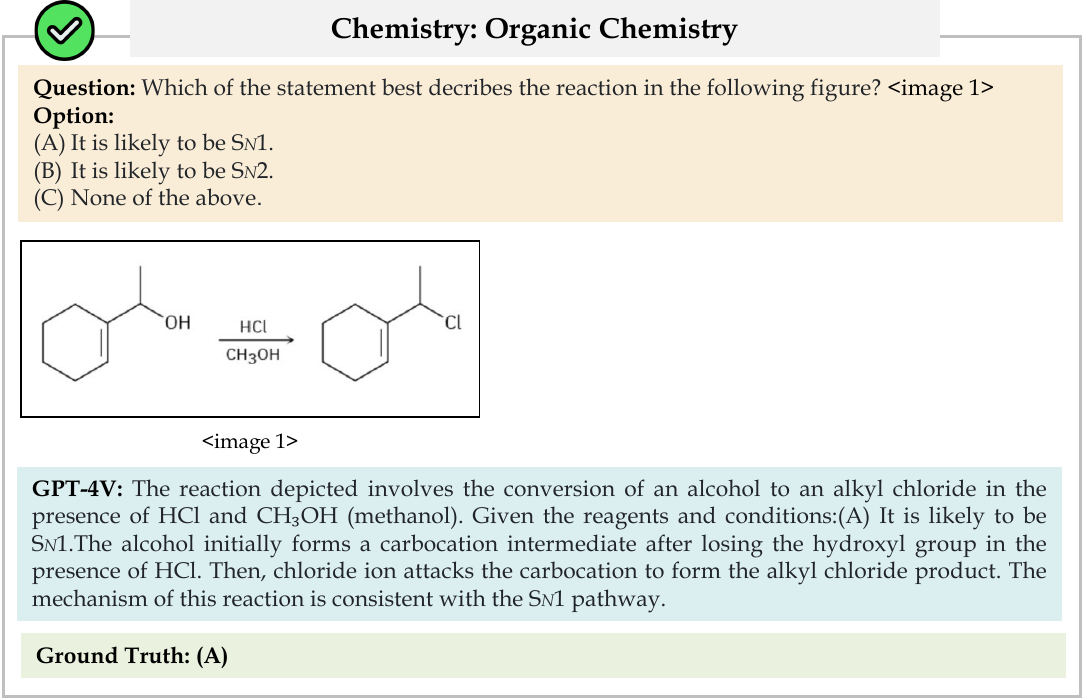}
    \caption{A sample correct case of Chemistry (subfield: Organic Chemistry).\\ \hyperref[list:list_of_figures]{Back to List of Figures} \textcolor{red}{$|$} \hyperref[tab:list_of_case_study_figures]{Back to Table Index}}
    \addcontentsline{afg}{appfigures}{\protect\numberline{\thefigure}Chemistry  2: Correct Case}
\label{fig:chemistry_2}
\end{figure*}
\newpage

\begin{figure*}[!htbp]
    \centering
\includegraphics[width=0.9\linewidth]{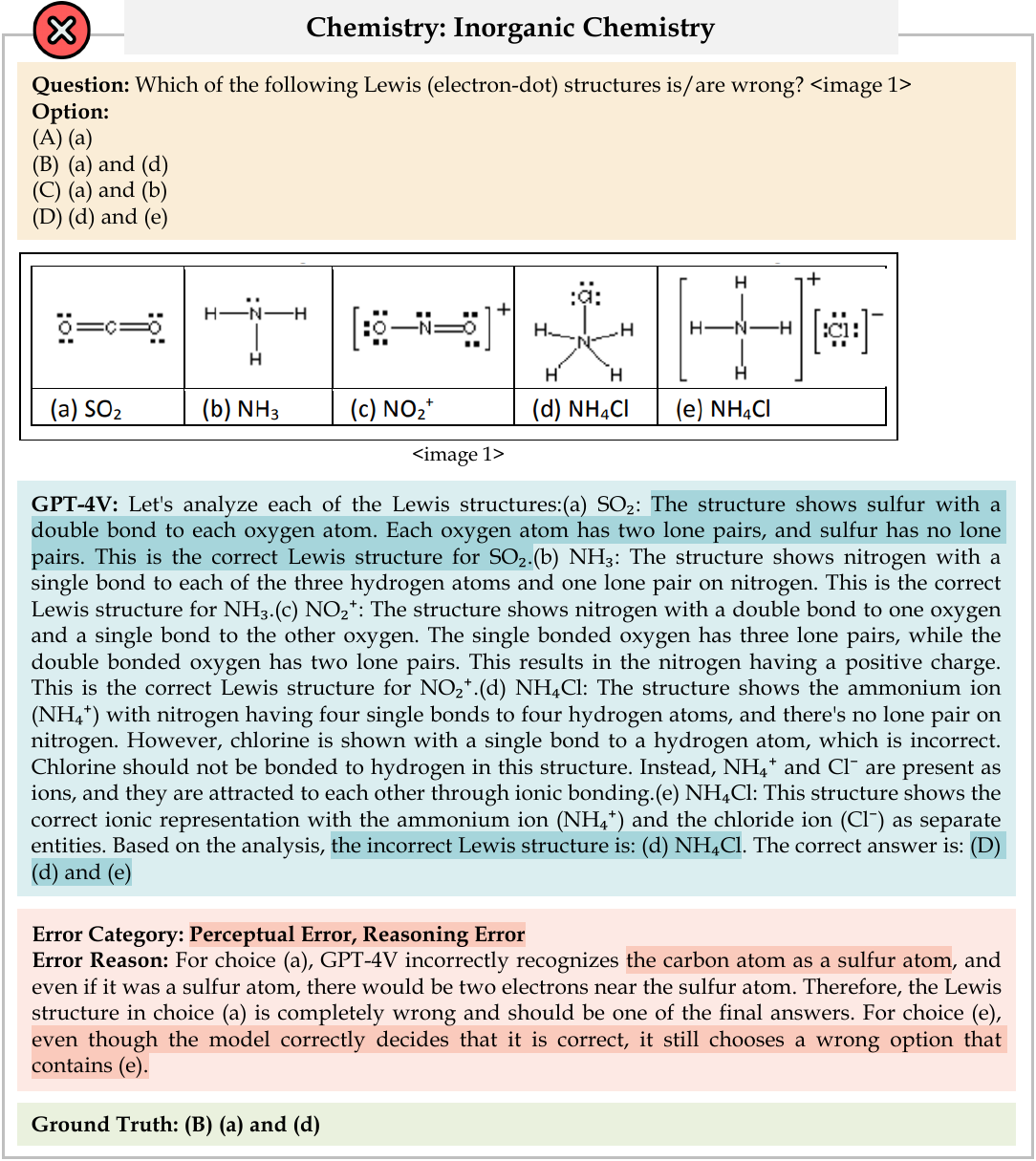}
    \caption{A sample error case of Chemistry (subfield: Inorganic Chemistry). Error category: Perceptual Error, Reasoning Error \newline \centering \hyperref[list:list_of_figures]{Back to List of Figures} \textcolor{red}{$|$} \hyperref[tab:list_of_case_study_figures]{Back to Table Index}}
    \addcontentsline{afg}{appfigures}{\protect\numberline{\thefigure}Chemistry  3: Perceptual Error, Reasoning Error}
\label{fig:chemistry_3}
\end{figure*}
\newpage

\begin{figure*}[!htbp]
    \centering
\includegraphics[width=0.9\linewidth]{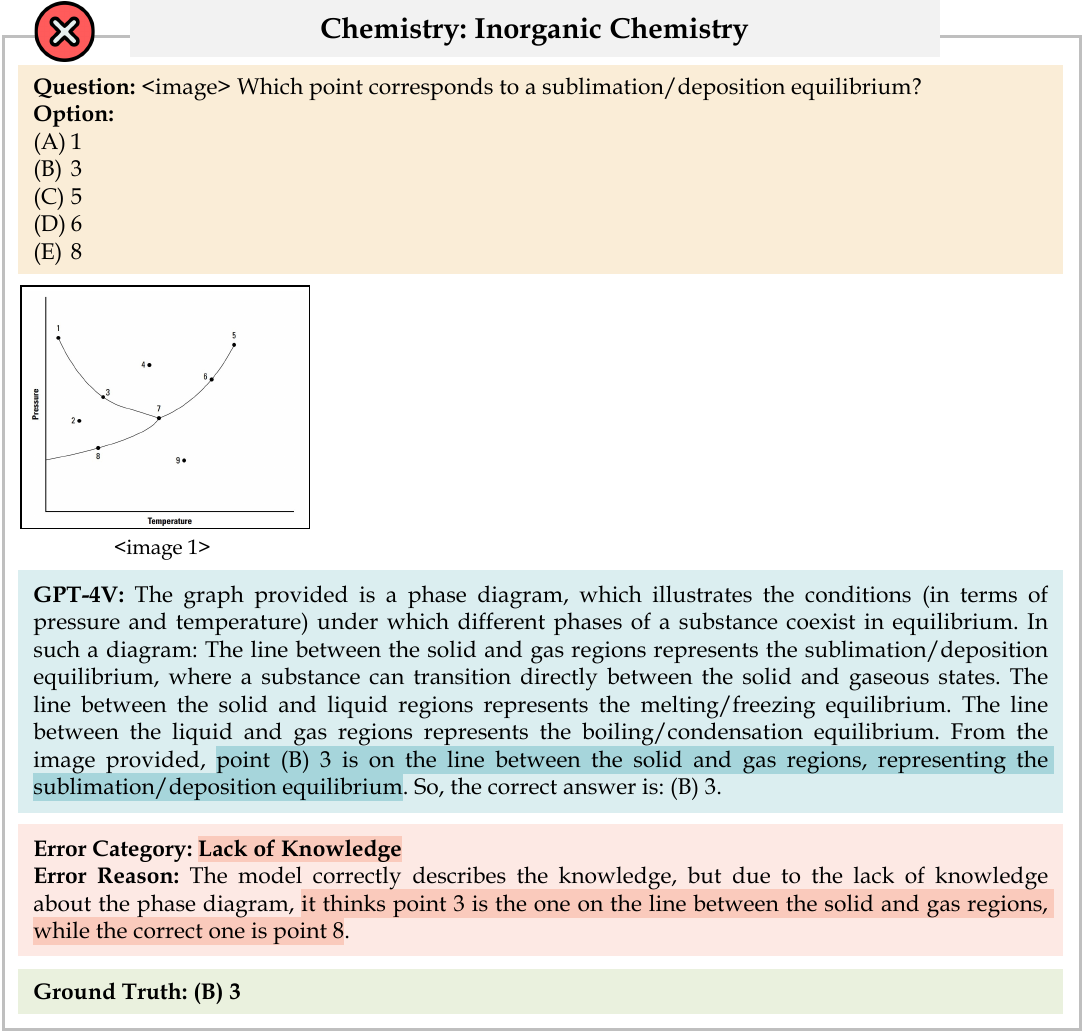}
    \caption{A sample error case of Chemistry (subfield: Inorganic Chemistry). Error category: Lack of Knowledge \newline \centering \hyperref[list:list_of_figures]{Back to List of Figures} \textcolor{red}{$|$} \hyperref[tab:list_of_case_study_figures]{Back to Table Index}}
    \addcontentsline{afg}{appfigures}{\protect\numberline{\thefigure}Chemistry  4: Lack of Knowledge}
\label{fig:chemistry_4}
\end{figure*}
\newpage

\begin{figure*}[!htbp]
    \centering
\includegraphics[width=0.9\linewidth]{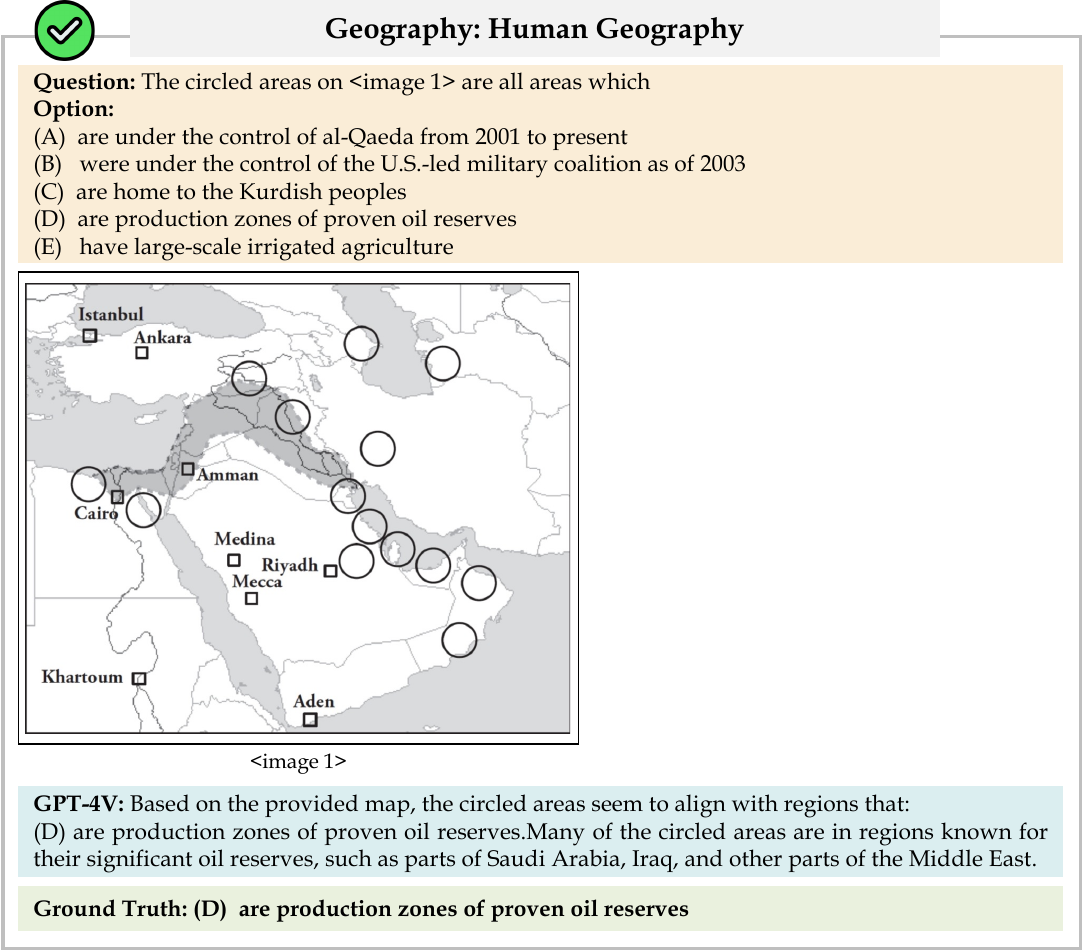}
    \caption{A sample correct case of Geography (subfield: Human Geography).\\ \hyperref[list:list_of_figures]{Back to List of Figures} \textcolor{red}{$|$} \hyperref[tab:list_of_case_study_figures]{Back to Table Index}}
    \addcontentsline{afg}{appfigures}{\protect\numberline{\thefigure}Geography  1: Correct Case}
\label{fig:geography_1}
\end{figure*}
\newpage

\begin{figure*}[!htbp]
    \centering
\includegraphics[width=0.9\linewidth]{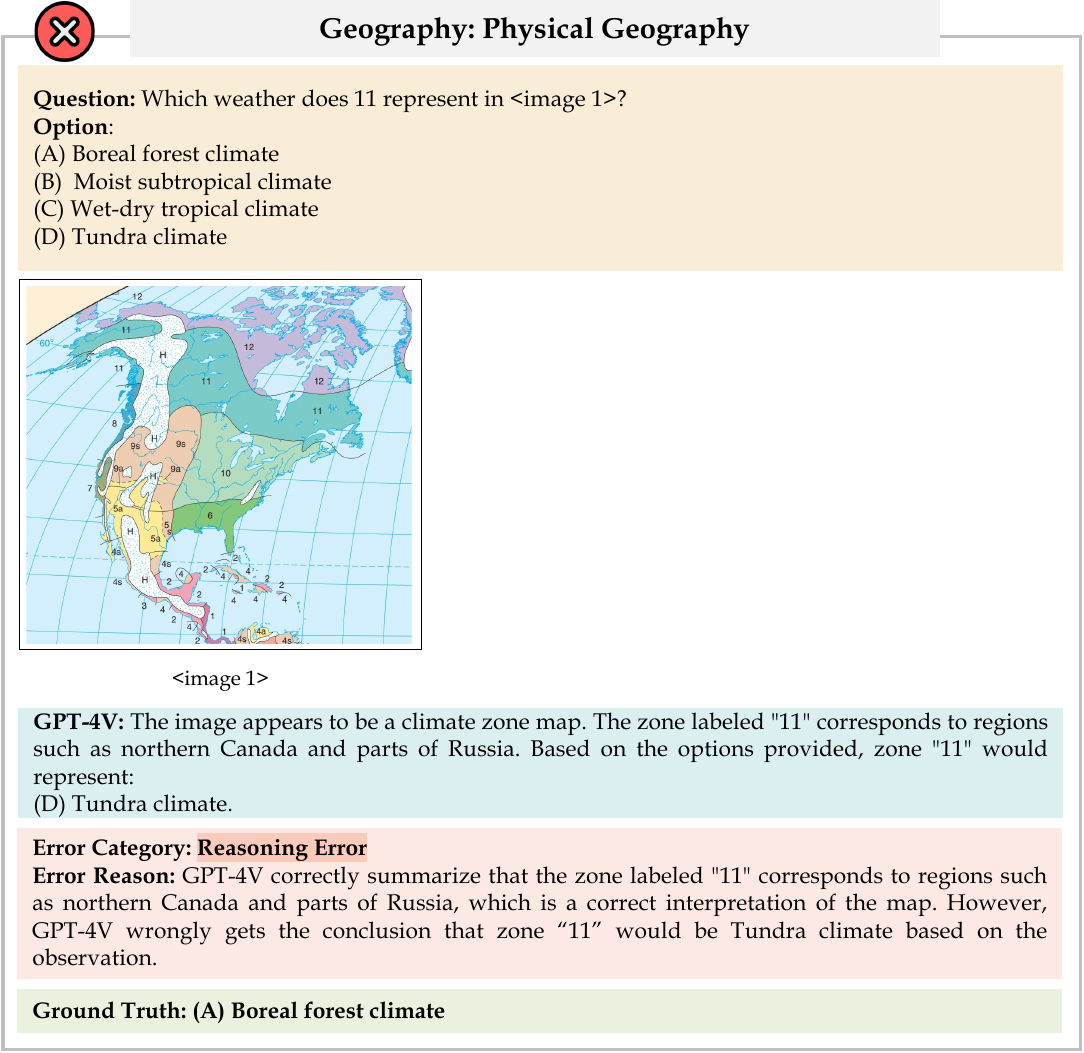}
    \caption{A sample error case of Geography (subfield: Physical Geography). Error category: Reasoning Error \newline \centering \hyperref[list:list_of_figures]{Back to List of Figures} \textcolor{red}{$|$} \hyperref[tab:list_of_case_study_figures]{Back to Table Index}}
    \addcontentsline{afg}{appfigures}{\protect\numberline{\thefigure}Geography  2: Reasoning Error}
\label{fig:geography_2}
\end{figure*}
\newpage

\begin{figure*}[!htbp]
    \centering
\includegraphics[width=0.9\linewidth]{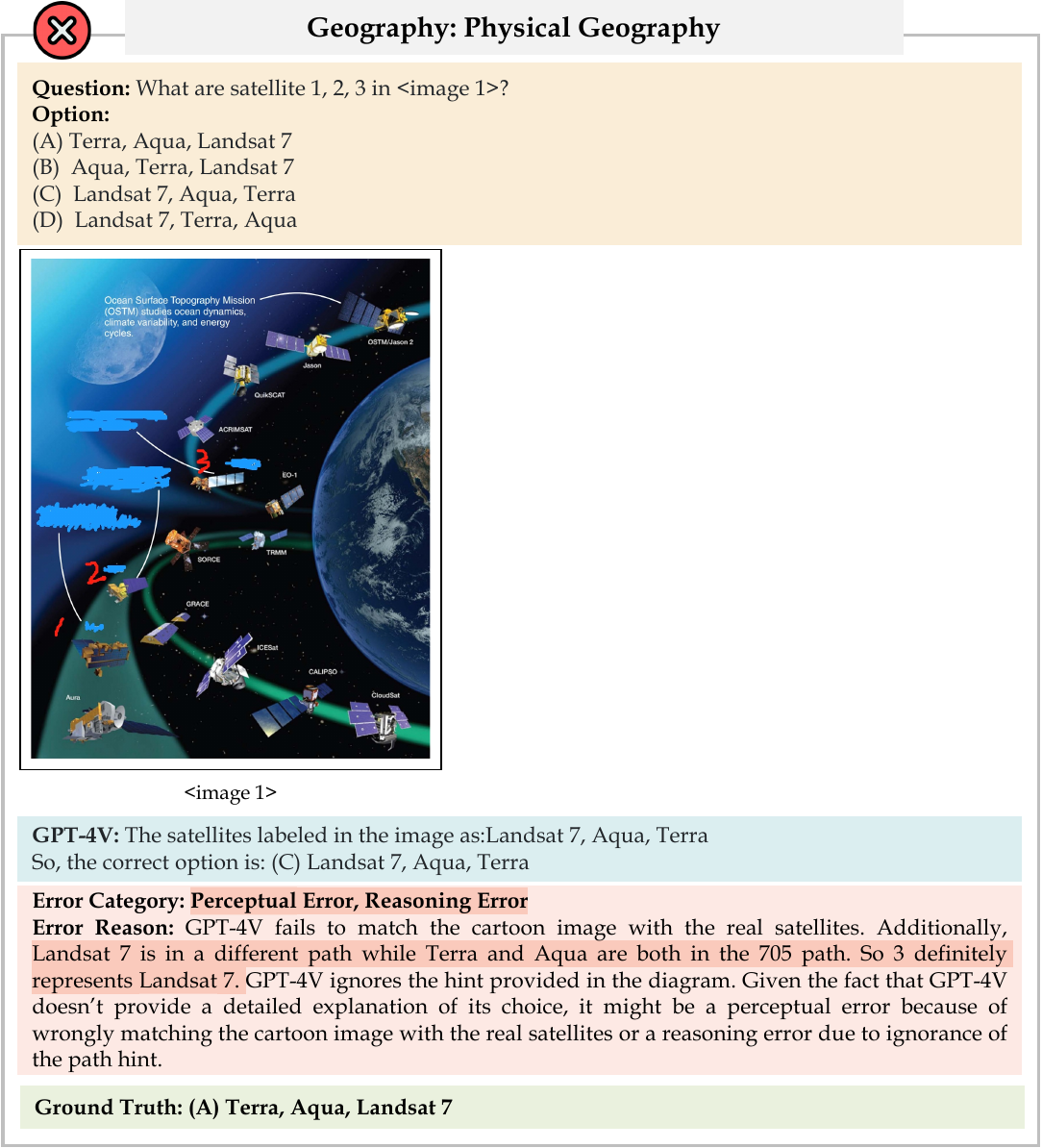}
    \caption{A sample error case of Geography (subfield: Physical Geography). Error category: Perceptual Error, Reasoning Error \newline \centering \hyperref[list:list_of_figures]{Back to List of Figures} \textcolor{red}{$|$} \hyperref[tab:list_of_case_study_figures]{Back to Table Index}}
    \addcontentsline{afg}{appfigures}{\protect\numberline{\thefigure}Geography  3: Perceptual Error, Reasoning Error}
\label{fig:geography_3}
\end{figure*}
\newpage

\begin{figure*}[!htbp]
    \centering
\includegraphics[width=0.9\linewidth]{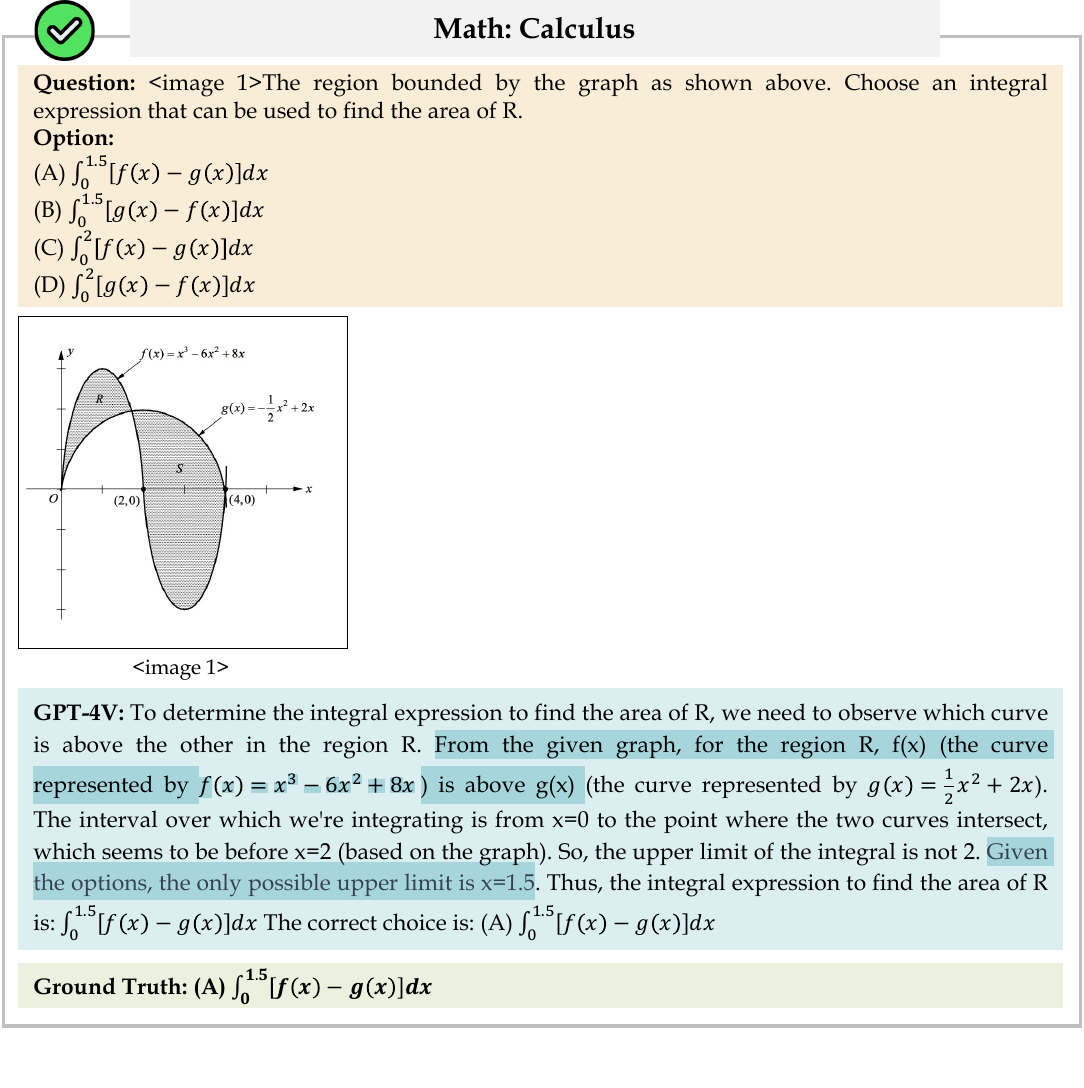}
    \caption{A sample correct case of Math (subfield: Calculus).\\ \hyperref[list:list_of_figures]{Back to List of Figures} \textcolor{red}{$|$} \hyperref[tab:list_of_case_study_figures]{Back to Table Index}}
    \addcontentsline{afg}{appfigures}{\protect\numberline{\thefigure}Math  1: Correct Case}
\label{fig:math_1}
\end{figure*}
\newpage

\begin{figure*}[!htbp]
    \centering
\includegraphics[width=0.9\linewidth]{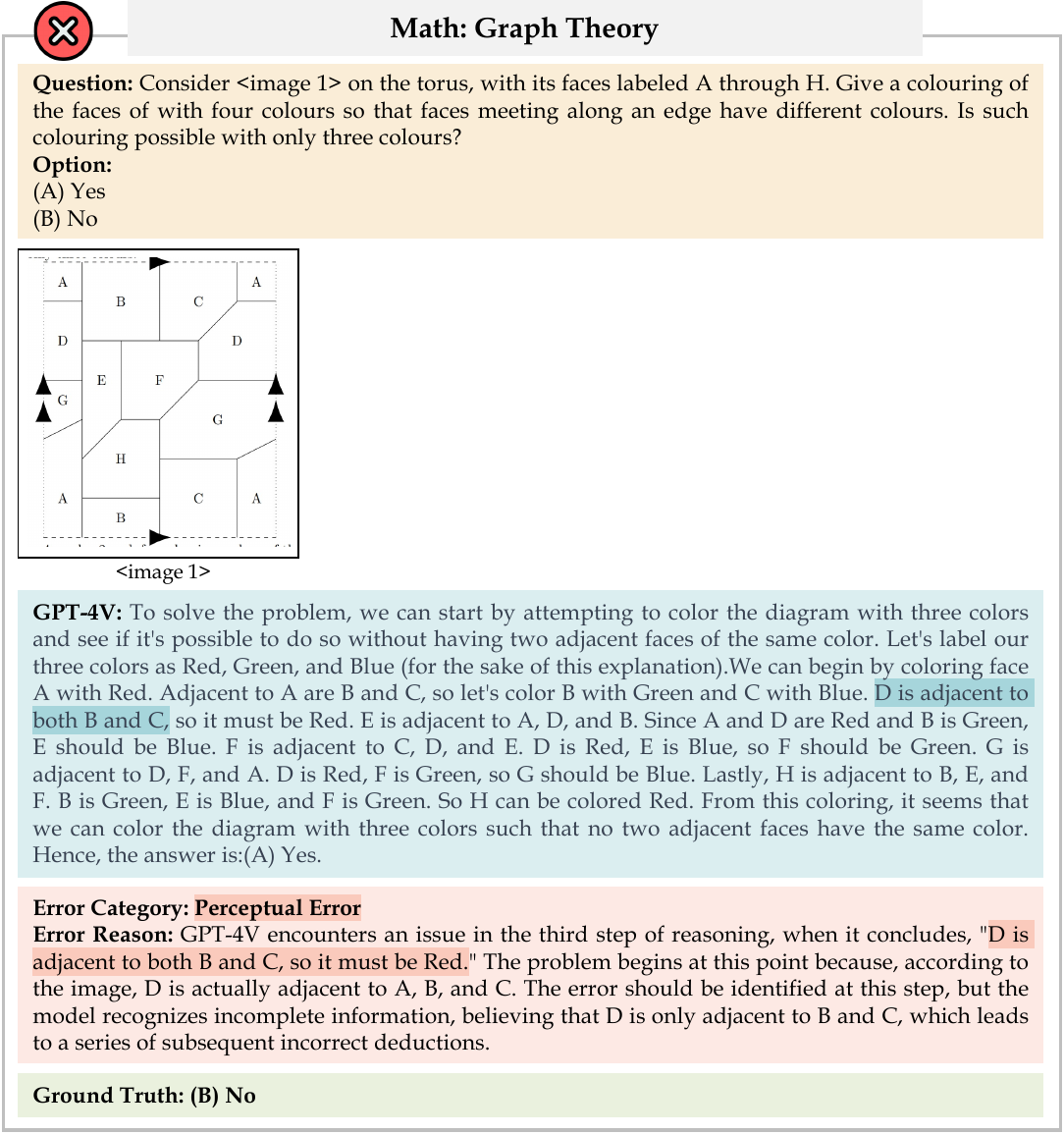}
    \caption{A sample error case of Math (subfield: Graph Theory). Error category: Perceptual Error \newline \centering \hyperref[list:list_of_figures]{Back to List of Figures} \textcolor{red}{$|$} \hyperref[tab:list_of_case_study_figures]{Back to Table Index}}
    \addcontentsline{afg}{appfigures}{\protect\numberline{\thefigure}Math  2: Perceptual Error}
\label{fig:math_2}
\end{figure*}
\newpage

\begin{figure*}[!htbp]
    \centering
\includegraphics[width=0.9\linewidth]{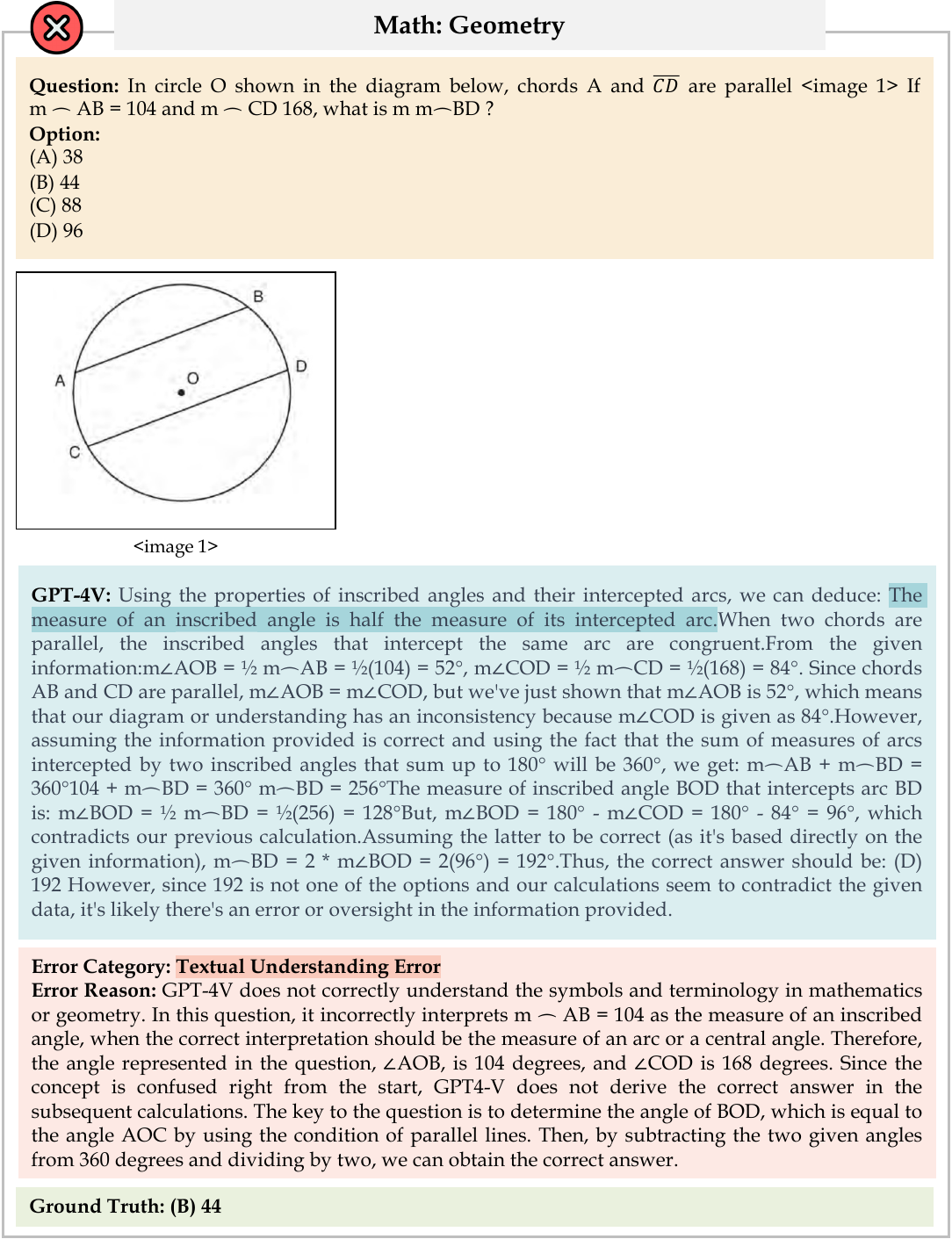}
    \caption{A sample error case of Math (subfield: Geometry). Error category: Textual Understanding Error \newline \centering \hyperref[list:list_of_figures]{Back to List of Figures} \textcolor{red}{$|$} \hyperref[tab:list_of_case_study_figures]{Back to Table Index}}
    \addcontentsline{afg}{appfigures}{\protect\numberline{\thefigure}Math  3: Textual Understanding Error}
\label{fig:math_3}
\end{figure*}
\newpage

\begin{figure*}[!htbp]
    \centering
\includegraphics[width=0.9\linewidth]{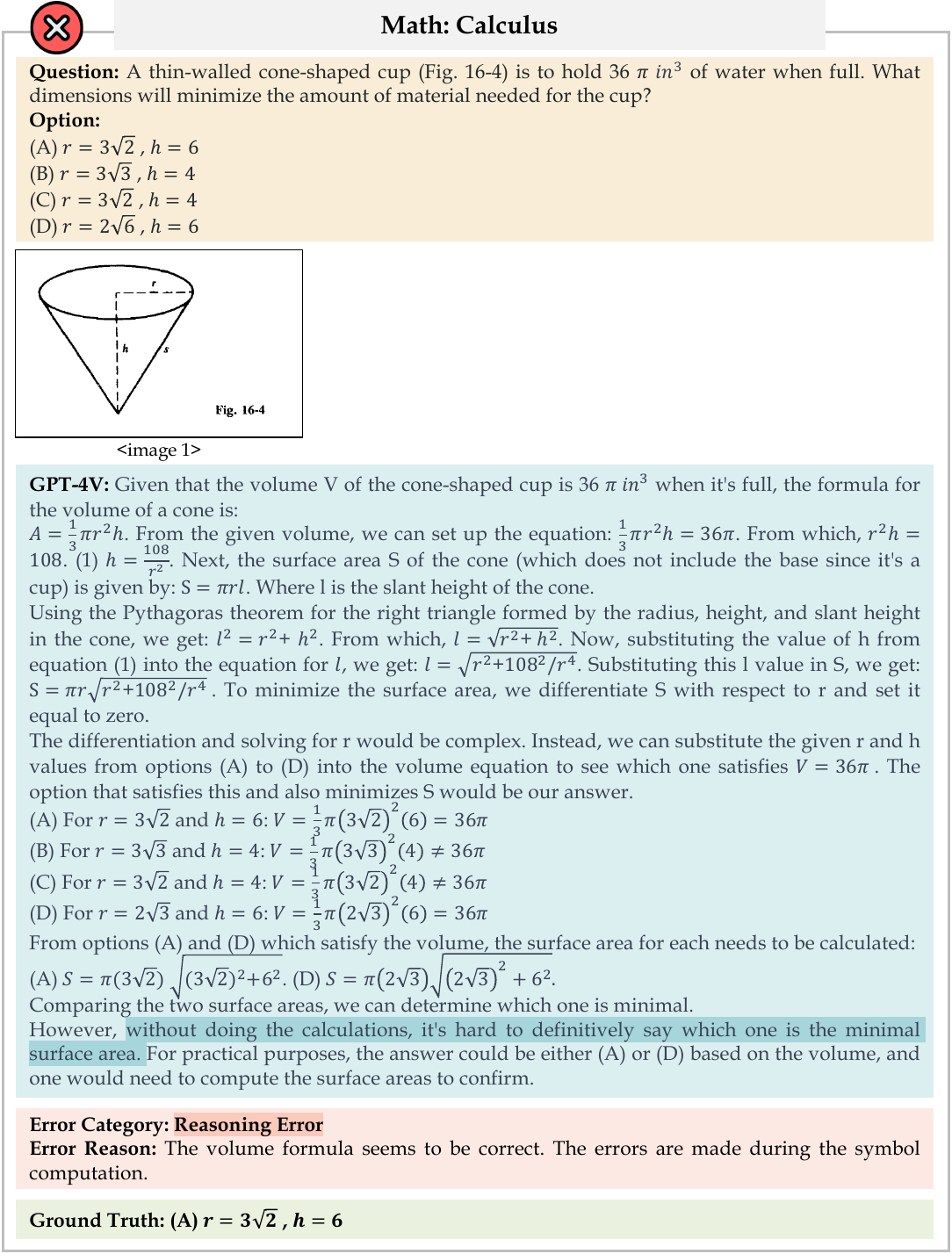}
    \caption{A sample error case of Math (subfield: Calculus). Error category: Reasoning Error \newline \centering \hyperref[list:list_of_figures]{Back to List of Figures} \textcolor{red}{$|$} \hyperref[tab:list_of_case_study_figures]{Back to Table Index}}
    \addcontentsline{afg}{appfigures}{\protect\numberline{\thefigure}Math  4: Reasoning Error}
\label{fig:math_4}
\end{figure*}
\newpage

\begin{figure*}[!htbp]
    \centering
\includegraphics[width=0.9\linewidth]{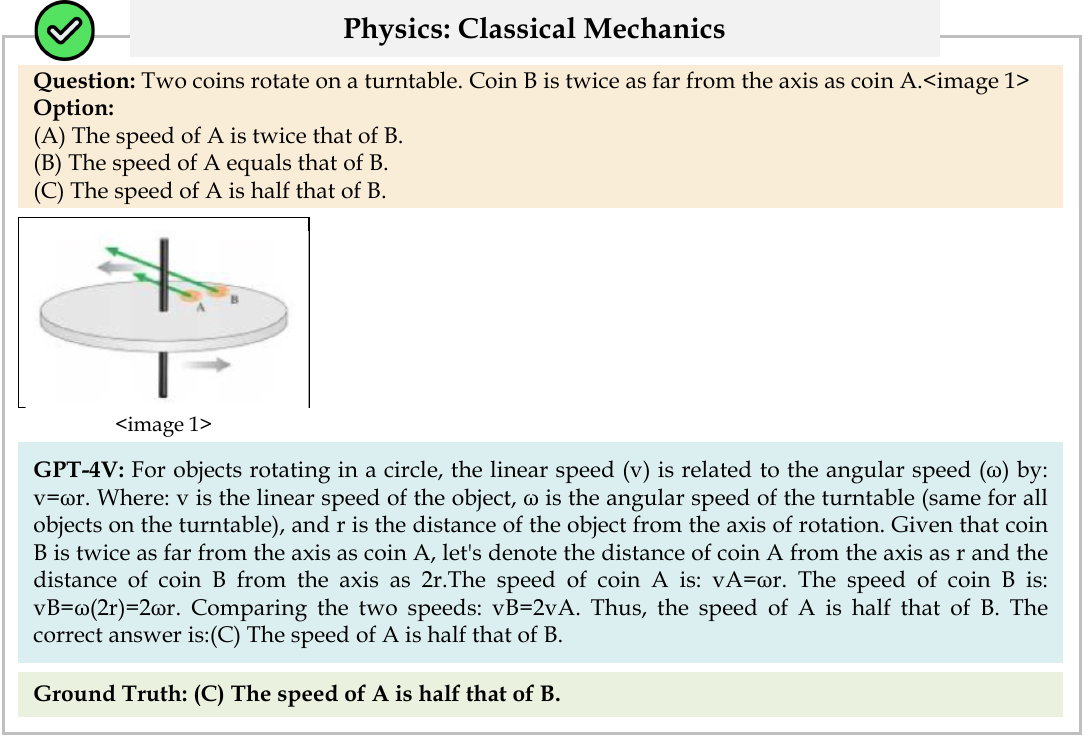}
    \caption{A sample correct case of Physics (subfield: Classical Mechanics).\\ \hyperref[list:list_of_figures]{Back to List of Figures} \textcolor{red}{$|$} \hyperref[tab:list_of_case_study_figures]{Back to Table Index}}
    \addcontentsline{afg}{appfigures}{\protect\numberline{\thefigure}Physics  1: Correct Case}
\label{fig:physics_1}
\end{figure*}
\newpage

\begin{figure*}[!htbp]
    \centering
\includegraphics[width=0.9\linewidth]{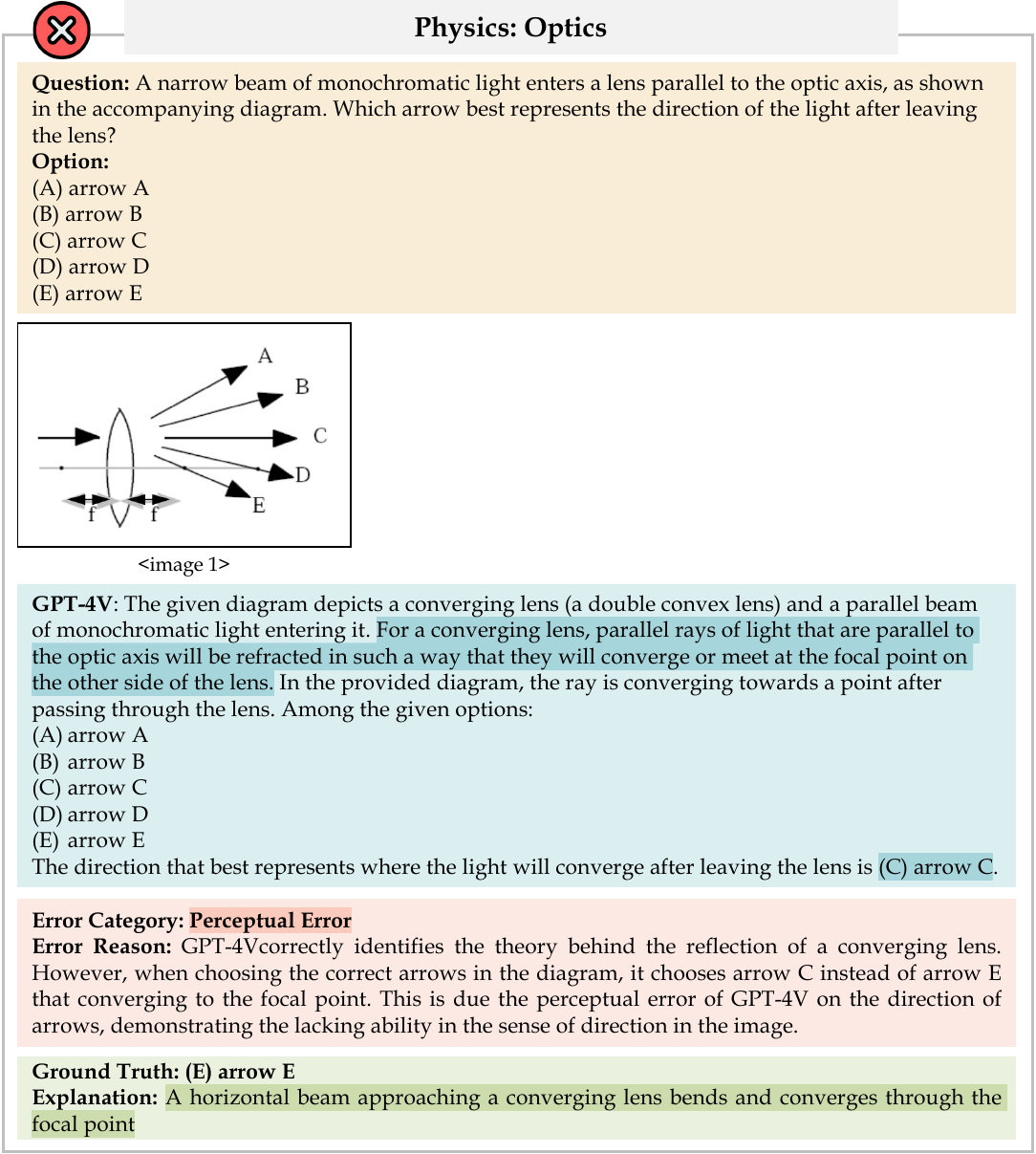}
    \caption{A sample error case of Physics (subfield: Optics). Error category: Perceptual Error \newline \centering \hyperref[list:list_of_figures]{Back to List of Figures} \textcolor{red}{$|$} \hyperref[tab:list_of_case_study_figures]{Back to Table Index}}
    \addcontentsline{afg}{appfigures}{\protect\numberline{\thefigure}Physics  2: Perceptual Error}
\label{fig:physics_2}
\end{figure*}
\newpage


\begin{figure*}[!htbp]
    \centering
\includegraphics[width=0.9\linewidth]{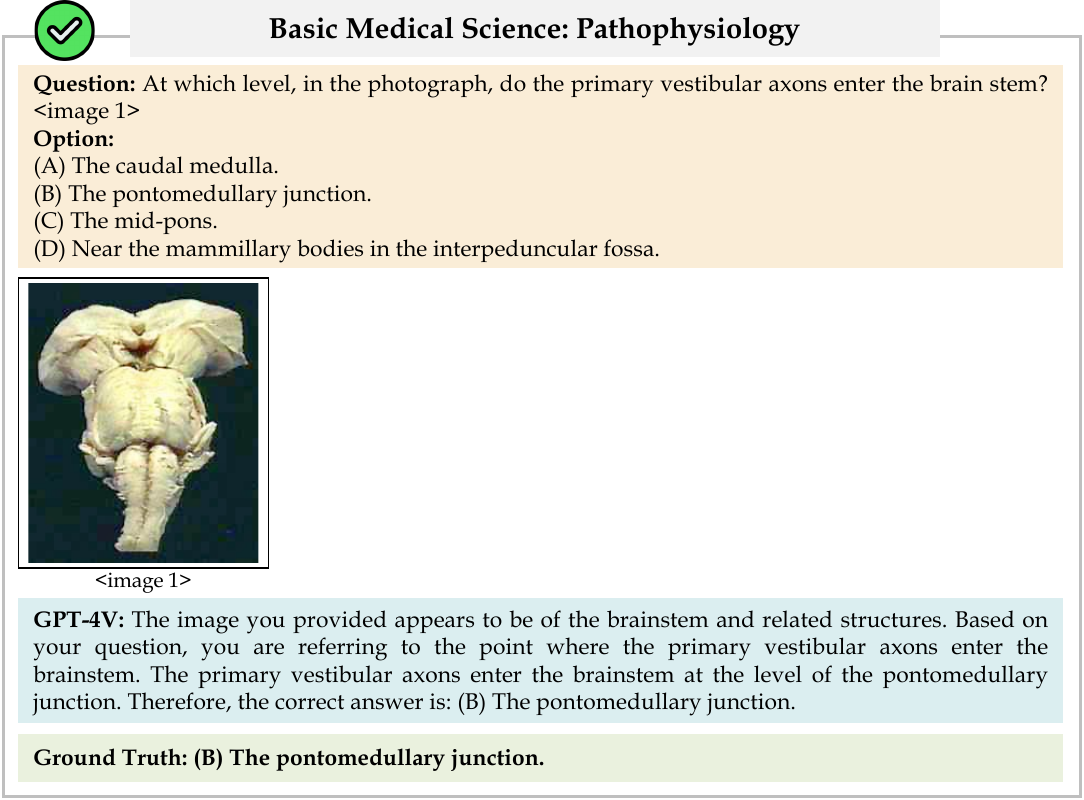}
    \caption{A sample correct case of Basic Medical Science (subfield: Pathophysiology).\\ \hyperref[list:list_of_figures]{Back to List of Figures} \textcolor{red}{$|$} \hyperref[tab:list_of_case_study_figures]{Back to Table Index}}
    \addcontentsline{afg}{appfigures}{\protect\numberline{\thefigure}Basic Medical Science  1: Correct Case}
\label{fig:basic_medical_science_1}
\end{figure*}
\newpage

\begin{figure*}[!htbp]
    \centering
\includegraphics[width=0.9\linewidth]{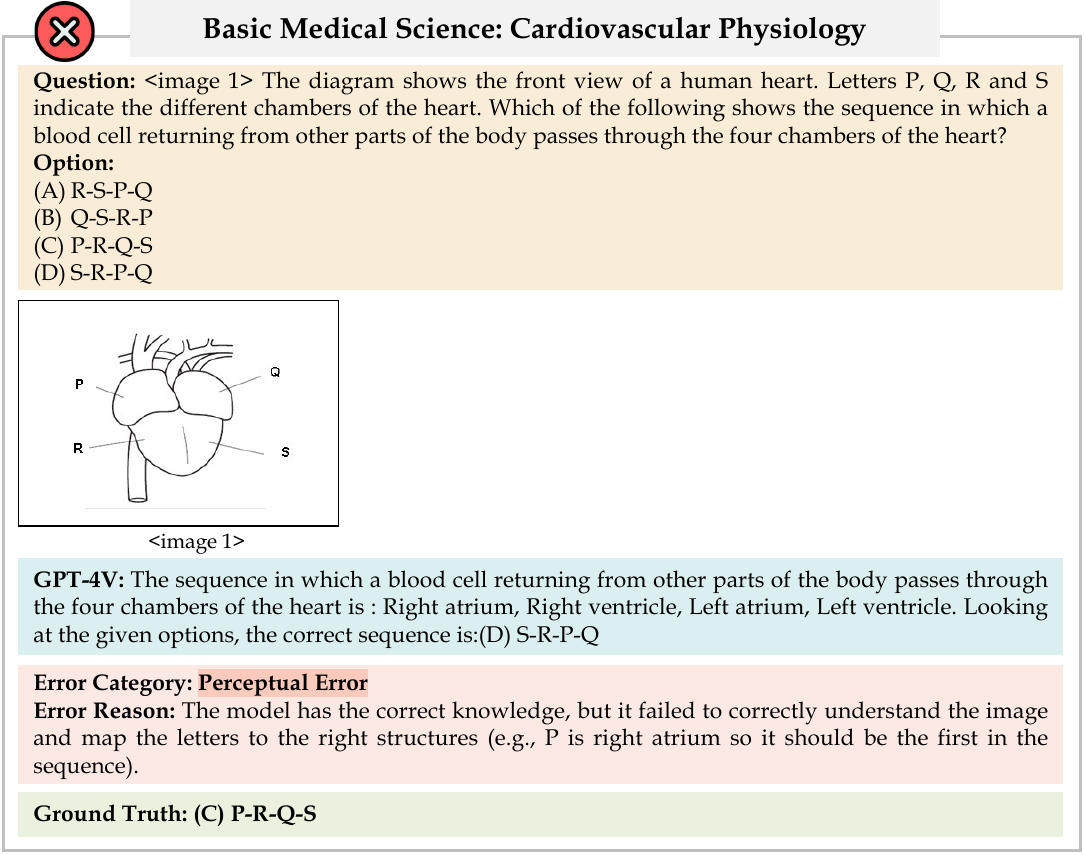}
    \caption{A sample error case of Basic Medical Science (subfield: Cardiovascular Physiology). Error category: Perceptual Error \newline \centering \hyperref[list:list_of_figures]{Back to List of Figures} \textcolor{red}{$|$} \hyperref[tab:list_of_case_study_figures]{Back to Table Index}}
    \addcontentsline{afg}{appfigures}{\protect\numberline{\thefigure}Basic Medical Science  2: Perceptual Error}
\label{fig:basic_medical_science_2}
\end{figure*}
\newpage


\begin{figure*}[!htbp]
    \centering
\includegraphics[width=0.9\linewidth]{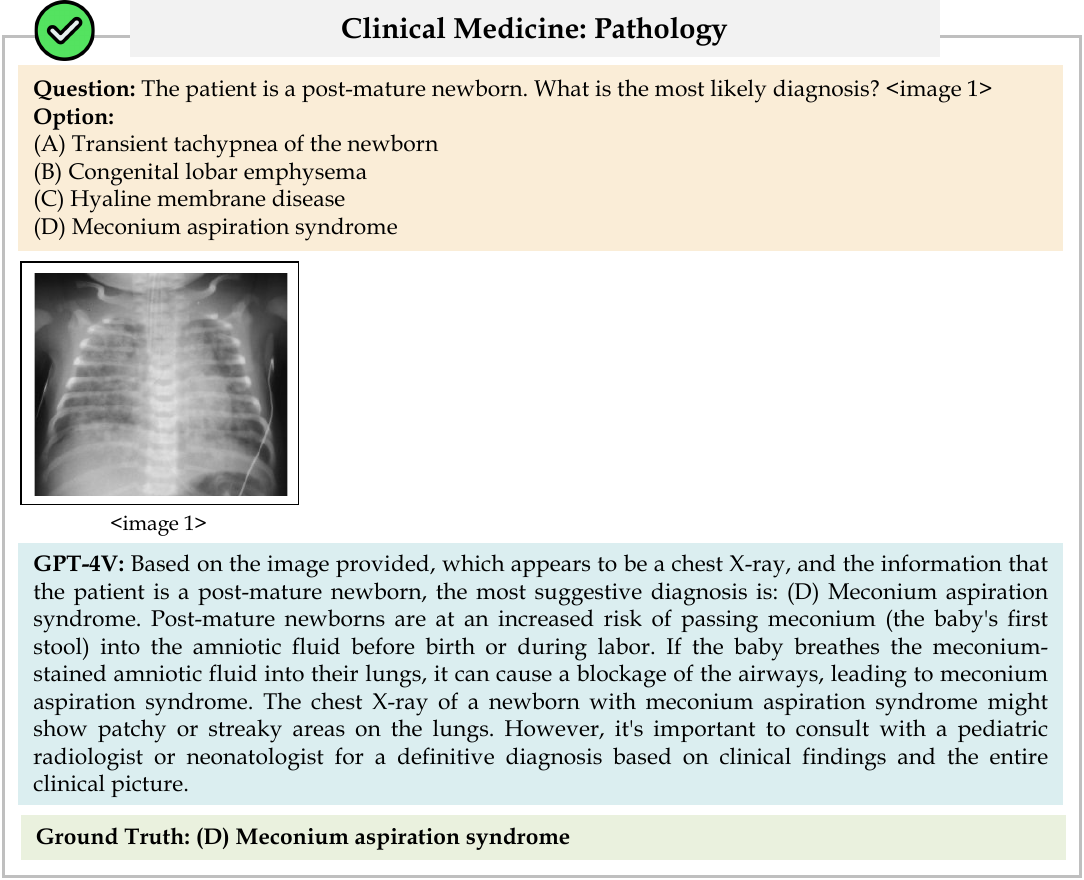}
    \caption{A sample correct case of Clinical Medicine (subfield: Pathology).\\ \hyperref[list:list_of_figures]{Back to List of Figures} \textcolor{red}{$|$} \hyperref[tab:list_of_case_study_figures]{Back to Table Index}}
    \addcontentsline{afg}{appfigures}{\protect\numberline{\thefigure}Clinical Medicine  1: Correct Case}
\label{fig:clinical_medicine_1}
\end{figure*}
\newpage

\begin{figure*}[!htbp]
    \centering
\includegraphics[width=0.9\linewidth]{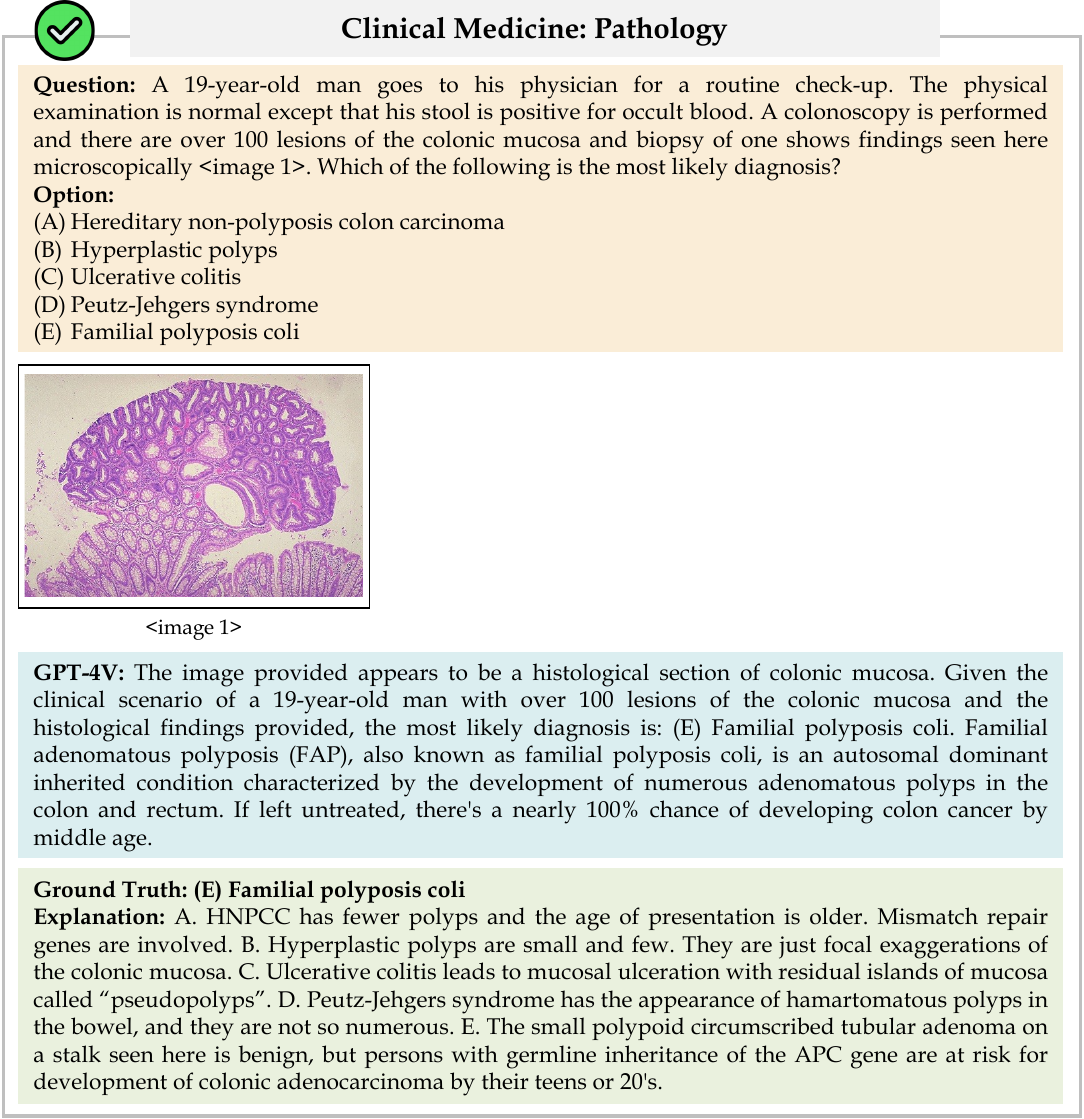}
    \caption{A sample correct case of Clinical Medicine (subfield: Pathology).\\ \hyperref[list:list_of_figures]{Back to List of Figures} \textcolor{red}{$|$} \hyperref[tab:list_of_case_study_figures]{Back to Table Index}}
    \addcontentsline{afg}{appfigures}{\protect\numberline{\thefigure}Clinical Medicine  2: Correct Case}
\label{fig:clinical_medicine_2}
\end{figure*}
\newpage

\begin{figure*}[!htbp]
    \centering
\includegraphics[width=0.9\linewidth]{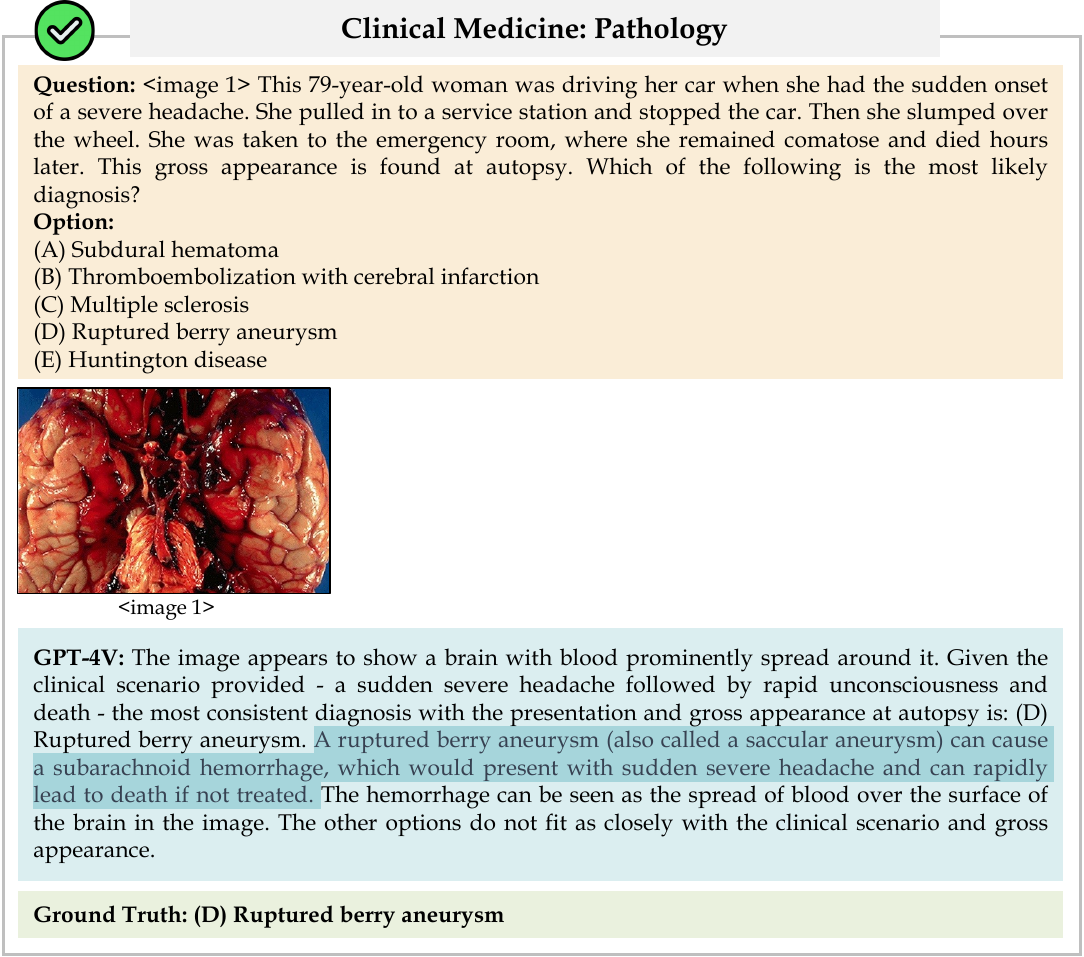}
    \caption{A sample correct case of Clinical Medicine (subfield: Pathology).\\ \hyperref[list:list_of_figures]{Back to List of Figures} \textcolor{red}{$|$} \hyperref[tab:list_of_case_study_figures]{Back to Table Index}}
    \addcontentsline{afg}{appfigures}{\protect\numberline{\thefigure}Clinical Medicine  3: Correct Case}
\label{fig:clinical_medicine_3}
\end{figure*}
\newpage

\begin{figure*}[!htbp]
    \centering
\includegraphics[width=0.9\linewidth]{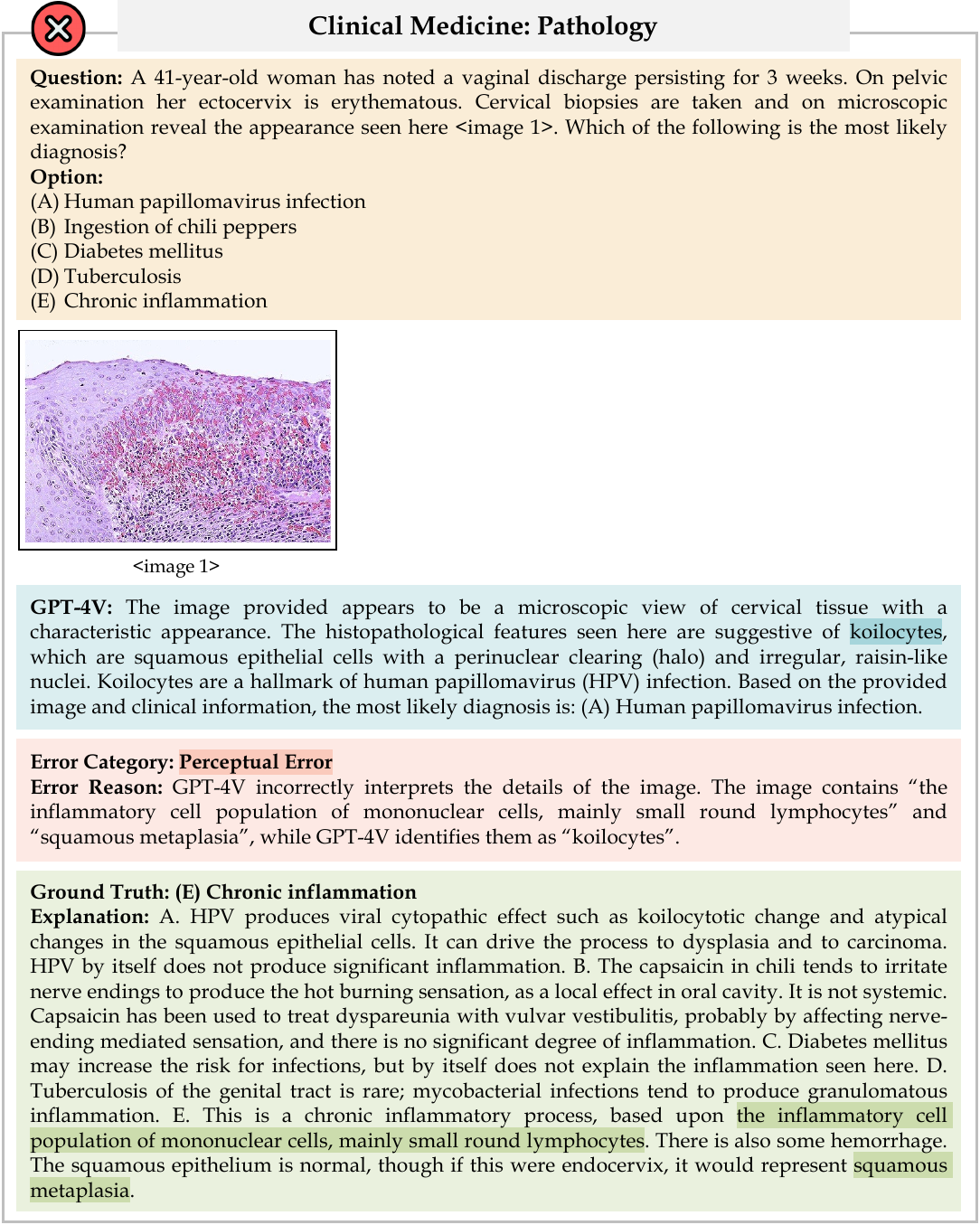}
    \caption{A sample error case of Clinical Medicine (subfield: Pathology). Error category: Perceptual Error \newline \centering \hyperref[list:list_of_figures]{Back to List of Figures} \textcolor{red}{$|$} \hyperref[tab:list_of_case_study_figures]{Back to Table Index}}
    \addcontentsline{afg}{appfigures}{\protect\numberline{\thefigure}Clinical Medicine  4: Perceptual Error}
\label{fig:clinical_medicine_4}
\end{figure*}
\newpage

\begin{figure*}[!htbp]
    \centering
\includegraphics[width=0.9\linewidth]{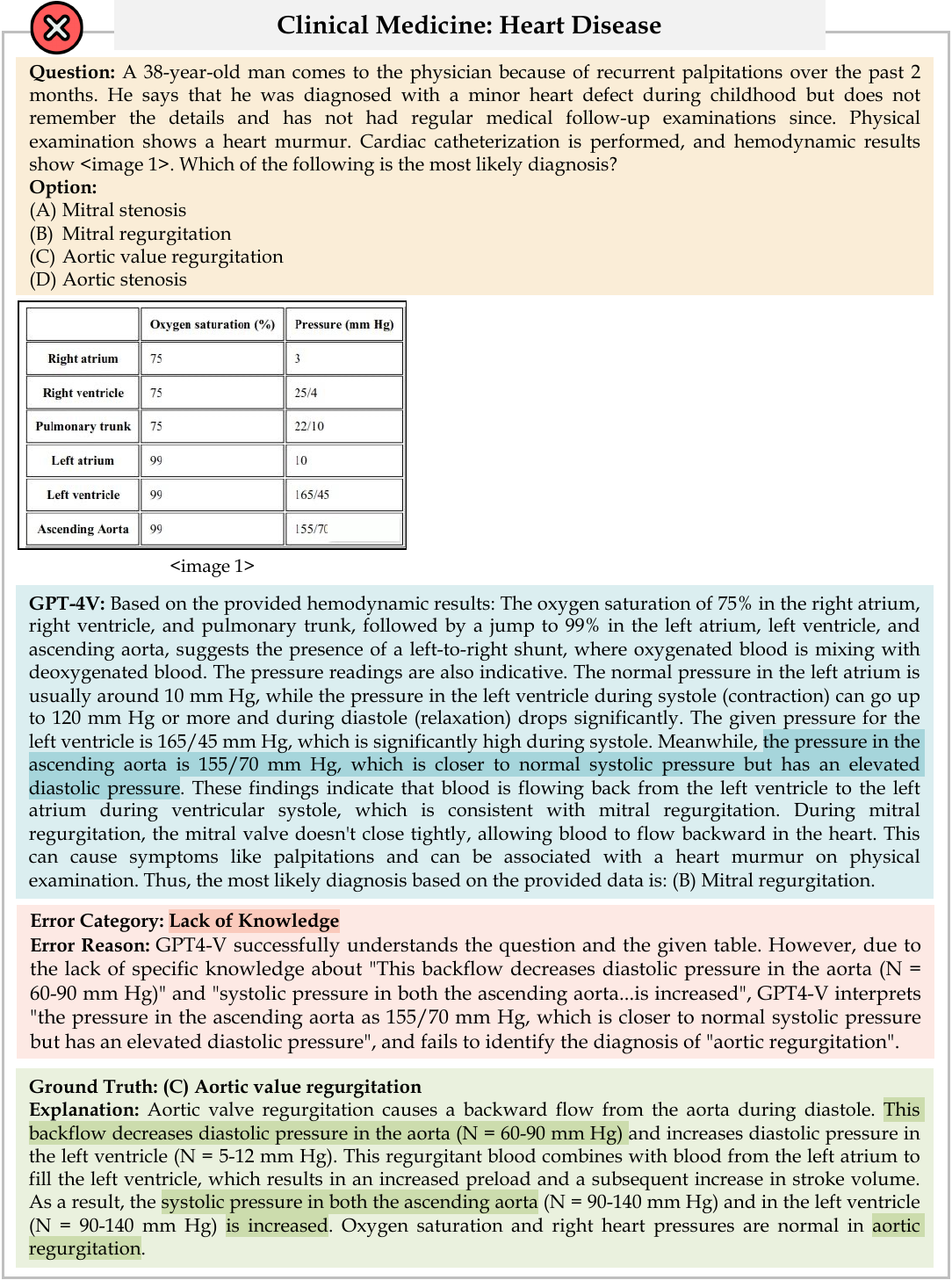}
    \caption{A sample error case of Clinical Medicine (subfield: Heart Disease). Error category: Lack of Knowledge \newline \centering \hyperref[list:list_of_figures]{Back to List of Figures} \textcolor{red}{$|$} \hyperref[tab:list_of_case_study_figures]{Back to Table Index}}
    \addcontentsline{afg}{appfigures}{\protect\numberline{\thefigure}Clinical Medicine  5: Lack of Knowledge}
\label{fig:clinical_medicine_5}
\end{figure*}
\newpage

\begin{figure*}[!htbp]
    \centering
\includegraphics[width=0.9\linewidth]{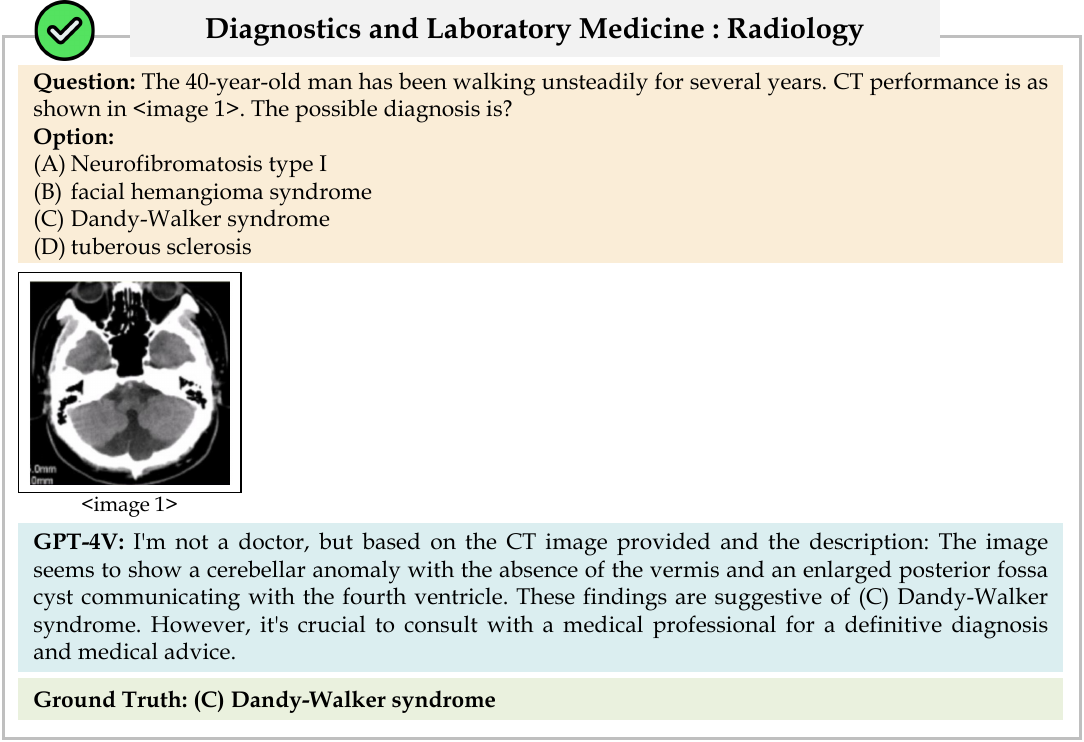}
    \caption{A sample correct case of Diagnostics and Laboratory Medicine (subfield: Radiology).\newline \centering \hyperref[list:list_of_figures]{Back to List of Figures} \textcolor{red}{$|$} \hyperref[tab:list_of_case_study_figures]{Back to Table Index}}
    \addcontentsline{afg}{appfigures}{\protect\numberline{\thefigure}Diagnostics and Lab Medicine  1: Correct Case}
\label{fig:diagnostics_and_laboratory_medicine_1}
\end{figure*}
\newpage

\begin{figure*}[!htbp]
    \centering
\includegraphics[width=0.9\linewidth]{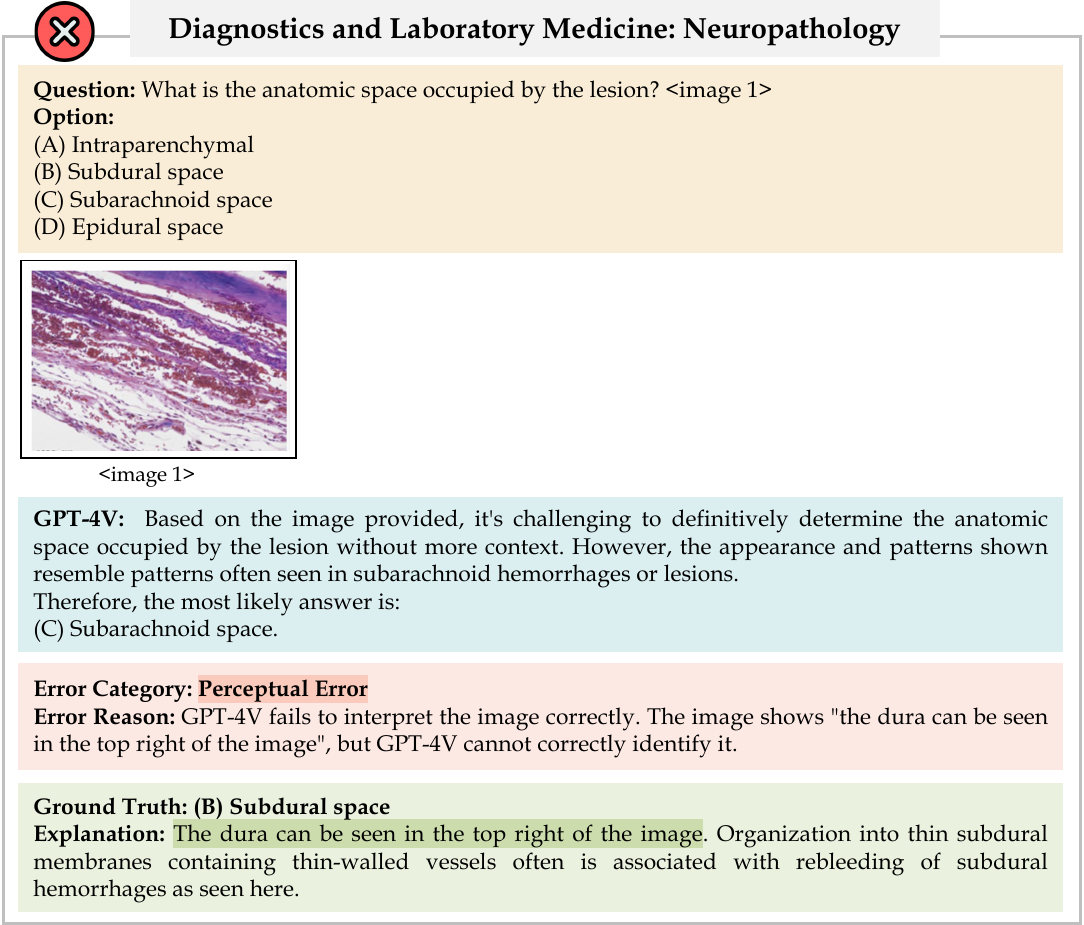}
    \caption{A sample error case of Diagnostics and Laboratory Medicine (subfield: Neuropathology). Error category: Perceptual Error \newline \centering \hyperref[list:list_of_figures]{Back to List of Figures} \textcolor{red}{$|$} \hyperref[tab:list_of_case_study_figures]{Back to Table Index}}
    \addcontentsline{afg}{appfigures}{\protect\numberline{\thefigure}Diagnostics and Lab Medicine  2: Perceptual Error}
\label{fig:diagnostics_and_laboratory_medicine_2}
\end{figure*}
\newpage

\begin{figure*}[!htbp]
    \centering
\includegraphics[width=0.9\linewidth]{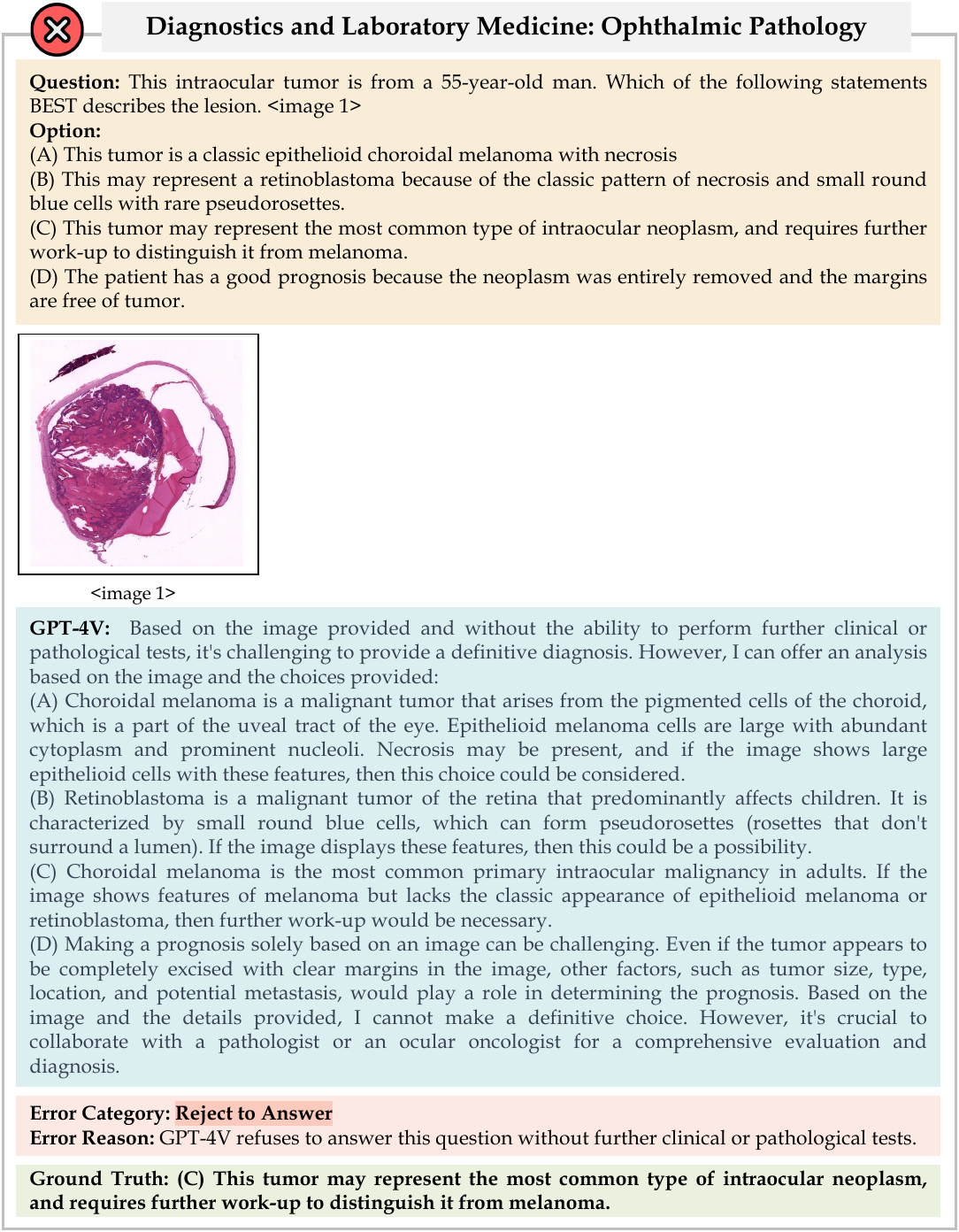}
    \caption{A sample error case of Diagnostics and Laboratory Medicine (subfield: Ophthalmic Pathology). Error category: Reject to Answer \newline \centering \hyperref[list:list_of_figures]{Back to List of Figures} \textcolor{red}{$|$} \hyperref[tab:list_of_case_study_figures]{Back to Table Index}}
    \addcontentsline{afg}{appfigures}{\protect\numberline{\thefigure}Diagnostics and Lab Medicine  3: Reject to Answer}
\label{fig:diagnostics_and_laboratory_medicine_3}
\end{figure*}
\newpage

\begin{figure*}[!htbp]
    \centering
\includegraphics[width=0.9\linewidth]{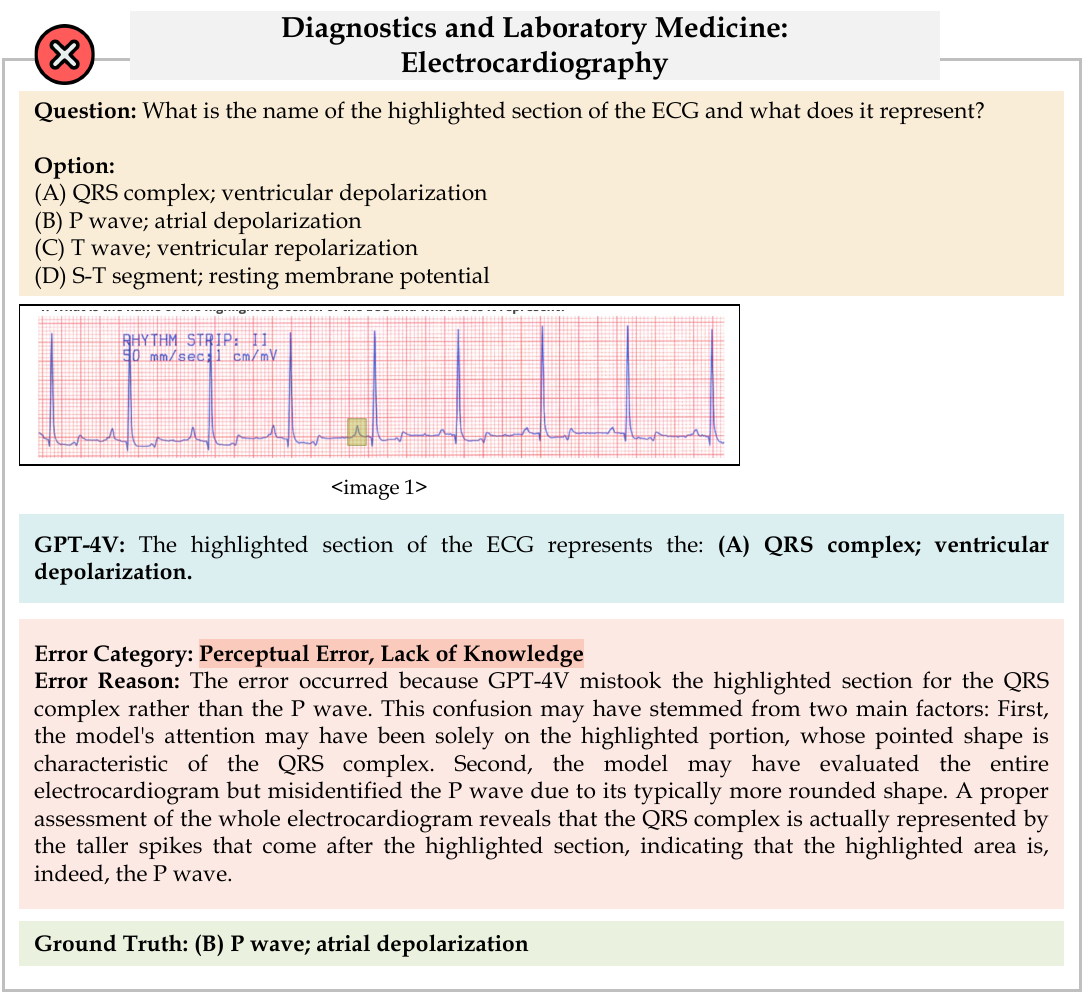}
    \caption{A sample error case of Diagnostics and Laboratory Medicine (subfield: Electrocardiography). Error category: Perceptual Error, Lack of Knowledge \newline \centering \hyperref[list:list_of_figures]{Back to List of Figures} \textcolor{red}{$|$} \hyperref[tab:list_of_case_study_figures]{Back to Table Index}}
    \addcontentsline{afg}{appfigures}{\protect\numberline{\thefigure}Diagnostics and Lab Medicine  4: Perceptual Error, Lack of Knowledge}
\label{fig:diagnostics_and_laboratory_medicine_4}
\end{figure*}
\newpage

\begin{figure*}[!htbp]
    \centering
\includegraphics[width=0.9\linewidth]{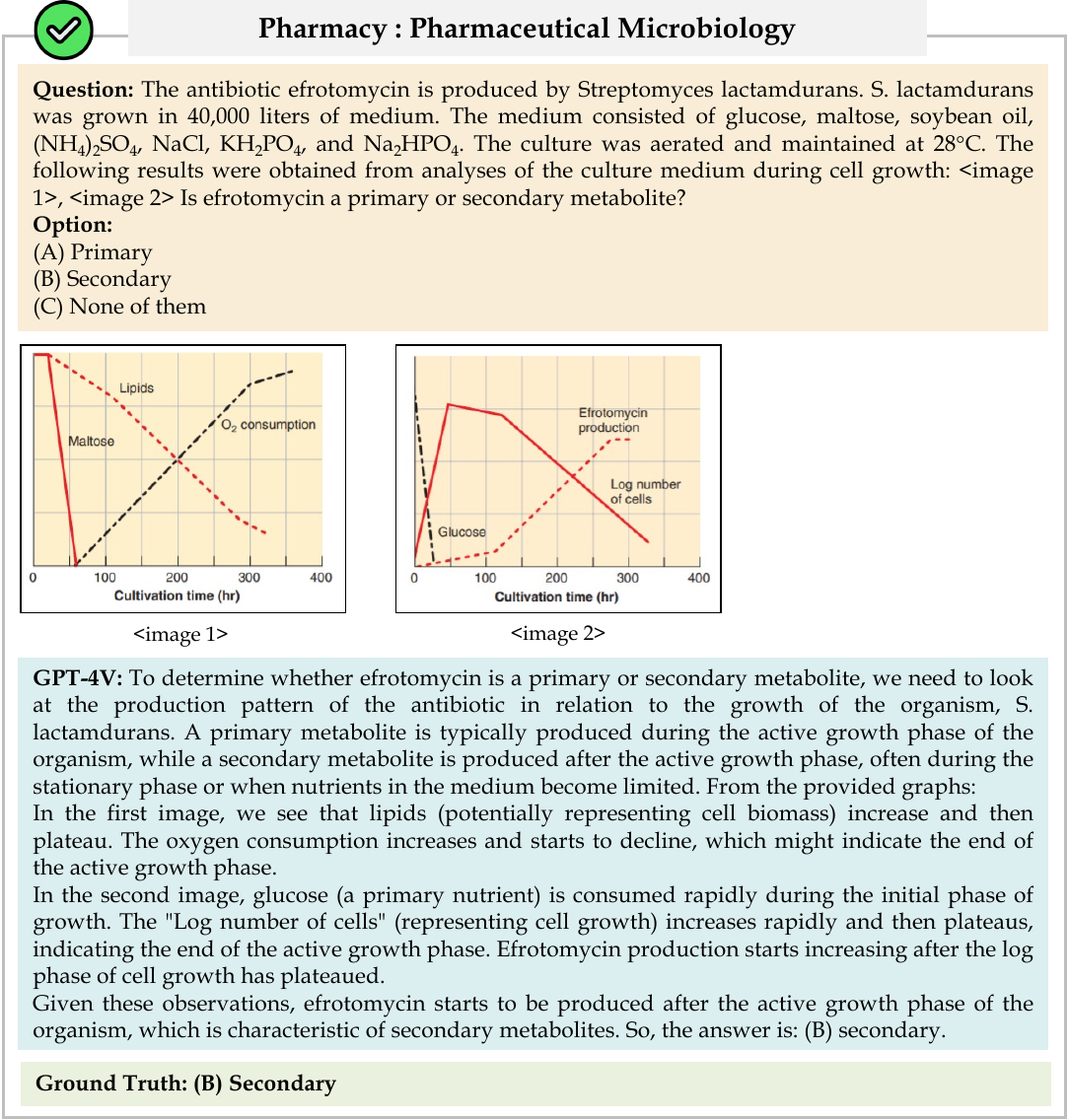}
    \caption{A sample correct case of Pharmacy (subfield: Pharmaceutical Microbiology).\newline \centering \hyperref[list:list_of_figures]{Back to List of Figures} \textcolor{red}{$|$} \hyperref[tab:list_of_case_study_figures]{Back to Table Index}}
    \addcontentsline{afg}{appfigures}{\protect\numberline{\thefigure}Pharmacy  1: Correct Case}
\label{fig:pharmacy_1}
\end{figure*}
\newpage

\begin{figure*}[!htbp]
    \centering
\includegraphics[width=0.9\linewidth]{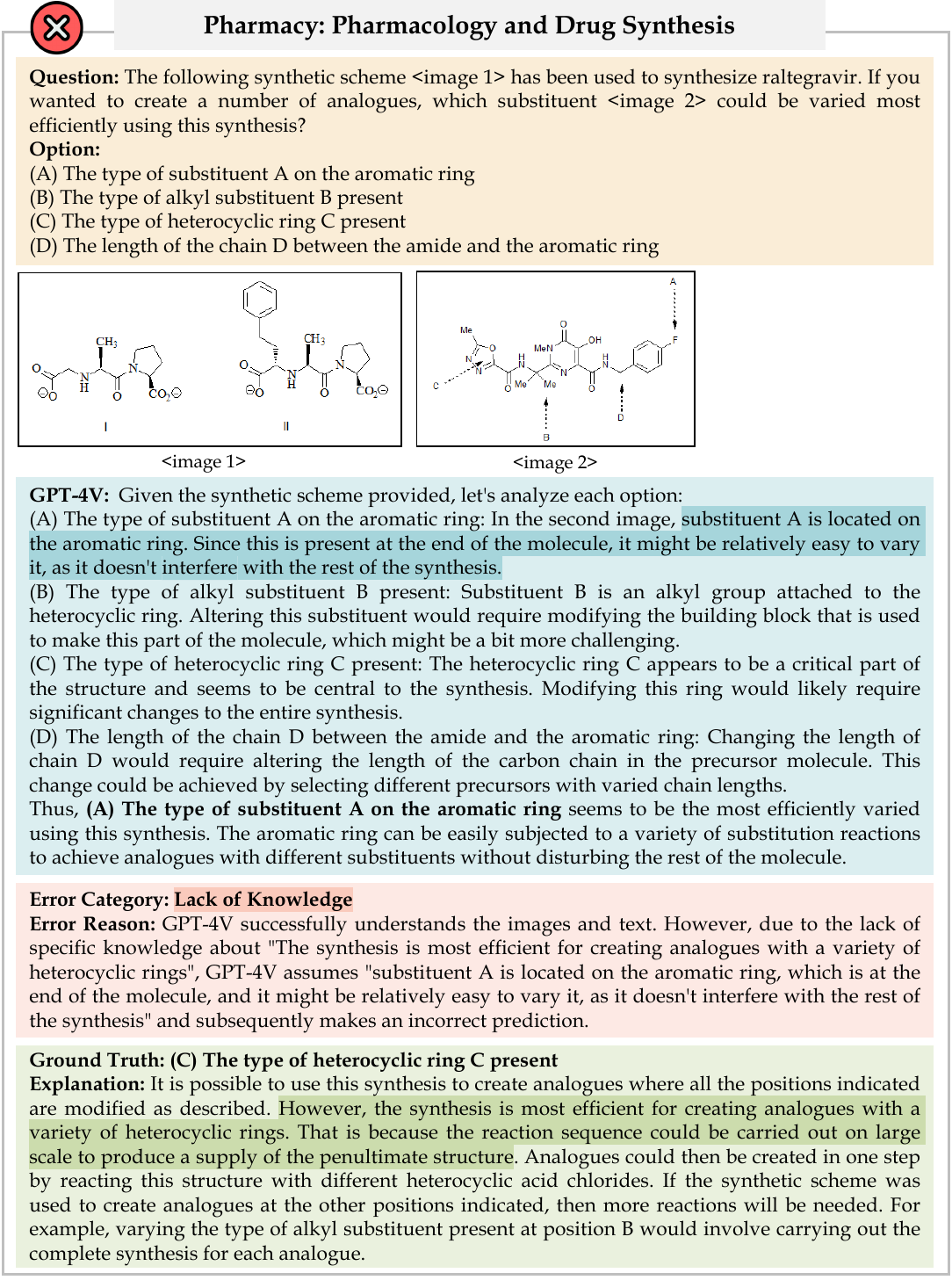}
    \caption{A sample error case of Pharmacy (subfield: Pharmacology and Drug Synthesis). Error category: Lack of Knowledge \newline \centering \hyperref[list:list_of_figures]{Back to List of Figures} \textcolor{red}{$|$} \hyperref[tab:list_of_case_study_figures]{Back to Table Index}}
    \addcontentsline{afg}{appfigures}{\protect\numberline{\thefigure}Pharmacy  2: Lack of Knowledge}
\label{fig:pharmacy_2}
\end{figure*}
\newpage

\begin{figure*}[!htbp]
    \centering
\includegraphics[width=0.9\linewidth]{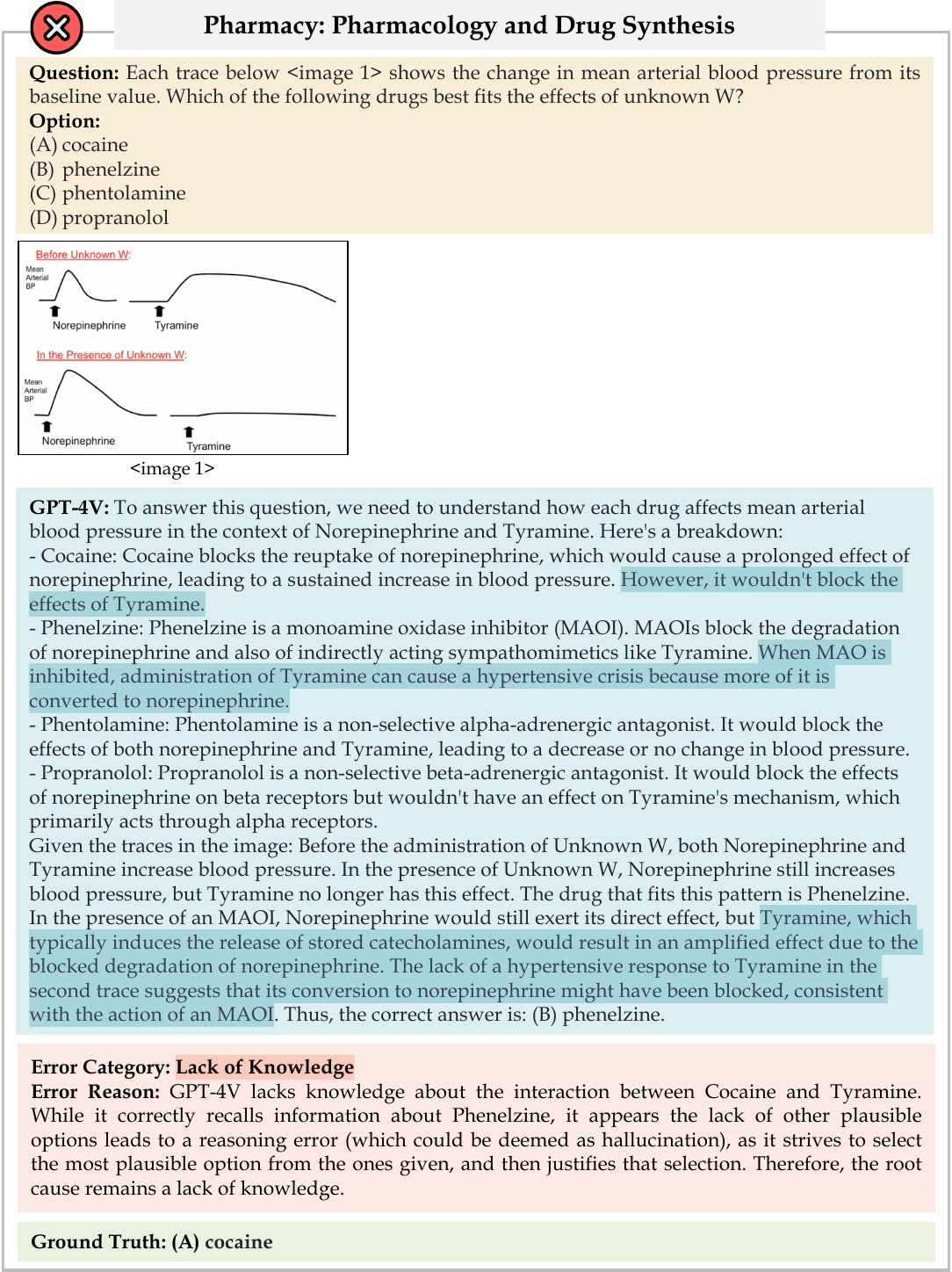}
    \caption{A sample error case of Pharmacy (subfield: Pharmacology and Drug Synthesis). Error category: Lack of Knowledge \newline \centering \hyperref[list:list_of_figures]{Back to List of Figures} \textcolor{red}{$|$} \hyperref[tab:list_of_case_study_figures]{Back to Table Index}}
    \addcontentsline{afg}{appfigures}{\protect\numberline{\thefigure}Pharmacy  3: Lack of Knowledge}
\label{fig:pharmacy_3}
\end{figure*}
\newpage

\begin{figure*}[!htbp]
    \centering
\includegraphics[width=0.9\linewidth]{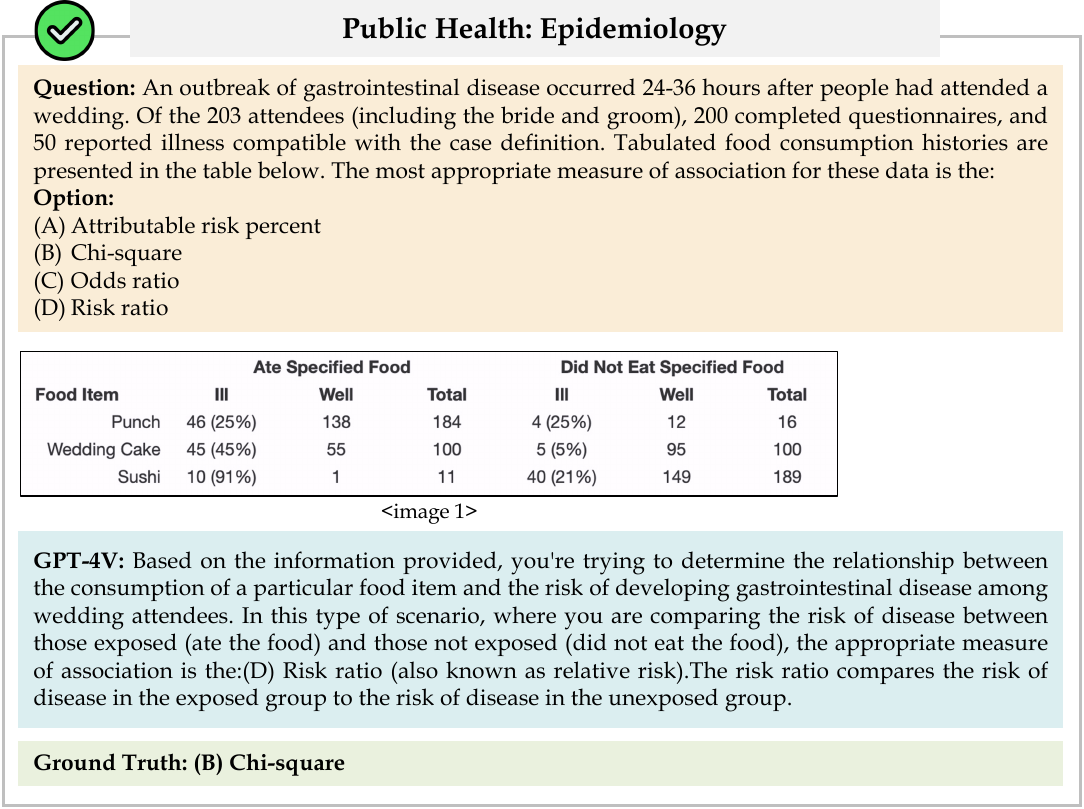}
    \caption{A sample correct case of Public Health (subfield: Epidemiology).\\ \hyperref[list:list_of_figures]{Back to List of Figures} \textcolor{red}{$|$} \hyperref[tab:list_of_case_study_figures]{Back to Table Index}}
    \addcontentsline{afg}{appfigures}{\protect\numberline{\thefigure}Public Health  1: Correct Case}
\label{fig:public_health_1}
\end{figure*}
\newpage

\begin{figure*}[!htbp]
    \centering
\includegraphics[width=0.9\linewidth]{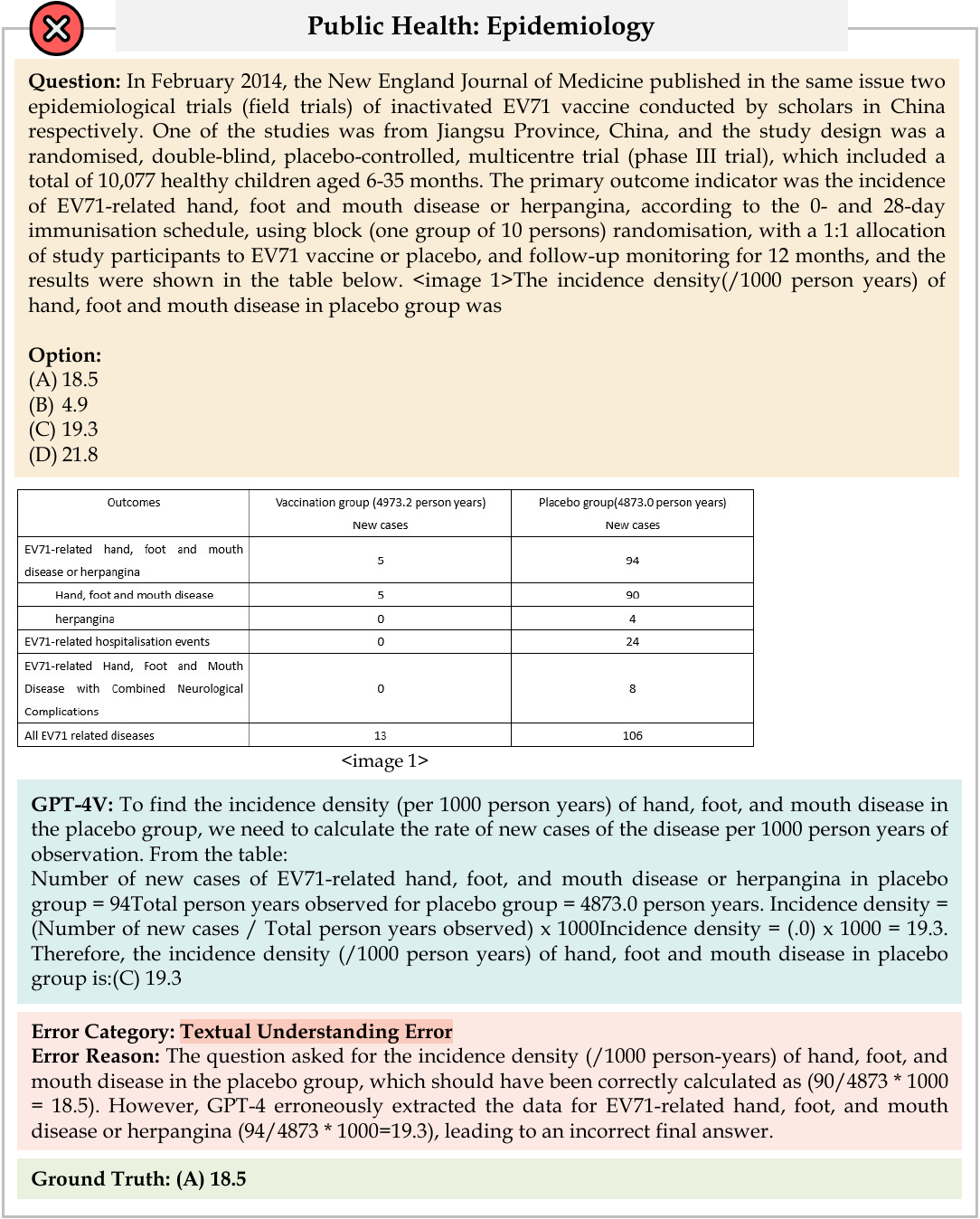}
    \caption{A sample error case of Public Health (subfield: Epidemiology). Error category: Textual Understanding Error \newline \centering \hyperref[list:list_of_figures]{Back to List of Figures} \textcolor{red}{$|$} \hyperref[tab:list_of_case_study_figures]{Back to Table Index}}
    \addcontentsline{afg}{appfigures}{\protect\numberline{\thefigure}Public Health  2: Textual Understanding Error}
\label{fig:public_health_2}
\end{figure*}
\newpage

\begin{figure*}[!htbp]
    \centering
\includegraphics[width=0.9\linewidth]{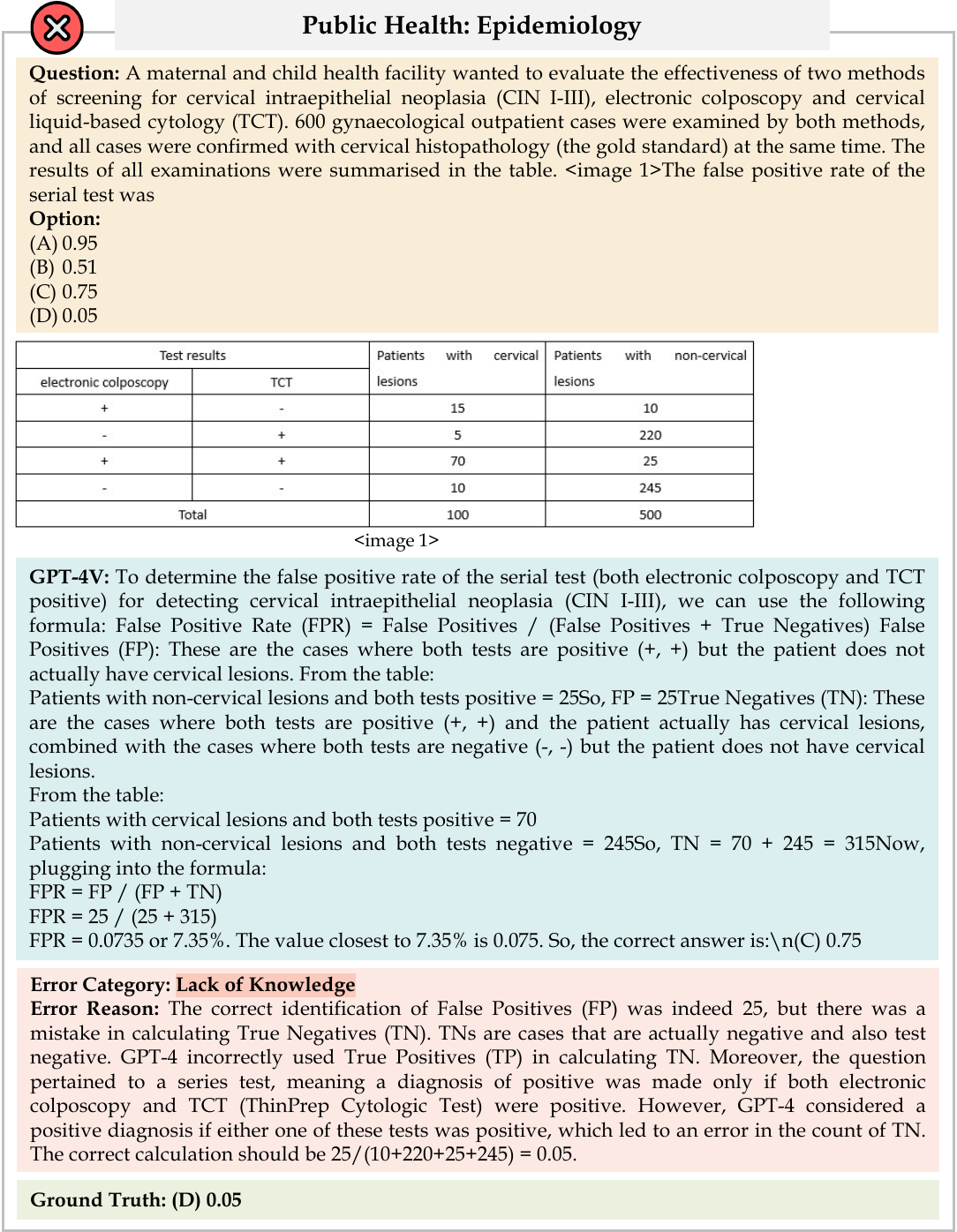}
    \caption{A sample error case of Public Health (subfield: Epidemiology). Error category: Lack of Knowledge \newline \centering \hyperref[list:list_of_figures]{Back to List of Figures} \textcolor{red}{$|$} \hyperref[tab:list_of_case_study_figures]{Back to Table Index}}
    \addcontentsline{afg}{appfigures}{\protect\numberline{\thefigure}Public Health  3: Lack of Knowledge}
\label{fig:public_health_3}
\end{figure*}
\newpage

\begin{figure*}[!htbp]
    \centering
\includegraphics[width=0.9\linewidth]{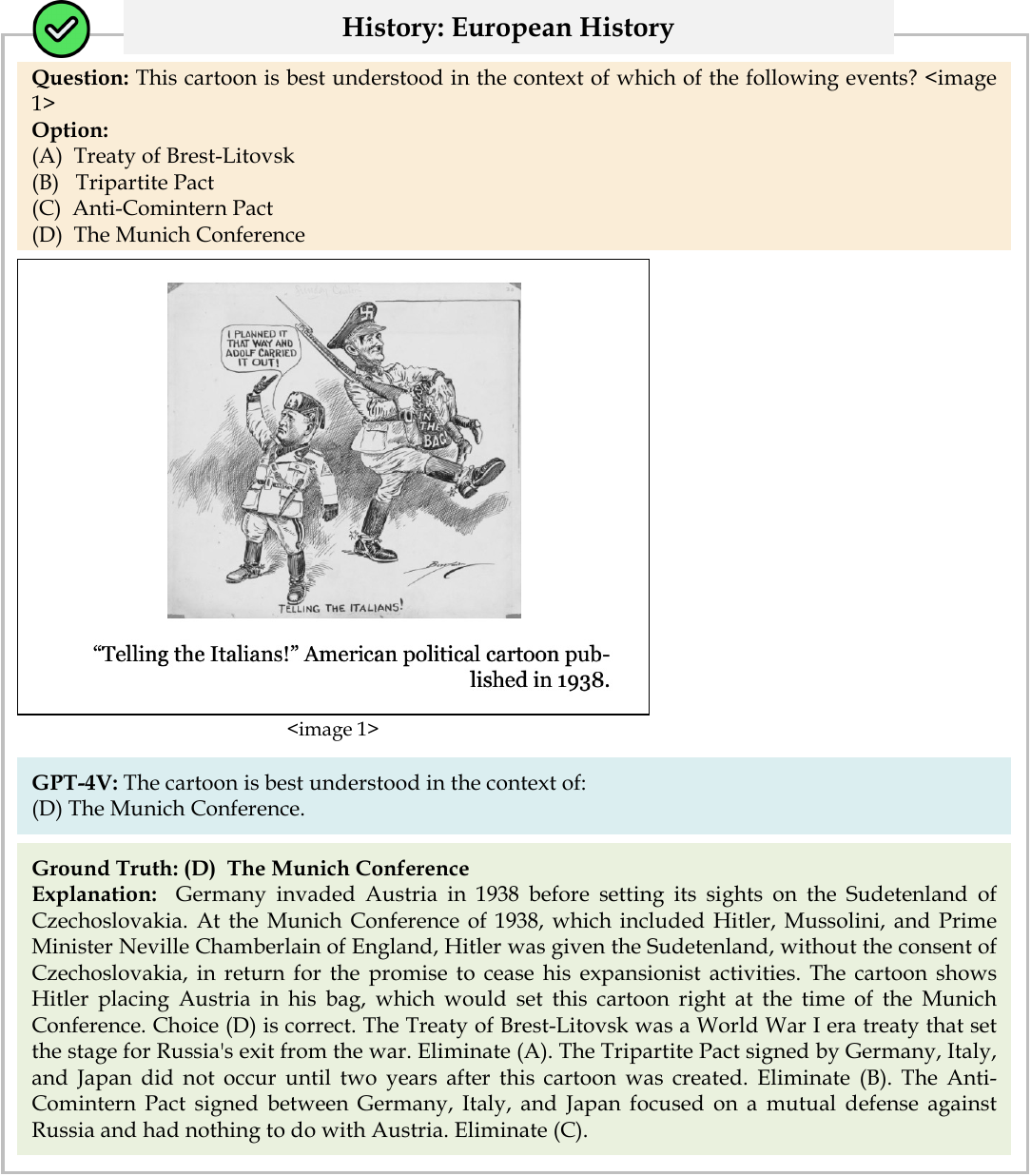}
    \caption{A sample correct case of History (subfield: European History).\\ \hyperref[list:list_of_figures]{Back to List of Figures} \textcolor{red}{$|$} \hyperref[tab:list_of_case_study_figures]{Back to Table Index}}
    \addcontentsline{afg}{appfigures}{\protect\numberline{\thefigure}History  1: Correct Case}
\label{fig:history_1}
\end{figure*}
\newpage

\begin{figure*}[!htbp]
    \centering
\includegraphics[width=0.9\linewidth]{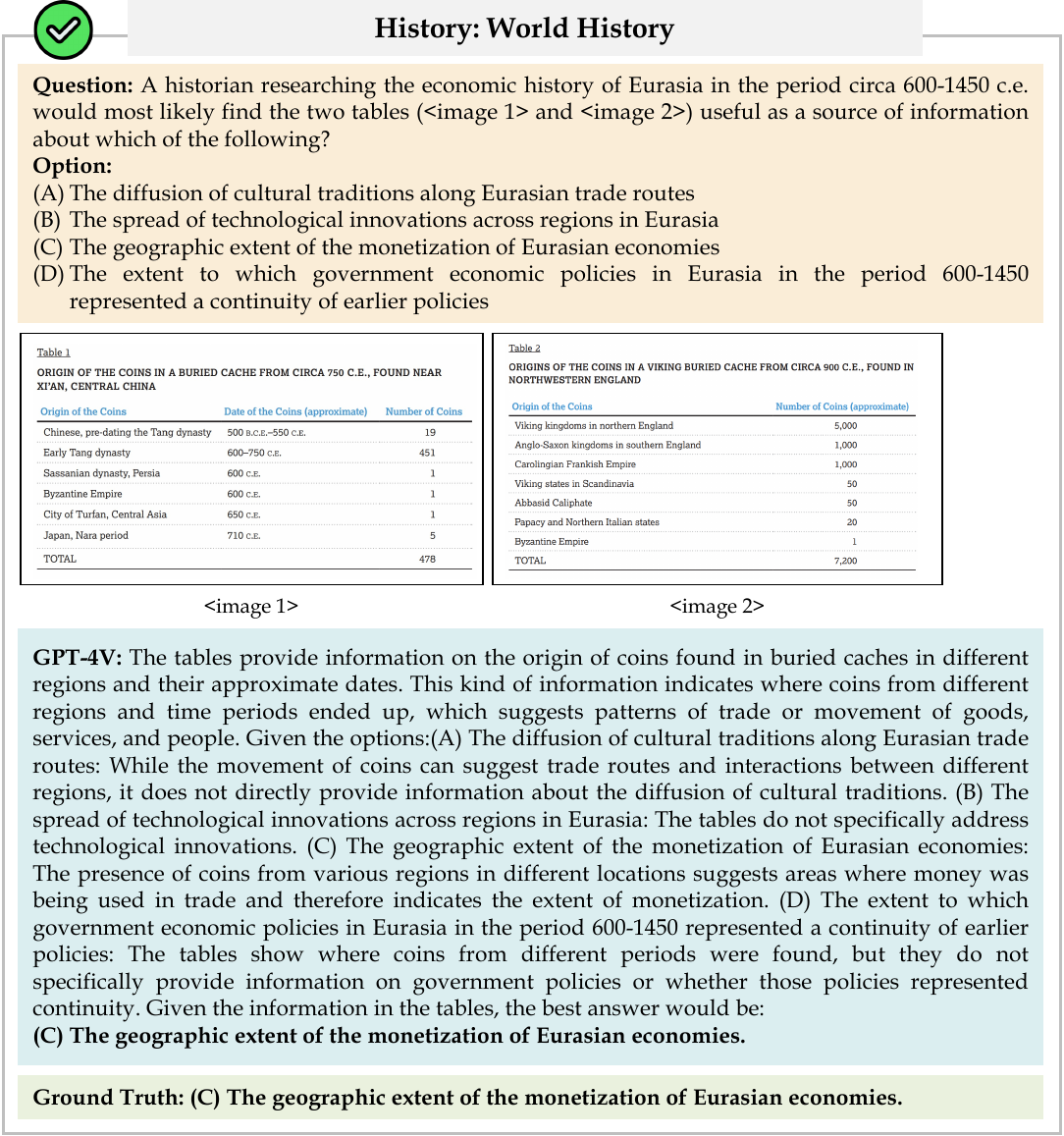}
    \caption{A sample correct case of History (subfield: World History).\\ \hyperref[list:list_of_figures]{Back to List of Figures} \textcolor{red}{$|$} \hyperref[tab:list_of_case_study_figures]{Back to Table Index}}
    \addcontentsline{afg}{appfigures}{\protect\numberline{\thefigure}History  2: Correct Case}
\label{fig:history_2}
\end{figure*}
\newpage

\begin{figure*}[!htbp]
    \centering
\includegraphics[width=0.9\linewidth]{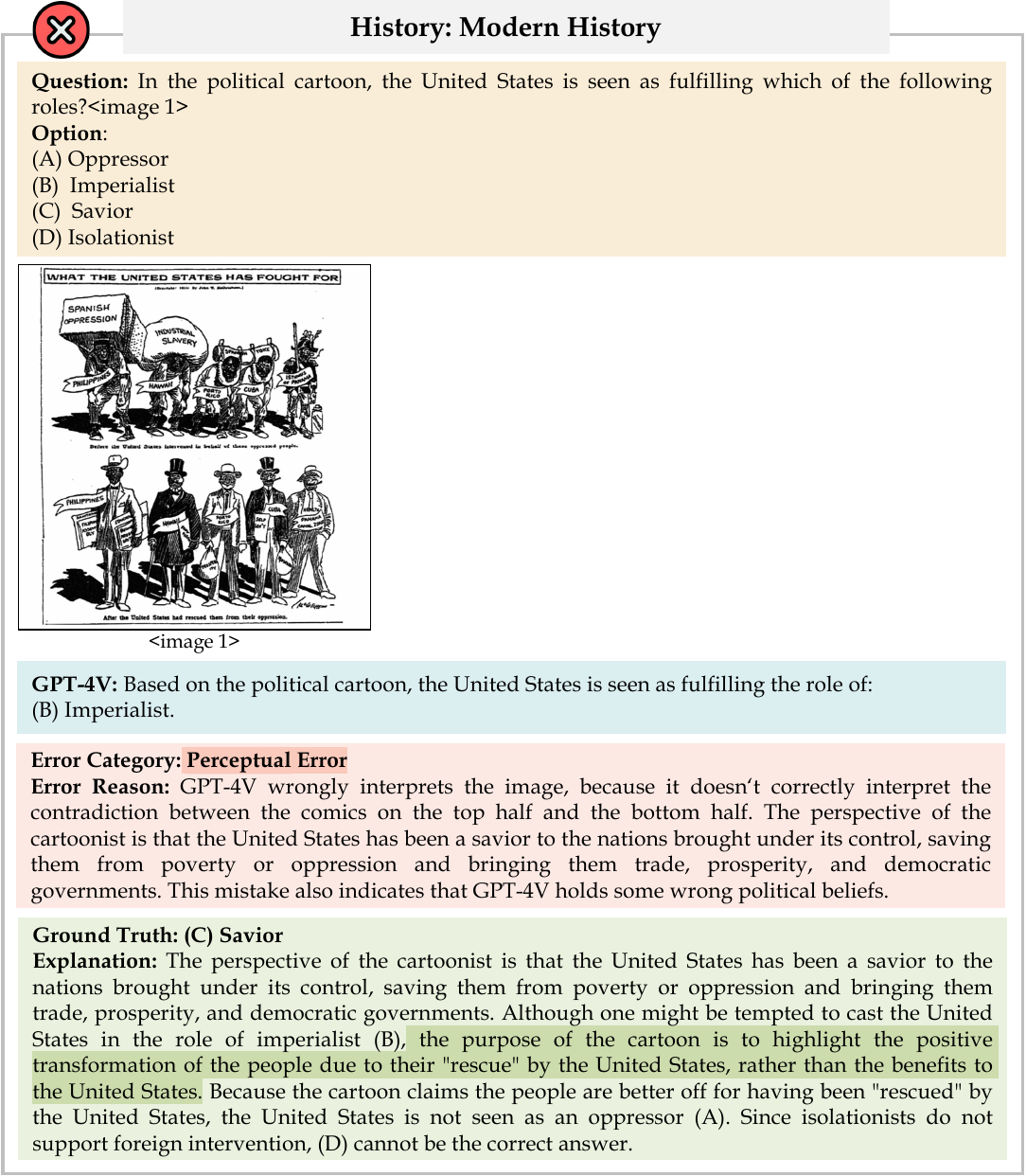}
    \caption{A sample error case of History (subfield: Modern History). Error category: Perceptual Error \newline \centering \hyperref[list:list_of_figures]{Back to List of Figures} \textcolor{red}{$|$} \hyperref[tab:list_of_case_study_figures]{Back to Table Index}}
    \addcontentsline{afg}{appfigures}{\protect\numberline{\thefigure}History  3: Perceptual Error}
\label{fig:history_3}
\end{figure*}
\newpage

\begin{figure*}[!htbp]
    \centering
\includegraphics[width=0.9\linewidth]{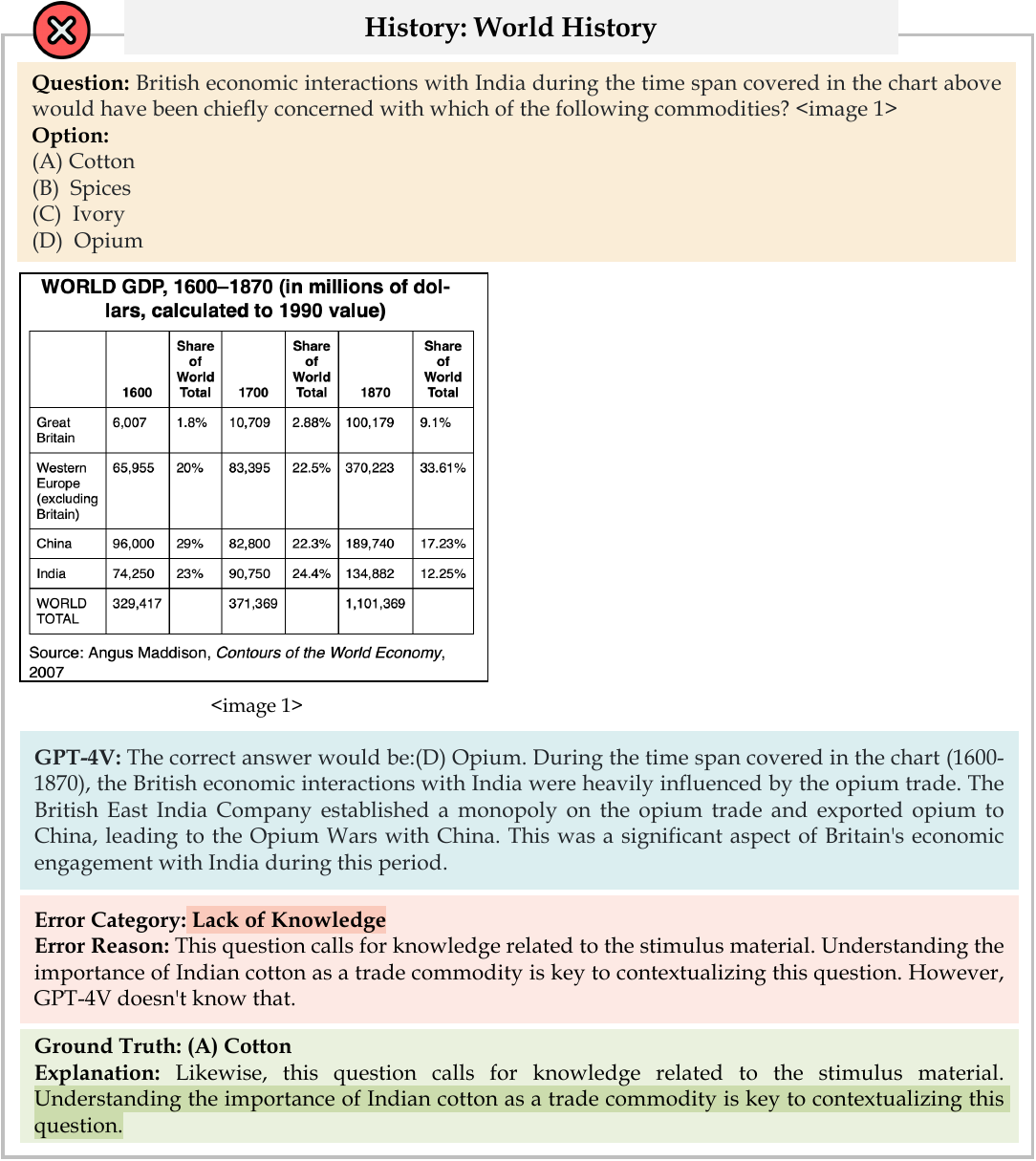}
    \caption{A sample error case of History (subfield: World History). Error category: Lack of Knowledge \newline \centering \hyperref[list:list_of_figures]{Back to List of Figures} \textcolor{red}{$|$} \hyperref[tab:list_of_case_study_figures]{Back to Table Index}}
    \addcontentsline{afg}{appfigures}{\protect\numberline{\thefigure}History  4: Lack of Knowledge}
\label{fig:history_4}
\end{figure*}
\newpage

\begin{figure*}[!htbp]
    \centering
\includegraphics[width=0.9\linewidth]{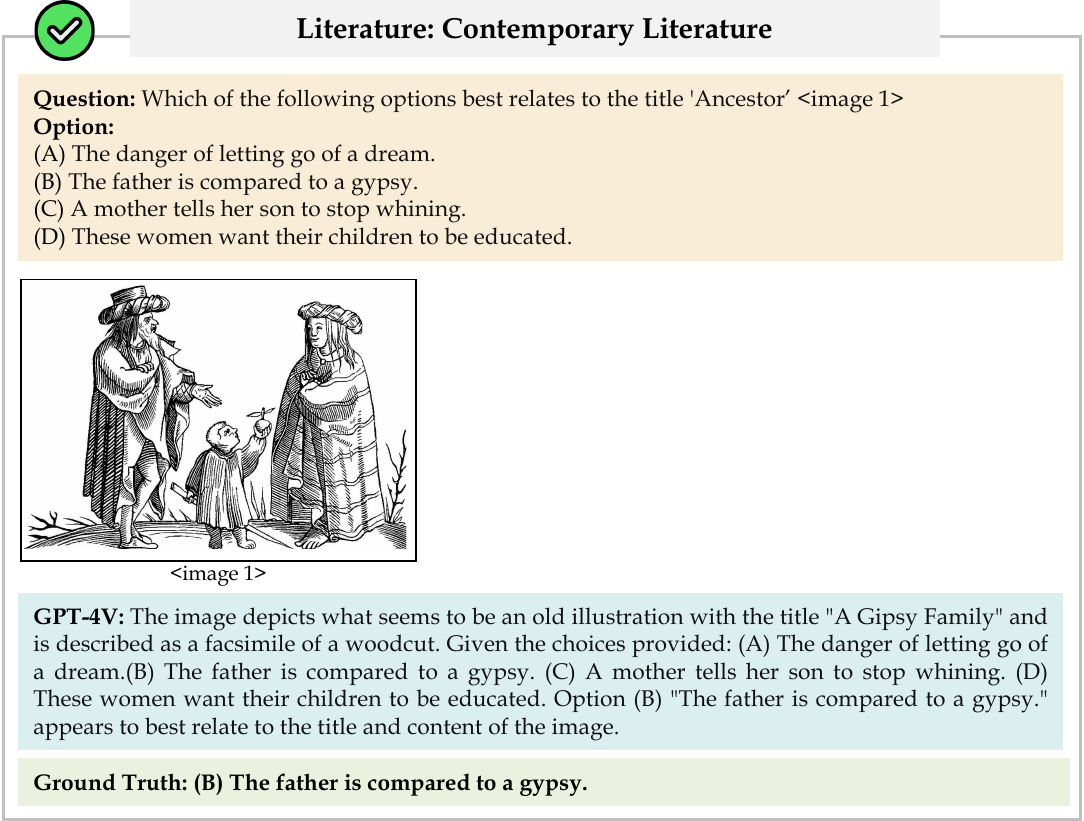}
    \caption{A sample correct case of Literature (subfield: Contemporary Literature).\\ \hyperref[list:list_of_figures]{Back to List of Figures} \textcolor{red}{$|$} \hyperref[tab:list_of_case_study_figures]{Back to Table Index}}
    \addcontentsline{afg}{appfigures}{\protect\numberline{\thefigure}Literature  1: Correct Case}
\label{fig:literature_1}
\end{figure*}
\newpage

\begin{figure*}[!htbp]
    \centering
\includegraphics[width=0.9\linewidth]{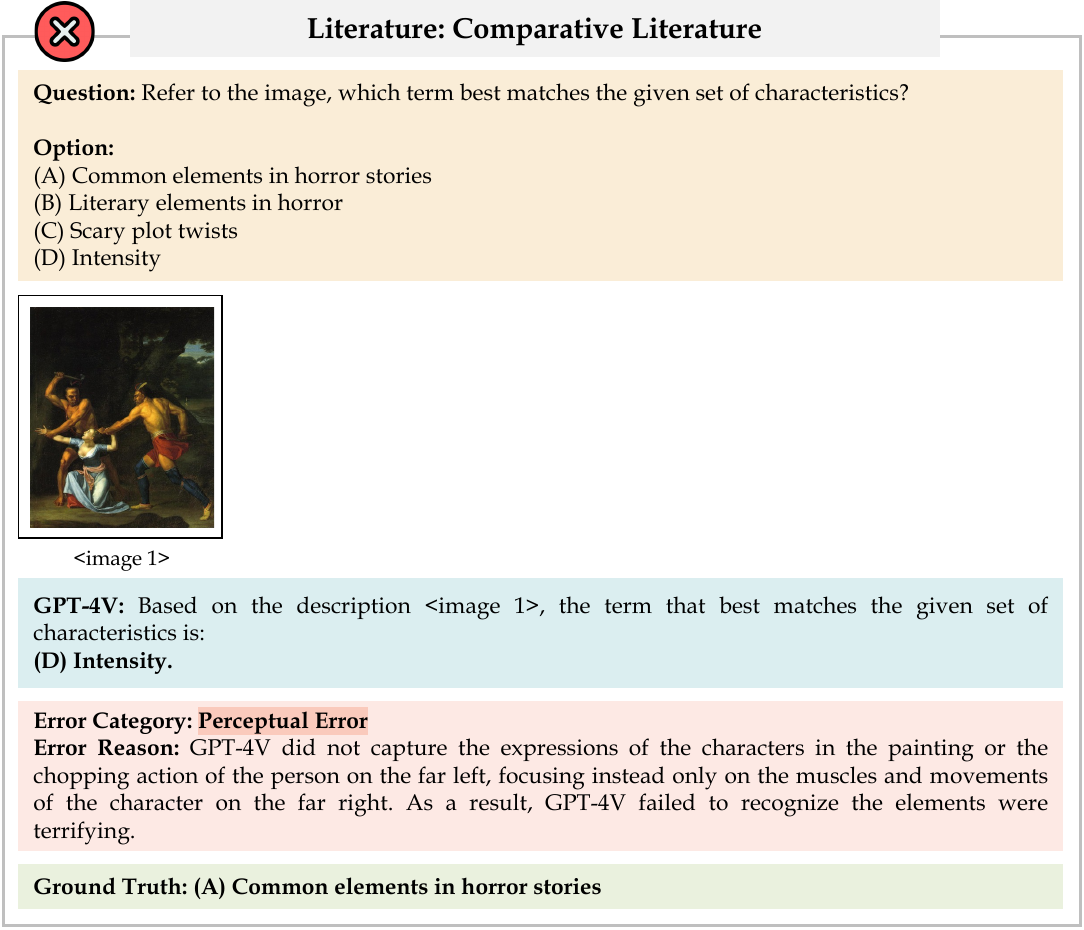}
    \caption{A sample error case of Literature (subfield: Comparative Literature). Error category: Perceptual Error}
    \addcontentsline{afg}{appfigures}{\protect\numberline{\thefigure}Literature  2: Perceptual Error}
\label{fig:literature_2}
\end{figure*}
\newpage


\begin{figure*}[!htbp]
    \centering
\includegraphics[width=0.9\linewidth]{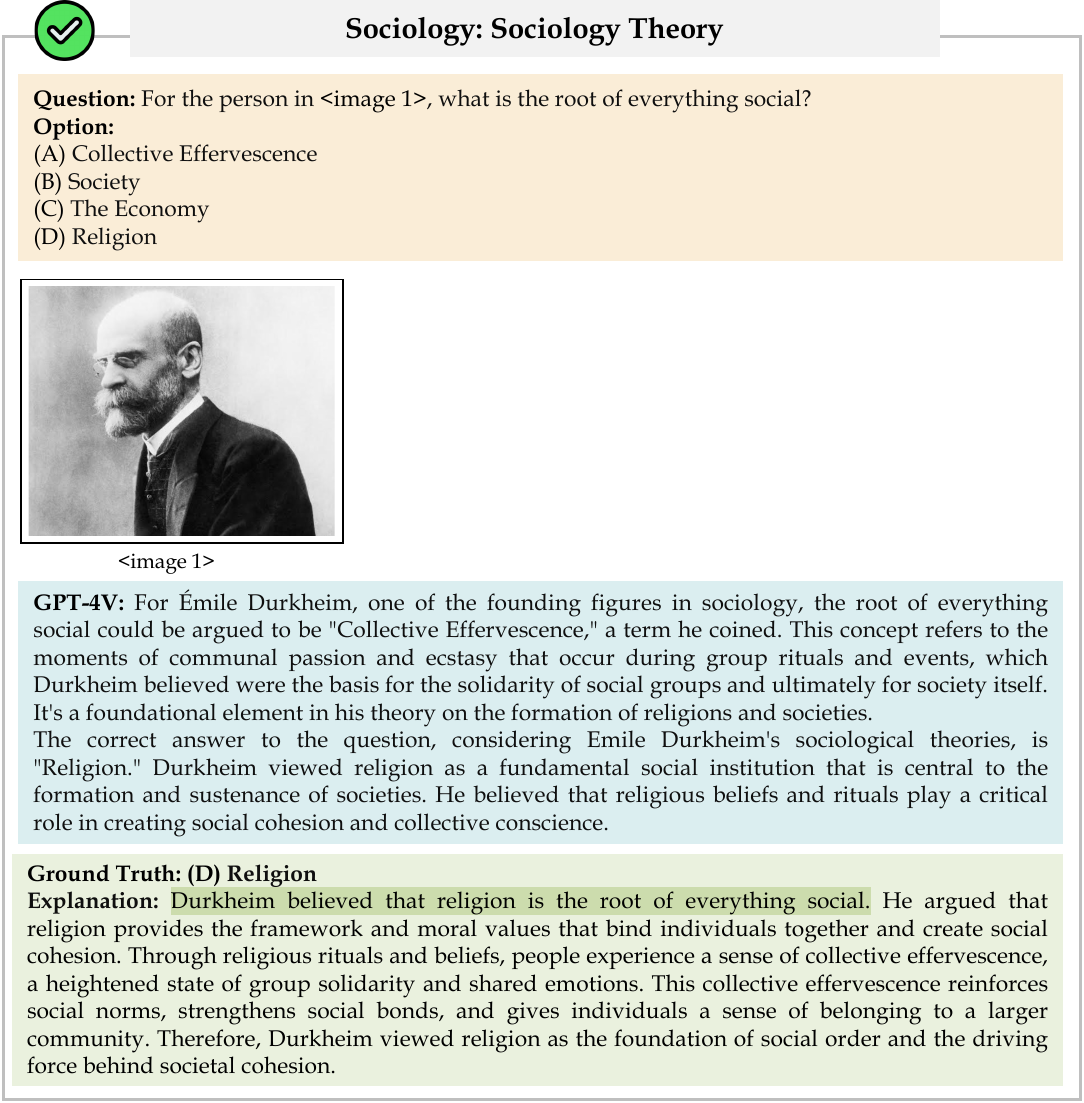}
    \caption{A sample correct case of Sociology (subfield: Sociology Theory).\\ \hyperref[list:list_of_figures]{Back to List of Figures} \textcolor{red}{$|$} \hyperref[tab:list_of_case_study_figures]{Back to Table Index}}
    \addcontentsline{afg}{appfigures}{\protect\numberline{\thefigure}Sociology  1: Correct Case}
\label{fig:sociology_1}
\end{figure*}
\newpage

\begin{figure*}[!htbp]
    \centering
\includegraphics[width=0.9\linewidth]{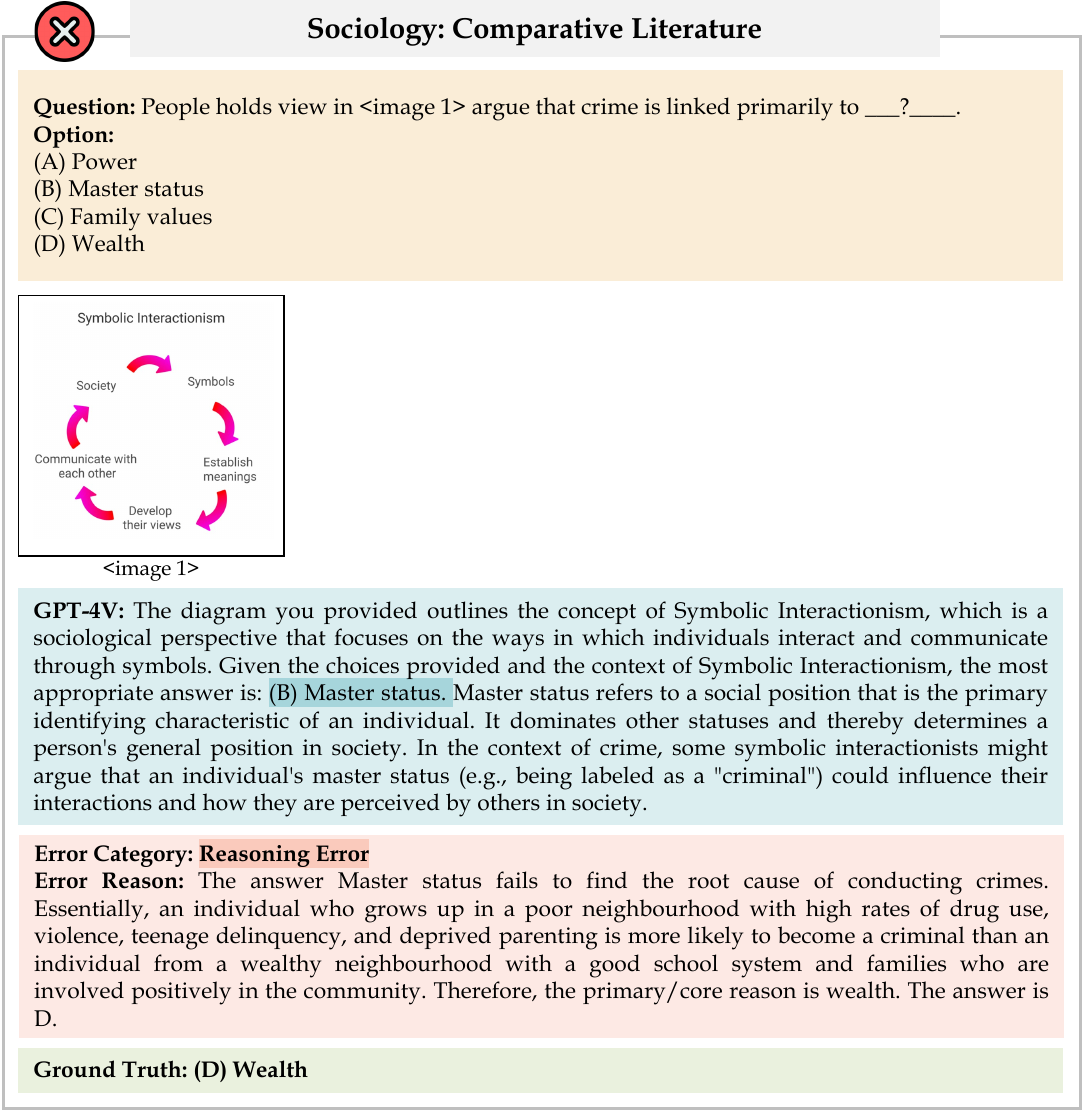}
    \caption{A sample error case of Sociology (subfield: Comparative Literature). Error category: Reasoning Error \newline \centering \hyperref[list:list_of_figures]{Back to List of Figures} \textcolor{red}{$|$} \hyperref[tab:list_of_case_study_figures]{Back to Table Index}}
    \addcontentsline{afg}{appfigures}{\protect\numberline{\thefigure}Sociology  2: Reasoning Error}
\label{fig:sociology_2}
\end{figure*}
\newpage


\begin{figure*}[!htbp]
    \centering
\includegraphics[width=0.9\linewidth]{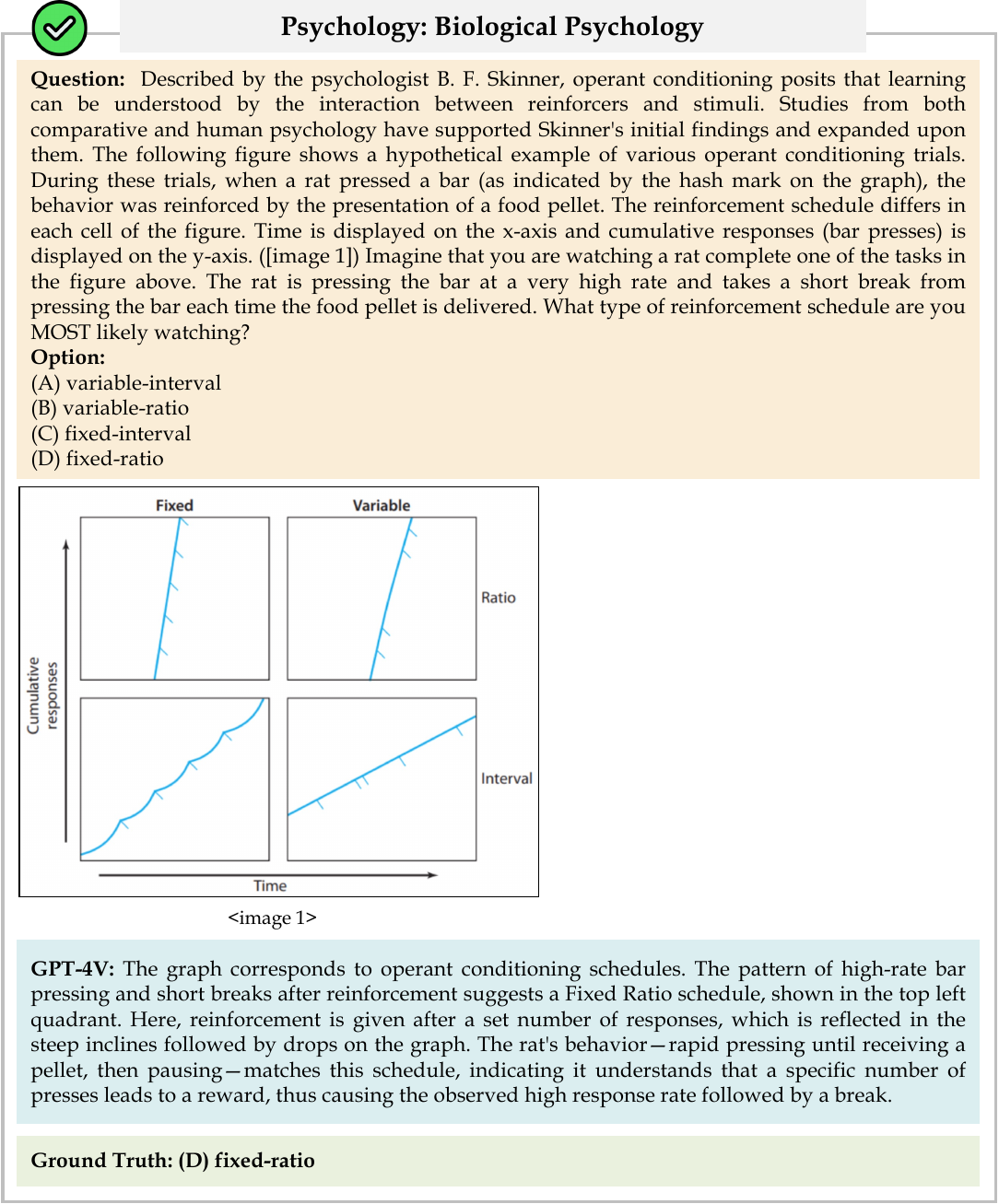}
    \caption{A sample correct case of Psychology (subfield: Biological Psychology).\\ \hyperref[list:list_of_figures]{Back to List of Figures} \textcolor{red}{$|$} \hyperref[tab:list_of_case_study_figures]{Back to Table Index}}
    \addcontentsline{afg}{appfigures}{\protect\numberline{\thefigure}Psychology  1: Correct Case}
\label{fig:psychology_1}
\end{figure*}
\newpage

\begin{figure*}[!htbp]
    \centering
\includegraphics[width=0.9\linewidth]{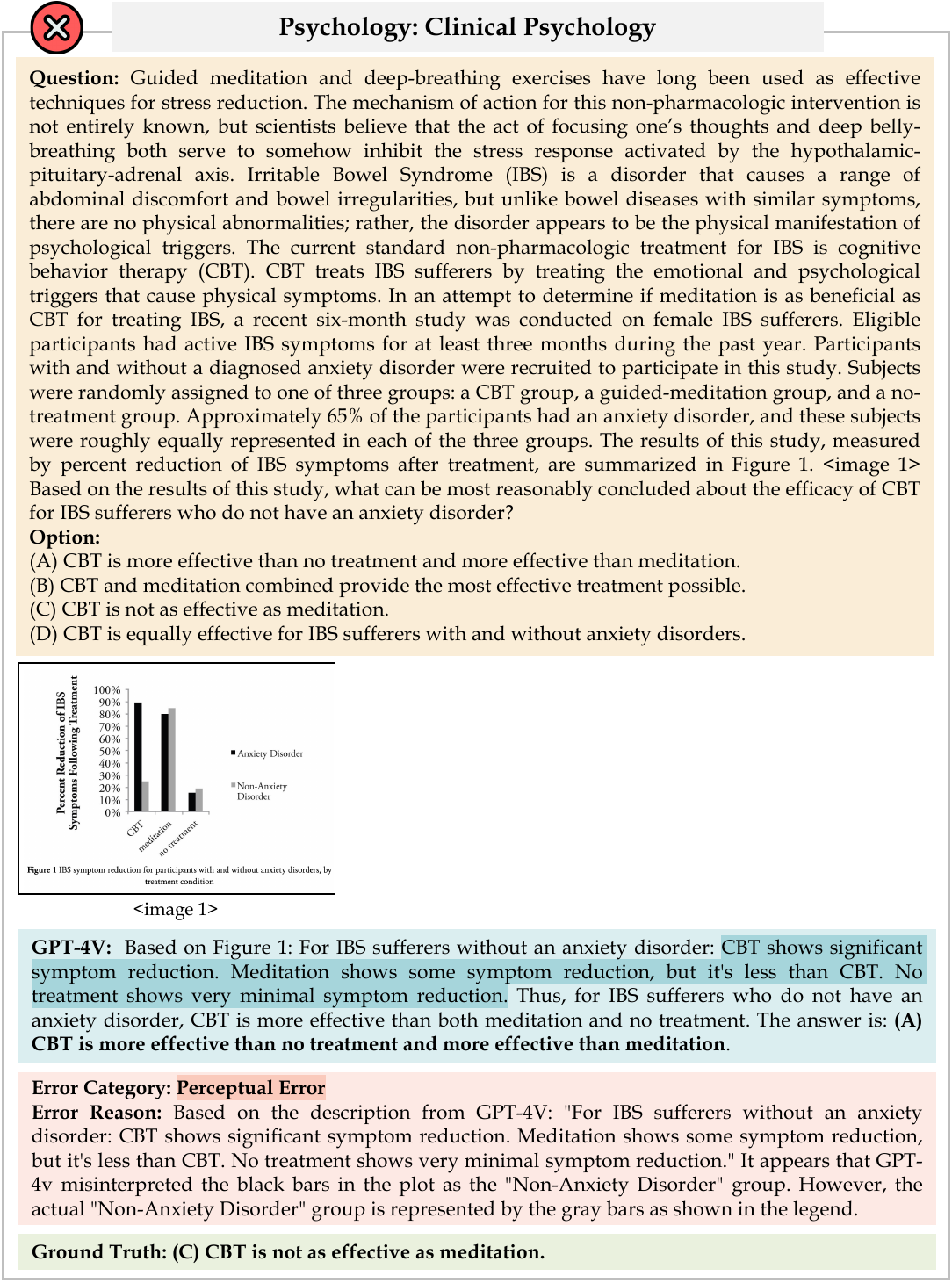}
    \caption{A sample error case of Psychology (subfield: Clinical Psychology). Error category: Perceptual Error \newline \centering \hyperref[list:list_of_figures]{Back to List of Figures} \textcolor{red}{$|$} \hyperref[tab:list_of_case_study_figures]{Back to Table Index}}
    \addcontentsline{afg}{appfigures}{\protect\numberline{\thefigure}Psychology  2: Perceptual Error}
\label{fig:psychology_2}
\end{figure*}
\newpage


\begin{figure*}[!htbp]
    \centering
\includegraphics[width=0.9\linewidth]{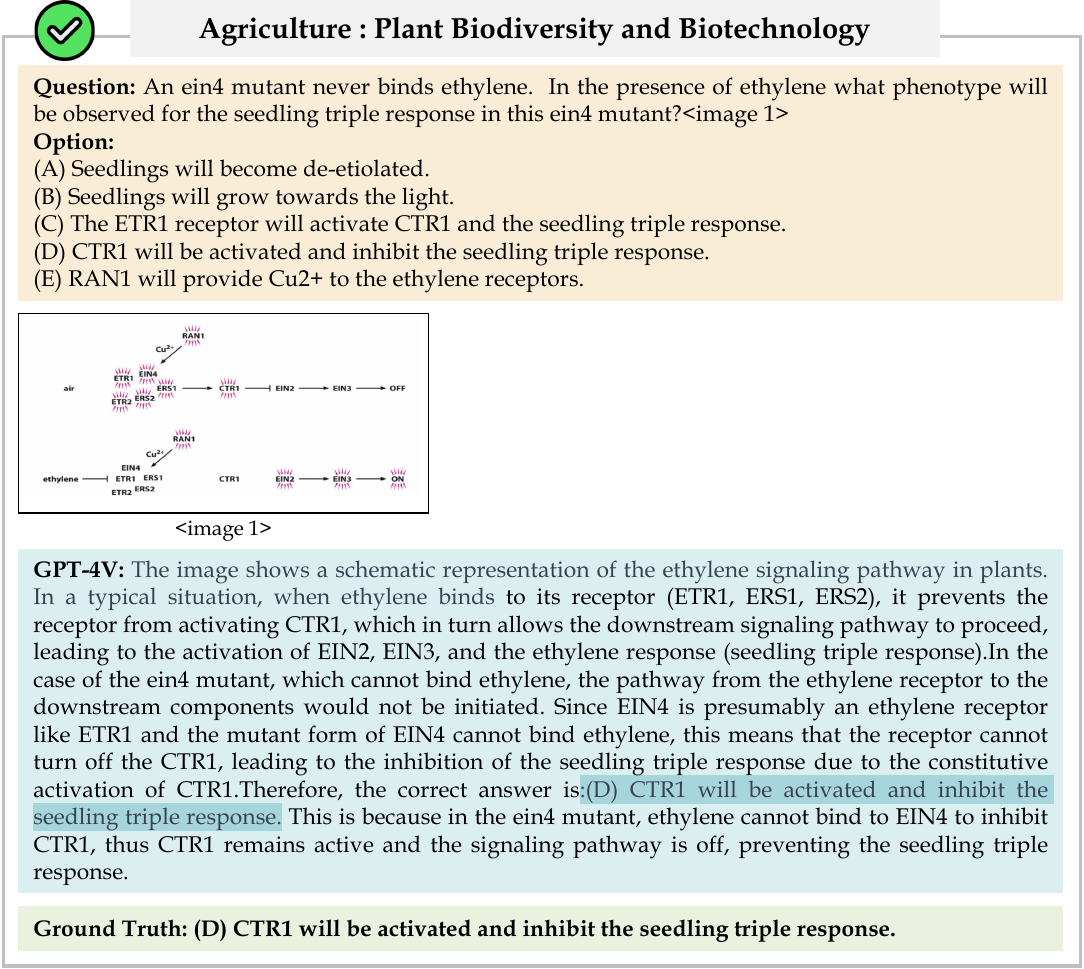}
    \caption{A sample correct case of Agriculture (subfield: Plant Biodiversity and Biotechnology).\newline \centering \hyperref[list:list_of_figures]{Back to List of Figures} \textcolor{red}{$|$} \hyperref[tab:list_of_case_study_figures]{Back to Table Index}}
    \addcontentsline{afg}{appfigures}{\protect\numberline{\thefigure}Agriculture  1: Correct Case}
\label{fig:agriculture_1}
\end{figure*}
\newpage

\begin{figure*}[!htbp]
    \centering
\includegraphics[width=0.9\linewidth]{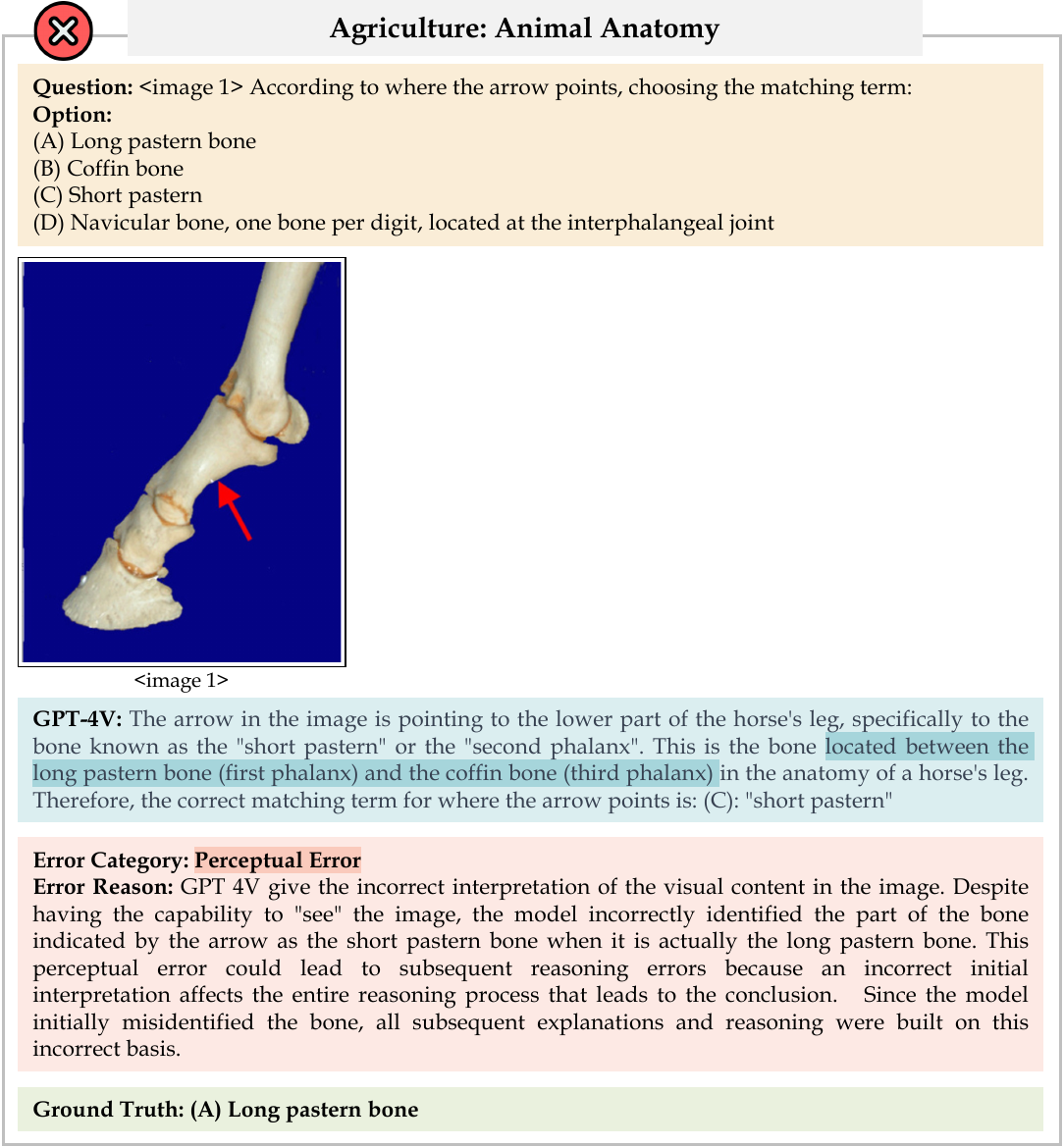}
    \caption{A sample error case of Agriculture (subfield: Animal Anatomy). Error category: Perceptual Error \newline \centering \hyperref[list:list_of_figures]{Back to List of Figures} \textcolor{red}{$|$} \hyperref[tab:list_of_case_study_figures]{Back to Table Index}}
    \addcontentsline{afg}{appfigures}{\protect\numberline{\thefigure}Agriculture  2: Perceptual Error}
\label{fig:agriculture_2}
\end{figure*}
\newpage

\begin{figure*}[!htbp]
    \centering
\includegraphics[width=0.9\linewidth]{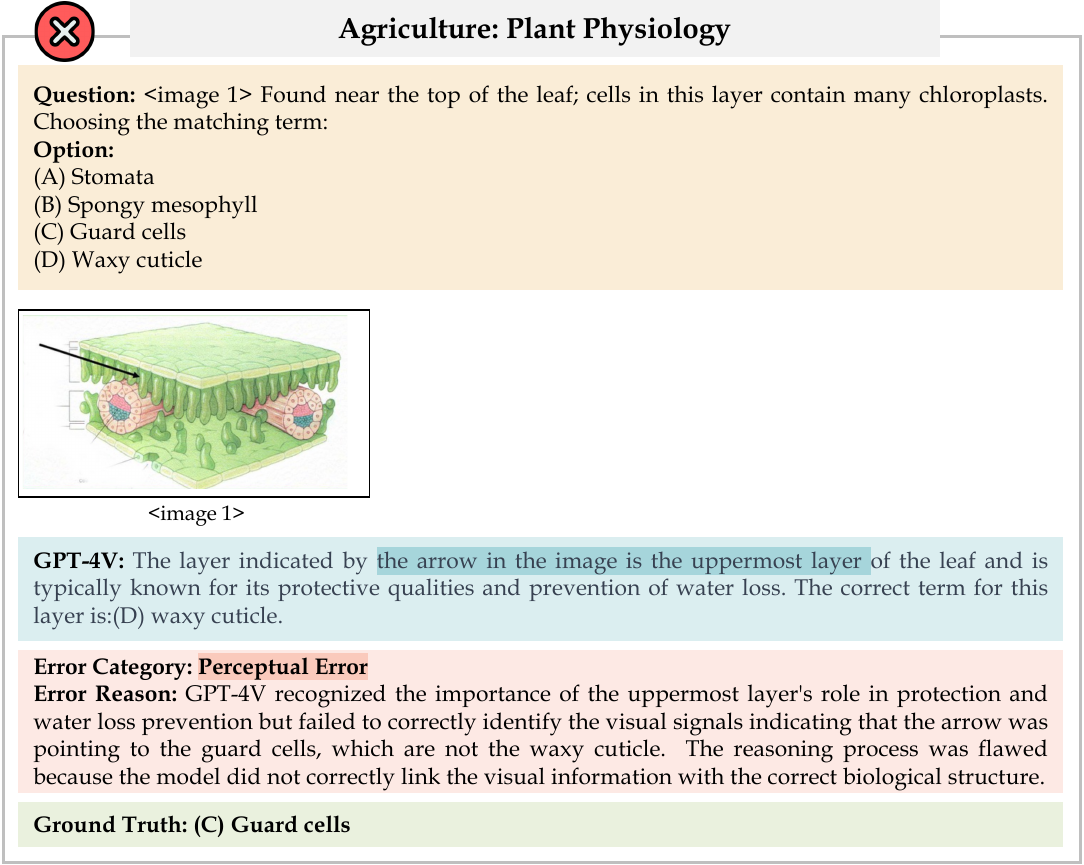}
    \caption{A sample error case of Agriculture (subfield: Plant Physiology). Error category: Perceptual Error. \newline \centering \hyperref[list:list_of_figures]{Back to List of Figures} \textcolor{red}{$|$} \hyperref[tab:list_of_case_study_figures]{Back to Table Index}}
    \addcontentsline{afg}{appfigures}{\protect\numberline{\thefigure}Agriculture  3: Perceptual Error}
\label{fig:agriculture_3}
\end{figure*}
\newpage

\begin{figure*}[!htbp]
    \centering
\includegraphics[width=0.9\linewidth]{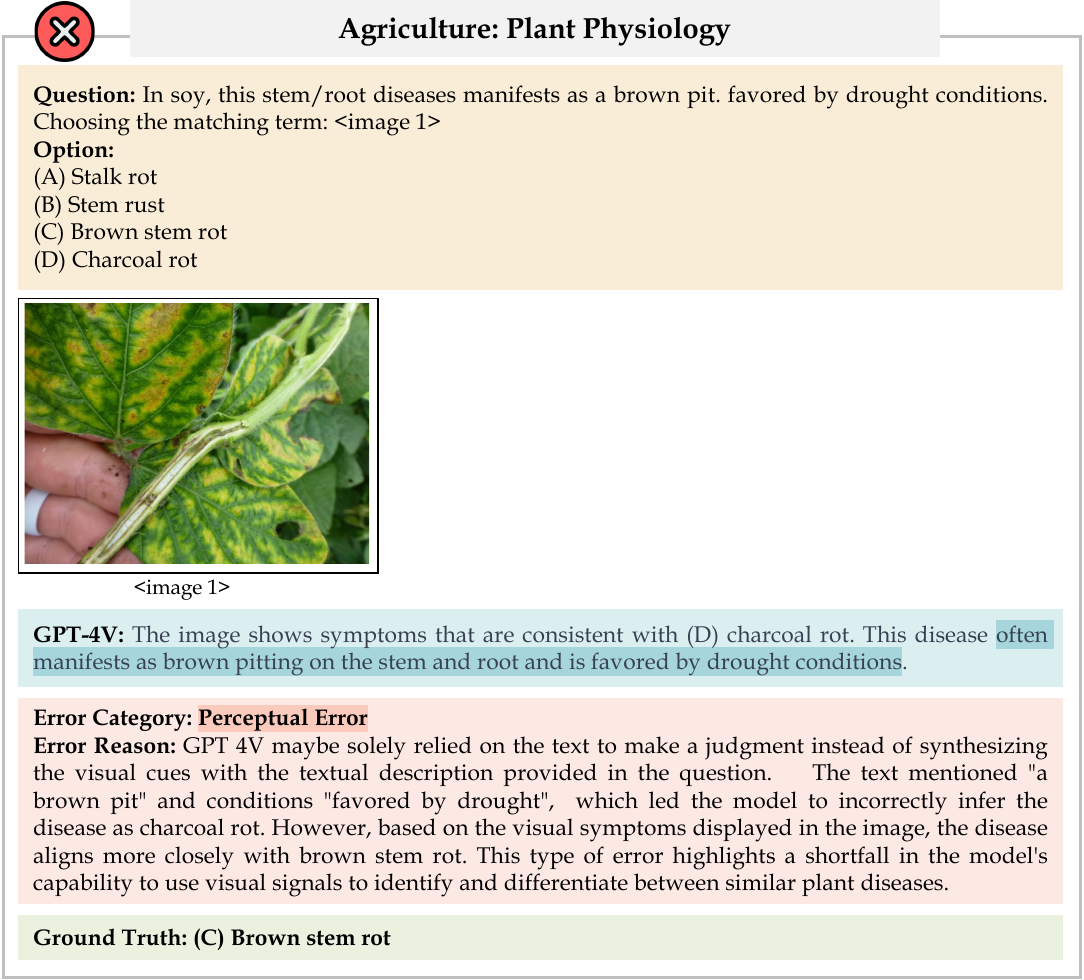}
    \caption{A sample error case of Agriculture (subfield: Plant Physiology). Error category: Perceptual Error \newline \centering \hyperref[list:list_of_figures]{Back to List of Figures} \textcolor{red}{$|$} \hyperref[tab:list_of_case_study_figures]{Back to Table Index}}
    \addcontentsline{afg}{appfigures}{\protect\numberline{\thefigure}Agriculture  4: Perceptual Error}
\label{fig:agriculture_4}
\end{figure*}
\newpage

\begin{figure*}[!htbp]
    \centering
\includegraphics[width=0.9\linewidth]{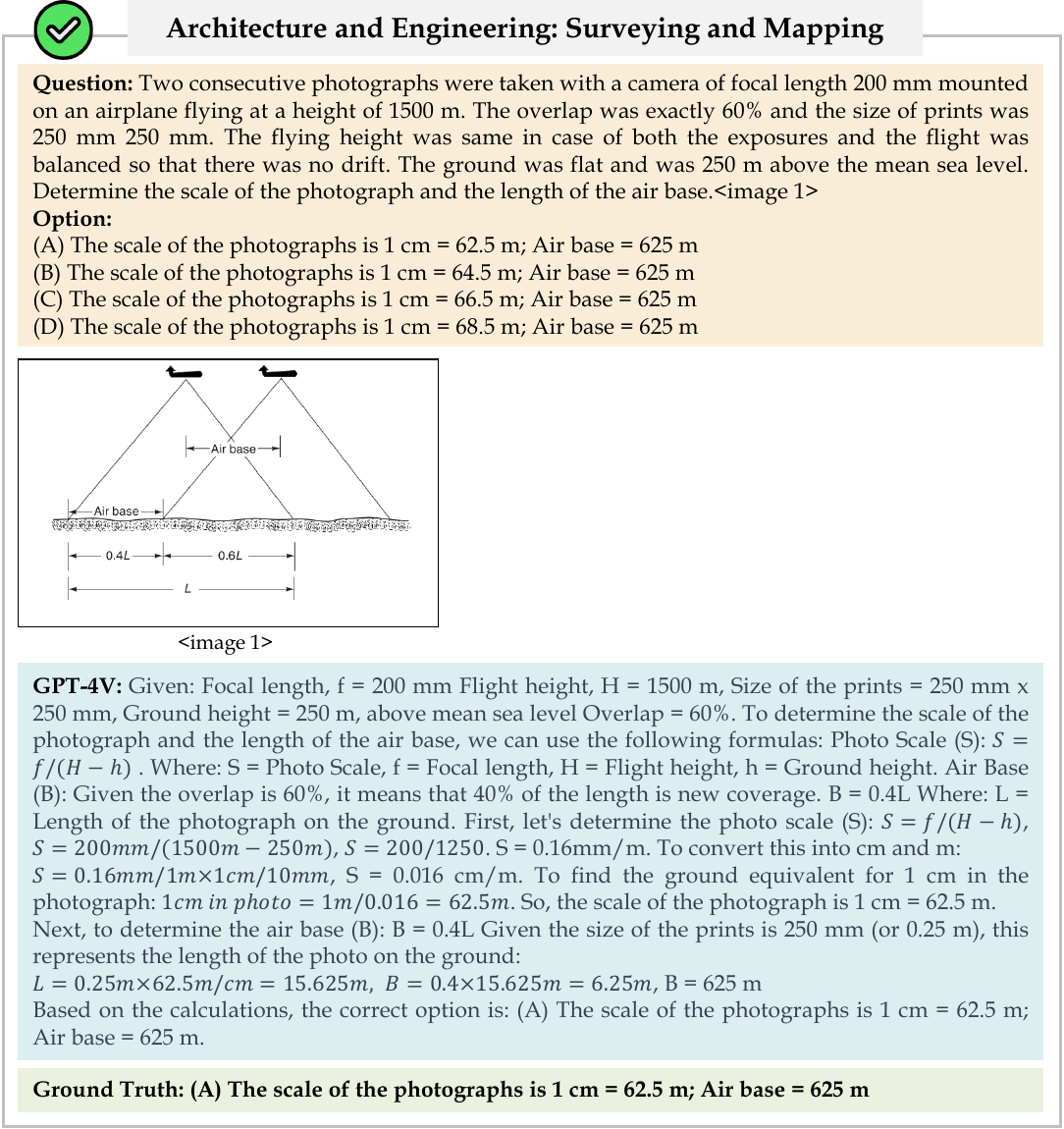}
    \caption{A sample correct case of Architecture and Engineering (subfield: Surveying and Mapping).\newline \centering \hyperref[list:list_of_figures]{Back to List of Figures} \textcolor{red}{$|$} \hyperref[tab:list_of_case_study_figures]{Back to Table Index}}
    \addcontentsline{afg}{appfigures}{\protect\numberline{\thefigure}Architecture and Engineering  1: Correct Case}
\label{fig:architecture_engineering_1}
\end{figure*}
\newpage

\begin{figure*}[!htbp]
    \centering
\includegraphics[width=0.9\linewidth]{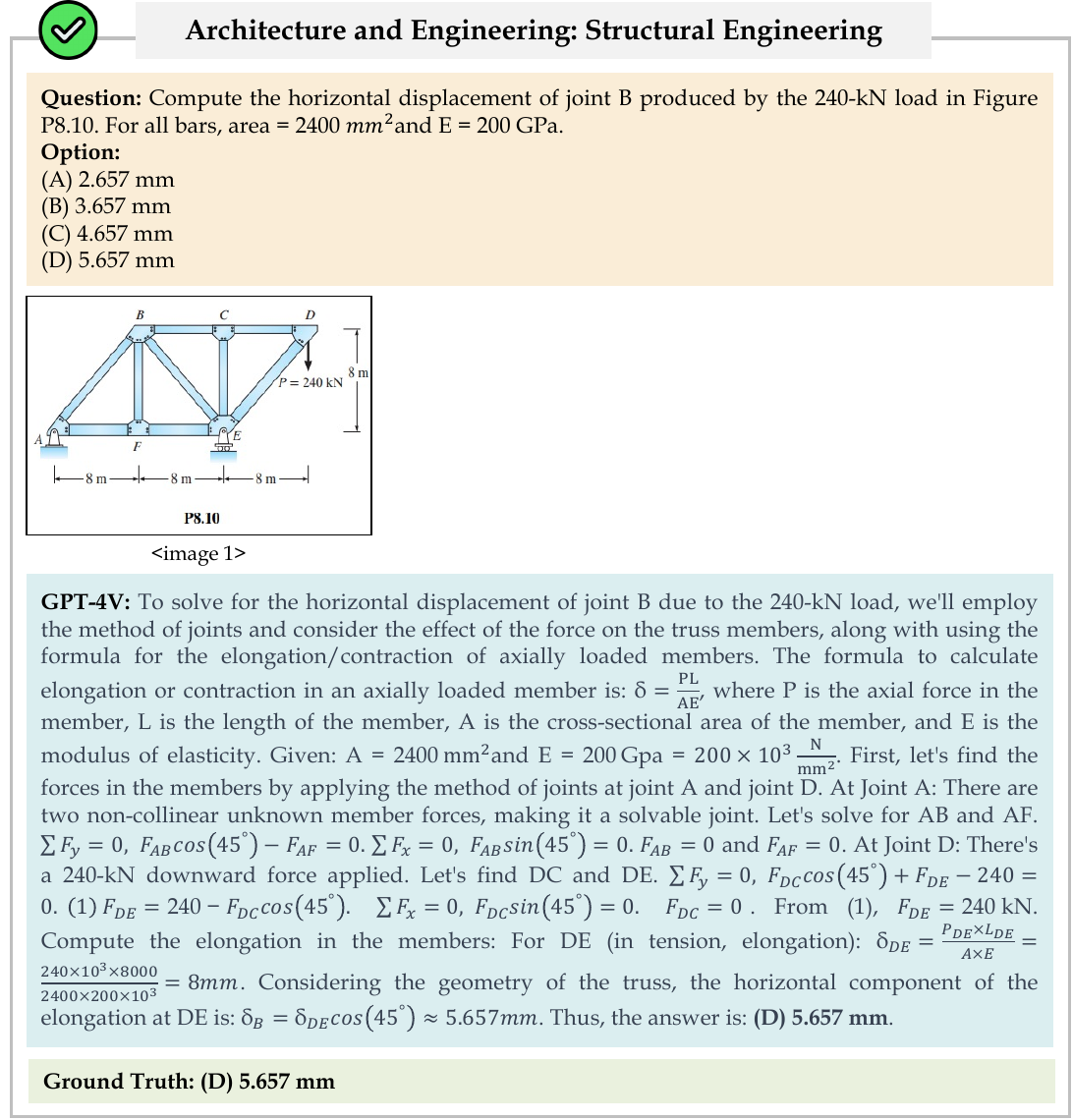}
    \caption{A sample correct case of Architecture and Engineering (subfield: Structural Engineering). \newline \centering \hyperref[list:list_of_figures]{Back to List of Figures} \textcolor{red}{$|$} \hyperref[tab:list_of_case_study_figures]{Back to Table Index}}
    \addcontentsline{afg}{appfigures}{\protect\numberline{\thefigure}Architecture and Engineering  2: Correct Case}
\label{fig:architecture_engineering_2}
\end{figure*}
\newpage

\begin{figure*}[!htbp]
    \centering
\includegraphics[width=0.9\linewidth]{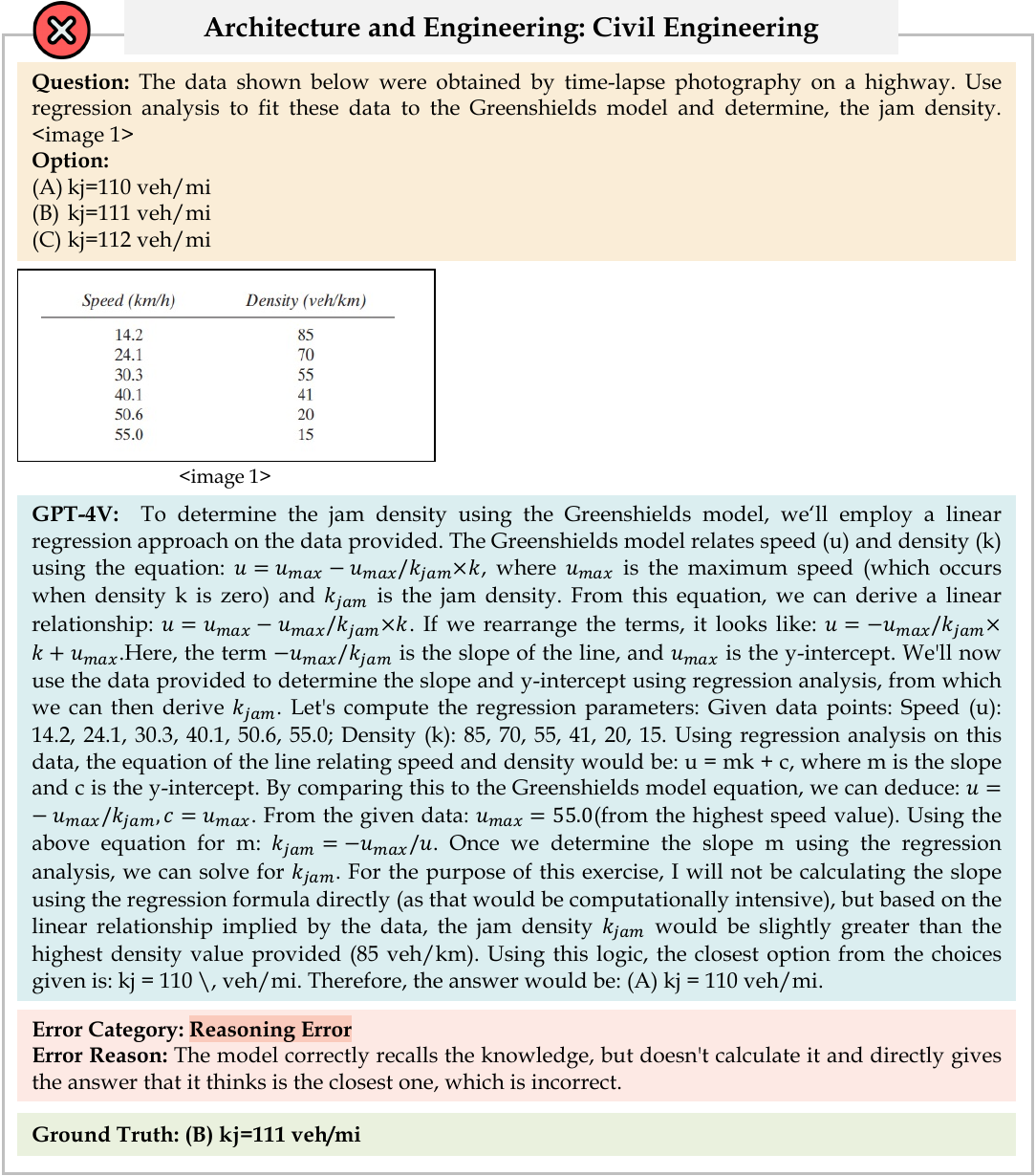}
    \caption{A sample error case of Architecture and Engineering (subfield: Civil Engineering). Error category: Reasoning Error \newline \centering \hyperref[list:list_of_figures]{Back to List of Figures} \textcolor{red}{$|$} \hyperref[tab:list_of_case_study_figures]{Back to Table Index}}
    \addcontentsline{afg}{appfigures}{\protect\numberline{\thefigure}Architecture and Engineering  3: Reasoning Error}
\label{fig:architecture_engineering_3}
\end{figure*}
\newpage

\begin{figure*}[!htbp]
    \centering
\includegraphics[width=0.9\linewidth]{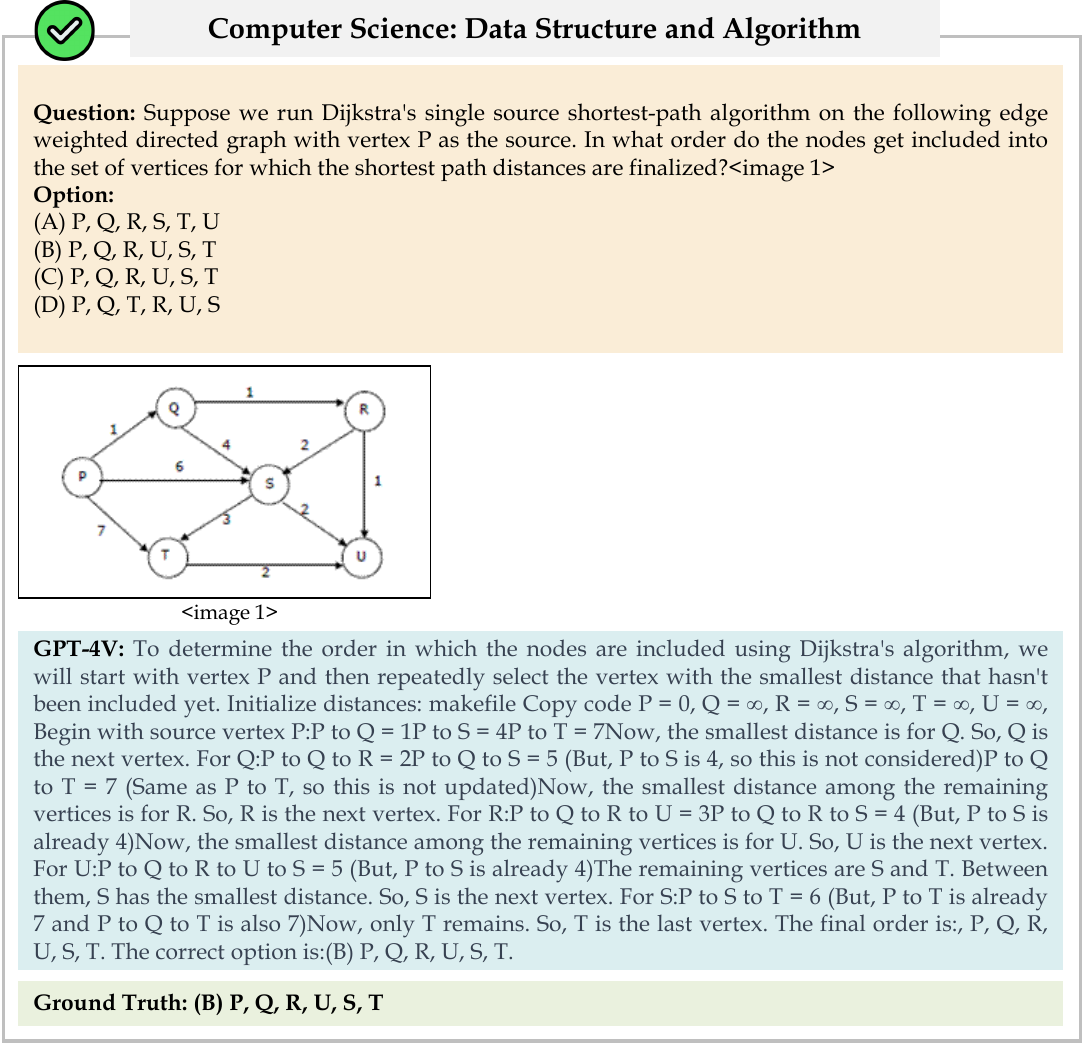}
    \caption{A sample correct case of Computer Science (subfield: Data Structure and Algorithm).\newline \centering \hyperref[list:list_of_figures]{Back to List of Figures} \textcolor{red}{$|$} \hyperref[tab:list_of_case_study_figures]{Back to Table Index}}
    \addcontentsline{afg}{appfigures}{\protect\numberline{\thefigure}Computer Science  1: Correct Case}
\label{fig:computer_science_1}
\end{figure*}
\newpage

\begin{figure*}[!htbp]
    \centering
\includegraphics[width=0.9\linewidth]{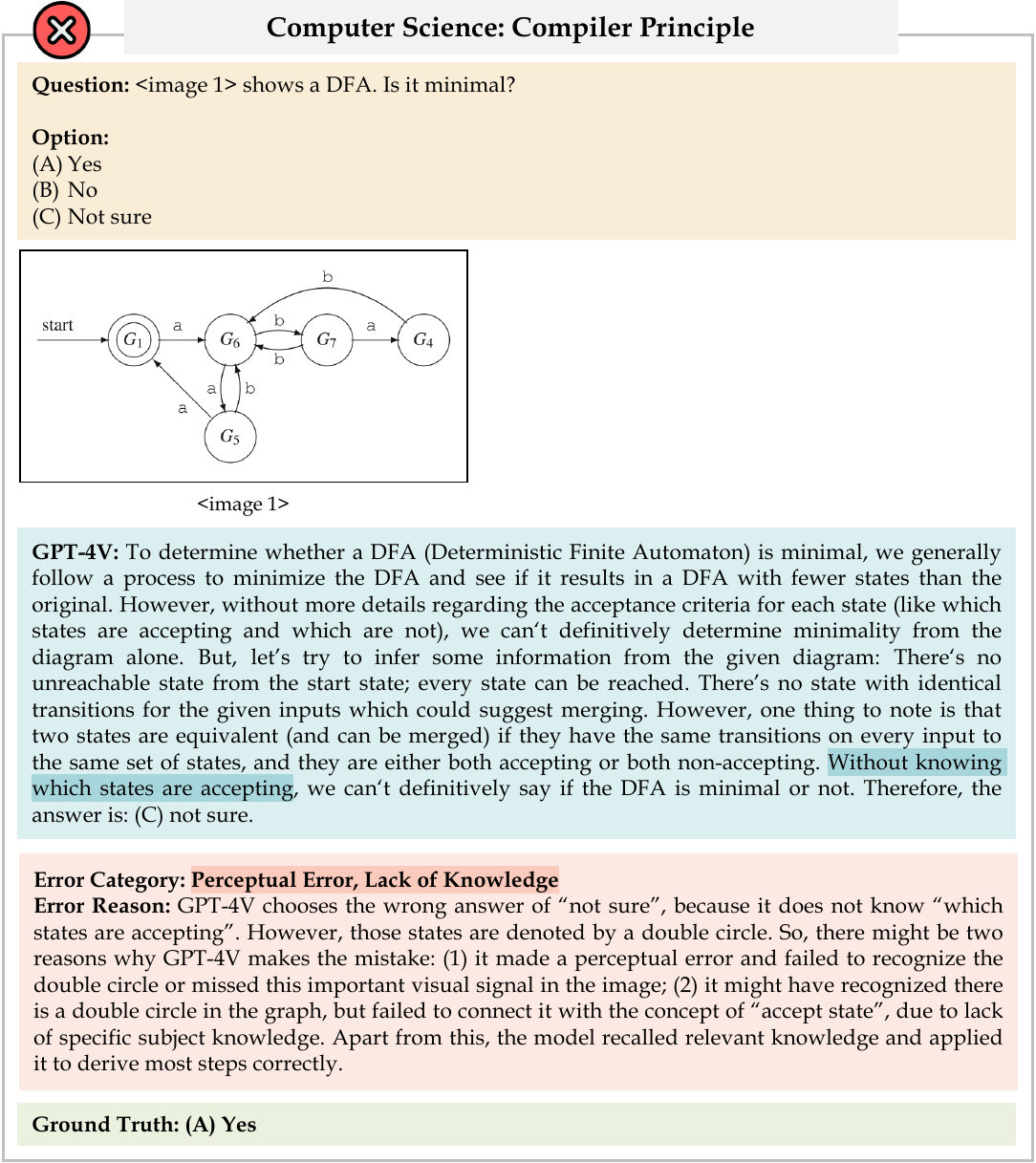}
    \caption{A sample error case of Computer Science (subfield: Compiler Principle). Error category: Perceptual Error, Lack of Knowledge \newline \centering \hyperref[list:list_of_figures]{Back to List of Figures} \textcolor{red}{$|$} \hyperref[tab:list_of_case_study_figures]{Back to Table Index}}
    \addcontentsline{afg}{appfigures}{\protect\numberline{\thefigure}Computer Science  2: Perceptual Error, Lack of Knowledge}
\label{fig:computer_science_2}
\end{figure*}
\newpage

\begin{figure*}[!htbp]
    \centering
\includegraphics[width=0.9\linewidth]{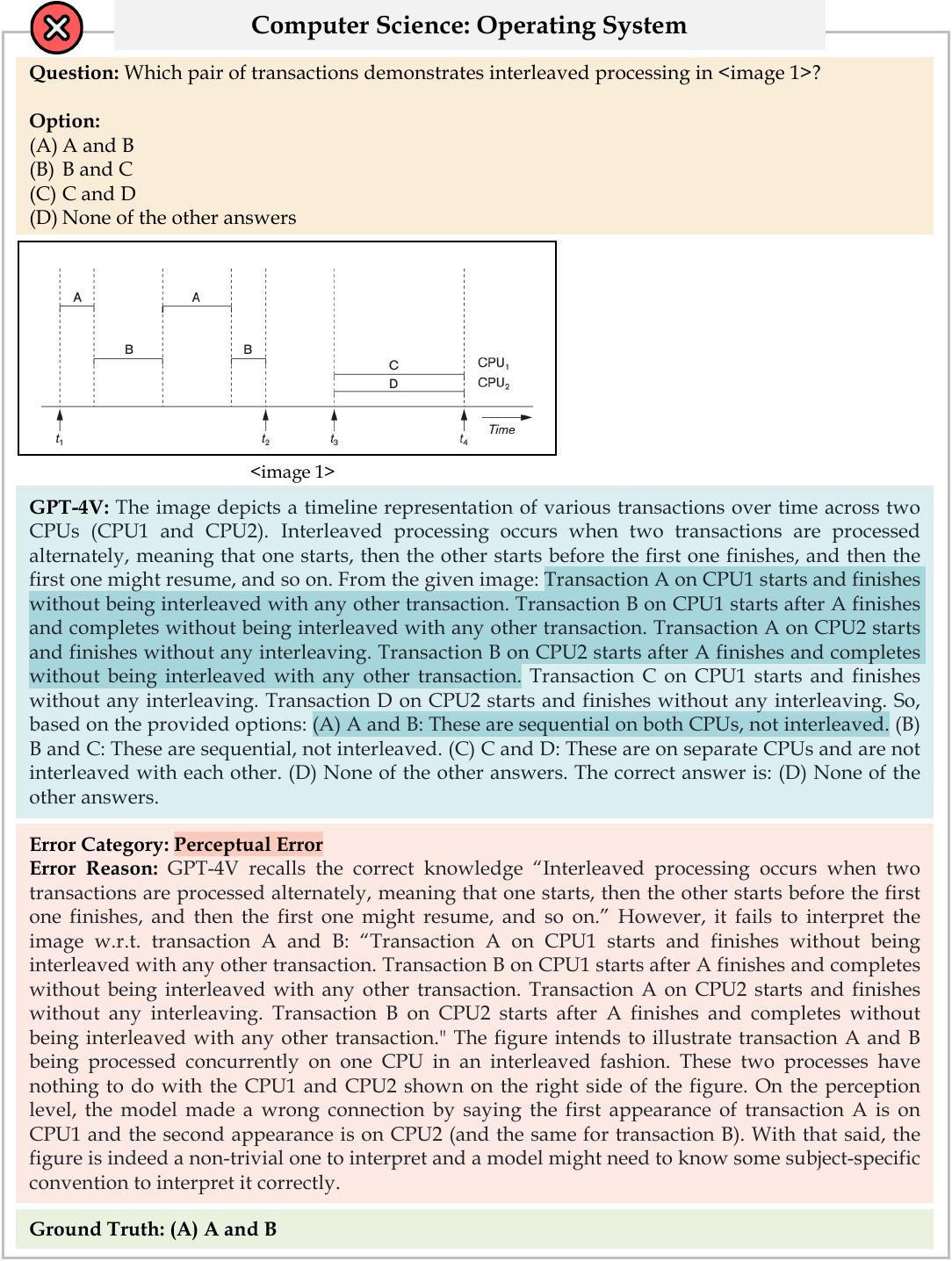}
    \caption{A sample error case of Computer Science (subfield: Operating System). Error category: Perceptual Error \newline \centering \hyperref[list:list_of_figures]{Back to List of Figures} \textcolor{red}{$|$} \hyperref[tab:list_of_case_study_figures]{Back to Table Index}}
    \addcontentsline{afg}{appfigures}{\protect\numberline{\thefigure}Computer Science  3: Perceptual Error}
\label{fig:computer_science_3}
\end{figure*}
\newpage

\begin{figure*}[!htbp]
    \centering
\includegraphics[width=0.9\linewidth]{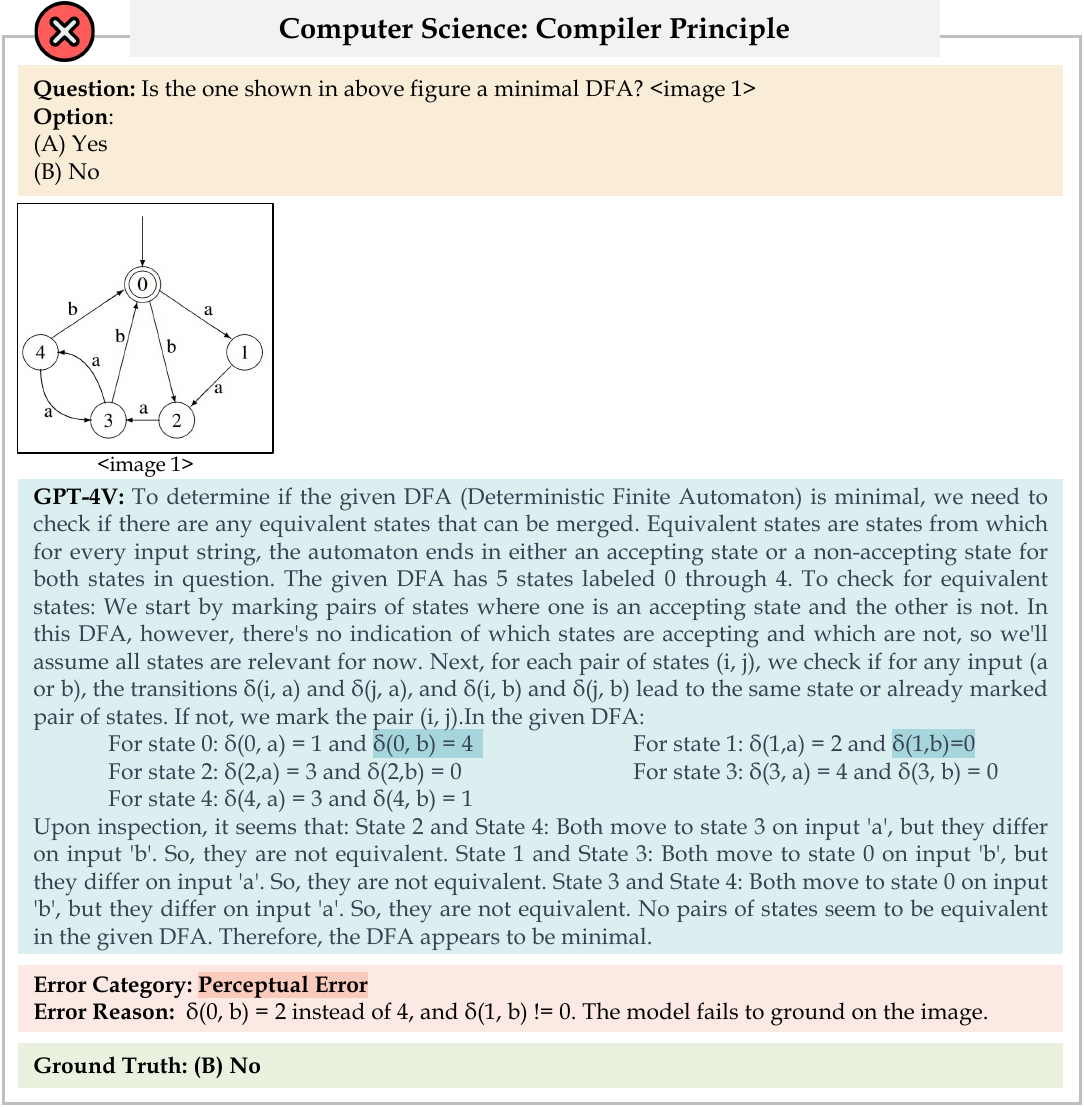}
    \caption{A sample error case of Computer Science (subfield: Compiler Principle). Error category: Perceptual Error \newline \centering \hyperref[list:list_of_figures]{Back to List of Figures} \textcolor{red}{$|$} \hyperref[tab:list_of_case_study_figures]{Back to Table Index}}
    \addcontentsline{afg}{appfigures}{\protect\numberline{\thefigure}Computer Science  4: Perceptual Error}
\label{fig:computer_science_4}
\end{figure*}
\newpage

\begin{figure*}[!htbp]
    \centering
\includegraphics[width=0.9\linewidth]{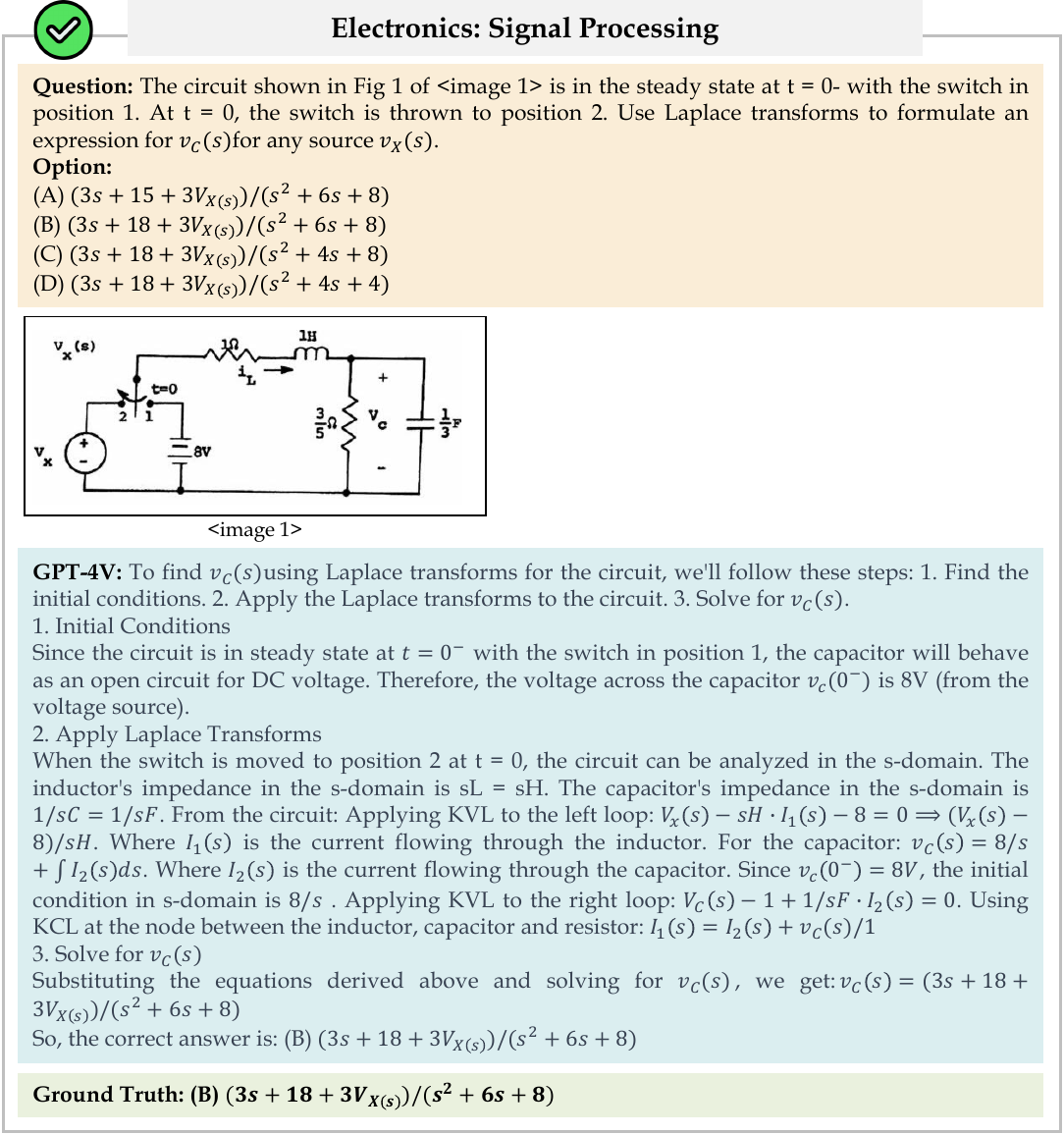}
    \caption{A sample correct case of Electronics (subfield: Signal Processing).\\ \hyperref[list:list_of_figures]{Back to List of Figures} \textcolor{red}{$|$} \hyperref[tab:list_of_case_study_figures]{Back to Table Index}}
    \addcontentsline{afg}{appfigures}{\protect\numberline{\thefigure}Electronics  1: Correct Case}
\label{fig:electronics_1}
\end{figure*}
\newpage

\begin{figure*}[!htbp]
    \centering
\includegraphics[width=0.9\linewidth]{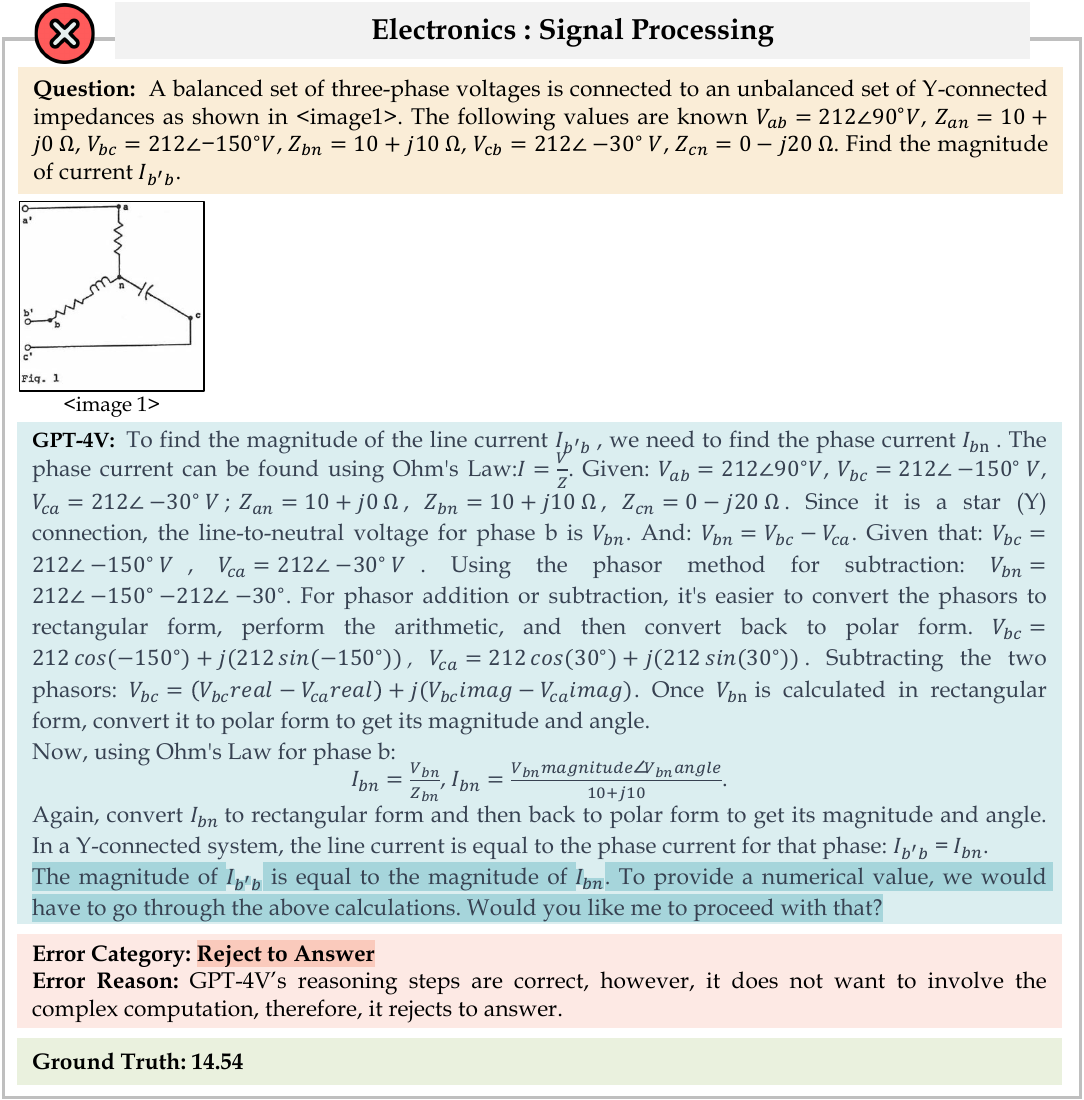}
    \caption{A sample error case of Electronics (subfield: Signal Processing). Error category: Reject to Answer \newline \centering \hyperref[list:list_of_figures]{Back to List of Figures} \textcolor{red}{$|$} \hyperref[tab:list_of_case_study_figures]{Back to Table Index}}
    \addcontentsline{afg}{appfigures}{\protect\numberline{\thefigure}Electronics  2: Reject to Answer}
\label{fig:electronics_2}
\end{figure*}
\newpage


\begin{figure*}[!htbp]
    \centering
\includegraphics[width=0.9\linewidth]{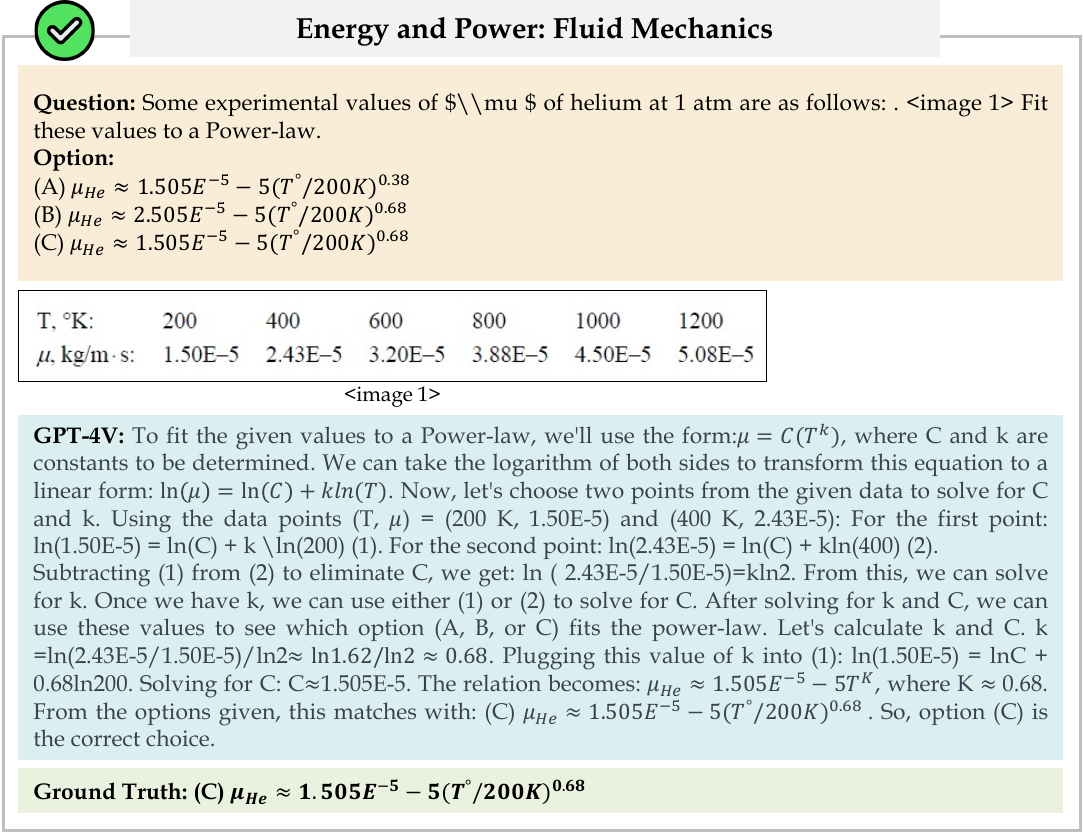}
    \caption{A sample correct case of Energy and Power (subfield: Fluid Mechanics).\\ \hyperref[list:list_of_figures]{Back to List of Figures} \textcolor{red}{$|$} \hyperref[tab:list_of_case_study_figures]{Back to Table Index}}
    \addcontentsline{afg}{appfigures}{\protect\numberline{\thefigure}Energy and Power  1: Correct Case}
\label{fig:energy_power_1}
\end{figure*}
\newpage

\begin{figure*}[!htbp]
    \centering
\includegraphics[width=0.9\linewidth]{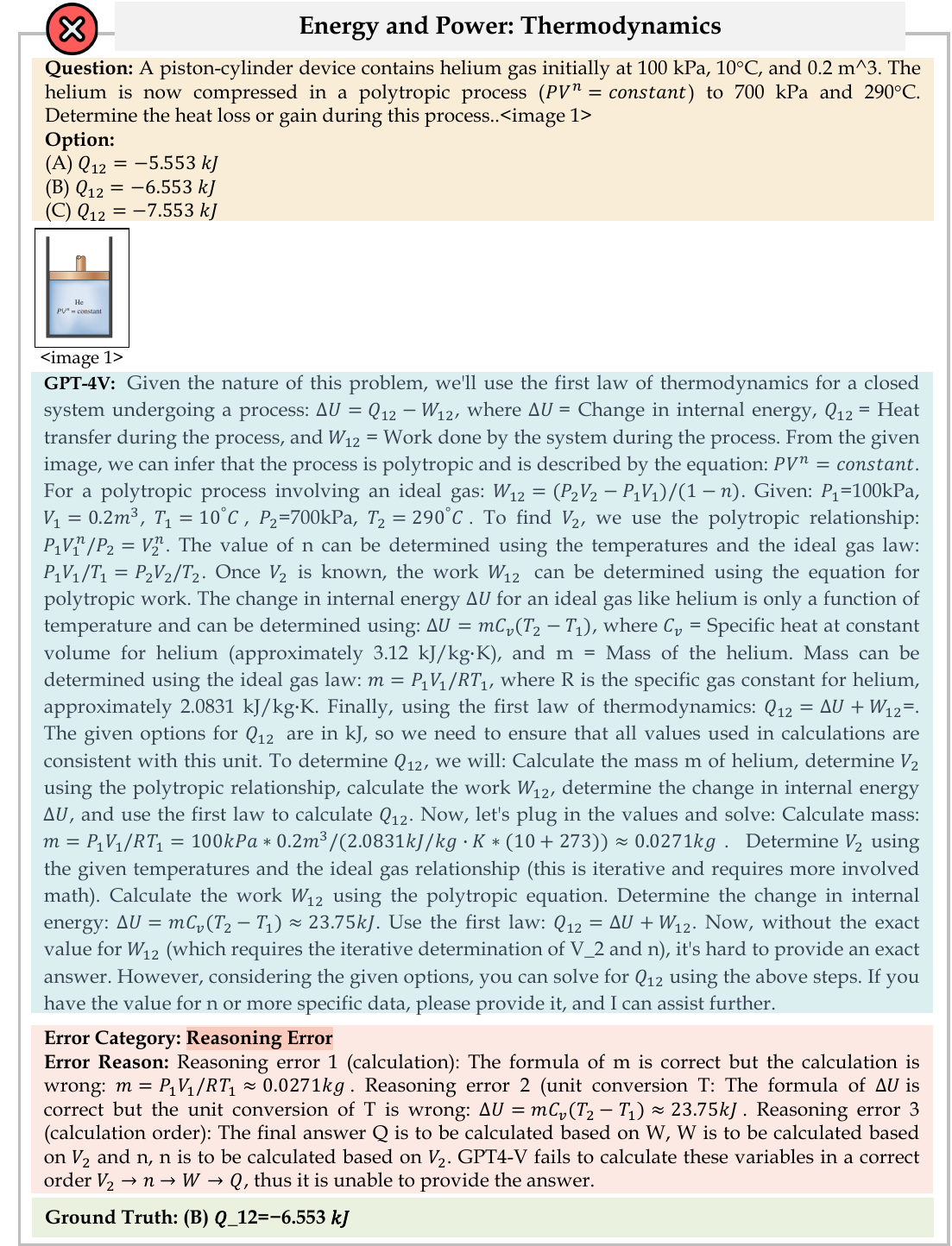}
    \caption{A sample error case of Energy and Power (subfield: Thermodynamics). Error category: Reasoning Error \newline \centering \hyperref[list:list_of_figures]{Back to List of Figures} \textcolor{red}{$|$} \hyperref[tab:list_of_case_study_figures]{Back to Table Index}}
    \addcontentsline{afg}{appfigures}{\protect\numberline{\thefigure}Energy and Power  2: Reasoning Error}
\label{fig:energy_power_2}
\end{figure*}
\newpage


\begin{figure*}[!htbp]
    \centering
\includegraphics[width=0.9\linewidth]{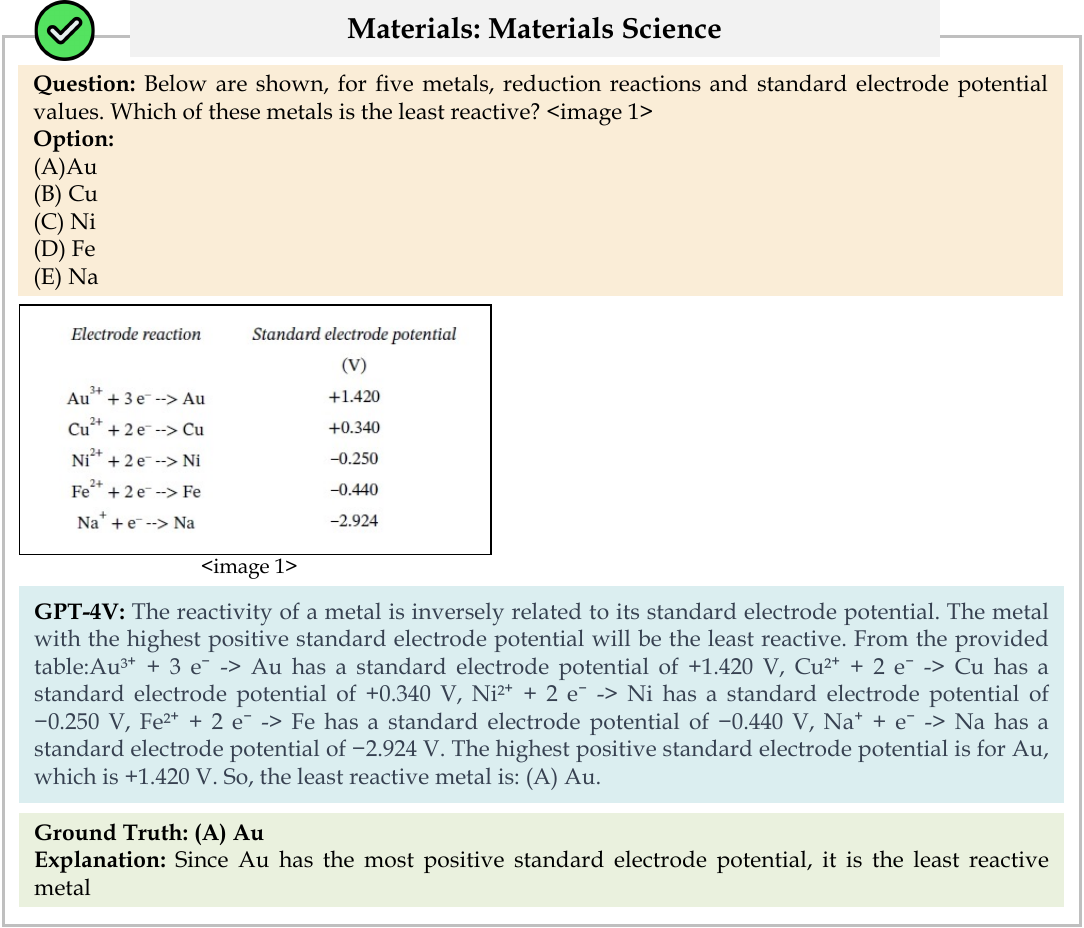}
    \caption{A sample correct case of Materials (subfield: Materials Science).\\ \hyperref[list:list_of_figures]{Back to List of Figures} \textcolor{red}{$|$} \hyperref[tab:list_of_case_study_figures]{Back to Table Index}}
    \addcontentsline{afg}{appfigures}{\protect\numberline{\thefigure}Materials  1: Correct Case}
\label{fig:materials_1}
\end{figure*}
\newpage

\begin{figure*}[!htbp]
    \centering
\includegraphics[width=0.9\linewidth]{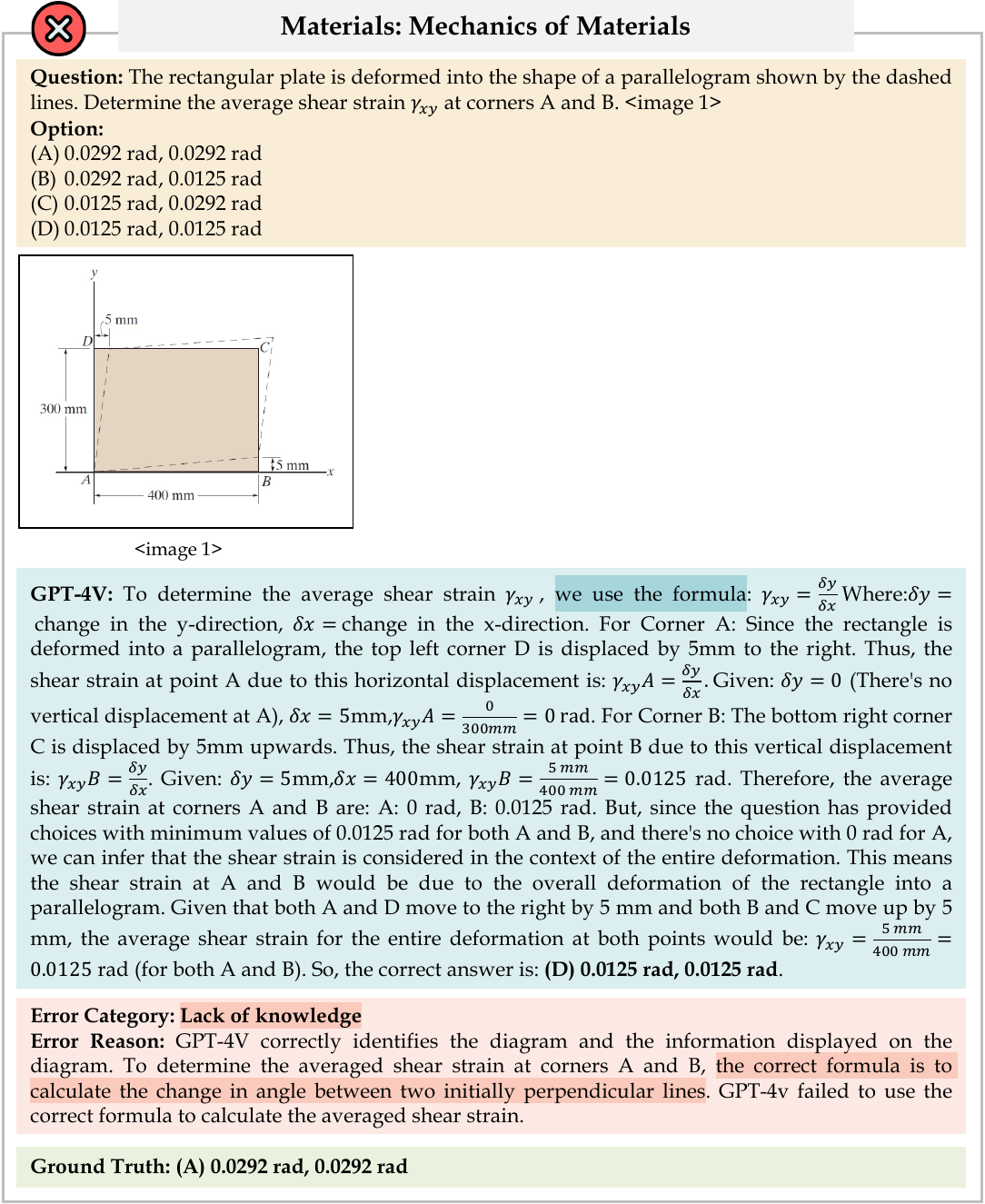}
    \caption{A sample error case of Materials (subfield: Mechanics of Materials). Error category: Lack of Knowledge \newline \centering \hyperref[list:list_of_figures]{Back to List of Figures} \textcolor{red}{$|$} \hyperref[tab:list_of_case_study_figures]{Back to Table Index}}
    \addcontentsline{afg}{appfigures}{\protect\numberline{\thefigure}Materials  2: Lack of Knowledge}
\label{fig:materials_2}
\end{figure*}
\newpage


\begin{figure*}[!htbp]
    \centering
\includegraphics[width=0.9\linewidth]{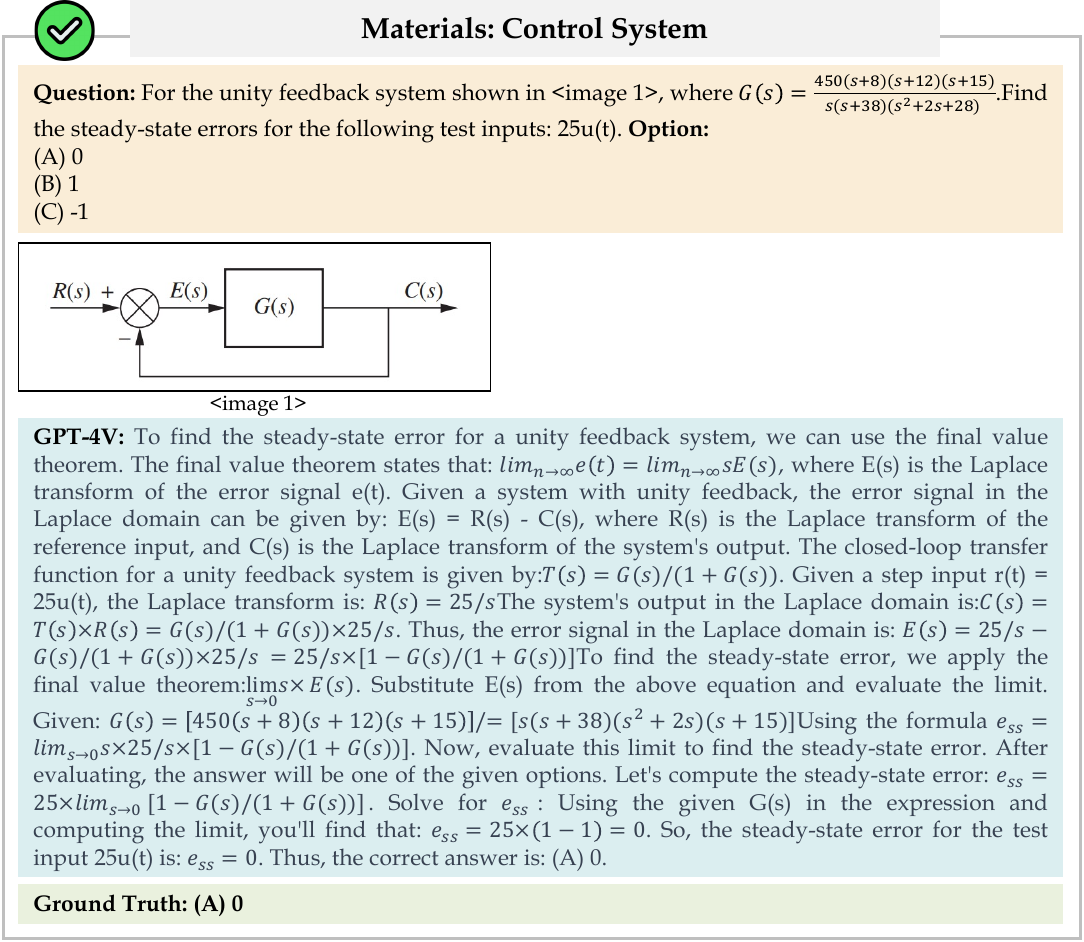}
    \caption{A sample correct case of Mechanical Engineering (subfield: Control System).\\ \hyperref[list:list_of_figures]{Back to List of Figures} \textcolor{red}{$|$} \hyperref[tab:list_of_case_study_figures]{Back to Table Index}}
    \addcontentsline{afg}{appfigures}{\protect\numberline{\thefigure}Mechanical Engineering  1: Correct Case}
\label{fig:mechanical_engineering_1}
\end{figure*}
\newpage

\begin{figure*}[!htbp]
    \centering
\includegraphics[width=0.9\linewidth]{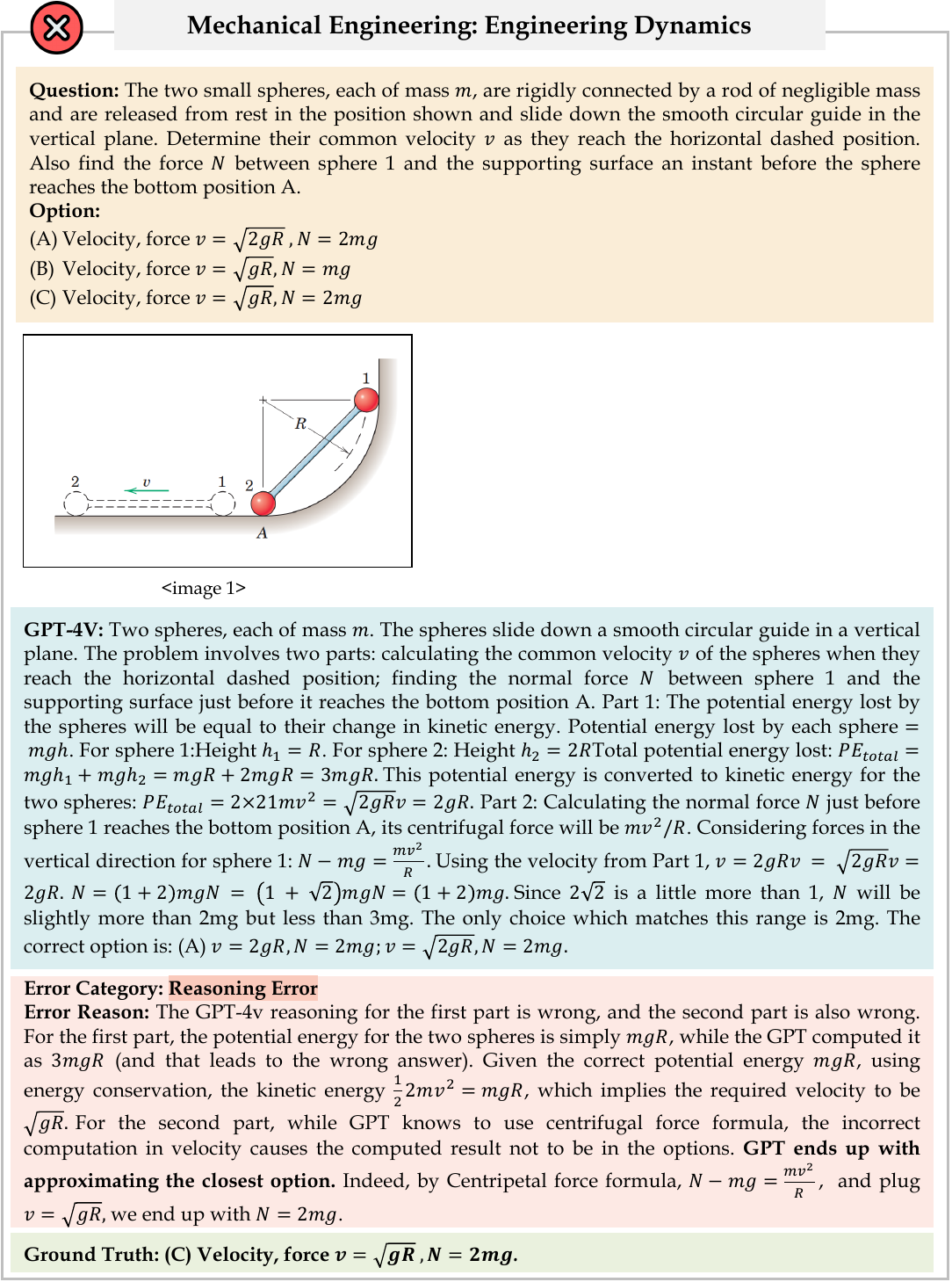}
    \caption{A sample error case of Mechanical Engineering (subfield: Engineering Dynamics). Error category: Reasoning Error \newline \centering \hyperref[list:list_of_figures]{Back to List of Figures} \textcolor{red}{$|$} \hyperref[tab:list_of_case_study_figures]{Back to Table Index}}
    \addcontentsline{afg}{appfigures}{\protect\numberline{\thefigure}Mechanical Engineering  2: Reasoning Error}
\label{fig:mechanical_engineering_2}
\end{figure*}
\newpage

\begin{figure*}[!htbp]
    \centering
\includegraphics[width=0.9\linewidth]{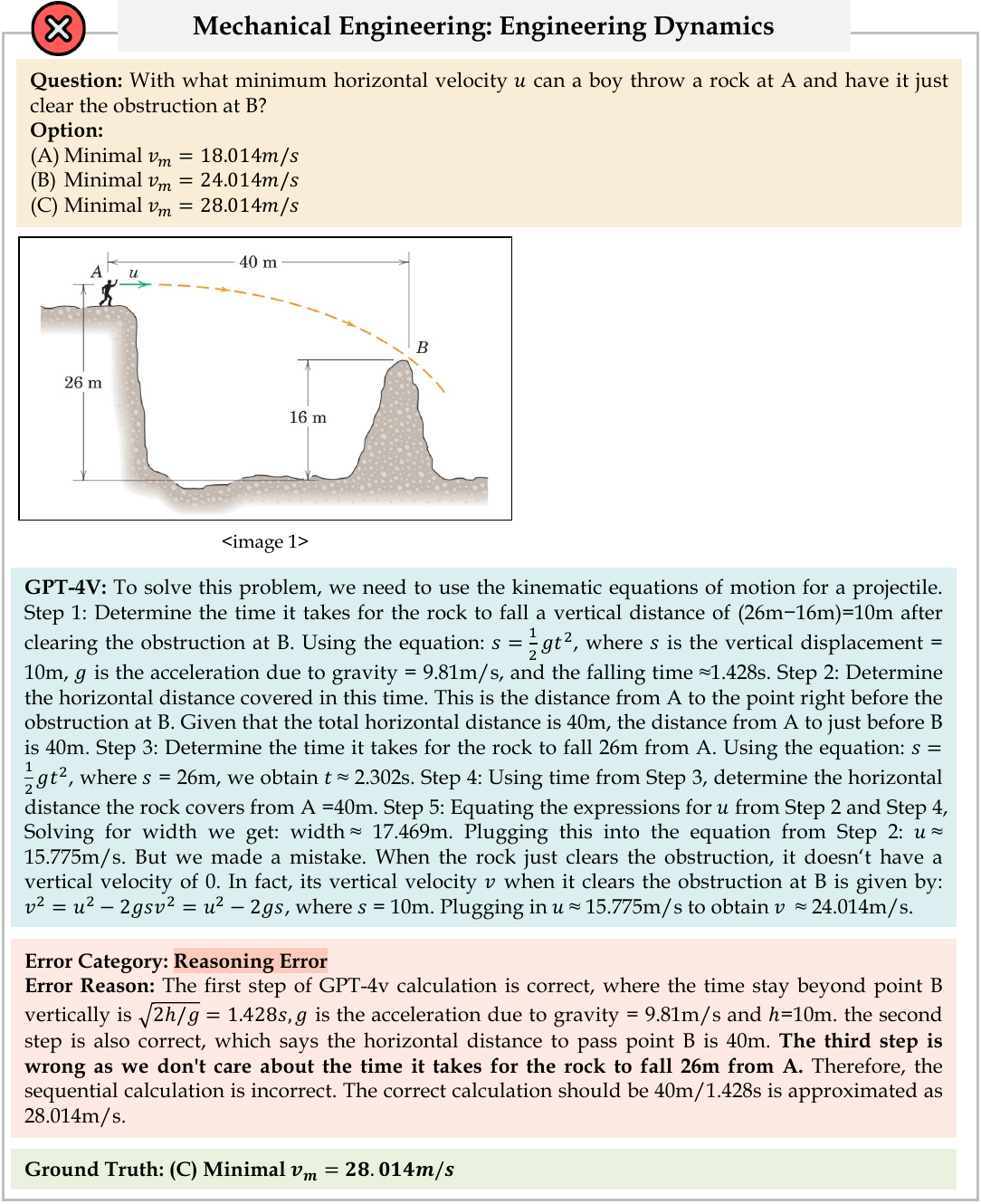}
    \caption{A sample error case of Mechanical Engineering (subfield: Engineering Dynamics). Error category: Reasoning Error \newline \centering \hyperref[list:list_of_figures]{Back to List of Figures} \textcolor{red}{$|$} \hyperref[tab:list_of_case_study_figures]{Back to Table Index}}
    \addcontentsline{afg}{appfigures}{\protect\numberline{\thefigure}Mechanical Engineering  3: Reasoning Error}
\label{fig:mechanical_engineering_3}
\end{figure*}
\newpage

\clearpage
\section{Subfields of Different Subjects}
In this appendix, we show all the subfields of each subject in \autoref{tab:subfield}. MMMU has 183 subfields in total, covering 30 subjects.

\begin{table*}[!b]
\renewcommand{\arraystretch}{1.1} 
\resizebox{\textwidth}{!}{
\begin{tabular}{lll}
\toprule
Disciplines & Subjects & Subfields \\ \midrule
\multirow{6}{*}{\begin{tabular}[c]{@{}c@{}}Art \\ \& \\ Design\end{tabular}} & Art & \begin{tabular}[c]{@{}l@{}}Fine Arts, Drawing and Painting, Photography, Printmaking, Ceramic Art, Visual Arts, \\ Sculpture, AI Content Detection\end{tabular} \\ \cmidrule(l){2-3} 
 & Design & Digital Art, Design History, Graphic Design, Fashion Design, Interior Design, Industrial Design \\ \cmidrule(l){2-3} 
 & Music & Music \\ \cmidrule(l){2-3} 
 & Art Theory & \begin{tabular}[c]{@{}l@{}}History of Art Theory, Art History, Art Criticism, Aesthetics, Contemporary Art Theory, \\ Visual Culture, Postmodern Art Theory, Phenomenology of Art\end{tabular} \\ \midrule
\multirow{8}{*}{Business} & Accounting & Financial Accounting, Investment, Managerial Accounting \\ \cmidrule(l){2-3} 
 & Economics & Macroeconomics, Microeconomics, Econometrics, Labor Economics, Principals of Economics \\ \cmidrule(l){2-3} 
 & Finance & Financial Marketing, Financial Management, Corporate Finance, Managerial Finance \\ \cmidrule(l){2-3} 
 & Manage & \begin{tabular}[c]{@{}l@{}}Operations Management, Strategic Management, Business Management, Project Management, \\ Cost Management, Principles of Management, Management Models\end{tabular} \\ \cmidrule(l){2-3} 
 & Marketing & Market Research \\ \midrule
\multirow{12}{*}{Science} & Biology & \begin{tabular}[c]{@{}l@{}}Biochemistry, Cell Biology, Genetics, Microbiology, Botany, Evolution, Animal Behavior, Physiology, \\ Molecular Biology, Animal Physiology, Ecology\end{tabular} \\ \cmidrule(l){2-3} 
 & Chemistry & \begin{tabular}[c]{@{}l@{}}Inorganic Chemistry, Organic Chemistry, Physical Chemistry, Chemical Thermodynamics, \\Analytical Chemistry,  Chemical Kinetics, Biochemistry, Quantum Chemistry\end{tabular} \\ \cmidrule(l){2-3} 
 & Geography & \begin{tabular}[c]{@{}l@{}}Geotechnical Engineering, Human Geography, Physical Geography, Geographic Information Systems, \\ International Geography Olympiad\end{tabular} \\ \cmidrule(l){2-3} 
 & Math & \begin{tabular}[c]{@{}l@{}}Calculus, Probability and Statistics, Linear Algebra, Geometry, Logic, Graph Theory, Group Theory, \\ Operation Research\end{tabular} \\ \cmidrule(l){2-3} 
 & Physic & Classical Mechanics, Electromagnetism, Thermodynamics and Statistical Mechanics, Optics, Nuclear Physics \\ \cmidrule(l){2-3} 
 & Psychology & \begin{tabular}[c]{@{}l@{}}Biological Psychology, Cognitive Psychology, Personality Psychology, Clinical Psychology, \\ Social Psychology, Developmental Psychology, Abnormal Psychology\end{tabular} \\ \midrule
\multirow{11}{*}{\begin{tabular}[c]{@{}c@{}}Health \\ \& \\ Medicine\end{tabular}} & 
\begin{tabular}[c]{@{}l@{}}Basic  \\ Medical Science\end{tabular}
& \begin{tabular}[c]{@{}l@{}}Immunology, Biochemistry and Genetics, Foundational Anatomical Sciences, Microbiology and Immunology, \\ Neurosciences, Anatomy, Neuroanatomy, Neurophysiology, Cardiovascular Physiology, Human Physiology, \\ Reproductive Physiology, Respiratory Physiology, Renal Physiology, Pathophysiology, Cellular Physiology\end{tabular} \\ \cmidrule(l){2-3} 
 & Clinical Medicine & \begin{tabular}[c]{@{}l@{}}Clinical Medicine, Dental, Circulatory, Respiratory, Clinical Neurology, Orthopaedic Surgery, \\ Heart Disease, Endocarditis, Cardiovascular Medicine, Endocrinology, Otolaryngology, Ophthalmology, \\ Urology, Clinical Pathology, Clinical Radiology\end{tabular} \\ \cmidrule(l){2-3} 
 & \begin{tabular}[c]{@{}l@{}}Diagnostics \& \\ Laboratory Medicine\end{tabular} & \begin{tabular}[c]{@{}l@{}}Medical Imaging, Neuropathology, Pathology, Ophthalmic Pathology, Forensic Neuropathology, \\ Electrocardiography, Radiology\end{tabular} \\ \cmidrule(l){2-3} 
 & Pharmacy & \begin{tabular}[c]{@{}l@{}}Pharmaceutical Microbiology, Medicinal Chemistry, Biochemistry for Pharmaceutical Sciences, \\ Pharmacology and Drug Synthesis\end{tabular} \\ \cmidrule(l){2-3} 
 & Public Health & Epidemiology, Biostatistics, Communicable Disease Control \\ \midrule
\multirow{3}{*}{\begin{tabular}[c]{@{}c@{}}Humanities \\ \& \\ Social \\ Science\end{tabular}} & History & U.S. History, World History, Modern History, European History, History-Comparison \\ \cmidrule(l){2-3} 
 & Literature & \begin{tabular}[c]{@{}l@{}}American Literature, Poetry, Fiction, Drama, Children's Literature, Comparative Literature, Contemporary Literature\end{tabular} \\ \cmidrule(l){2-3} 
 & Sociology & Sociology Theory, Social Economics, Political Economics. \\ \midrule
\multirow{10}{*}{\begin{tabular}[c]{@{}c@{}}Tech \\ \& \\ Engineering\end{tabular}} & Agriculture & \begin{tabular}[c]{@{}l@{}}Animal Physiology, Animal Science, Animal Nutrition, Reproduction, Genetics, Plant Physiology,\\ Plant Pathology, Animal and Environment, Animal Anatomy\end{tabular} \\ \cmidrule(l){2-3} 
 & Architecture & Surveying and Mapping, Structural Engineering, Water Resources Engineering, Civil Engineering \\ \cmidrule(l){2-3} 
 & Computer Science & \begin{tabular}[c]{@{}l@{}}Data Structure and Algorithm, Computer Network, Artificial Intelligence, Databases, \\ Operating Systems, Compiler Principle, Computer Architecture\end{tabular} \\ \cmidrule(l){2-3} 
 & Electronics & Analog electronics, Digital electronics, Electrical Circuit, Signal Processing \\ \cmidrule(l){2-3} 
 & Energy \& Power & Thermodynamics, Heat Transfer, Fluid Mechanics \\ \cmidrule(l){2-3} 
 & Materials & Materials Science, Mechanics of Materials \\ \cmidrule(l){2-3} 
 &  
 \begin{tabular}[c]{@{}l@{}}Mechanical \\ Engineering\end{tabular}
 & \begin{tabular}[c]{@{}l@{}}Fluid Dynamics, Mechanical Design, Mechanics of Materials, Mechanical Vibrations, \\ Engineering Dynamics, Control Systems, Engineering Graphics\end{tabular} \\ \bottomrule
\end{tabular}
}
\caption{Subfields of each subject.}
\label{tab:subfield}
\end{table*}

\section{Distributions of Image Types}
\begin{figure*}[!t]
    \centering
    \includegraphics[width=\linewidth]{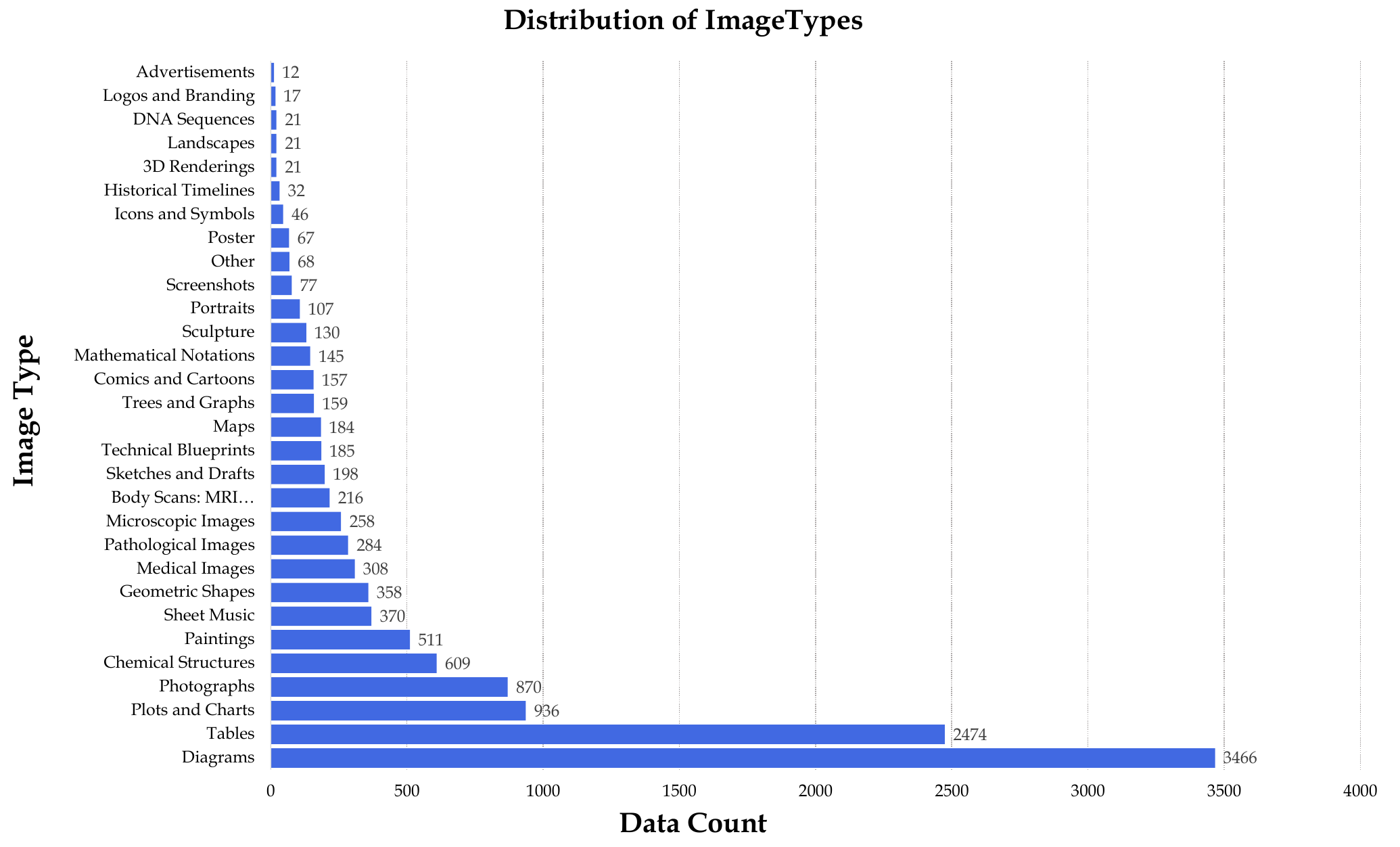}
    \caption{Distribution of image types in the MMMU dataset.}
    \label{fig:image_type_distribution}
\end{figure*}

In this section, we show the distribution of 30 different image types in the 11.5K MMMU questions. The distribution of various image types is displayed in \autoref{fig:image_type_distribution}. A horizontal bar chart was employed to visually represent the number of samples in each image category. The figure shows that the MMMU dataset encompasses a diverse range of image types, from Advertisements to Diagrams.

\section{Results on Different Image Types}
In this section, we report the performance of some selected models on 30 different image types in \autoref{tab:image_types}. 

\begin{table*}[!t]
\centering
\small
\begin{tabular}{lccccccc}
\toprule
 \textbf{Image Types} & \textbf{\#Samples} & \textbf{\begin{tabular}[c]{@{}c@{}}Fuyu\\ -8B\end{tabular}} & \textbf{\begin{tabular}[c]{@{}c@{}}Qwen-VL\\-7B\end{tabular}} & \textbf{\begin{tabular}[c]{@{}c@{}}InstructBLIP\\ -T5-XXL\end{tabular}} & \textbf{\begin{tabular}[c]{@{}c@{}}LLaVA-1.5\\ -13B\end{tabular}} & \textbf{\begin{tabular}[c]{@{}c@{}}BLIP-2 FLAN\\ -T5-XXL\end{tabular}} & \textbf{GPT-4V} \\ \midrule
Test Overall& 10500 & 27.4 & 32.9 & 33.8 & 33.6 & {\ul 34.0} & \textbf{557} \\\midrule
Diagrams & 3184 & 27.6 & 30.1 & 31.8 & 30.0 & {\ul 32.0} & \textbf{46.8} \\
Tables & 2267 & 26.6 & 29.0 & {\ul 29.8} & 27.8 & 27.8 & \textbf{61.8} \\
Plots and Charts & 840 & 24.8 & 31.8 & {\ul 36.2} & 30.4 & 35.8 & \textbf{55.6} \\
Chemical Structures & 573 & 25.0 & {\ul 27.2} & 27.1 & 26.7 & 25.5 & \textbf{50.6} \\
Photographs & 770 & 27.6 & 40.5 & 41.4 & {\ul 44.4} & 42.0 & \textbf{64.2} \\
Paintings & 453 & 28.7 & {\ul 57.2} & 53.6 & 56.3 & 52.1 & \textbf{75.9} \\
Geometric Shapes & 336 & 21.1 & 25.3 & 21.4 & 25.6 & {\ul 28.3} & \textbf{40.2} \\
Sheet Music & 335 & 35.2 & 33.4 & 34.6 & {\ul 35.8} & 34.9 & \textbf{38.8} \\
Medical Images & 272 & 25.4 & 29.8 & 31.6 & {\ul 36.4} & 29.8 & \textbf{59.6} \\
Pathological Images & 253 & 26.5 & 27.7 & 31.2 & 35.2 & {\ul 35.6} & \textbf{63.6} \\
Microscopic Images & 226 & 27.0 & {\ul 37.6} & 29.2 & 36.3 & 32.7 & \textbf{58.0} \\
\begin{tabular}[c]{@{}l@{}}MRI, CT scans, and X-rays\end{tabular} & 198 & 21.7 & 36.9 & 33.3 & {\ul 39.4} & 29.8 & \textbf{50.0} \\
Sketches and Drafts & 184 & 37.0 & 32.1 & 29.9 & {\ul 38.0} & 33.7 & \textbf{55.4} \\
Maps & 170 & 38.2 & 36.5 & 45.9 & {\ul 47.6} & 43.5 & \textbf{61.8} \\
Technical Blueprints & 162 & 24.7 & 25.9 & {\ul 28.4} & 25.3 & 27.8 & \textbf{38.9} \\
Trees and Graphs & 146 & 30.1 & 28.1 & 28.8 & 28.8 & {\ul 34.9} & \textbf{50.0} \\
Mathematical Notations & 133 & 15.8 & {\ul 27.1} & 22.6 & 21.8 & 21.1 & \textbf{45.9} \\
Comics and Cartoons & 131 & 29.0 & 51.9 & 49.6 & {\ul 54.2} & 51.1 & \textbf{68.7} \\
Sculpture & 117 & 30.8 & 46.2 & 49.6 & 51.3 & {\ul 53.0} & \textbf{76.1} \\
Portraits & 91 & 20.9 & 52.7 & 46.2 & {\ul 54.9} & 47.3 & \textbf{70.3} \\
Screenshots & 70 & 38.6 & 35.7 & 38.6 & 34.3 & {\ul 47.1} & \textbf{65.7} \\
Other & 60 & 28.3 & 38.3 & 50.0 & 51.7 & {\ul 58.3} & \textbf{68.3} \\
Poster & 57 & 38.6 & 50.9 & 52.6 & 61.4 & {\ul 64.9} & \textbf{80.7} \\
Icons and Symbols & 42 & 23.8 & {\ul 66.7} & 57.1 & 59.5 & 59.5 & \textbf{78.6} \\
Historical Timelines & 30 & 30.0 & 36.7 & 40.0 & {\ul 43.3} & {\ul 43.3} & \textbf{63.3} \\
3D Renderings & 21 & 33.3 & 28.6 & \textbf{57.1} & 38.1 & {\ul 47.6} & {\ul 47.6} \\
DNA Sequences & 20 & 20.0 & {\ul 45.0} & 25.0 & 25.0 & {\ul 45.0} & \textbf{55.0} \\
Landscapes & 16 & 43.8 & 43.8 & 50.0 & 31.2 & {\ul 62.5} & \textbf{68.8} \\
Logos and Branding & 14 & 21.4 & 57.1 & {\ul 64.3} & 35.7 & 50.0 & \textbf{85.7} \\
Advertisements & 10 & 30.0 & 60.0 & 50.0 & 60.0 & {\ul 70.0} & \textbf{100.0} \\ \bottomrule
\end{tabular}
\caption{Selected models' performance on 30 different image types. Note that a single image may have multiple image types.}
\label{tab:image_types}
\end{table*}

\clearpage

\section{Few-shot Results}
As existing models like OpenFlamingo and Otter support few-shot or in-context learning, we report their few-shot performance using the dev set as the in-context learning examples. 

As shown in \autoref{tab:few_shot}, OpenFlamingo shows a decrease in performance when moving from 0-shot to 1-shot and 3-shot learning (from 0.263 to 0.256) and there is a slight increase when moving to 5-shot.  Otter shows a consistent decline as more shots are introduced, dropping to 0.276 in 1-shot and further down to 0.258 in 3-shot and 5-shot. This trend suggests that existing open-source models' few-shot learning ability is very weak. And it additionally shows that our data samples might be too hard for these models to understand the underlying patterns or context.

\begin{table}[!h]
\centering
\small
\begin{tabular}{@{}lcccc@{}}
\toprule
 & 0shot & 1shot & 3shot & 5shot \\ \midrule
OpenFlamingo & 0.263 & 0.256 & 0.259 & 0.264 \\
Otter & 0.291 & 0.276 & 0.258 & 0.258 \\ \bottomrule
\end{tabular}%
\caption{Few-shot results of OpenFlamingo and Otter.}
\label{tab:few_shot}
\end{table}

\clearpage

\section{Data Annotation Protocol}
This document describes a comprehensive protocol for annotating a dataset comprising college-level multimodal questions  (i.e., questions that incorporate images).

\subsection{Data Collection}

\paragraph{Sources of Data:} Data is primarily collected from free online resources, quizzes, textbooks, and other study materials. When collecting questions, the annotators should strictly adhere to copyright and licensing regulations on the source sites. Data from sources that prohibit copying or redistribution MUST be explicitly avoided. Besides, the annotators should try to find diverse sources instead of collecting questions from a single source.

\vspace{3pt}
\noindent\textbf{Types of Questions:}
\vspace{1pt}

\begin{itemize}
    \item \textbf{Multiple-Choice Questions:} Including standard multiple-choice questions and true/false questions. These are characterized by a question followed by several answer choices, with only one correct option.
    \item \textbf{Open-Ended Questions:} Encompassing formats like factoid, fill-in-the-blank, calculation-based, and short descriptive responses. Avoid collecting questions that have very long answers.
\end{itemize}
 
\noindent\textbf{Image Types:} The annotators should find various types of images (e.g., diagrams, charts, photographs)

\subsection{General Guidelines}

\begin{itemize}
    \item \textbf{General Principles:} Annotations must be accurate, consistent, and adhere to a high standard of academic rigor.
    
    \item \textbf{Specific Instructions:}
    \vspace{1pt}
        \begin{itemize}
            \item All questions must contain one or more images.
            \item All questions should be written in English.
            \item All questions should meet the college-level difficulty. 
            \item The question should not be ambiguous and can be answered with one of the given options or a short answer. 
            \item Clearly categorize each question as either multiple-choice or open-ended.
            \item Annotate all fields, including the question, answer options for multiple-choice questions, the correct answer, image types, question difficulty, and explanation (if there exists).  
        \end{itemize}
\end{itemize}

\subsection{Data Format and Structure}

\begin{itemize}
    \item \textbf{JSON File Format:} The structured JSON format will include fields for number, question type, question text, answer options (for multiple-choice), correct answer, question difficulty, and explanation (if there exists).
    
    \item \textbf{Naming Conventions:}
    \vspace{1pt}
        \begin{itemize}
            \item Each collected sample will be stored in a separate JSON file following a standard naming rule: \textbf{subject\_\{Number\}}.json
            \item Image Files: \textbf{image\_\{QuesNum\}\_\{ImageNum\}}.png
        \end{itemize}

    \item \textbf{Interleaving Question with Images:} The images should be inserted as a file path in the question/options/explanations. 
\end{itemize}

\subsection{Quality Control and Validation}

\begin{itemize}
    \item A secondary review team will rigorously vet annotations for quality and guideline adherence.
    \item Regular audits of random samples from the dataset will be conducted to ensure sustained quality and consistency.
\end{itemize}

\subsection{Handling Ambiguities}
Ambiguities or unclear data instances should be flagged for a detailed review process. These questions will be collaboratively examined in team meetings to establish a standardized approach for annotation.

\subsection{Ethical Considerations}

\begin{itemize}
    \item \textbf{Copyright and Licensing:} Strict adherence to copyright and licensing regulations is mandatory. Data from sources that prohibit copying or redistribution will be explicitly avoided.
    \item \textbf{Data Privacy:} Compliance with privacy laws and ethical standards in data handling is paramount. The annotators should avoid collecting questions that contain any private information.
\end{itemize}

\subsection{Data Contamination Considerations}
In the construction of benchmarks for evaluating foundation models, it is essential to consider the risk of data contamination. To address this, annotators should be tasked with carefully selecting questions that go beyond straightforward queries with easily accessible answers. Instead, the focus should be on questions whose answers are tucked away in less obvious locations, such as in separate documents or hidden in the concluding sections of extensive textbooks. This approach is beneficial for constructing benchmarks that truly test the model's ability to comprehend and synthesize information from diverse and challenging sources.

\subsection{Example Questions}
Detailed examples of annotated questions are provided in an appendix to serve as a reference for the annotators.
\begin{itemize}
    \item \textbf{Multiple-choice Questions:}  \autoref{fig:example_multiple_choice} shows an example of a multiple-choice question.

        \begin{figure*}[!t]
            \centering
            \begin{subfigure}[c]{0.25\linewidth}
                \includegraphics[width=\linewidth]{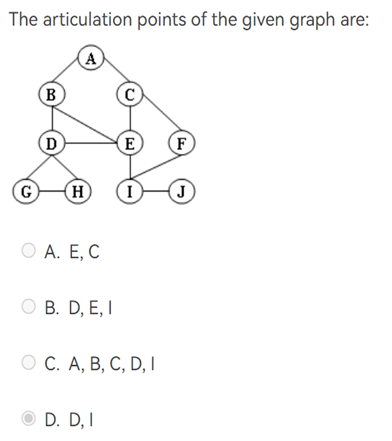}
                \label{fig:example_multiple_choice_a}
            \end{subfigure}
            \begin{subfigure}[c]{0.35\linewidth}
                \includegraphics[width=\linewidth]{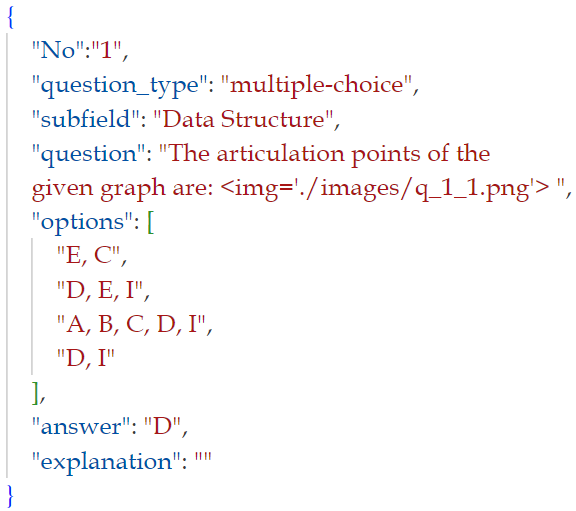}
            \end{subfigure}
            \caption{Multiple-choice question and its JSON representation.}
            \label{fig:example_multiple_choice}
        \end{figure*}

    \item \textbf{Open-ended Questions:} 
    \autoref{fig:example_open} shows an example of the open-ended question.


        \begin{figure*}[!t]
            \centering
            \begin{subfigure}[c]{0.35\linewidth} 
                \centering
                \includegraphics[width=\linewidth]{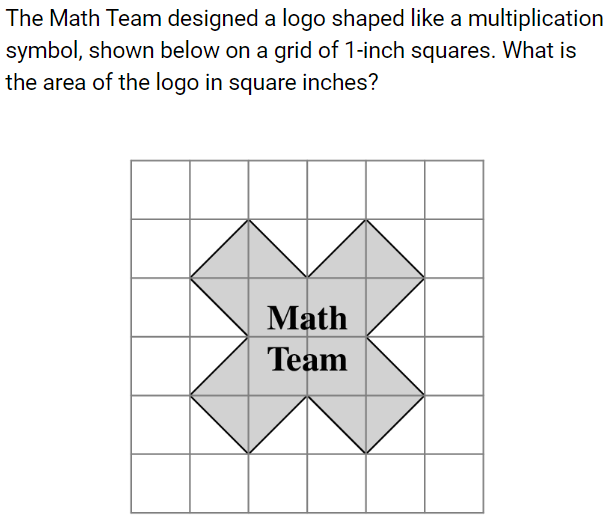}
            \end{subfigure}
            \begin{subfigure}[c]{0.35\linewidth} 
                \centering
                \includegraphics[width=\linewidth]{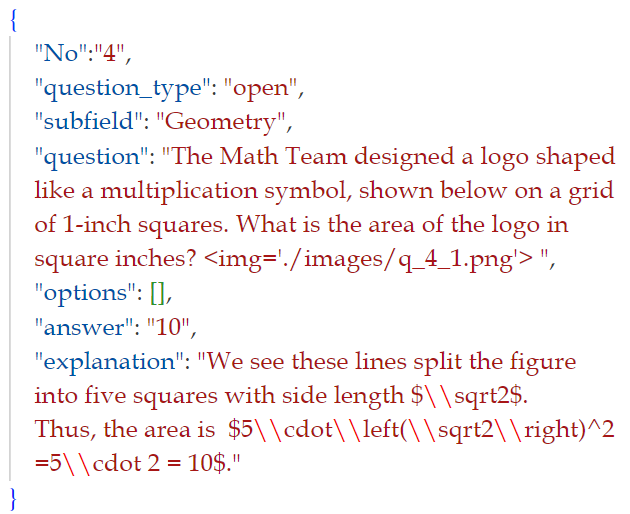}
            \end{subfigure}
            \caption{Open question and its JSON representation.}
            \label{fig:example_open}
        \end{figure*}

        \begin{figure*}[!t]
            \centering
            \begin{subfigure}[c]{0.4\linewidth} 
                \centering
                \includegraphics[width=\linewidth]{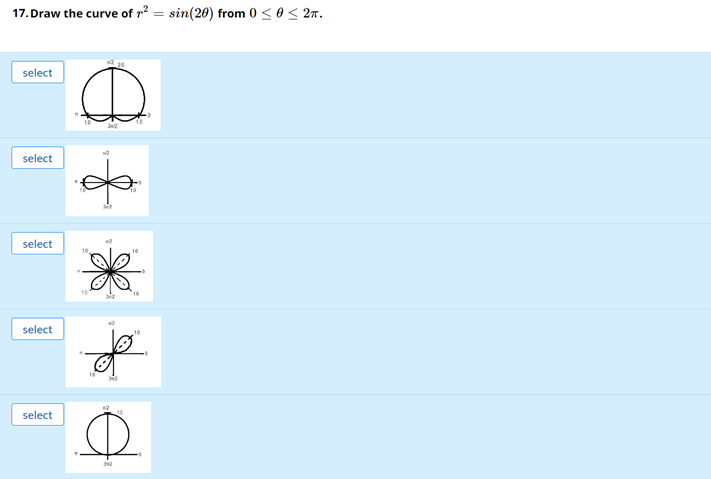}
            \end{subfigure}
            \begin{subfigure}[c]{0.4\linewidth} 
                \centering
                \includegraphics[width=\linewidth]{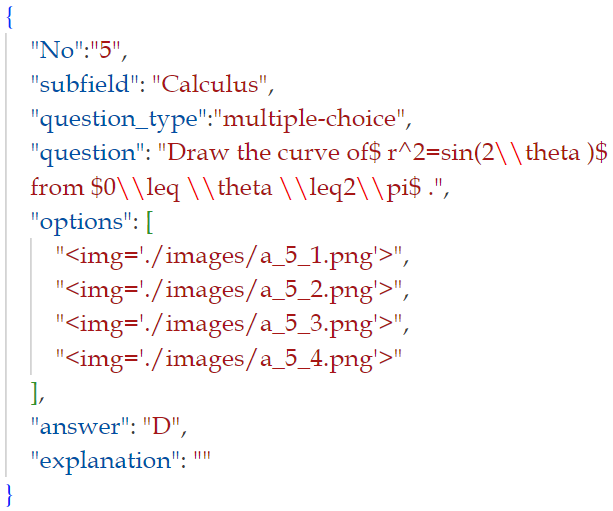}
            \end{subfigure}
            \caption{Multiple image question and its JSON representation.}
            \label{fig:example_multiple_image}
        \end{figure*}
        
\end{itemize}

Besides, the annotators are encouraged to collect questions that contain multiple images within a single example. This type of question requires special attention to file naming so that each image can be correctly referenced. \autoref{fig:example_multiple_image} shows an example of a multiple-image question along with its JSON representation.

\section{Author Contribution Statement}
\label{Appendix:Author_Conntribution}
All authors made significant contributions to data collection, annotation, and validation. We authors contributed to 1/3 of the MMMU examples. Additionally, all authors contributed to the case study and error analysis, plotting case study figures in the Appendix. Besides, all authors participated in the discussion of data annotation, provided feedback on the project, and proofread the paper. The following authors made additional contributions:

\vspace{5pt}
\noindent
\textbf{Xiang Yue} conceived and led the project, outlining the primary objectives, establishing the data collection methodology and protocol, designing and running experiments, as well as doing follow-up analysis. Xiang Yue also took the lead in writing the manuscript, drafting the original text, and incorporating revisions from co-authors. In addition, Xiang Yue managed project administration and coordinated the collaboration between 20+ coauthors and 30+ student annotators, ensuring the project's milestones were met and facilitating communication among team members. Xiang Yue also took the lead in the dataset release. 

\vspace{5pt}
\noindent \textbf{Yuansheng Ni} co-led the data curation process with Xiang Yue. Specifically, Yuansheng Ni developed the protocols for data quality assurance, standardizing the data annotation procedures, and supervising the team of data annotators to ensure consistency and accuracy. In addition to data curation, Yuansheng Ni also played a collaborative role in data analysis, offering critical insights that shaped the interpretation and presentation of the dataset's characteristics.

\vspace{5pt}
\noindent \textbf{Kai Zhang} played a crucial role in the empirical evaluation of the dataset by building the evaluation pipeline to assess various LMMs. Kai Zhang carefully executed different models and analyzed their performance metrics. Kai Zhang also contributed to the manuscript by documenting the evaluation process and implementation details. The thorough model evaluation conducted by Kai Zhang has been fundamental in demonstrating the utility of the dataset. 

\vspace{5pt}
\noindent
\textbf{Tianyu Zheng} made significant contributions to the project by participating in the evaluation of text-only, OCR-augmented and caption-augmented baselines. In addition, Tianyu Zheng developed a user-friendly web interface for data annotation and verification. The interface design significantly improved the workflow for data curation.

\vspace{5pt}
\noindent
\textbf{Ruoqi Liu} plotted or helped revise Figures 1, 2, and 3. Ruoqi Liu designed the prototype figure template for the case study figures in the Appendix.

\vspace{5pt}
\noindent
\textbf{Boyuan Zheng} participated in part of the evaluation. 

\vspace{5pt}
\noindent
\textbf{Huan Sun} and \textbf{Yu Su} provided overarching and insightful discussions and comments throughout the development and execution of the project. Huan Sun and Yu Su contributed to the conceptualization of the research by helping to refine the research questions and by providing critical insights into the design of the dataset. They offered strategic direction and expert advice that significantly enhanced the dataset and the follow-up analysis. Huan Sun and Yu Su also contributed to the initial writing of the paper. 

\vspace{5pt}
\noindent
\textbf{Wenhu Chen} conceived the project with Xiang Yue. Wenhu Chen contributed to the conceptualization of the research by helping to refine the research questions and by providing critical insights into the design of the project. Besides, Wenhu Chen contributed to a significant amount of initial writing of the draft and offered strategic direction and expert advice that significantly enhanced the dataset and the follow-up analysis.

\onecolumn
\section{Version Change Log}
\textbf{CHANGES TO V2 (Dec.18) }FROM V1 (Nov.27)
\begin{itemize}
\setlength{\itemindent}{1em}
    \item We added Qwen-VL-PLUS results from the author-provided outputs. (Table \ref{tab:overall_results}, \ref{tab:overall_art_design}, \ref{tab:overall_Business_results}, \ref{tab:overall_Science_results}, \ref{tab:overall_Health_results}, \ref{tab:overall_Humanities_results}, \ref{tab:overall_Tech_results})
    \item We added SPHINX results from the author-provided outputs. (Table \ref{tab:overall_results}, \ref{tab:overall_art_design}, \ref{tab:overall_Business_results}, \ref{tab:overall_Science_results}, \ref{tab:overall_Health_results}, \ref{tab:overall_Humanities_results}, \ref{tab:overall_Tech_results})
    \item We added Gemini Ultra results from the Gemini report~\cite{deepmind_gemini_report}. (Table \ref{tab:overall_results}, \ref{tab:overall_art_design}, \ref{tab:overall_Business_results}, \ref{tab:overall_Science_results}, \ref{tab:overall_Health_results}, \ref{tab:overall_Humanities_results}, \ref{tab:overall_Tech_results})
    \item We added Gemini Pro \& Nano2 results from the Gemini report~\cite{deepmind_gemini_report}. (Table \ref{tab:overall_results})
    \item We added a section of the author contribution statement. (Appendix \ref{Appendix:Author_Conntribution})
    \item We updated mPLUG-Owl2 results with author-provided prompt. (Table \ref{tab:overall_results}, \ref{tab:overall_art_design}, \ref{tab:overall_Business_results}, \ref{tab:overall_Science_results}, \ref{tab:overall_Health_results}, \ref{tab:overall_Humanities_results}, \ref{tab:overall_Tech_results})
    \item We fixed text box dimensions in Appendix \ref{Appendix:Case_Study}:
    \begin{itemize}
    \setlength{\itemindent}{1em}
        \item Figure \ref{fig:art_theory_3}. A sample error case of Art Theory
    \end{itemize}
    \item We fixed the typo in Appendix \ref{Appendix:Case_Study}:
    \begin{itemize}
    \setlength{\itemindent}{1em}
        \item Figure \ref{fig:biology_5}. A sample error case of Biology
        \item Figure \ref{fig:math_3}. A sample error case of Math
        \item Figure \ref{fig:agriculture_4}. A sample error case of Agriculture
        \item Figure \ref{fig:mechanical_engineering_3}. A sample error case of Mechanical Engineering
    \end{itemize}
\end{itemize}

\vspace{5pt}
\noindent
\textbf{CHANGES TO V3 (Dec.21)} FROM V2 (Dec.18)
\begin{itemize}
\setlength{\itemindent}{1em}
    \item We fixed the typo in the homepage URL.
\end{itemize}

\vspace{5pt}
\noindent
\textbf{CHANGES TO V4 (June.13)} FROM V3 (Dec.21)
\begin{itemize}
\setlength{\itemindent}{1em}
    \item We Updated the name of QwenVL-7B to QwenVL-7B-Chat.
    \item We Updated the name of Fuyu-8B to Adept Fuyu-8B.
    \item We Updated the name of Gemini Ultra \& PRo to Gemini 1.0 Ultra \& Pro.
    \item We adjusted the overall results in \autoref{tab:overall_results} and added the main results table in \autoref{tab:main_results}.
    \item We Updated SPHINX Test set results from the \href{https://eval.ai/web/challenges/challenge-page/2179/overview}{EvalAI}. (Table \ref{tab:overall_results}, \ref{tab:main_results}, \ref{tab:overall_art_design}, \ref{tab:overall_Business_results}, \ref{tab:overall_Science_results}, \ref{tab:overall_Health_results}, \ref{tab:overall_Humanities_results}, \ref{tab:overall_Tech_results})
    \item We added the validation set results from the author-provided outputs. (Table \ref{tab:overall_results}, \ref{tab:main_results}, \ref{tab:overall_art_design}, \ref{tab:overall_Business_results}, \ref{tab:overall_Science_results}, \ref{tab:overall_Health_results}, \ref{tab:overall_Humanities_results}, \ref{tab:overall_Tech_results})
    \begin{itemize}
        \setlength{\itemindent}{1em}
        \item MiniCPM-V \& V2
        \item OmniLMM-12B
        \item HPT Air \& Pro
        \item Adept Fuyu-Heavy
    \end{itemize}
    \item We added the test \& validation set results from the author-provided and \href{https://eval.ai/web/challenges/challenge-page/2179/overview}{EvalAI}. (Table \ref{tab:overall_results}, \ref{tab:main_results}, \ref{tab:overall_art_design}, \ref{tab:overall_Business_results}, \ref{tab:overall_Science_results}, \ref{tab:overall_Health_results}, \ref{tab:overall_Humanities_results}, \ref{tab:overall_Tech_results})
    \begin{itemize}
        \setlength{\itemindent}{1em}
        \item Emu2-Chat
        \item SVIT
        \item Yi-VL-6B \& 34B
        \item Bunny-3B
        \item InternVL-Chat-V1.1 \& V1.2
        \item InfiMM-Zephyr-7B
        \item InternLM-XComposer2-VL
        \item LLaVA-1.6-34B
        \item Marco-VL \& PLUS
        \item Qwen-VL-MAX
        \item SenseChat-Vision-0423-Preview
        \item VILA1.5
        \item Skywork-VL
    \end{itemize}
    \item We added Reka Edge \& Flash \& Core results from the Reka report \cite{Reka}. (Table \ref{tab:main_results})
    \item We added Claude 3 Haiku \& Sonnet \& Opus results from the Claude 3 report \cite{Claude3}. (Table \ref{tab:overall_results}, \ref{tab:main_results})
    \item We added Gemini 1.5 Flash \& Pro results from the Gemini report \cite{deepmind_gemini1.5_report}. (Table \ref{tab:main_results}) 
    \item We added GPT-4o results from the OpenAI report \cite{gpt-4o}. (Table \ref{tab:main_results})
    \item We added the human experts' results.(Table \ref{tab:overall_results}, \ref{tab:main_results}, \ref{tab:overall_art_design}, \ref{tab:overall_Business_results}, \ref{tab:overall_Science_results}, \ref{tab:overall_Health_results}, \ref{tab:overall_Humanities_results}, \ref{tab:overall_Tech_results})
    \item We removed an erroneous biology case and renamed.
\end{itemize}

\end{document}